\pdfoutput=1
\documentclass[10pt,twocolumn,letterpaper]{article}
\usepackage[pagenumbers]{paperlayout}
\usepackage{amssymb,amsthm,graphicx,booktabs,array,tabularx,colortbl,multirow}
\usepackage{needspace}
\usepackage{listings}
\usepackage{float}
\usepackage{xr-hyper}
\definecolor{rowtint}{rgb}{0.94,0.97,0.995}
\definecolor{summarytint}{gray}{0.935}
\newcommand{\hlrow}{\rowcolor{rowtint}}
\newcommand{\sumrow}{\rowcolor{summarytint}}
\newcommand{\best}[1]{\textbf{#1}}
\newcommand{\second}[1]{\underline{#1}}
\definecolor{rescuecol}{HTML}{168979}
\definecolor{harmcol}{HTML}{A04C25}
\definecolor{keptcol}{HTML}{3C444C}
\definecolor{failcol}{HTML}{8C939A}
\definecolor{recfull}{rgb}{0.082,0.275,0.431}
\definecolor{recadd}{rgb}{0.188,0.529,0.686}
\definecolor{recbnd}{rgb}{0.412,0.459,0.553}
\newcommand{\rescuec}[1]{\textcolor{rescuecol}{#1}}
\newcommand{\harmc}[1]{\textcolor{harmcol}{#1}}
\newcommand{\keptc}[1]{\textcolor{keptcol}{#1}}
\newcommand{\failc}[1]{\textcolor{failcol}{#1}}
\newcommand{\swatch}[1]{\textcolor{#1}{\rule[-0.05em]{0.62em}{0.62em}}}

\usepackage{amsmath,amsfonts,bm}

\def\eqref#1{equation~\ref{#1}}

\def\1{\bm{1}}

\DeclareMathAlphabet{\mathsfit}{\encodingdefault}{\sfdefault}{m}{sl}
\SetMathAlphabet{\mathsfit}{bold}{\encodingdefault}{\sfdefault}{bx}{n}

\newcommand{\SFourBaselineSR}{86.69}
\newcommand{\SFourRegularizedSR}{90.81}
\newcommand{\SFourGain}{4.11}
\newcommand{\SFourHammerGain}{6.93}
\newcommand{\SFourSwitchGain}{9.49}

\newcommand{\LiberoHeldoutGain}{26.67}
\newcommand{\LiberoHeldoutCILow}{10.22}
\newcommand{\LiberoHeldoutCIHigh}{43.78}

\definecolor{paperlayoutblue}{rgb}{0.21,0.49,0.74}
\usepackage[pagebackref,breaklinks,colorlinks,allcolors=paperlayoutblue]{hyperref}
\newcommand{\PaperTitle}{Sharing Corrections across Tasks: A Study of Generalizable Action Editing for Pretrained Robot Policies}
\title{\PaperTitle}
\author{Guangyu Huang$^{1,2,*,\ddagger}$\quad Ye Lin$^{1,*}$\quad Mingyu Liu$^{1,*}$\quad Yue Hu$^{1}$\quad Haokai Zhang$^{1}$\quad Jinze Lu$^{3}$\\
Hengyu Zhao$^{1}$\quad Liyang Li$^{1}$\quad Jing Liu$^{2,4,\dagger}$\quad Hao Chen$^{1,\dagger}$\quad Chunhua Shen$^{1,\dagger}$\\[2pt]
{\small $^{1}$Zhejiang University, State Key Lab of CAD \& CG\quad $^{2}$Zhongguancun Academy\quad $^{3}$Lanzhou University}\\[-1pt]
{\small $^{4}$Institute of Automation, Chinese Academy of Sciences\qquad $^{*}$Equal contribution\quad $^{\ddagger}$Project lead\quad $^{\dagger}$Corresponding authors}\\[4pt]
{\small\textbf{Project Page:}~\href{https://huang2202.github.io/sharing-corrections/}{\fontsize{8.6}{10.2}\selectfont\sffamily\bfseries https://huang2202.github.io/sharing-corrections}}}
\hypersetup{pdfauthor={Guangyu Huang, Ye Lin, Mingyu Liu, Yue Hu, Haokai Zhang, Jinze Lu, Hengyu Zhao, Liyang Li, Jing Liu, Hao Chen, Chunhua Shen},pdftitle={\PaperTitle}}
\makeatletter
\patchcmd{\@maketitle}{\vspace*{12pt}}{\vspace*{0pt}}{}{\errmessage{title patch failed}}
\patchcmd{\@maketitle}{\vspace*{24pt}}{\vspace*{8pt}}{}{\errmessage{title patch failed}}
\patchcmd{\@maketitle}{\vskip .375in}{\vskip .04in}{}{\errmessage{title patch failed}}
\patchcmd{\abstract}{\vspace*{12pt}}{\vspace*{6pt}}{}{\errmessage{abstract start patch failed}}
\patchcmd{\endabstract}{\vspace*{12pt}}{\vspace*{2pt}}{}{\errmessage{abstract patch failed}}
\makeatother
\makeatletter\providecommand{\swinput}[1]{\input{#1} }\makeatother

\usepackage{cuted}
\begin{document}
\twocolumn[{%
\renewcommand\twocolumn[1][]{#1}%
\maketitle
\begin{center}
\vspace{-16pt}
\includegraphics[width=0.955\textwidth]{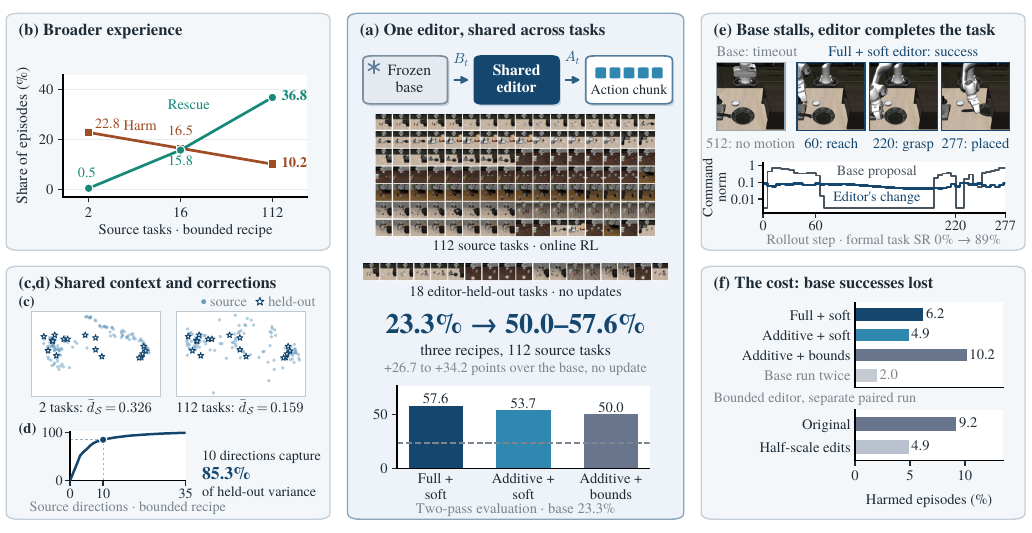}
\captionsetup{hypcap=false}
\captionof{figure}[Cross-task action editing.]{\textbf{Sharing corrections across tasks.}
(a)~One editor trained on 112 source tasks raises success on 18 editor-held-out tasks
from 23.3\% to 50.0--57.6\% across three recipes.
(b)~More source tasks raise \rescuec{rescue} and lower \harmc{harm}.
(c,\,d)~Broad source experience covers held-out contexts, and ten source correction
directions capture 85.3\% of the bounded editor's held-out correction variance.
(e)~On a held-out task, the base stalls and the editor completes the task.
(f)~Soft-penalty recipes harm fewer episodes.
}
\label{fig:correction-support}
\captionsetup{hypcap=true}
\end{center}%
\vspace{-10pt}%
}]

\begin{abstract}
Action-editing methods refine frozen robot policies via online reinforcement
learning, typically one task at a time. We test whether a single shared editor
transfers to \emph{editor-held-out tasks}, on which the base policy was
fine-tuned but no editor was trained. On LIBERO-130, editors trained on 112
source tasks under three recipes raise mean held-out success from 23.3\% to
50.0--57.6\% with no update on those tasks.
Each base--editor episode pair is rescued, kept, harmed, or still failed, and
the gain is rescue minus harm. Rescue grows and harm falls with source
breadth, while the source-task ranking of six 16-task recipes does not carry
over to held-out tasks. Gains concentrate on tasks the base rarely solves,
and harm on tasks it solves at least half the time. Harm follows correction
size: soft-penalty recipes harm 4.9--6.2\% of held-out episodes, against
10.2\% with output bounds, and halving the correction at inference lowers
harm. A fixed source-mean correction nearly matches the
bounded editor but falls 12.0--18.1 points below the soft-penalty editors,
which depend on live context. In single-task adaptation, pretraining halves
critic loss under every recipe; offline source replay more than doubles
retained source success under the additive soft penalty without lowering
target success, while the bounded recipe stays below the base. With a different base and simulator
(RoboTwin), a four-task editor repeats the narrow-source pattern of two-task
LIBERO editors: it raises source-task success from 74.6\% to 84.1\% and
lowers it on 12 others. ReSOP, our asynchronous system, trains the
112-task bounded editor in 167~GPU-hours.
\end{abstract}

\section{Introduction}
\label{sec:introduction}

Imitation pretraining provides broad manipulation
skills~\citep{team2024octo,kim2024openvla,black2024pi0,intelligence2025pi05,wang2025vqvla,liu2026gaeexpert,huang2026notvla,liu2026stamo,yang2026como,liu2026perturbot},
yet precise manipulation remains sensitive to execution error~\citep{liu2026perfectdemo}. Online action
editing over a frozen base offers lightweight
refinement~\citep{yuan2024policy,ankile2025residual,xu2026rltoken}, but prior
work trains one editor per task, leaving cross-task transfer unexamined. We
evaluate whether a shared editor helps \emph{editor-held-out tasks}, on which
the base policy was fine-tuned but no editor was trained, with no update on
those tasks.

On LIBERO-130, editors trained on 112 source tasks under three recipes raise
mean success rate (SR) on 18 held-out tasks from 23.3\% to 50.0--57.6\%
(Figure~\ref{fig:correction-support}a).
The gain is the share of episodes rescued minus the share harmed, and
soft-penalty recipes harm fewer: 4.9\% and 6.2\%, against 10.2\% for the
output-bounded recipe.

We hypothesize that corrections recur across tasks in local contexts such as
contact and alignment, and study what transfers. Four findings follow.
\begin{itemize}[leftmargin=*,nosep]
    \item \textbf{Transfer grows with source breadth, and source success does
    not predict it}
    (Sections~\ref{sec:experiments:libero-reference} and~\ref{sec:experiments:design}).
    A two-task editor reaches 93.5\% on its source tasks yet 1.0\% on held-out
    tasks (Figure~\ref{fig:correction-support}b).
    \item \textbf{Gains concentrate where the base fails}
    (Sections~\ref{sec:experiments:libero-reference}--\ref{sec:experiments:transfer-structure}).
    The editors gain most on tasks the base rarely solves; harm arises on
    tasks it solves at least half the time.
    \item \textbf{Harm follows correction size}
    (Sections~\ref{sec:experiments:transfer-structure}--\ref{sec:experiments:design}).
    It is lower under soft penalties, and halving the correction at inference
    lowers it under all three recipes (9.2\% to 4.9\% in one batch of the
    bounded editor).
    \item \textbf{A fixed offset explains the bounded editor's transfer;
    soft-penalty editors depend on live context}
    (Section~\ref{sec:experiments:transfer-structure}). A source-mean correction
    falls 12.0--18.1 points below them (Figure~\ref{fig:libero-controls}).
\end{itemize}
We further show that in single-task adaptation pretraining halves critic loss
under every recipe and source replay preserves far more source skill under the
additive soft penalty than under output bounds
(Section~\ref{sec:experiments:adaptation}), and that
the narrow-source pattern recurs with a different base and simulator: a
four-task RoboTwin editor improves its source tasks (from 74.6\% to 84.1\%)
yet lowers success on 12 others (Section~\ref{sec:experiments:robotwin}). Our system \emph{ReSOP}
(Section~\ref{sec:method:resop}), built on the RLinf codebase~\citep{yu2026rlinf} and adopting
the decoupled, task-balanced replay design of SOP~\citep{pan2026sop}, trains the 112-task bounded editor in
167~GPU-hours and removes 79\% of render calls in command replay.

\section{Related Work}
\label{sec:related-work}

\paragraph{Online improvement of pretrained policies.}
Residual reinforcement learning (RL) adds a trainable correction to an existing controller~\citep{silver2018residual,johannink2019residual} or imitation policy~\citep{yuan2024policy,ankile2025imitation,ankile2025residual}, extended via value learning~\citep{ma2026dawn,dodeja2026uncertainty}, subtask decomposition~\citep{liu2026foresight}, object-centric features~\citep{kim2026objectcentric}, human feedback~\citep{muttaqien2026orpa}, and latency handling~\citep{dong2026realtime}. Other methods steer diffusion or flow policies~\citep{wagenmaker2025dsrl,tang2026frs,nakamoto2024steering}, feed compact RL tokens to anchored actors~\citep{xu2026rltoken}, regularize toward the base's support~\citep{xu2026gsbr}, or update the base policy directly~\citep{Li2025SimpleVLA_RL,perry2026expoft,chen2025conrft,guo2025irevla,hu2025flare,ren2025dppo,lu2025vlarl,dong2026expo,peng2026visuallyrobust}. We adopt RL Token and ResFiT as starting points to study how representation, prediction, and anchoring shape shared editing.

\paragraph{Multi-task improvement and transfer.}
RL post-training of the base policy can generalize beyond supervised fine-tuning~\citep{liu2025rlvla,chu2025sft,li2026drive}, but it does not learn a lightweight editor over a frozen base across tasks. MT-Opt shares experience across tasks~\citep{kalashnikov2022scaling}; DyGRO-VLA routes residual experts via task-supervised embeddings on source tasks~\citep{lin2026dygrovla}; PLD distills task-specific rollouts back into the generalist~\citep{xiao2025self}; Korniak et al.~\citep{korniak2026generalist} extract or rank actions for held-out tasks. We transfer one shared editor over a frozen base without distillation or execution-time critics, decompose its gain into rescue and harm, and study how transfer varies with task breadth.

\paragraph{Joint optimization and asynchronous systems.}
Parameter sharing across tasks~\citep{yu2019metaworld,nauman2025brc} risks competing updates~\citep{yu2020gradient,liu2023famo}; unlike continual retention studies~\citep{lopezpaz2017gradient,lesort2020continual,liu2026pretrained,shenfeld2026razor}, we pair one shared actor with per-task critics and measure gradient relations, mixture-of-experts (MoE) routing, and downstream adaptation without and with source replay. On systems, large-scale robotic RL pairs distributed collection with off-policy replay~\citep{kalashnikov2018qtopt,luo2024serl,luo2024hilserl}, SOP separates collection from balanced sampling~\citep{pan2026sop}, and SimpleVLA-RL/RLinf-VLA parallelize simulation~\citep{Li2025SimpleVLA_RL,zang2026rlinfvla}; ReSOP builds on RLinf~\citep{yu2026rlinf} for lightweight shared-editor training.

\section{Action Editing with a Shared Policy}
\label{sec:method}

\begin{figure*}[t]
    \centering
    \includegraphics[width=0.83\textwidth]{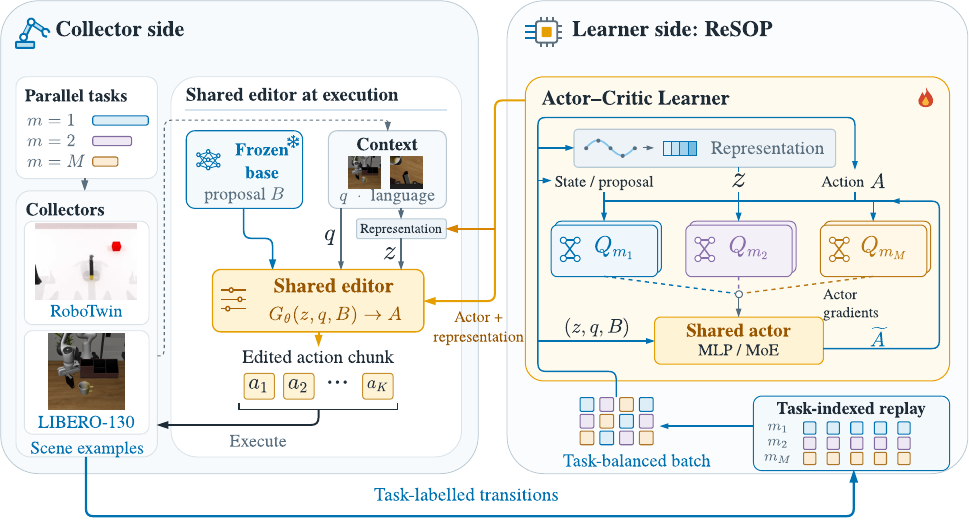}
    \caption{\textbf{Shared action editing in ReSOP.}
    Collectors execute edited chunks; the learner trains a shared actor with
    per-task critics (blue: transitions; orange: snapshots; dashed: gradients).
    Main LIBERO runs use MLP actors; routing studies use MoE actors.}
    \label{fig:method-overview}
\end{figure*}

We study a design space for modifying a frozen base policy with one editor
shared across tasks: a representation, an action prediction and an anchoring
choice, trained with ReSOP (Figure~\ref{fig:method-overview}).

\subsection{Problem setting and context coverage}
\label{sec:method:interface}

Let $o_t$, $q_t$, and $\ell$ denote the visual observation, proprioceptive
state, and instruction at decision time $t$. A frozen base policy proposes
an action plan, of which $B_t\in\mathbb R^{K\times d_a}$ is supplied to
the current editor decision. The editor combines a representation $z_t$
of the observation and instruction with $q_t$ and $B_t$:
\begin{equation}
    c_t=(z_t,q_t,B_t), \qquad A_t=G_\theta(c_t).
    \label{eq:shared-editor}
\end{equation}
$A_t$ is the noise-free edited proposal. Each new decision uses a fresh
observation after the preceding chunk ends.

An editor learns from source tasks $\mathcal S\subset\mathcal T$, which
supply demonstrations, online interaction, and reward supervision.
Editor-held-out tasks $\mathcal H\subset\mathcal T$, with
$\mathcal S\cap\mathcal H=\varnothing$, supply none of these training data.
We evaluate transfer with the editor fixed.
Task indices identify replay partitions and critics during training;
the shared policy acts from its context $c_t$.

We hypothesize that broader source experience supports transfer by covering
local contexts that require similar corrections. We measure visual
\emph{context coverage} by nearest-source distance,
\begin{equation}
    d_{\mathcal S}(c)=
    \min_{c'\in\mathcal C_{\mathcal S}}
        \|\phi(c)-\phi(c')\|_2,
    \label{eq:context-coverage}
\end{equation}
where $\mathcal C_{\mathcal S}$ contains sampled source contexts and
$\phi$ is a fixed visual feature map shared by the compared conditions.
We use the 112-task editor's encoder at 241.25k updates
(Appendix~\ref{app:libero-breadth}). As a coarse scene-level proxy, we group
editor-held-out tasks into \emph{source-covered scenes} (sharing a named
scene or single-scene suite with $\mathcal S$) and \emph{uncovered scenes}.
To separate context-dependent edits from a static action bias, we
define a fixed $K\times d_a$ \emph{source-mean template} by averaging
chunk corrections over sampled source-task rollouts.

\subsection{Recipe space}
\label{sec:method:recipes}

Table~\ref{tab:editing-recipes} crosses two representations with three
editing choices. R1 adapts the RL Token approach to multi-task learning,
and L3 adapts ResFiT; the other recipes connect their representation and
editing choices.

\begin{table}[t]
\centering
\caption{\textbf{Six recipes for shared action editing.}
R and L denote the RL-token readout (main camera) and the
online-learned encoder (main and wrist cameras), as in their source methods.
All LIBERO recipes use MLP actors.
Swatches give the color of each editing choice in all figures.}
\label{tab:editing-recipes}
\small\setlength{\tabcolsep}{3pt}
\begin{tabularx}{\linewidth}{@{}l>{\raggedright\arraybackslash}Xl@{}}
\toprule
Recipes & Actor output $A_t$ & Objective \\
\midrule
\swatch{recfull}~R1, L1 & $\pi_\theta(c_t)$ & Eq.~\ref{eq:soft-actor} \\
\swatch{recadd}~R2, L2 & $B_t+\Delta_\theta(c_t)$ & Eq.~\ref{eq:soft-actor} \\
\swatch{recbnd}~R3, L3 & $B_t+\Lambda\odot r_\theta(c_t)$ & Eq.~\ref{eq:resfit-actor} \\
\bottomrule
\end{tabularx}
\end{table}

All LIBERO recipes use end-effector coordinates and five-action chunks;
additive editors predict one correction per action. The base policy's native eight-action
execution is reported as a reference in Section~\ref{sec:experiments}; other
platforms retain their base policy's action coordinates.

\subsection{Representation}
\label{sec:method:rlt}

Following RL Token~\citep{xu2026rltoken}, the R recipes use a readout trained
to extract a compact RL token from the frozen base's final-layer
visual--language token features. Both the base and readout remain fixed
during online RL, while the actor and critics learn from this representation.

The L recipes instead train an independent visual encoder on the current
camera inputs, with language and proprioception supplying additional
context, following ResFiT~\citep{ankile2025residual}. Critic gradients
update the encoder online; actor updates use detached features.

\subsection{Editing, anchoring, and value learning}
\label{sec:method:resfit}

For full-action recipes, $A_t=\pi_\theta(c_t)$; for additive soft-penalty recipes,
$A_t=B_t+\Delta_\theta(c_t)$. At zero output the latter reproduces the base,
whereas a full-action actor must learn that mapping, so R1 and L1 use a
proposal-fitting stage.

\paragraph{Soft penalty.}
R1, R2, L1, and L2 penalize departure from the base through
\begin{equation}
    \mathcal L^{\mathrm{soft}}_{\pi,m}
       = \mathcal L^{\mathrm{value}}_{\pi,m}
       + \beta\,\mathbb E_m[
           \operatorname{MSE}(\widetilde A_t,B_t)].
    \label{eq:soft-actor}
\end{equation}
Here $\widetilde A_t$ is the recipe's training action obtained from $A_t$
with its training noise and command limits.
The value term uses the mean critic score $\overline Q_m$,
$\mathcal L^{\mathrm{value}}_{\pi,m}=-w_m\mathbb E_m[\overline Q_m(x_t,\widetilde A_t)]$,
where $w_m$ is the recipe's value weight. The critic context $x_t$ combines the
observation representation, proprioception, and base-plan context.
$\mathbb E_m$ averages over task $m$'s batch and the penalty over chunk entries; $\beta=5$ in LIBERO.
We compare complete recipes, including their associated optimization choices.

\paragraph{Bounded output.}
R3 and L3 produce a unit-bounded residual mean $r_\theta$:
\begin{equation}
    A_t=B_t+\Lambda\odot r_\theta(c_t).
    \label{eq:resfit-action}
\end{equation}
The per-coordinate scale $\Lambda$, broadcast over the chunk, bounds the
noise-free correction. Without a deviation penalty, the actor maximizes the
mean of the selected task's critic heads, $\overline Q_m$, up to the output
scale $\Lambda$:
\begin{equation}
    \mathcal L^{\mathrm{bound}}_{\pi,m}
       =-\mathbb E_m[\overline Q_m(x_t,\widetilde A_t)].
    \label{eq:resfit-actor}
\end{equation}

\paragraph{Task-specific value learning.}
Each source task maintains online and target copies of a critic ensemble
$\{Q_{m,h}\}_{h=1}^{H_Q}$ ($H_Q=10$ in LIBERO) trained on demonstrations and
online transitions, evaluating proposed chunks against $L$-action physical
returns $R_t^{(L)}=\sum_{j=0}^{L-1}\gamma^j r_{t+j}$ ($L=3K$ in LIBERO) and terminal flag
$d_t$~\citep{li2025qchunking}:
\begin{equation}
    y_t=R_t^{(L)}+(1-d_t)\gamma^L
        Q_m^-(x_{t+L},\widetilde A^-_{t+L}),
    \label{eq:editor-backup}
\end{equation}
where $Q_m^-$ minimizes over two sampled target heads~\citep{chen2021randomized}
and $\widetilde A^-$ is the backup action (target-smoothed in
TD3~\citep{fujimoto2018addressing} for L3). LIBERO uses terminal success
rewards ($0/1$), delayed actor updates ($1:4$), equal demonstration/online
batches after critic warm-up~\citep{ball2023rlpd,nakamoto2023calql}, and no
entropy backup. Deployment uses only the actor, its representation, and
the frozen base; editor-held-out tasks need no critic.

\section{ReSOP: Training One Editor from Many Tasks}
\label{sec:method:resop}

To avoid per-task training cost, ReSOP is built on the RLinf
codebase~\citep{yu2026rlinf} and adopts the decoupled collection and
task-balanced replay design of SOP~\citep{pan2026sop}, adding task-specific critics and
chunked execution (Figure~\ref{fig:method-overview}). Collectors store
transitions per task and learners draw balanced demonstration/online batches,
keeping replay sampling at $1.06\times$ max/min ratio across 112 tasks despite
a $7.07\times$ interaction ratio. All
1,120 critic heads, targets, and optimizers (156.1\,GB full learner state)
stay on the learner, while collectors pull only the 22.7\,MB inference
snapshot (actor and encoder/readout). Cameras render only at $K$-action
boundaries and episode ends while physics, proprioception, rewards, and
success checks run every step: on $K=5$ two-task command replay, this matches
states, rewards, and boundary images while lowering render calls by 79\% and
raising speed by 38.5\%. ReSOP trains the 112-task
bounded editor to 481,250 actor updates (12.1M actions) in 41.8 hours on four
GPUs (167~GPU-hours, 1.49 per task); a third seed with later learner
optimizations takes 36.31~GPU-hours to 121.25k updates, and each single-task
adaptation takes 0.94--1.60~GPU-hours on one GPU
(Appendix~\ref{app:sw-system}).

\section{Experiments}
\label{sec:experiments}
\label{sec:experiments:setup}

Table~\ref{tab:what-matters} summarizes the findings. LIBERO runs freeze the OpenVLA-OFT 130-task
checkpoint~\citep{kim2025oft,rlinfLibero130Base,liu2023libero} and execute five
physical actions per base query.

\paragraph{Evaluation.}
Success rate (SR) averages episodes per task and then tasks equally. Each
frozen checkpoint is paired with the matching base by task, initial state,
and pass, under 50 states$\times$2 passes unless specified otherwise. Every
pair has one of four outcomes, colored as in
Figure~\ref{fig:libero-reference-training}b: \rescuec{\textbf{rescued}}
(the base fails, the editor succeeds), \keptc{\textbf{kept}} (both
succeed), \harmc{\textbf{harmed}} (the base succeeds, the editor fails),
or \failc{\textbf{still failed}} (both fail).
All four are shares of a cohort's paired episodes ($r+k+h+f=1$): editor SR
is $r+k$, base SR $k+h$, and the gain $r-h$. \emph{Rescue} and \emph{harm}
always mean $r$ and $h$, not rates conditioned on the base outcome. We write
held-out for editor-held-out below. Reported runs use one
training seed unless noted; Appendix~\ref{app:sw-robustness} repeats the
evaluations and adds training seeds (overview in
Appendix~\ref{app:sw-guide}). 95\% confidence intervals (CIs) use task-level
or intra-task initial-state-cluster bootstrap.

\subsection{Does learned action editing transfer across tasks?}
\label{sec:experiments:libero-reference}

\emph{Yes: all three 112-task recipes improve held-out tasks without any
update, and transfer grows with source breadth.}

We evaluate the three online-encoder recipes (\swatch{recfull}~L1,
\swatch{recadd}~L2, \swatch{recbnd}~L3; Table~\ref{tab:editing-recipes})
trained on 112 source tasks to 481,250 actor updates, on their source tasks
and on 18 editor-held-out tasks without target updates, alongside 2-task and
16-task L3 editors.

\paragraph{All three recipes transfer without target updates.}
On the 112 source tasks the editors reach 93.55--94.72\% from a base SR of
19.09\% (Table~\ref{tab:learner-classes}).

Without target updates on the 18 held-out tasks, L1 raises SR from 23.33\%
to \textbf{57.56\%} (+34.22 points). Under the same pairing, L2 reaches
\textbf{53.67\%} and L3 reaches \textbf{50.00\%} (+\LiberoHeldoutGain\ points;
Table~\ref{tab:learner-classes}), with 95\% whole-task gain CI
[\LiberoHeldoutCILow, \LiberoHeldoutCIHigh]. All pairwise recipe differences
exclude zero: L1 exceeds L3 by +7.56 points [4.72, 10.33], L2 exceeds L3 by
+3.67 [0.78, 6.61], and L1 exceeds L2 by +3.89 [1.11, 6.61]. Success improves
on 16 of the 18 held-out tasks under L1 and L2, and on 12 under L3.
At 121.25k updates both soft-penalty recipes exceed L3 in each of three
training seeds; one L3 seed is 11.94 points below the first, and the L1--L2 order depends on
training length and seed (Appendix~\ref{app:sw-robustness-seeds112}).

\begin{figure}[t]
    \centering
    \includegraphics[width=\linewidth]{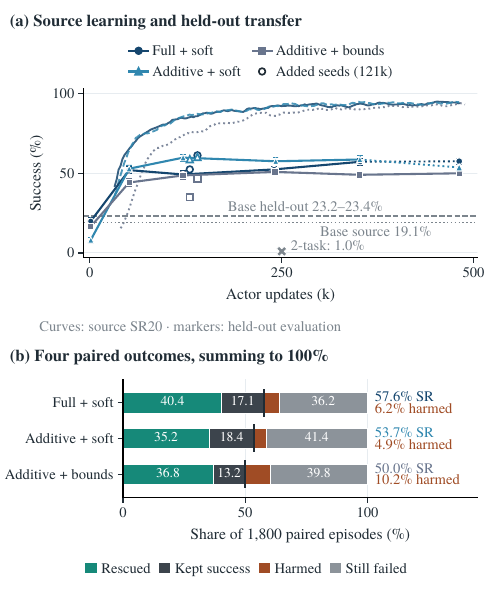}
    \caption{\textbf{LIBERO learning and paired outcomes.}
    (a)~Task-equal rolling 20-episode source SR and held-out SR across
    training for three 112-task recipes. Dotted segments join L1/L2 to
    endpoints from a separate batch; hollow markers are two added seeds at 121k; whiskers
    are 95\% initial-state bootstrap intervals; the cross is the two-task L3 editor.
    (b)~Shares of 1,800 paired held-out episodes \rescuec{rescued},
    \keptc{kept}, \harmc{harmed}, or \failc{still failed}; black ticks mark
    editor SR.}
    \label{fig:libero-reference-training}
\end{figure}

\paragraph{Soft penalties harm less; gains concentrate where the base fails.}
As Figure~\ref{fig:libero-reference-training}b illustrates across 1,800
episodes, harm is 4.9\% (L2) and 6.2\% (L1) under soft penalties and
10.2\% under bounds (L3). Rescue spans 35.2--40.4\% across recipes
(Table~\ref{tab:learner-classes}). Repeating the base alone gives 2.0\% harm, below all three editors (Appendix~\ref{app:sw-robustness-repeat}).

On the nine tasks the base solves at most
7\% of the time, the editors add 31--42 points with at most 2\% harm. On the
five tasks it solves at least half the time, L1 and L2 still add 15--18
points, whereas L3 loses 2 points. Success on placing wine onto the cabinet, for example, falls from
55\% to 22--24\% under L1 and L3 and stays at 55\% under L2.

Source SR varies by 1.17 points across the three recipes, whereas
held-out SR differs by 7.56 points. Transfer appears early: held-out SR
reaches 44.22\% under L3, 52.00\% under L1 and 52.89\% under L2 at 51.25k updates
(Figure~\ref{fig:libero-reference-training}a; Appendix~\ref{app:sw-completions-evolution}).

\begin{table}[t]
\centering
\caption{\textbf{What matters for transfer of a shared editor}, with the
section that gives the evidence. AUC: area under the success curve.}
\label{tab:what-matters}
\footnotesize\setlength{\tabcolsep}{3pt}
\begin{tabularx}{\linewidth}{@{}l>{\raggedright\arraybackslash}Xl@{}}
\toprule
Factor & Finding & Sec. \\
\midrule
Source breadth & Rescue grows and harm falls with it & \ref{sec:experiments:libero-reference} \\
Source-task success & Does not predict the held-out ranking & \ref{sec:experiments:design} \\
Base competence & Gains concentrate where the base fails & \ref{sec:experiments:libero-reference} \\
Correction size & Halving it lowers harm under every recipe & \ref{sec:experiments:transfer-structure} \\
Penalty or bounds & Soft penalties harm less & \ref{sec:experiments:design} \\
Fixed offset & Explains the bounded editor's transfer only & \ref{sec:experiments:transfer-structure} \\
Live vision & Every editor depends on it & \ref{sec:experiments:transfer-structure} \\
Instruction & No measurable effect on the correction & \ref{sec:experiments:transfer-structure} \\
Pretraining & Halves critic loss; AUC rises only with bounds & \ref{sec:experiments:adaptation} \\
Source replay & Doubles L2 retention; L3 below base & \ref{sec:experiments:adaptation} \\
MoE vs.\ MLP actor & No effect established (16 tasks) & \ref{sec:experiments:design} \\
\bottomrule
\end{tabularx}
\end{table}

\paragraph{Rescue grows and harm falls with breadth.}
Widening L3 source training across 2, 16 and 112 tasks raises
rescue from 0.5\% through 15.8\% to 36.8\%. Harm falls from 22.8\% through 16.5\% to 10.2\%
(Figure~\ref{fig:correction-support}b).

Solving source tasks alone does not produce transfer: the two-task L3 editor
attains 93.50\% source SR yet 1.00\% held-out SR, and two-task L1 and L2
likewise fall below the base
(Appendix~\ref{app:sw-completions-twosource}). Breadth also brings source
contexts closer to held-out ones: nearest-source visual distance falls from
0.326 to 0.159 at matched early interaction budgets
(Figure~\ref{fig:correction-support}c).

\subsection{Where does transfer rescue vs.\ harm?}
\label{sec:experiments:transfer-structure}

\emph{Rescue is higher in source-covered scenes, harm follows correction
size, and soft-penalty editors depend on live context.}

\paragraph{Rescue is higher in source-covered scenes, with a wider gap for L2 and L3.}
Ten held-out tasks share scenes with source tasks while eight form a cohort
fixed before training from two scenes with no source task. Under L2, rescue reaches 42.0\% on
covered scenes vs.\ 26.8\% uncovered. Bounded L3 shows a similar
44.5\% vs.\ 27.3\% gap. Under the 16-task split, all six 16-task editors also rescue more in covered scenes
(28.0--43.6\% vs.\ 8.0--25.9\%; Table~\ref{tab:learner-classes}).

Full-action L1 transfers more evenly, with 42.4\% rescue in covered scenes
and 38.0\% in uncovered ones. Uncovered-scene harm is 3.3--3.4\% under the
soft-penalty recipes and 10.5\% under bounded L3. For the 16-task editors
on 96 further unseen tasks, the scene grouping separates rescue by at most
3.5 points, and the gain of the soft-penalty editors instead falls with
feature distance to source contexts (Appendix~\ref{app:sw-unseen96-distance}).

Source directions describe the bounded editor's held-out corrections best:
ten capture 85.3\% of their variance (Figure~\ref{fig:correction-support}d),
against 76.1\% for L2 and 68.8\% for L1
(Appendix~\ref{app:sw-corrsize-subspace}). L3's corrections are large:
69.8\% of its recorded gripper commands are clipped to the native range.

\begin{figure}[t]
\centering
\includegraphics[width=\linewidth]{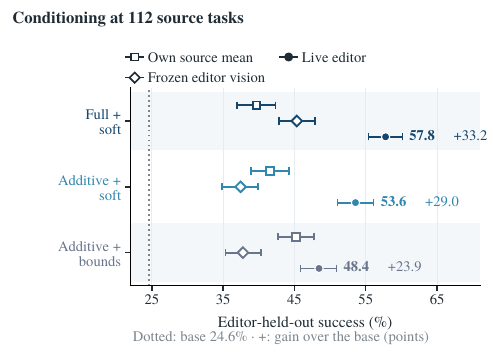}
\caption{\textbf{Live editing, static corrections, and frozen editor vision.}
Each row is a 112-task recipe in its color from Table~\ref{tab:editing-recipes},
on 18 held-out tasks (50 states, first pass); marker shape gives the condition.
Whiskers are pointwise 95\% within-task state-bootstrap intervals. The dotted
line marks the base; labels give each live editor's SR and its gain.}
\label{fig:libero-controls}
\end{figure}

\paragraph{A static correction explains only the bounded editor.}
First-pass controls test what each editor conditions on
(Figure~\ref{fig:libero-controls}). Replacing each editor's corrections by its own source-mean template reduces
L1 SR by 18.11 points [14.78, 21.22] and L2 SR by 12.00 points [8.56, 15.67]. The L3 difference is smaller:
3.22 points, with a paired 95\% interval spanning zero. A context-free
correction thus accounts for most of the bounded editor's gain, but not for
the soft-penalty editors'. All three editors also stay 7.22--16.56 points
above the base's native eight-action execution (41.22\% on this pass).

\paragraph{Editors use live vision, not the instruction.}
Freezing the editor's frames at step~0 lowers SR under every recipe relative
to live vision: by 12.44 points [9.22, 15.67] under L1, 16.11 [12.78, 19.34] under
L2, and 10.67 [7.33, 14.11] under L3. Replacing only the editor's
instruction changes SR by at most 1.67 points under any recipe, with every
interval including zero (Appendix~\ref{app:sw-completions-instruction}).

\paragraph{Smaller corrections lower harm under every recipe.}
Every reduced scale lowers harm (Table~\ref{tab:libero-controls}), and half
scale raises SR by 4.39, 3.39 and 8.94 points for L1, L2 and L3. At quarter
scale L1 and L2 fall below their unscaled SR, while L3 stays near its
half-scale SR. A separate L3 batch agrees; there, zeroing the gripper correction
instead lowers SR (Appendix~\ref{app:sw-corrsize-dose}).

A soft penalty in training reaches similar harm (4.9--6.2\%): both routes
tie harm to correction size.

\begin{table}[t]
\centering
\caption{\textbf{Scaling the correction at inference.} The executed chunk
is $B+\alpha(A-B)$. SR and harm share (H) on 18 held-out tasks: mean of two
900-pair batches for the 112-task editors (base SR 23.89\%), and a separate
L3 batch (base SR 23.56\%) whose last row removes the gripper correction.
Bold: best per column.}
\label{tab:libero-controls}
\footnotesize\setlength{\tabcolsep}{3.4pt}
\begin{tabularx}{\linewidth}{@{}>{\raggedright\arraybackslash}Xrrrrrrrr@{}}
\toprule
 & \multicolumn{2}{c}{L1} & \multicolumn{2}{c}{L2} & \multicolumn{2}{c}{L3} & \multicolumn{2}{c}{L3, sep.} \\
\cmidrule(lr){2-3} \cmidrule(lr){4-5} \cmidrule(lr){6-7} \cmidrule(l){8-9}
Scale $\alpha$ & SR & H & SR & H & SR & H & SR & H \\
\midrule
1 & 57.11 & 7.1 & 53.83 & 5.3 & 49.33 & 11.5 & 48.22 & 9.2 \\
0.75 & 60.39 & 5.6 & 56.78 & 4.2 & 56.83 & 7.6 & -- & -- \\
0.5 & \best{61.50} & \best{4.2} & \best{57.22} & \best{3.7} & \best{58.28} & 5.6 & \best{57.44} & \best{4.9} \\
0.25 & 51.94 & 4.3 & 51.56 & 3.8 & 56.33 & \best{5.4} & 56.56 & 5.3 \\
1, no grip. &  &  &  &  &  &  & 39.89 & 11.4 \\
 
\bottomrule
\end{tabularx}
\end{table}

\subsection{Which recipes preserve success and transfer?}
\label{sec:experiments:design}

\emph{Soft-penalty recipes harm less at both scales, and source-task success
does not predict the held-out ranking.}

Table~\ref{tab:learner-classes} compares all six recipes after 100k online
actor updates on 16 source tasks and the three online-encoder recipes after
481.25k updates on 112 source tasks, each on its source tasks and the same
18 held-out tasks.

\begin{table*}[tb]
\centering
\caption{\textbf{Recipe comparison at two source-task scales.}
R, K, H and F are the rescued, kept, harmed and still-failed shares of a cohort's paired episodes, in percent
(SR\,=\,R\,+\,K). Source/held-out base SR is 22.06/23.83\% for 16-task runs and
19.09/23.33\% for 112-task runs. Covered/uncovered split the held-out pairs
by whether a task's scene has a source task (7/11 and 10/8 tasks). Actors are MLPs except in the bottom block,
which compares actor routing by source SR. ``Base, repeated run'' scores a
second base run against a first (mean of three pairings). Bold/underlining:
best/second-best within a block; shading: highest held-out SR.}
\label{tab:learner-classes}
\footnotesize\setlength{\tabcolsep}{2.5pt}\renewcommand{\arraystretch}{1.10}
\newcommand{\hd}[1]{\makebox[0pt][r]{#1}}%
\begin{tabularx}{\textwidth}{@{}l>{\raggedright\arraybackslash}X*{14}{>{\raggedleft\arraybackslash}p{21.5pt}}@{}}
\toprule
 & & \multicolumn{5}{c}{Source tasks} & \multicolumn{5}{c}{Held-out tasks (18)} & \multicolumn{2}{c}{Covered} & \multicolumn{2}{c}{Uncovered} \\
\cmidrule(lr){3-7} \cmidrule(lr){8-12} \cmidrule(lr){13-14} \cmidrule(l){15-16}
ID & Editing choice & \hd{SR\,$\uparrow$} & \hd{R\,$\uparrow$} & K & \hd{H\,$\downarrow$} & F & \hd{SR\,$\uparrow$} & \hd{R\,$\uparrow$} & K & \hd{H\,$\downarrow$} & F & \hd{R\,$\uparrow$} & \hd{H\,$\downarrow$} & \hd{R\,$\uparrow$} & \hd{H\,$\downarrow$} \\
\midrule
\multicolumn{16}{@{}l}{\emph{16 source tasks; 100k online actor updates}} \\
R1 & \swatch{recfull}~Full action + soft penalty & 61.75 & 42.4 & 19.3 & 2.8 & 35.5 & 40.28 & 26.4 & 13.8 & 10.0 & 49.7 & 36.7 & 12.1 & \second{19.9} & 8.6 \\
R2 & \swatch{recadd}~Additive + soft penalty & 77.00 & 56.0 & 21.0 & 1.1 & 21.9 & \second{41.06} & \second{27.6} & 13.5 & 10.3 & 48.6 & 30.1 & 13.6 & \best{25.9} & 8.3 \\
R3 & \swatch{recbnd}~Additive + output bounds & 56.88 & 39.7 & 17.2 & 4.9 & 38.3 & 39.22 & 24.1 & 15.1 & 8.7 & 52.1 & 38.4 & 10.9 & 15.0 & \second{7.4} \\
\midrule
L1 & \swatch{recfull}~Full action + soft penalty & \best{93.00} & \best{71.9} & 21.1 & \second{1.0} & 6.0 & 40.28 & 24.9 & 15.4 & \second{8.4} & 51.3 & \best{43.6} & \best{7.0} & 13.0 & 9.4 \\
\hlrow
L2 & \swatch{recadd}~Additive + soft penalty & \second{91.63} & \second{70.1} & 21.5 & \best{0.6} & 7.8 & \best{45.06} & \best{27.9} & 17.1 & \best{6.7} & 48.2 & \second{42.6} & \second{10.7} & 18.6 & \best{4.2} \\
L3 & \swatch{recbnd}~Additive + output bounds & 85.63 & 65.1 & 20.6 & 1.5 & 12.9 & 23.11 & 15.8 & 7.3 & 16.5 & 60.4 & 28.0 & 15.7 & 8.0 & 17.0 \\
\midrule
\multicolumn{16}{@{}l}{\emph{112 source tasks; 481.25k online actor updates}} \\
\hlrow
L1 & \swatch{recfull}~Full action + soft penalty & \best{94.72} & \best{75.9} & 18.8 & \best{0.3} & 5.0 & \best{57.56} & \best{40.4} & 17.1 & \second{6.2} & 36.2 & \second{42.4} & \second{8.6} & \best{38.0} & \best{3.3} \\
L2 & \swatch{recadd}~Additive + soft penalty & \second{94.70} & \best{75.9} & 18.8 & \best{0.3} & 5.0 & \second{53.67} & 35.2 & 18.4 & \best{4.9} & 41.4 & 42.0 & \best{6.1} & 26.8 & \second{3.4} \\
L3 & \swatch{recbnd}~Additive + output bounds & 93.55 & 75.0 & 18.6 & 0.5 & 5.9 & 50.00 & \second{36.8} & 13.2 & 10.2 & 39.8 & \best{44.5} & 9.9 & \second{27.3} & 10.5 \\
\addlinespace[1.5pt]
 & Base, repeated run &  &  &  &  &  & 23.57 & 2.2 & 21.4 & 2.0 & 74.4 & 3.8 & 3.5 & 0.1 & 0.1 \\
\midrule
\multicolumn{2}{@{}l}{\emph{Actor routing, 16 tasks (SR\,$\uparrow$)}} & MLP & MoE & \hd{+\,sup.} &  &  &  &  &  &  &  &  &  &  & \\
R2 & \swatch{recadd}~Additive + soft penalty & 77.00 & 75.75 & 68.31 &  &  &  &  &  &  &  &  &  &  & \\
L2 & \swatch{recadd}~Additive + soft penalty & 91.63 & 90.25 & \best{93.75} &  &  &  &  &  &  &  &  &  &  & \\
\bottomrule
\end{tabularx}
\end{table*}

\paragraph{Source-task success does not predict transfer.}
Five 16-task recipes achieve 39.22--45.06\% held-out SR (Table~\ref{tab:learner-classes}).
Across all six runs, source SR spans 56.88--93.00\%, but its ordering does not
carry over to held-out SR.

Bounded L3 stays at the base level on these tasks (23.11\%) despite higher
source scores than all readout runs, and it is below both soft-penalty
encoder recipes in each of three training seeds
(Appendix~\ref{app:sw-robustness-seeds16}). On 96 further unseen tasks L3
also gains least, 9.44 points against 21.79 and 21.56 for L1 and L2
(Appendix~\ref{app:sw-unseen96}). L2 has the highest held-out SR among the
reported 16-task runs (45.06\%), but L1 and L2 are not separated across
seeds.

The online encoder leads the readout by 14.63--31.25 points on the 16 source
tasks under every editing choice. That lead shrinks on held-out tasks: L1 and
R1 are level (0.0 [$-$2.8, 2.9]) and L2 leads R2 by 4.0 [1.1, 6.9].

\paragraph{Soft penalties retain base successes.}
On 16 source tasks, additive soft-penalty recipes incur 0.6--1.1\% harm,
compared with 1.5--4.9\% for bounded recipes. On a fixed panel of observations, bounded
recipes place an output coordinate within 10\% of its bound on almost every decision and move the
gripper 50--110$\times$ as much (Appendix~\ref{app:six-recipes}). At 112 source tasks the same ordering holds
for harm on held-out tasks (4.9--6.2\% vs.\ 10.2\%; Table~\ref{tab:learner-classes}).

\paragraph{Matched cameras remove the encoder's held-out advantage.}
Removing wrist images lowers 16-task L2 source SR from 91.63\% to 79.94\%
while its held-out SR falls less, to 40.56\%. Over the base, main-camera L1 gains
10.67 points on held-out tasks and main-camera L3 gains 2.78. With matched cameras,
no 16-task encoder exceeds its readout counterpart on held-out tasks
(Appendix~\ref{app:sw-completions-cameras}).

\paragraph{Expert routing matches a shared MLP on source tasks.}
On 16 source tasks, a four-expert top-two MoE matches the shared MLP within 1.4
points (Table~\ref{tab:learner-classes}). With task-cluster supervision L2-MoE reaches 93.75\% and the readout MoE
falls by 7.44 points; single runs do not establish an effect
(Appendix~\ref{app:sw-completions-moe}). Task
gradients of the 112-task MLP show little cancellation under all three
recipes (0.98--1.07$\times$ the orthogonal reference;
Appendix~\ref{app:libero-gradients}).

\subsection{Do pretraining and replay help adaptation?}
\label{sec:experiments:adaptation}

\emph{Pretraining halves critic loss under every recipe and raises AUC only with
output bounds. Source SR stays above the base only with source replay, and
then only under soft-penalty L2, not under bounded L3.}

\begin{figure}[t]
    \centering
    \includegraphics[width=0.84\columnwidth]{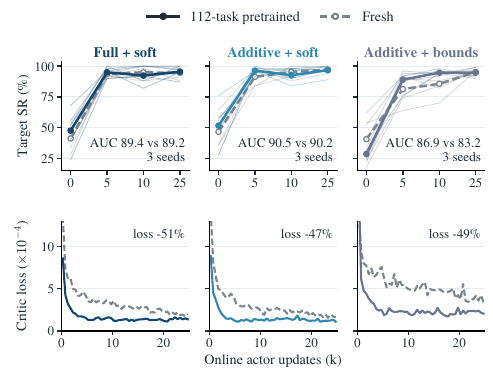}
    \caption{\textbf{Single-task adaptation from pretrained and fresh editors.}
    Mean over two targets and three seeds per recipe. Top: target SR (thin
    lines: single runs) and AUC, pretrained vs.\ fresh. Bottom: critic loss
    (500-update bins) and its mean change.}
    \label{fig:libero-adaptation}
\end{figure}

We initialize each recipe from its 112-task snapshot or fresh weights and
adapt without source replay for 25k online actor updates
(Figure~\ref{fig:libero-adaptation}), with three seeds per condition on two
targets, bowl-onto-plate and wine-onto-cabinet placement. AUC is taken over
the four checkpoints.

\paragraph{Pretraining halves critic loss; AUC rises only for the bounded recipe.}
Pretraining lowers mean online critic loss by 47--51\% under every recipe,
and in all 18 run pairs matched on recipe, target and seed. For L3, mean AUC
is higher under pretraining on both targets (86.93 vs.\ 83.15) and in four of
the six seed-matched pairs; the gain is early, 94.67\% vs.\ 85.67\% at 10k
updates. For L1 and L2 the two initializations are level (mean AUC differs by 0.13
and 0.28 points), and a third target
(Task~39) shows the same critic-loss drop and rapid convergence
(Appendix~\ref{app:sw-adaptation}).

\paragraph{Source replay doubles L2's retained source SR; L3 stays below the base.}
Without source replay, source SR after adaptation is 22.71\% (L1), 24.58\%
(L2) and 5.42\% (L3) on 16 source tasks. Mixing each batch half-and-half with
frozen replay of those tasks, while collecting only on the target, raises it
to 54.17\% for L2 (base 24.38\%) and to 20.63\% for L3, still below the base.
Target SR at 25k is 98.00\% and 95.67\%, not lower than without replay
(Appendix~\ref{app:sw-adaptation-replay}).

\subsection{Does the narrow-source pattern recur beyond LIBERO?}
\label{sec:experiments:robotwin}

\emph{Yes: with a different base and simulator, a four-task editor improves
its source tasks through state-dependent routing and lowers success outside
them, at full and at half correction size.}

RoboTwin~\citep{chen2025robotwin2} changes the base (frozen
LingBot-VLA~\citep{wu2026lingbotvla}), simulator and embodiment (two arms,
joint coordinates); its simulation throughput limited source training to
four tasks (S4). The editor is a four-expert top-two MoE (RL-token readout,
full action, soft penalty) with task-contrastive
supervision~\citep{lin2026dygrovla}.

\begin{figure}[t]
    \centering
    \includegraphics[width=\columnwidth]{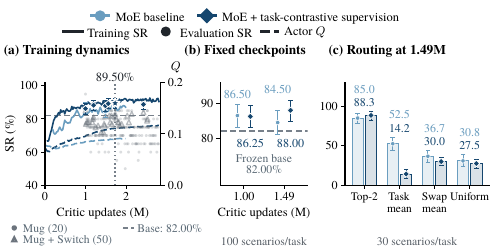}
    \caption{\textbf{RoboTwin S4 training, checkpoints, and routing.}
    (a) Rolling 100-episode SR and $Q$ (star: 89.50\% at 1.73M); (b) matched
    100-scene checks (base 82.00\%); (c) 1.49M routing
    interventions (95\% CIs).}
    \label{fig:s4-extended-training}
\end{figure}

\paragraph{The editor improves its source tasks.}
On 200 paired scenes per task it reaches 84.13\% against 74.63\% for the
base (+9.50 points; Figure~\ref{fig:s4-extended-training}a,b). Rescue is
15.0\% and harm 5.5\%.

\paragraph{The gain relies on state-dependent routing.}
Replacing dynamic gating with fixed task-mean mixtures lowers SR from
88.33\% to 14.17\% (Figure~\ref{fig:s4-extended-training}c). For an editor
trained without task supervision it falls from 85.00\% to 52.50\%. Routing
follows state within a task: its association with phase labels exceeds
circular-shift nulls in all six tested task--model pairs.

\paragraph{Outside its source tasks the editor lowers success.}
On 12 other tasks without updates, SR falls from 71.00\% to 25.17\% (harm
50.0\%), as with two-task editors on LIBERO. Harm is 6--24\% on three
press and click tasks and 34--98\% on the nine others. In a matched
600-scene follow-up the editor without task supervision also falls below
the base (30.50\%), and the two editors lose the same 233 base successes.
Halving the correction there lowers harm from 48.3\% to 40.5\% while transfer
stays negative (Appendix~\ref{app:s4-followup}).

\Needspace{5\baselineskip}
\section{Conclusion}
\label{sec:conclusion}
\label{sec:limitations}

Shared action editing transfers across tasks as rescue minus harm: three
112-task editors raise 18-task held-out SR from 23.3\% to 50.0--57.6\% with
no update on those tasks. Four findings follow. \emph{Transfer grows with
source breadth}, and source success does not predict it. \emph{Gains
concentrate where the base fails}, while harm arises on tasks the base solves
at least half the time. \emph{Harm follows correction size}: it is lower under soft
penalties and after shrinking a correction at inference.
\emph{A fixed offset explains the bounded editor's transfer}; soft-penalty
editors depend on live context. With matched cameras, online encoders hold no held-out advantage over the
fixed readout. In single-task adaptation, pretraining halves
critic loss under every recipe; source replay more than doubles the source
success retained under the additive soft penalty without lowering target
success, while the bounded recipe stays below the base. With a different base and
simulator, a four-task editor repeats the narrow-source pattern of two-task
LIBERO editors: it improves its source tasks and lowers success on 12
others. ReSOP trains the 112-task bounded editor in 167~GPU-hours.

\textbf{Limitations.}
All experiments are in simulation. The growth of transfer with source
breadth is shown on LIBERO only; on RoboTwin, editors are trained on four
source tasks. The 481k-update endpoints use one training run per recipe; three-seed runs at
121k updates show that fine recipe rankings vary with seed and training
length. Editors remain active at every step on a fixed source-task set.
Adaptation with source replay covers two recipes and two targets, replays 16
source tasks, and retains source skill only in part.

\clearpage
\section*{Acknowledgments}
This work was supported by Zhongguancun Academy (Project No.~C20250505).
Computing resources were provided by Chunhua Shen's group at Zhejiang
University and by the Zenith computing platform of Zhongguancun Academy.

{\small
\setlength{\bibsep}{0pt plus 1pt}

}
\newcommand{\AppendixInOnePDF}{}
\ifdefined\AppendixInOnePDF
  \clearpage
  \twocolumn[{\centering\Large\bfseries Appendix\par\vspace{1.2em}}]
\else
  \maketitleappendix
\fi
\appendix
\setcounter{table}{0}
\setcounter{figure}{0}
\setcounter{equation}{0}
\renewcommand{\thetable}{S\arabic{table}}
\renewcommand{\thefigure}{S\arabic{figure}}
\renewcommand{\theequation}{S\arabic{equation}}
\renewcommand{\theHtable}{supp.\arabic{table}}
\renewcommand{\theHfigure}{supp.\arabic{figure}}
\renewcommand{\theHequation}{supp.\arabic{equation}}
\makeatletter\providecommand{\swinput}[1]{\input{#1} }\makeatother
\section{Overview of the Appendix}
\label{app:sw-guide}

\noindent\textbf{What is here.}
Appendices~\ref{app:sw-robustness}--\ref{app:sw-completions} take up
questions that the main text answers for one recipe, one training seed or
one cohort, all on LIBERO. Does editor harm exceed what repeating an
evaluation produces, and does the recipe comparison hold across training
seeds (Appendix~\ref{app:sw-robustness})? Does transfer extend beyond the 18
held-out tasks (Appendix~\ref{app:sw-unseen96})? Do correction size and
correction structure behave alike for all three recipes
(Appendix~\ref{app:sw-corrsize})? How do the comparisons that the main text
details for one recipe, including single-task adaptation, look for the others
(Appendix~\ref{app:sw-completions})? Table~\ref{tab:sw-guide-questions}
answers each question in one line. Every evaluation in these appendices
uses the checkpoints and configurations of the main text, and every added
training run repeats one of those configurations with a different seed, or
with the source tasks, camera input or replay mixture stated in its section.
Appendix~\ref{app:sw-system} describes the training system, its cost and
later engineering checks. Appendices~\ref{app:method}--\ref{app:system}
then follow the order of the main text: learners and runtime, evaluation
protocols, the 112-task and 16-task LIBERO studies, RoboTwin S4, and the
system measurements (Table~\ref{tab:supp-roadmap}).
Appendix~\ref{app:sw-videos} describes the video replays shown on the
project page.

Recipe names follow Table~\ref{tab:editing-recipes}: R and L denote the
RL-token readout and the online-learned encoder; 1, 2, and 3 denote full
action with a soft penalty, additive correction with a soft penalty, and
additive correction with output bounds. A held-out task lies in a
source-covered scene of an editor if its scene contains one of that
editor's source tasks, and in an uncovered scene otherwise. In the tables
of this appendix, unless a caption says otherwise, bold marks the best and
underlining the second-best value among the entries compared, and light
shading marks the row discussed in the text or a summary row. These marks
order point estimates; they do not indicate a tested difference.

\paragraph{How rescue and harm are counted.}
Each of the $N$ matched base and editor episodes in a cohort is rescued
(base fails, editor succeeds), kept (both succeed), harmed (base succeeds,
editor fails), or still failed. As in the main text, the four are reported
as shares of $N$, so that $r+k+h+f=1$, the editor SR is $r+k$, the base SR
is $b=k+h$, and the SR gain is $r-h$. The same counts can be expressed as
conditional rates: rescue among base failures, $r/(1-b)$, and harm among
base successes, $h/b$. For example, the 112-task L2 editor harms 88 of the
1,800 held-out pairs. That is 4.9\% of all pairs, or 21.0\% of the 420
base successes. Table~\ref{tab:outcome-shares-rates} lists both forms for
the main comparisons.

\paragraph{Which base each number is paired with.}
Every editor SR is paired with a base run from its own evaluation batch on
the same tasks, initial states, and passes. The base SR on the 18 held-out
tasks therefore differs slightly between evaluations in the main text:
23.33\% over two passes, 24.56\% on the first pass alone, and 23.56\% in the
separate batch of the scaling study.
Appendix~\ref{app:sw-robustness} measures this variation directly: three
repeated base runs reach 23.22\%, 23.44\%, and 23.83\%. Across the
evaluation batches of Appendices~\ref{app:sw-robustness}--\ref{app:sw-adaptation}
the base SR on these tasks lies between 22.89\% and 25.22\%, and each table
states the base SR of its batch. A checkpoint that is evaluated in two
batches therefore appears with two slightly different SRs.

\begingroup
\newcommand{\swrBaseNullHarm}{1.8--2.3}
\newcommand{\swrBaseNullRescue}{1.9--2.5}
\newcommand{\swrBaseNullDifferCov}{6.8--7.8}
\newcommand{\swrBaseNullDifferUnc}{0.1--0.4}
\newcommand{\swrBaseNullHarmUncMax}{0.3}
\newcommand{\swrHarmLOne}{6.4--6.7}
\newcommand{\swrRepDeltaOne}{1.00}
\newcommand{\swrHarmLTwo}{4.8--5.7}
\newcommand{\swrRepDeltaTwo}{0.89}
\newcommand{\swrHarmLThree}{10.2--10.9}
\newcommand{\swrRepDeltaThree}{0.33}
\newcommand{\swrSoftHarmUnc}{3.0--3.1}
\newcommand{\swrBoundHarmUnc}{9.8--10.1}
\newcommand{\swrMinRecipeGap}{2.56}
\newcommand{\swrPaperReproMax}{0.83}
\newcommand{\swrMaxBaseDelta}{0.61}
\newcommand{\swrBaseSrFirstPass}{23.78--24.89}
\newcommand{\swrBaseLenDiffer}{7.7--8.3}
\newcommand{\swrTwoTaskSame}{24--25}
\newcommand{\swrTwoTaskCross}{23--29}
\newcommand{\swrHotTasks}{four}
\newcommand{\swrHotShare}{220 of the 226}
\newcommand{\swrOtherTasks}{14}
\newcommand{\swrSxBoundSr}{20.00--27.44}
\newcommand{\swrSxSoftSr}{39.67--47.33}
\newcommand{\swrSxBoundHarm}{14.7--16.8}
\newcommand{\swrSxSoftHarm}{5.4--9.4}
\newcommand{\swrSxLtwoMinusLone}{+0.69}
\newcommand{\swrSxLtwoMinusLoneLo}{$-$2.94}
\newcommand{\swrSxLtwoMinusLoneHi}{+2.89}
\newcommand{\swrSxSdLone}{4.16}
\newcommand{\swrSxSdLtwo}{1.34}
\newcommand{\swrSxBoundSource}{85.63--89.19}
\newcommand{\swrSxSoftSource}{88.69--93.00}
\newcommand{\swrSxSeedZeroHere}{40.72, 43.61 and 24.00}
\newcommand{\swrSxSeedZeroPaper}{40.28, 45.06 and 23.11}
\newcommand{\swrHxBaseUnc}{24.50}
\newcommand{\swrHxHarmZeroOne}{10.6}
\newcommand{\swrHxSeedDiffOne}{+4.39}
\newcommand{\swrHxSeedCiOne}{[+1.56, +7.11]}
\newcommand{\swrHxSeedDiffTwo}{$-$0.11}
\newcommand{\swrHxSeedCiTwo}{[$-$2.94, +2.78]}
\newcommand{\swrHxHarmZeroThree}{8.6}
\newcommand{\swrHxSeedDiffThree}{$-$13.44}
\newcommand{\swrHxSeedAbsThree}{13.44}
\newcommand{\swrHxSeedCiThree}{[$-$16.33, $-$10.61]}
\newcommand{\swrHxSourceOneThree}{76.43}
\newcommand{\swrHxUncOneThree}{20.00}
\newcommand{\swrHxTwoMinusThreeOne}{23.89}
\newcommand{\swrHxTwoMinusThreeOneCi}{[+21.00, +26.89]}
\newcommand{\swrHxTwoMinusOneOne}{6.44}
\newcommand{\swrHxTwoMinusOneOneCi}{[+3.50, +9.44]}
\newcommand{\swrHxTwoMinusThreeZero}{10.56}
\newcommand{\swrHxTwoMinusThreeZeroCi}{[+7.72, +13.39]}
\newcommand{\swrHxTwoMinusOneZero}{10.94}
\newcommand{\swrHxTwoMinusOneZeroCi}{[+8.22, +13.61]}
\newcommand{\swrHxOneMinusThreeZero}{$-$0.39}
\newcommand{\swrHxOneMinusThreeZeroCi}{[$-$3.28, +2.61]}
\newcommand{\swrThreeBase}{23.17}
\newcommand{\swrThreeBaseCov}{22.30}
\newcommand{\swrThreeBaseUnc}{24.25}
\newcommand{\swrThreeBaseSrc}{20.71}
\newcommand{\swrThreeSrLoneZero}{49.94}
\newcommand{\swrThreeSrLoneOne}{50.78}
\newcommand{\swrThreeSrLoneTwo}{61.28}
\newcommand{\swrThreeSrLtwoZero}{59.72}
\newcommand{\swrThreeSrLtwoOne}{58.11}
\newcommand{\swrThreeSrLtwoTwo}{60.28}
\newcommand{\swrThreeSrLthreeZero}{46.78}
\newcommand{\swrThreeSrLthreeOne}{34.83}
\newcommand{\swrThreeSrLthreeTwo}{46.44}
\newcommand{\swrThreeDiffLoneOne}{+0.83}
\newcommand{\swrThreeCiLoneOne}{[$-$1.95, +3.61]}
\newcommand{\swrThreeDiffLoneTwo}{+11.33}
\newcommand{\swrThreeCiLoneTwo}{[+8.67, +13.94]}
\newcommand{\swrThreeDiffLtwoOne}{$-$1.61}
\newcommand{\swrThreeCiLtwoOne}{[$-$4.44, +1.11]}
\newcommand{\swrThreeDiffLtwoTwo}{+0.56}
\newcommand{\swrThreeCiLtwoTwo}{[$-$2.11, +3.28]}
\newcommand{\swrThreeDiffLthreeOne}{$-$11.94}
\newcommand{\swrThreeAbsLthreeOne}{11.94}
\newcommand{\swrThreeCiLthreeOne}{[$-$14.89, $-$9.00]}
\newcommand{\swrThreeDiffLthreeTwo}{$-$0.33}
\newcommand{\swrThreeCiLthreeTwo}{[$-$3.22, +2.67]}
\newcommand{\swrThreeDiffLthreeTwoVsOne}{+11.61}
\newcommand{\swrThreeCiLthreeTwoVsOne}{[+8.78, +14.44]}
\newcommand{\swrThreeTwoMinusThreeZero}{+12.94}
\newcommand{\swrThreeTwoMinusThreeZeroCi}{[+10.00, +15.83]}
\newcommand{\swrThreeTwoMinusThreeOne}{+23.28}
\newcommand{\swrThreeTwoMinusThreeOneCi}{[+20.50, +26.06]}
\newcommand{\swrThreeTwoMinusThreeTwo}{+13.83}
\newcommand{\swrThreeTwoMinusThreeTwoCi}{[+11.11, +16.61]}
\newcommand{\swrThreeOneMinusThreeZero}{+3.17}
\newcommand{\swrThreeOneMinusThreeZeroCi}{[+0.11, +6.22]}
\newcommand{\swrThreeOneMinusThreeOne}{+15.94}
\newcommand{\swrThreeOneMinusThreeOneCi}{[+13.11, +18.83]}
\newcommand{\swrThreeOneMinusThreeTwo}{+14.83}
\newcommand{\swrThreeOneMinusThreeTwoCi}{[+12.11, +17.56]}
\newcommand{\swrThreeTwoMinusOneZero}{+9.78}
\newcommand{\swrThreeTwoMinusOneZeroCi}{[+7.11, +12.44]}
\newcommand{\swrThreeTwoMinusOneOne}{+7.33}
\newcommand{\swrThreeTwoMinusOneOneCi}{[+4.28, +10.22]}
\newcommand{\swrThreeTwoMinusOneTwo}{$-$1.00}
\newcommand{\swrThreeTwoMinusOneTwoCi}{[$-$3.61, +1.61]}
\newcommand{\swrThreeSrcTwoLone}{85.00}
\newcommand{\swrThreeSrcTwoLtwo}{84.46}
\newcommand{\swrThreeSrcTwoLthree}{78.84}
\newcommand{\swrThreeUncOneLthree}{19.75}
\newcommand{\swrThreeUncTwoLthree}{39.63}
\newcommand{\swrThreeHalfLthreeTwo}{39.83}
\newcommand{\swrThreeHalfLoneTwo}{46.33}
\newcommand{\swrThreeHalfLtwoTwo}{55.28}
\newcommand{\UnsPairs}{4,800}
\newcommand{\UnsBase}{19.15}
\newcommand{\UnsHeldBase}{24.89}
\newcommand{\UnsSRLa}{40.94}
\newcommand{\UnsGainLa}{+21.79}
\newcommand{\UnsGainCILa}{[20.54, 23.04]}
\newcommand{\UnsHeldSRLa}{40.56}
\newcommand{\UnsHeldGainLa}{+15.67}
\newcommand{\UnsHeldGainCILa}{[12.56, 18.78]}
\newcommand{\UnsLowGainLa}{+23.36}
\newcommand{\UnsLowHarmLa}{1.9}
\newcommand{\UnsHighGainLa}{+6.67}
\newcommand{\UnsHighGainCILa}{[2.00, 11.33]}
\newcommand{\UnsHighHarmLa}{13.8}
\newcommand{\UnsGoalGainLa}{$-$11.33}
\newcommand{\UnsGoalGainCILa}{[$-$15.67, $-$6.67]}
\newcommand{\UnsNegLa}{23}
\newcommand{\UnsPosLa}{65}
\newcommand{\UnsRhoLa}{$-$0.37}
\newcommand{\UnsRhoCILa}{[$-$0.54, $-$0.18]}
\newcommand{\UnsRhoHarmLa}{0.36}
\newcommand{\UnsNearLa}{+34.44}
\newcommand{\UnsFarLa}{+6.00}
\newcommand{\UnsCovRescueLa}{30.5}
\newcommand{\UnsUncRescueLa}{27.0}
\newcommand{\UnsSRLb}{40.71}
\newcommand{\UnsGainLb}{+21.56}
\newcommand{\UnsGainCILb}{[20.25, 22.94]}
\newcommand{\UnsHeldSRLb}{45.78}
\newcommand{\UnsHeldGainLb}{+20.89}
\newcommand{\UnsHeldGainCILb}{[17.67, 24.11]}
\newcommand{\UnsLowGainLb}{+23.32}
\newcommand{\UnsLowHarmLb}{2.2}
\newcommand{\UnsHighGainLb}{+6.67}
\newcommand{\UnsHighGainCILb}{[2.00, 11.56]}
\newcommand{\UnsHighHarmLb}{12.2}
\newcommand{\UnsGoalGainLb}{$-$4.67}
\newcommand{\UnsGoalGainCILb}{[$-$9.67, +0.33]}
\newcommand{\UnsNegLb}{20}
\newcommand{\UnsPosLb}{62}
\newcommand{\UnsRhoLb}{$-$0.33}
\newcommand{\UnsRhoCILb}{[$-$0.52, $-$0.12]}
\newcommand{\UnsRhoHarmLb}{0.32}
\newcommand{\UnsNearLb}{+31.94}
\newcommand{\UnsFarLb}{+11.94}
\newcommand{\UnsCovRescueLb}{29.6}
\newcommand{\UnsUncRescueLb}{27.6}
\newcommand{\UnsSRLc}{28.58}
\newcommand{\UnsGainLc}{+9.44}
\newcommand{\UnsGainCILc}{[8.19, 10.69]}
\newcommand{\UnsHeldSRLc}{23.89}
\newcommand{\UnsHeldGainLc}{$-$1.00}
\newcommand{\UnsHeldGainCILc}{[$-$3.89, +2.00]}
\newcommand{\UnsLowGainLc}{+17.74}
\newcommand{\UnsLowHarmLc}{2.5}
\newcommand{\UnsHighGainLc}{$-$20.89}
\newcommand{\UnsHighGainCILc}{[$-$25.56, $-$16.22]}
\newcommand{\UnsHighHarmLc}{31.6}
\newcommand{\UnsGoalGainLc}{$-$16.67}
\newcommand{\UnsGoalGainCILc}{[$-$22.00, $-$11.33]}
\newcommand{\UnsNegLc}{37}
\newcommand{\UnsPosLc}{48}
\newcommand{\UnsRhoLc}{$-$0.18}
\newcommand{\UnsRhoCILc}{[$-$0.37, +0.03]}
\newcommand{\UnsRhoHarmLc}{0.20}
\newcommand{\UnsNearLc}{+20.81}
\newcommand{\UnsFarLc}{+6.31}
\newcommand{\UnsCovRescueLc}{20.4}
\newcommand{\UnsUncRescueLc}{20.3}
\newcommand{\UnsHighBase}{71.33}
\newcommand{\UnsHighTasks}{9}
\newcommand{\UnsLowTasks}{47}
\newcommand{\UnsLowHarmMax}{2.5}
\newcommand{\UnsGoalBase}{51.67}
\newcommand{\UnsDiffLbLa}{$-$0.23}
\newcommand{\UnsDiffLbLaCI}{[$-$1.52, +1.10]}
\newcommand{\UnsDiffLcLb}{$-$12.13}
\newcommand{\UnsDiffLcLbCI}{[$-$13.46, $-$10.83]}
\newcommand{\UnsCovTasks}{34}
\newcommand{\UnsUncTasks}{62}
\newcommand{\UnsCovBase}{27.12}
\newcommand{\UnsUncBase}{14.77}
\newcommand{\UnsRhoBaseMax}{0.05}
\newcommand{\UnsHeldHighTasks}{5}
\newcommand{\UnsSingleSceneTasks}{18}
\newcommand{\UnsDevTasks}{25}
\newcommand{\SWcsBaseOne}{23.78}
\newcommand{\SWcsBaseTwo}{24.00}
\newcommand{\SWcsHalfPooledLone}{+4.39}
\newcommand{\SWcsHalfPooledCILone}{[+1.67, +7.11]}
\newcommand{\SWcsHarmDropLone}{2.9}
\newcommand{\SWcsRescueChangeLone}{+1.5}
\newcommand{\SWcsSRLoneFullOne}{57.56}
\newcommand{\SWcsSRLoneHalfOne}{61.89}
\newcommand{\SWcsSRLoneQuarterOne}{51.78}
\newcommand{\SWcsSRLoneFullTwo}{56.67}
\newcommand{\SWcsSRLoneHalfTwo}{61.11}
\newcommand{\SWcsSRLoneQuarterTwo}{52.11}
\newcommand{\SWcsHalfPooledLtwo}{+3.39}
\newcommand{\SWcsHalfPooledCILtwo}{[+0.50, +6.28]}
\newcommand{\SWcsHarmDropLtwo}{1.6}
\newcommand{\SWcsRescueChangeLtwo}{+1.8}
\newcommand{\SWcsSRLtwoFullOne}{54.11}
\newcommand{\SWcsSRLtwoHalfOne}{57.22}
\newcommand{\SWcsSRLtwoQuarterOne}{50.67}
\newcommand{\SWcsSRLtwoFullTwo}{53.56}
\newcommand{\SWcsSRLtwoHalfTwo}{57.22}
\newcommand{\SWcsSRLtwoQuarterTwo}{52.44}
\newcommand{\SWcsHalfPooledLthree}{+8.94}
\newcommand{\SWcsHalfPooledCILthree}{[+6.17, +11.67]}
\newcommand{\SWcsHarmDropLthree}{5.9}
\newcommand{\SWcsRescueChangeLthree}{+3.0}
\newcommand{\SWcsSRLthreeFullOne}{49.22}
\newcommand{\SWcsSRLthreeHalfOne}{58.22}
\newcommand{\SWcsSRLthreeQuarterOne}{56.89}
\newcommand{\SWcsSRLthreeFullTwo}{49.44}
\newcommand{\SWcsSRLthreeHalfTwo}{58.33}
\newcommand{\SWcsSRLthreeQuarterTwo}{55.78}
\newcommand{\SWcsHalfLtwoOneCI}{[$-$0.22, +6.44]}
\newcommand{\SWcsQuarterLoneOne}{$-$5.78 [$-$9.11, $-$2.44]}
\newcommand{\SWcsQuarterLoneTwo}{$-$4.56 [$-$7.78, $-$1.33]}
\newcommand{\SWcsLthreeQuarterGap}{2.56}
\newcommand{\SWcsArmNormLone}{0.075}
\newcommand{\SWcsArmNormLtwo}{0.074}
\newcommand{\SWcsArmNormLthree}{0.150}
\newcommand{\SWcsMainBaseMean}{23.89}
\newcommand{\SWcsSubTenLone}{68.8}
\newcommand{\SWcsSubNeedLone}{19}
\newcommand{\SWcsSubBlockLone}{56.4}
\newcommand{\SWcsSubTenLtwo}{76.1}
\newcommand{\SWcsSubNeedLtwo}{15}
\newcommand{\SWcsSubBlockLtwo}{65.1}
\newcommand{\SWcsSubTenLthree}{85.3}
\newcommand{\SWcsSubNeedLthree}{10}
\newcommand{\SWcsSubBlockLthree}{76.2}
\newcommand{\SWcsSubRawLthree}{99.4}
\newcommand{\SWcsSubSrcTenLone}{72.3}
\newcommand{\SWcsSubNsrcLone}{10,963}
\newcommand{\SWcsSubNheldLone}{57,446}
\newcommand{\SWcsSubSrcTenLtwo}{77.0}
\newcommand{\SWcsSubNsrcLtwo}{10,285}
\newcommand{\SWcsSubNheldLtwo}{58,478}
\newcommand{\SWcsSubSrcTenLthree}{88.9}
\newcommand{\SWcsSubNsrcLthree}{8,705}
\newcommand{\SWcsSubNheldLthree}{58,728}
\newcommand{\SWcsAucArmLone}{0.82}
\newcommand{\SWcsAucArmLtwo}{0.83}
\newcommand{\SWcsAucGripLone}{0.80}
\newcommand{\SWcsAucGripLtwo}{0.89}
\newcommand{\SWcsAucArmLthree}{0.62}
\newcommand{\SWcsAucGripLthree}{0.52}
\newcommand{\SWcsAucDistLthree}{0.75}
\newcommand{\SWcsAucSnapLthree}{0.75}
\newcommand{\SWcsAucDistMaxLone}{0.64}
\newcommand{\SWcsAucDistMaxLtwo}{0.69}
\newcommand{\SWcsAucDistMeanLone}{0.51}
\newcommand{\SWcsAucDistMeanLtwo}{0.61}
\newcommand{\SWcsAucRescueLow}{0.28}
\newcommand{\SWcsAucRescueHigh}{0.38}
\newcommand{\SWcsNBaseSucc}{214}
\newcommand{\SWcsNBaseFail}{686}
\newcommand{\SWcsGateAlphaLow}{0.87}
\newcommand{\SWcsGateAlphaHigh}{0.94}
\newcommand{\SWcsGateCapLone}{3.67}
\newcommand{\SWcsGateCapLoneCI}{[+0.89, +6.33]}
\newcommand{\SWcsGateMatchedLow}{$-$1.77}
\newcommand{\SWcsGateMatchedHigh}{+2.13}
\newcommand{\SWcsDevTasks}{25}
\newcommand{\SWcsDevCandidates}{27}
\newcommand{\SWcsDevBelow}{7}
\newcommand{\SWagSRFull}{48.06}
\newcommand{\SWagSRBoth}{56.11}
\newcommand{\SWagSRArm}{58.33}
\newcommand{\SWagSRGrip}{37.78}
\newcommand{\SWagDArm}{+10.28}
\newcommand{\SWagDArmCI}{[+5.56, +15.28]}
\newcommand{\SWagDGrip}{$-$10.28}
\newcommand{\SWagDGripCI}{[$-$14.72, $-$5.83]}
\newcommand{\SWagInter}{+8.06}
\newcommand{\SWagInterCI}{[+1.94, +14.17]}
\newcommand{\SWagDBoth}{+8.06}
\newcommand{\SWagDBothCI}{[+3.06, +13.06]}
\newcommand{\SWagDArmAbs}{10.28}
\newcommand{\SWagDGripAbs}{10.28}
\newcommand{\SWagGripGivenArm}{$-$2.22}
\newcommand{\SWagGainBoth}{18.3}
\newcommand{\SWagLostBoth}{10.3}
\newcommand{\SWagGainArm}{18.6}
\newcommand{\SWagLostArm}{8.3}
\newcommand{\SWagGainGrip}{5.8}
\newcommand{\SWagLostGrip}{16.1}
\newcommand{\SWagTasksUpArm}{13}
\newcommand{\SWagTasksDownGrip}{11}
\newcommand{\SWagTabThreeOrig}{48.22}
\newcommand{\SWagTabThreeNoGrip}{39.89}
\begin{table*}[t]
\centering\footnotesize
\caption{\textbf{Questions answered in
Appendices~\ref{app:sw-robustness}--\ref{app:sw-completions}.} Each answer is
stated with its evidence in the section listed. Rescue and harm are shares
of all paired episodes. Unless a row says otherwise, answers
refer to the 18 held-out tasks and to editors at the end of training.}
\label{tab:sw-guide-questions}
\setlength{\tabcolsep}{5pt}
\renewcommand{\arraystretch}{1.12}
\begin{tabularx}{\textwidth}{@{}>{\raggedright\arraybackslash}p{0.265\textwidth}>{\raggedright\arraybackslash}Xl@{}}
\toprule
Question & Answer & Appendix \\
\midrule
Is editor harm more than a repeated evaluation produces? & Yes. In \swrBaseNullHarm\% of pairs a repeated base run fails where the first run succeeded; the 112-task editors harm \swrHarmLTwo\% (L2) to \swrHarmLThree\% (L3) of pairs. & \ref{app:sw-robustness-repeat} \\
Do the recipe differences hold across training seeds? & At 16 source tasks, L3 is below L1 and L2 in all three seeds. At 112 source tasks and 121.25k updates, L1 and L2 exceed L3 in all three seeds; L2 stays consistent (\swrThreeSrLtwoOne--\swrThreeSrLtwoTwo\%), L3 seed~2 (\swrThreeSrLthreeTwo\%) matches seed~0 (\swrThreeSrLthreeZero\%) while seed~1 is lower (\swrThreeSrLthreeOne\%), and L1 reaches \swrThreeSrLoneTwo\% in seed~2. & \ref{app:sw-robustness-seeds16}, \ref{app:sw-robustness-seeds112} \\
Does transfer extend beyond the 18 held-out tasks? & Yes. On 96 further unseen tasks the 16-task L1 and L2 editors raise SR from \UnsBase\% to \UnsSRLa\% and \UnsSRLb\%; L3 reaches \UnsSRLc\%. & \ref{app:sw-unseen96-cohort} \\
Is transfer higher where source experience is closer? & For the 16-task L1 and L2 editors on the 96 tasks, gain falls as feature distance to source contexts rises (Spearman \UnsRhoLa{} and \UnsRhoLb); for L3 the interval includes zero. & \ref{app:sw-unseen96-distance} \\
Does a smaller correction help every recipe? & Yes for the three 112-task editors: half scale changes held-out SR by \SWcsHalfPooledLone, \SWcsHalfPooledLtwo{} and \SWcsHalfPooledLthree{} points for L1, L2 and L3, and lowers harm for all three. & \ref{app:sw-corrsize-dose} \\
Do the arm and the gripper correction respond alike to scaling? & No. For the 112-task L3 editor, halving the arm correction raises held-out SR by \SWagDArmAbs{} points and halving the gripper correction lowers it by \SWagDGripAbs{} points. & \ref{app:sw-corrsize-armgrip} \\
Do source correction directions explain held-out corrections for every recipe? & For the 112-task editors, ten source directions capture \SWcsSubTenLthree\% of held-out correction variance for L3, \SWcsSubTenLtwo\% for L2 and \SWcsSubTenLone\% for L1. & \ref{app:sw-corrsize-subspace} \\
Do context-dependent gates exceed uniform scaling? & No. None of the support, magnitude-cap or cross-recipe disagreement gates exceeds uniform scaling at its own mean scale; cross-recipe gating falls below uniform scaling for L3 ($-$2.63 points) and the mean ensemble ($-$4.85 points). & \ref{app:sw-corrsize-gates} \\
Does narrow-source negative transfer hold for every recipe? & Yes. Two-source L1 and L2 also drop below the base on the 18 held-out tasks (5.44\% and 19.72\%, against 1.00\% for L3), despite 87--94\% source SR. & \ref{app:sw-completions-twosource} \\
Does held-out transfer appear early for the soft-penalty recipes? & Yes. L1 and L2 reach 52.00\% and 52.89\% held-out SR at 51.25k updates (46.33--56.72\% across added seeds). L2 is highest near 121k updates and lower at its endpoint, so the L1--L2 order depends on training length and seed. & \ref{app:sw-completions-evolution} \\
Do expert routing, the editor's instruction, or the camera input change held-out transfer? & Routing: single runs do not establish an effect. Instruction: SR changes by at most 1.67 points for any recipe. Camera: with the main camera only, none of the three 16-task encoder editors (L1, L2 or L3) exceeds its readout counterpart. & \ref{app:sw-completions-moe}--\ref{app:sw-completions-cameras} \\
Does pretraining speed adaptation under the soft-penalty recipes? & Not consistently. Pretraining lowers online critic loss by 45--60\% across all three targets (including Task~39), but pretrained and fresh runs are level from 5k updates and lose source SR without replay. & \ref{app:sw-adaptation} \\
Does source replay preserve source skills during adaptation? & For L2 in part: with half of every batch replayed from the 16 source tasks, source SR after adaptation is 54.17\% against 24.58\% without replay, at 98.00\% target SR. L3 reaches 20.63\%, below the base (24.38\%). & \ref{app:sw-adaptation-replay} \\
 
\bottomrule
\end{tabularx}
\end{table*}
\endgroup

\begin{table}[t]
\centering\footnotesize
\caption{\textbf{Where each result of the main text is detailed in the appendix.}}
\label{tab:supp-roadmap}
\setlength{\tabcolsep}{3pt}\renewcommand{\arraystretch}{1.13}
\begin{tabularx}{\linewidth}{@{}>{\raggedright\arraybackslash}X>{\raggedright\arraybackslash}p{0.40\linewidth}@{}}
\toprule
Result in the main text & Appendix \\
\midrule
112-task source and held-out success, pairwise recipe intervals (Section~\ref{sec:experiments:libero-reference}) & Appendices~\ref{app:libero-reference} and~\ref{app:libero-112-soft} \\
Source breadth (Figure~\ref{fig:correction-support}b) and held-out evolution (Figure~\ref{fig:libero-reference-training}a) & Appendices~\ref{app:libero-breadth} and~\ref{app:libero-evolution} \\
Context coverage and shared correction directions (Figure~\ref{fig:correction-support}c,d) & Appendices~\ref{app:libero-breadth} and~\ref{app:correction-structure} \\
Native-chunk base, source-mean and frozen-frame controls (Figure~\ref{fig:libero-controls}) & Appendices~\ref{app:libero-controls} and~\ref{app:libero-112-soft} \\
Correction scaling and gripper control (Table~\ref{tab:libero-controls}) & Appendices~\ref{app:sw-corrsize-dose} and~\ref{app:libero-controls} \\
Six recipes, camera ablation, routing (Table~\ref{tab:learner-classes}) & Appendices~\ref{app:six-recipes}, \ref{app:six-recipe-heldout}, \ref{app:sw-completions-moe} and~\ref{app:sw-completions-cameras} \\
Gradient probe (Section~\ref{sec:experiments:design}) & Appendices~\ref{app:libero-gradients} and~\ref{app:gradient-phases} \\
Adaptation and source retention (Section~\ref{sec:experiments:adaptation}) & Appendices~\ref{app:libero-adaptation} and~\ref{app:sw-adaptation}--\ref{app:sw-adaptation-replay} \\
Narrow-source pattern on RoboTwin S4 (Section~\ref{sec:experiments:robotwin}, Figure~\ref{fig:s4-extended-training}) & Appendices~\ref{app:s4-matched-mechanisms}--\ref{app:s4-followup} \\
ReSOP design, cost and measurements (Section~\ref{sec:method:resop}) & Appendices~\ref{app:sw-system} and~\ref{app:system-performance} \\
\bottomrule
\end{tabularx}
\end{table}

\section{Repeat Evaluations and Training Seeds}
\label{app:sw-robustness}

The main text reports one training run per configuration, evaluated once
against the base run of its batch. This section repeats the evaluation of the
112-task checkpoints and adds training seeds under the same configurations;
the method and its hyperparameters are unchanged. Rescue and
harm are shares of all paired episodes (Appendix~\ref{app:sw-guide}).

\subsection{How much do paired outcomes change when the same policy is evaluated twice?}
\label{app:sw-robustness-repeat}

\emph{In \swrBaseNullHarm\% of pairs a repeated base run fails where the
first run succeeded; the harm of every editor is at least twice that level.}

One batch runs the base three times and each 112-task editor twice on the 18
held-out tasks. Table~\ref{tab:sw-robustness-repeat} compares a second run
with the first as if the first were the base, so its rescue and harm measure
outcome changes that occur without any edit. Every editor run is within
\swrPaperReproMax{} points of the held-out SR that
Table~\ref{tab:learner-classes} lists for its recipe.

\begin{table*}[t]
\centering\footnotesize
\caption{\textbf{Repeated evaluations of the same policy.} 18 held-out tasks, 50 states$\times$2 passes
(1,800 pairs per comparison; 10 tasks in source-covered scenes, 8 in uncovered scenes), all in one
batch with the 112-task editors of the main text. Left: the second run is compared with the first as if the first
were the base, so rescue, harm and their sum (outcome differs) are shares of pairs that changed without any
edit; $\Delta$ is the second run minus the first in points, with a 95\% interval from 5,000 within-task
state resamples. Right: each editor against the base of this batch, as the range over the six pairings of three
base runs and two editor runs. Cov.\ and Unc.: source-covered and uncovered scenes. Entries are
percentages; $\Delta$ and SR gain are in points.}
\label{tab:sw-robustness-repeat}
\setlength{\tabcolsep}{3.2pt}
\begin{tabularx}{\textwidth}{@{}>{\raggedright\arraybackslash}Xrrrrrrrrrrr@{}}
\toprule
 & \multicolumn{3}{c}{SR} & \multicolumn{2}{c}{Second vs.\ first run} & \multicolumn{3}{c}{Outcome differs} & \multicolumn{3}{c}{Editor vs.\ base, six pairings} \\
\cmidrule(lr){2-4} \cmidrule(lr){5-6} \cmidrule(lr){7-9} \cmidrule(l){10-12}
Policy and runs & First & Second & $\Delta$ [95\% CI] & Rescue & Harm & All 18 & Cov.\,10 & Unc.\,8 & SR gain & Rescue & Harm \\
\midrule
Base, runs 1 and 2 & 23.22 & 23.83 & +0.61 [$-$0.33, +1.56] & 2.5 & 1.9 & 4.4 & 7.8 & 0.1 & -- & -- & -- \\
Base, runs 1 and 3 & 23.22 & 23.44 & +0.22 [$-$0.67, +1.17] & 2.1 & 1.8 & 3.9 & 6.8 & 0.3 & -- & -- & -- \\
Base, runs 2 and 3 & 23.83 & 23.44 & $-$0.39 [$-$1.39, +0.61] & 1.9 & 2.3 & 4.3 & 7.4 & 0.4 & -- & -- & -- \\
\midrule
L1, runs 1 and 2 & 57.83 & 56.83 & $-$1.00 [$-$2.33, +0.28] & 3.8 & 4.8 & 8.7 & 11.7 & 4.9 & 33.00--34.61 & 39.7--41.2 & 6.4--6.7 \\
L2, runs 1 and 2 & 53.39 & 54.28 & +0.89 [$-$0.28, +2.06] & 4.3 & 3.4 & 7.8 & 11.1 & 3.6 & 29.56--31.06 & 35.2--35.9 & 4.8--5.7 \\
L3, runs 1 and 2 & 50.83 & 50.50 & $-$0.33 [$-$1.50, +0.83] & 2.9 & 3.3 & 6.2 & 9.6 & 2.0 & 26.67--27.61 & 37.6--38.1 & 10.2--10.9 \\
\bottomrule
\end{tabularx}
\end{table*}

\paragraph{Editor harm is above the repeat level.}
Between two base runs, \swrBaseNullHarm\% of pairs turn from success to
failure and \swrBaseNullRescue\% from failure to success, and every interval
on the SR difference contains zero. Against the base, harm is
\swrHarmLTwo\% for L2, \swrHarmLOne\% for L1 and \swrHarmLThree\% for L3
across the six pairings of base and editor runs. The simulation is not
exactly repeatable: episode length differs in \swrBaseLenDiffer\% of repeated
base pairs.

\paragraph{Four tasks carry the variation of the base.}
Across the three base comparisons, \swrHotShare{} changed outcomes fall on
\swrHotTasks{} tasks, all in source-covered scenes; the other
\swrOtherTasks{} tasks repeat almost exactly. Between two base runs,
\swrBaseNullDifferCov\% of pairs in source-covered scenes therefore change
outcome, against \swrBaseNullDifferUnc\% in uncovered scenes. In uncovered scenes the repeat level of harm is at most
\swrBaseNullHarmUncMax\%, while the editors harm \swrSoftHarmUnc\% of pairs
under L1 and L2 and \swrBoundHarmUnc\% under L3.

\paragraph{Recipe gaps exceed repeat differences.}
A second run of the same editor moves held-out SR by \swrRepDeltaOne{} points
for L1, \swrRepDeltaTwo{} for L2 and \swrRepDeltaThree{} for L3. The recipes
keep the order L1, L2, L3 in every combination of runs, and the smallest gap
between adjacent recipes is \swrMinRecipeGap{} points.

The three base runs differ by at most \swrMaxBaseDelta{} points over two
passes and span \swrBaseSrFirstPass\% on the first pass alone, a spread
comparable to the differences among the base SRs that the main text lists for
its separate evaluations. On two of the four variable tasks, run for the
base and L2 with 50 states and one pass, repeats on the same machine change
\swrTwoTaskSame\% of outcomes and repeats on a second machine
\swrTwoTaskCross\%. The variation is therefore not tied to one machine.
Harm measured for the three editors is above the repeat level in every
pairing, and a repeated evaluation reproduces the recipe order on the
held-out tasks.

\subsection{Do the recipe differences at 16 source tasks persist across training seeds?}
\label{app:sw-robustness-seeds16}

\emph{The bounded recipe is below both soft-penalty recipes in every seed,
and the two soft-penalty recipes are within seed variation of each other.}

Table~\ref{tab:learner-classes} reports one training run per 16-task recipe.
Table~\ref{tab:sw-robustness-seeds16} adds seeds 1 and 2 for L1, L2 and L3
under the same configuration. Seed 0 was evaluated again on the held-out
tasks with its own base run and reaches \swrSxSeedZeroHere\% for L1, L2 and
L3. Table~\ref{tab:learner-classes} lists \swrSxSeedZeroPaper\% for the same
checkpoints, so two evaluations of one checkpoint differ by up to 1.45
points here.

\begin{table}[t]
\centering\footnotesize
\caption{\textbf{Three training seeds at 16 source tasks.} 100k online actor updates; 16 source tasks
(1,600 pairs) and 18 held-out tasks (1,800 pairs; 7 in source-covered scenes, 11 in uncovered scenes), 50
states$\times$2 passes. Rescue and harm are shares of held-out pairs. Seed 0 is the run of Table~\ref{tab:learner-classes}; its
source SR is from that evaluation and its held-out rows come from a repeated evaluation. Held-out base SR is
23.50\% for seed 0 and 22.89\% for seeds 1 and 2; source base SR is 22.06\% and
23.25\%. SD is the sample standard deviation over seeds. All entries are percentages.}
\label{tab:sw-robustness-seeds16}
\setlength{\tabcolsep}{3.0pt}
\begin{tabular}{@{}llrrrrrr@{}}
\toprule
 & & Source & \multicolumn{3}{c}{Held-out (18)} & \multicolumn{2}{c}{Held-out SR} \\
\cmidrule(lr){4-6} \cmidrule(l){7-8}
 & Seed & SR & SR & Rescue & Harm & Cov.\,7 & Unc.\,11 \\
\midrule
L1 & 0 & 93.00 & 40.72 & 24.9 & 7.7 & 66.71 & 24.18 \\
 & 1 & 88.69 & 47.33 & 33.9 & 9.4 & 70.00 & 32.91 \\
 & 2 & 92.13 & 39.67 & 22.2 & 5.4 & 59.57 & 27.00 \\
 & Mean & 91.27 & 42.57 & 27.0 & 7.5 & 65.43 & 28.03 \\
 & SD & 2.28 & 4.16 & 6.1 & 2.0 & 5.33 & 4.45 \\
\midrule
L2 & 0 & 91.63 & 43.61 & 27.6 & 7.5 & 58.43 & 34.18 \\
 & 1 & 89.31 & 44.39 & 28.8 & 7.3 & 53.43 & 38.64 \\
 & 2 & 92.25 & 41.78 & 27.6 & 8.7 & 59.57 & 30.45 \\
 & Mean & 91.06 & 43.26 & 28.0 & 7.8 & 57.14 & 34.42 \\
 & SD & 1.55 & 1.34 & 0.7 & 0.7 & 3.27 & 4.10 \\
\midrule
L3 & 0 & 85.63 & 24.00 & 16.2 & 15.7 & 42.86 & 12.00 \\
 & 1 & 89.19 & 20.00 & 13.9 & 16.8 & 36.29 & 9.64 \\
 & 2 & 86.31 & 27.44 & 19.2 & 14.7 & 49.29 & 13.55 \\
 & Mean & 87.04 & 23.81 & 16.5 & 15.7 & 42.81 & 11.73 \\
 & SD & 1.89 & 3.73 & 2.6 & 1.1 & 6.50 & 1.97 \\
\bottomrule
\end{tabular}
\end{table}

\paragraph{The bounded recipe is lower in every seed.}
Held-out SR is \swrSxBoundSr\% over the three L3 seeds and \swrSxSoftSr\%
over the six L1 and L2 seeds, so the two ranges do not overlap. Harm
separates in the same way, with \swrSxBoundHarm\% for L3 against
\swrSxSoftHarm\% for L1 and L2.

\paragraph{L1 and L2 are not separated.}
The difference of L2 minus L1 in held-out SR, taken between runs with the
same seed index, averages \swrSxLtwoMinusLone{} points and ranges from
\swrSxLtwoMinusLoneLo{} to \swrSxLtwoMinusLoneHi. L1 varies more between
seeds than L2, with standard deviations of \swrSxSdLone{} and \swrSxSdLtwo{}
points. On the source tasks the recipes are closer: L3 reaches
\swrSxBoundSource\% and the soft-penalty recipes \swrSxSoftSource\%. At 16
source tasks, the lower held-out SR and the higher harm of the bounded recipe
hold in all three seeds, and the data do not order L1 and L2.

\subsection{Do the recipe differences at 112 source tasks persist across three training seeds?}
\label{app:sw-robustness-seeds112}

\emph{At 121.25k updates both soft-penalty recipes exceed the bounded recipe in all three seeds; L2 stays consistent across seeds, L3 seed~2 reproduces seed~0 while seed~1 is lower, and the order of L1 and L2 depends on both training length and seed.}

The main text reports one run of each 112-task recipe, trained to 481.25k
online actor updates. Two additional seeds (seeds~1 and~2) of each recipe were
trained under the same configuration to 121.25k updates, and all nine
checkpoints (with L2 seed~0 at its nearest saved snapshot, 121.54k) were
evaluated together in one matched batch (Table~\ref{tab:sw-robustness-seeds112}).

\begin{table*}[t]
\centering\footnotesize
\caption{\textbf{Three training seeds at 112 source tasks (121.25k updates).} Seeds~1 and~2 use the
configuration of seed~0 and stop at 121.25k online actor updates; all nine checkpoints (with L2 seed~0 at its
nearest saved snapshot, 121.54k) are evaluated together in one matched batch. Held-out: 18 tasks, 50
states$\times$2 passes, 1,800 pairs per row, base SR 23.17\% (22.30\% on the 10
tasks in source-covered scenes, 24.25\% on the 8 in uncovered scenes). Source: 112 tasks, 10
states, one pass, 1,120 pairs, base SR 20.71\%, evaluated in this batch for seed~2 only; seed~0
and seed~1 source SRs appear in Table~\ref{tab:sw-completions-evolution} and the earlier two-seed batch
(86.07\%, 86.61\% and 76.43\% for L1, L2 and L3 seed~1). Early snapshots (1.25k and 51.25k) for seeds~1 and~2
appear in Table~\ref{tab:sw-completions-evolution}. Rescue and harm are shares of held-out pairs. Intervals
are 95\% from 5,000 within-task state resamples and do not include variation between training runs. All entries
are percentages.}
\label{tab:sw-robustness-seeds112}
\setlength{\tabcolsep}{2pt}
\begin{tabular*}{\textwidth}{@{\extracolsep{\fill}}llrrrrrrrrr@{}}
\toprule
 & & & \multicolumn{4}{c}{Held-out (18)} & \multicolumn{2}{c}{Held-out SR} & Source & Seed $k$ $-$ seed 0, \\
\cmidrule(lr){4-7} \cmidrule(lr){8-9}
 & Seed & Updates & SR & Rescue & Harm & SR gain over base [95\% CI] & Cov.\,10 & Unc.\,8 & SR & held-out SR [95\% CI] \\
\midrule
L1 & 0 & 121.25k & 49.94 & 37.4 & 10.7 & +26.78 [+23.89, +29.67] & 50.50 & 49.25 & -- & -- \\
 & 1 & 121.25k & 50.78 & 35.4 & 7.8 & +27.61 [+24.78, +30.56] & 57.60 & 42.25 & -- & +0.83 [$-$1.95, +3.61] \\
 & 2 & 121.25k & 61.28 & 44.4 & 6.3 & +38.11 [+35.22, +40.89] & 63.60 & 58.38 & 85.00 & +11.33 [+8.67, +13.94] \\
\midrule
L2 & 0 & 121.54k & 59.72 & 41.3 & 4.8 & +36.56 [+33.83, +39.33] & 65.60 & 52.38 & -- & -- \\
 & 1 & 121.25k & 58.11 & 42.6 & 7.7 & +34.94 [+32.00, +37.78] & 60.20 & 55.50 & -- & $-$1.61 [$-$4.44, +1.11] \\
 & 2 & 121.25k & 60.28 & 42.4 & 5.3 & +37.11 [+34.17, +40.06] & 65.90 & 53.25 & 84.46 & +0.56 [$-$2.11, +3.28] \\
\midrule
L3 & 0 & 121.25k & 46.78 & 33.3 & 9.7 & +23.61 [+20.56, +26.61] & 55.30 & 36.13 & -- & -- \\
 & 1 & 121.25k & 34.83 & 25.7 & 14.1 & +11.67 [+8.61, +14.61] & 46.90 & 19.75 & -- & $-$11.94 [$-$14.89, $-$9.00] \\
 & 2 & 121.25k & 46.44 & 33.8 & 10.6 & +23.28 [+20.33, +26.22] & 51.90 & 39.63 & 78.84 & $-$0.33 [$-$3.22, +2.67] \\
\bottomrule
\end{tabular*}
\end{table*}

\paragraph{L2 is consistent across seeds; L1 and L3 vary across seeds.}
In the matched three-seed batch (base SR \swrThreeBase\%), L2 stays within
2.17 points across seeds: \swrThreeSrLtwoZero\%, \swrThreeSrLtwoOne\% and
\swrThreeSrLtwoTwo\% (seed~1 vs.\ seed~0: \swrThreeDiffLtwoOne{}
\swrThreeCiLtwoOne; seed~2 vs.\ seed~0: \swrThreeDiffLtwoTwo{}
\swrThreeCiLtwoTwo). For bounded L3, seed~2 (\swrThreeSrLthreeTwo\%) matches
seed~0 (\swrThreeSrLthreeZero\%; difference \swrThreeDiffLthreeTwo{}
\swrThreeCiLthreeTwo), whereas seed~1 is \swrThreeAbsLthreeOne{} points lower
(\swrThreeSrLthreeOne\%, \swrThreeCiLthreeOne; in the earlier two-seed batch
with base 23.44\%, L3 seed~1 scored 35.00\% against 48.44\% for seed~0,
\swrHxSeedDiffThree{} \swrHxSeedCiThree, differing by repeat-evaluation
variation). L3 seed~1 falls below the base in uncovered scenes
(\swrThreeUncOneLthree\% vs.\ \swrThreeBaseUnc\%), whereas seed~2 reaches
\swrThreeUncTwoLthree\%. For L1, seed~1 (\swrThreeSrLoneOne\%) matches seed~0
(\swrThreeSrLoneZero\%; \swrThreeDiffLoneOne{} \swrThreeCiLoneOne), while
seed~2 rises to \swrThreeSrLoneTwo\% (\swrThreeDiffLoneTwo{}
\swrThreeCiLoneTwo). On the 112 source tasks, seed~2 reaches
\swrThreeSrcTwoLone\% (L1), \swrThreeSrcTwoLtwo\% (L2) and
\swrThreeSrcTwoLthree\% (L3), consistent with the earlier seed-1 source
evaluation (86.07\%, 86.61\% and \swrHxSourceOneThree\%).

\paragraph{What holds in all three seeds.}
Both soft-penalty recipes exceed bounded L3 in held-out SR in every seed: L2
minus L3 is \swrThreeTwoMinusThreeZero{} \swrThreeTwoMinusThreeZeroCi{} in
seed~0, \swrThreeTwoMinusThreeOne{} \swrThreeTwoMinusThreeOneCi{} in seed~1,
and \swrThreeTwoMinusThreeTwo{} \swrThreeTwoMinusThreeTwoCi{} in seed~2; L1
minus L3 is \swrThreeOneMinusThreeZero{} \swrThreeOneMinusThreeZeroCi,
\swrThreeOneMinusThreeOne{} \swrThreeOneMinusThreeOneCi{} and
\swrThreeOneMinusThreeTwo{} \swrThreeOneMinusThreeTwoCi. L2 also incurs less
harm than L3 in all three seeds (4.8--7.7\% vs.\ 9.7--14.1\%), and every
112-task L3 seed remains above all three 16-task L3 seeds
(Table~\ref{tab:sw-robustness-seeds16}).

\paragraph{What depends on training length or seed.}
At 121.25k updates, L2 leads L1 in seeds~0 and~1 (\swrThreeTwoMinusOneZero{}
\swrThreeTwoMinusOneZeroCi{} and \swrThreeTwoMinusOneOne{}
\swrThreeTwoMinusOneOneCi), whereas seed~2 leaves them statistically level
(L2 minus L1: \swrThreeTwoMinusOneTwo{} \swrThreeTwoMinusOneTwoCi).
Appendix~\ref{app:sw-completions-evolution} further shows that L2 peaks near
121.5k updates and is lower at 481.25k (59.83\% vs.\ 53.67\%, separate
batches), while L1 continues rising to
481.25k. Full-budget 481.25k endpoints remain single-seed runs, so the fine
ordering between L1 and L2 depends on both training length and random seed,
whereas the advantage of soft penalties over output bounds holds across all
three seeds at both 16 and 112 source tasks.

\section{Transfer to 96 Further Unseen Tasks}
\label{app:sw-unseen96}

The main text measures transfer on 18 held-out tasks. The 16-source-task editors
of Table~\ref{tab:learner-classes} were not trained on 96 further
LIBERO tasks, which are neither source tasks nor held-out tasks. This section
evaluates the L1, L2 and L3 checkpoints of that study on those 96
tasks, with no change to the method or hyperparameters; each recipe has one
training run. As for the 18 held-out tasks, the base was fine-tuned on all 130
LIBERO tasks, so these tasks are unseen by the editors, not by the base.
The evaluation plan fixed \UnsDevTasks{} of the 96 tasks, those with a task
index divisible by four, as development tasks for the gates of
Appendix~\ref{app:sw-corrsize}; all 96 are outside the training of these
16-task editors, although they are source tasks of the 112-task editors.

\begin{table}[t]
\centering\footnotesize
\caption{\textbf{The 16-source-task encoder editors on tasks outside their source set} (one training run each, 100k
online actor updates). One pass over 50 initial states per task; each cohort is paired with the base run of
its own batch. Rescue, kept, harm and fail are shares of all pairs in the cohort; gain is rescue minus harm,
with a 95\% interval from 5,000 within-task resamples of initial states.}
\label{tab:sw-unseen96-cohorts}
\setlength{\tabcolsep}{2.4pt}
\begin{tabular*}{\linewidth}{@{\extracolsep{\fill}}lrrrrrr@{}}
\toprule
Recipe & SR & Rescue & Kept & Harm & Fail & Gain [95\% CI] \\
\midrule
\multicolumn{7}{@{}l}{\emph{96 further tasks; 4,800 pairs; base SR 19.15\%}} \\
L1 & 40.94 & 28.3 & 12.7 & 6.5 & 52.6 & +21.79 [20.54, 23.04] \\
L2 & 40.71 & 28.3 & 12.4 & 6.8 & 52.5 & +21.56 [20.25, 22.94] \\
L3 & 28.58 & 20.3 & 8.3 & 10.9 & 60.5 & +9.44 [8.19, 10.69] \\
\midrule
\multicolumn{7}{@{}l}{\emph{18 held-out tasks, separate batch; 900 pairs; base SR 24.89\%}} \\
L1 & 40.56 & 23.6 & 17.0 & 7.9 & 51.6 & +15.67 [12.56, 18.78] \\
L2 & 45.78 & 28.3 & 17.4 & 7.4 & 46.8 & +20.89 [17.67, 24.11] \\
L3 & 23.89 & 15.8 & 8.1 & 16.8 & 59.3 & $-$1.00 [$-$3.89, +2.00] \\
\bottomrule
\end{tabular*}
\par\smallskip\raggedright
Paired SR contrasts on the 96 tasks: L2$-$L1 = $-$0.23 [$-$1.52, +1.10]; L3$-$L2 = $-$12.13 [$-$13.46, $-$10.83].
\end{table}

\begin{table*}[t]
\centering\footnotesize
\caption{\textbf{The 96 further tasks by scene, by base SR and by LIBERO suite} (50 pairs per task and recipe). Entries
are percentages; R and H are rescue and harm as shares of all pairs in the row, and Gain is R$-$H in points
(kept is SR$-$R; fail is 100$-$SR$-$H). $^{\circ}$ marks a gain whose 95\% interval includes zero. Scene
groups follow the definition of the main text for the 16 source tasks; base-SR groups use each task's base SR in this batch.}
\label{tab:sw-unseen96-breakdown}
\setlength{\tabcolsep}{3.4pt}
\begin{tabularx}{\textwidth}{@{}>{\raggedright\arraybackslash}Xrr@{\hspace{10pt}}rrrr@{\hspace{10pt}}rrrr@{\hspace{10pt}}rrrr@{}}
\toprule
 & & & \multicolumn{4}{c}{L1: full action + soft penalty} & \multicolumn{4}{c}{L2: additive + soft penalty} & \multicolumn{4}{c}{L3: additive + output bounds} \\
\cmidrule(r{8pt}){4-7} \cmidrule(r{8pt}){8-11} \cmidrule{12-15}
Task group & Tasks & Base SR & SR & R & H & Gain & SR & R & H & Gain & SR & R & H & Gain \\
\midrule
All 96 tasks & 96 & 19.15 & 40.94 & 28.3 & 6.5 & +21.79 & 40.71 & 28.3 & 6.8 & +21.56 & 28.58 & 20.3 & 10.9 & +9.44 \\
\midrule
Source-covered scenes & 34 & 27.12 & 50.53 & 30.5 & 7.1 & +23.41 & 48.76 & 29.6 & 8.0 & +21.65 & 33.59 & 20.4 & 13.9 & +6.47 \\
Uncovered scenes & 62 & 14.77 & 35.68 & 27.0 & 6.1 & +20.90 & 36.29 & 27.6 & 6.1 & +21.52 & 25.84 & 20.3 & 9.2 & +11.06 \\
\midrule
Base SR $\le$10\% & 47 & 3.45 & 26.81 & 25.3 & 1.9 & +23.36 & 26.77 & 25.5 & 2.2 & +23.32 & 21.19 & 20.3 & 2.5 & +17.74 \\
Base SR $>$10 to $<$50\% & 40 & 25.85 & 49.20 & 33.6 & 10.2 & +23.35 & 48.70 & 33.7 & 10.9 & +22.85 & 32.35 & 22.6 & 16.1 & +6.50 \\
Base SR $\ge$50\% & 9 & 71.33 & 78.00 & 20.4 & 13.8 & +6.67 & 78.00 & 18.9 & 12.2 & +6.67 & 50.44 & 10.7 & 31.6 & $-$20.89 \\
\midrule
LIBERO-90 & 72 & 14.69 & 37.03 & 28.3 & 5.9 & +22.33 & 37.94 & 29.3 & 6.1 & +23.25 & 27.94 & 21.8 & 8.6 & +13.25 \\
LIBERO-10 (long horizon) & 6 & 9.67 & 24.00 & 17.3 & 3.0 & +14.33 & 23.67 & 16.7 & 2.7 & +14.00 & 3.67 & 2.7 & 8.7 & $-$6.00 \\
LIBERO-Spatial & 6 & 36.33 & 70.33 & 43.7 & 9.7 & +34.00 & 74.67 & 46.3 & 8.0 & +38.33 & 50.33 & 26.7 & 12.7 & +14.00 \\
LIBERO-Object & 6 & 32.33 & 76.00 & 46.0 & 2.3 & +43.67 & 50.67 & 29.0 & 10.7 & +18.33 & 33.00 & 24.3 & 23.7 & +0.67$^{\circ}$ \\
LIBERO-Goal & 6 & 51.67 & 40.33 & 6.3 & 17.7 & $-$11.33 & 47.00 & 9.0 & 13.7 & $-$4.67$^{\circ}$ & 35.00 & 9.7 & 26.3 & $-$16.67 \\
\bottomrule
\end{tabularx}
\end{table*}

\subsection{Does transfer from 16 source tasks extend beyond the 18 held-out tasks?}
\label{app:sw-unseen96-cohort}

\emph{Yes: on the 96 further tasks the two soft-penalty editors raise SR by
about 22 points and the bounded editor by 9.}
Each task is evaluated on its 50 native initial states in one pass, which
gives \UnsPairs{} base--editor pairs per recipe
(Table~\ref{tab:sw-unseen96-cohorts}). The base and the three editors of a
task share one GPU: an RTX~5090 or an RTX~4090 for the 96 tasks, and an A100
for the separate batch on the 18 held-out tasks.

\paragraph{The soft-penalty recipes are level and both exceed the bounded recipe.}
Base SR on the 96 tasks is \UnsBase\%. L1 reaches \UnsSRLa\% and L2
\UnsSRLb\%, and their paired difference is \UnsDiffLbLa{} points
\UnsDiffLbLaCI. L3 reaches \UnsSRLc\%; its paired
difference to L2 is \UnsDiffLcLb{} points \UnsDiffLcLbCI.

\paragraph{The bounded recipe gains on the 96 tasks but not on the 18.}
The same three checkpoints were re-evaluated on the 18 held-out tasks in one
pass, where base SR is \UnsHeldBase\%. L1, L2 and L3 reach \UnsHeldSRLa\%,
\UnsHeldSRLb\% and \UnsHeldSRLc\% there. The soft-penalty recipes are above
the bounded recipe on both task sets. The L3 gain is \UnsHeldGainLc{} points
\UnsHeldGainCILc{} on the 18 tasks and \UnsGainLc{} points \UnsGainCILc{} on
the 96.

\paragraph{Harm is highest where the base already succeeds, most of all for the bounded recipe.}
Table~\ref{tab:sw-unseen96-breakdown} groups the 96 tasks by their base SR in
this batch. Harm is a share of all pairs and cannot exceed base SR, and the
groups are formed on the base run that also serves as the paired reference,
so harm rises with base SR partly by construction. Comparisons between
recipes within a group are not affected by either. On the \UnsLowTasks{}
tasks with base SR of at most 10\%, harm is at most \UnsLowHarmMax\% of pairs
for every recipe and L3 gains \UnsLowGainLc{} points.

On the \UnsHighTasks{} tasks with base SR of at least 50\%, base SR averages
\UnsHighBase\%. There L3 harms \UnsHighHarmLc\% of pairs and its gain is
\UnsHighGainLc{} points \UnsHighGainCILc. L1 and L2 each gain
\UnsHighGainLa{} points on these tasks, with intervals \UnsHighGainCILa{} and
\UnsHighGainCILb, although their harm is also highest in this group. Such
tasks make up \UnsHeldHighTasks{} of the 18 held-out tasks and
\UnsHighTasks{} of the 96, which is consistent with the lower L3 gain on the
18 tasks.

\paragraph{LIBERO-Goal shows negative transfer for L1 and L3.}
For L1 and L2 the gain is positive in LIBERO-90, LIBERO-10, LIBERO-Spatial
and LIBERO-Object. On the six LIBERO-Goal tasks, where base SR is
\UnsGoalBase\%, the L1 and L3 gains are \UnsGoalGainLa{} and
\UnsGoalGainLc{} points. The L2 gain there is \UnsGoalGainLb{} points
\UnsGoalGainCILb. L3 is also below the base on the six LIBERO-10 tasks. Per
task, SR falls on \UnsNegLa{} (L1), \UnsNegLb{} (L2) and \UnsNegLc{} (L3) of
the 96 tasks.

Transfer from 16 source tasks holds on a set of unseen tasks more than five
times the size of the 18-task held-out set, with the soft-penalty recipes
above the bounded recipe in every base-SR group.

\subsection{Is transfer higher where source experience is closer?}
\label{app:sw-unseen96-distance}

\emph{For the soft-penalty editors, yes: across the 96 tasks, gain falls as
the mean nearest-source distance of a task rises; for the bounded editor the
interval includes zero.}
For every editor decision we evaluate the nearest-source distance of
Equation~\ref{eq:context-coverage} in that editor's own features (mean-pooled
main- and wrist-camera encoder features, concatenated and L2-normalized).
The reference set holds decisions recorded on three initial states of each
of its 16 source tasks, and the distance is averaged over all decisions of
a task. The feature map therefore
differs from the fixed 112-task encoder of
Figure~\ref{fig:correction-support}c, and distances compare tasks within one
recipe only. The relation below is observational.

\begin{figure}[t]
\centering
\includegraphics[width=\linewidth]{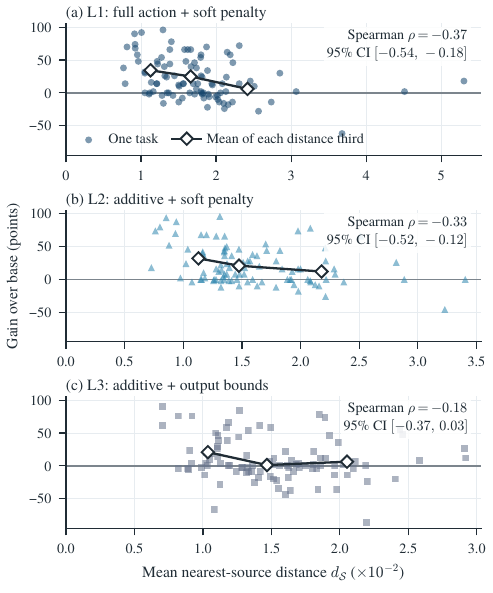}
\caption{\textbf{Gain over base against mean nearest-source distance on the 96
further tasks} (one point per task, 50 pairs each). Distances use each
editor's own features, so the horizontal axes are not comparable across
panels. Diamonds mark the mean distance and mean gain of the 32 nearest, 32
middle and 32 farthest tasks. Intervals resample tasks.}
\label{fig:sw-unseen96-distance}
\end{figure}

\begin{table}[t]
\centering\footnotesize
\caption{\textbf{Mean nearest-source distance and transfer across the 96 further tasks.} Left: Spearman correlation
between a task's distance and its rescue, harm and gain; $^{*}$ marks a 95\% interval that excludes zero
(5,000 resamples of tasks). Right: gain in points over the 32 nearest, 32 middle and 32 farthest tasks under
each editor's own distance.}
\label{tab:sw-unseen96-distance}
\setlength{\tabcolsep}{2.6pt}
\begin{tabular*}{\linewidth}{@{\extracolsep{\fill}}lrrrrrr@{}}
\toprule
 & \multicolumn{3}{c}{Correlation with distance} & \multicolumn{3}{c}{Gain by distance third} \\
\cmidrule(r){2-4} \cmidrule(l){5-7}
Recipe & Rescue & Harm & Gain & Nearest & Middle & Farthest \\
\midrule
L1 & $-$0.29$^{*}$ & 0.36$^{*}$ & $-$0.37$^{*}$ & +34.44 & +24.94 & +6.00 \\
L2 & $-$0.26$^{*}$ & 0.32$^{*}$ & $-$0.33$^{*}$ & +31.94 & +20.81 & +11.94 \\
L3 & $-$0.10 & 0.20 & $-$0.18 & +20.81 & +1.19 & +6.31 \\
\bottomrule
\end{tabular*}
\end{table}

\paragraph{Gain falls and harm rises with distance for L1 and L2.}
Figure~\ref{fig:sw-unseen96-distance} shows every task. The Spearman
correlation between distance and gain is \UnsRhoLa{} for L1 \UnsRhoCILa{} and
\UnsRhoLb{} for L2 \UnsRhoCILb; for L3 it is \UnsRhoLc{} \UnsRhoCILc. Harm
is higher on more distant tasks for L1 and L2, with correlations of
\UnsRhoHarmLa{} and \UnsRhoHarmLb{}
(Table~\ref{tab:sw-unseen96-distance}). Distance is not associated with base
SR on these tasks (absolute correlation of at most \UnsRhoBaseMax{} for the
three editors), so this relation does not restate the base-SR grouping above.

\paragraph{The nearest third of the tasks gains most.}
From the 32 nearest to the 32 farthest tasks, the L1 gain falls from
\UnsNearLa{} to \UnsFarLa{} points. The L2 gain falls from \UnsNearLb{} to
\UnsFarLb{} points. The L3 gain is \UnsNearLc{} points on its nearest third
and \UnsFarLc{} on its farthest, with the middle third lowest.

\paragraph{The scene grouping separates rescue by at most 3.5 points on these tasks.}
The scene grouping of the main text, applied to the 16 source tasks, places
\UnsCovTasks{} of the 96 tasks in source-covered scenes and \UnsUncTasks{} in
uncovered scenes, with base SR of \UnsCovBase\% and \UnsUncBase\%. All
\UnsSingleSceneTasks{} tasks of the three single-scene suites count as
source-covered.

L1 rescues \UnsCovRescueLa\% of pairs in source-covered scenes and
\UnsUncRescueLa\% in uncovered scenes. For L2 the two shares are
\UnsCovRescueLb\% and \UnsUncRescueLb\%. For L3 they are \UnsCovRescueLc\%
and \UnsUncRescueLc\%. On the 96 tasks, the scene grouping separates rescue
by at most 3.5 points, while the gain of the soft-penalty editors falls with
feature-space distance to source contexts.

\newcommand{\SWtrGScenes}{600}
\newcommand{\SWtrGBaseSucc}{412}
\newcommand{\SWtrGBothHarm}{233}
\newcommand{\SWtrGBothHarmShare}{38.8}
\newcommand{\SWtrGOnlyTCHarm}{57}
\newcommand{\SWtrGOnlyMoEHarm}{34}
\newcommand{\SWtrGBothKeep}{88}
\newcommand{\SWtrGHarmTC}{290}
\newcommand{\SWtrGHarmMoE}{267}
\newcommand{\SWtrGOverlapTC}{80}
\newcommand{\SWtrGOverlapMoE}{87}
\newcommand{\SWtrGBothRescue}{16}
\newcommand{\SWtrGOnlyTCRescue}{21}
\newcommand{\SWtrGOnlyMoERescue}{22}
\newcommand{\SWtrGBothFail}{129}
\newcommand{\SWtrGNetSucc}{$-$23}
\newcommand{\SWtrGNetFail}{$-$1}
\newcommand{\SWtrGNet}{$-$24}
\newcommand{\SWtrGateLoneNewRescue}{49}
\newcommand{\SWtrGateLoneLostRescue}{49}
\newcommand{\SWtrGateLoneRepaired}{16}
\newcommand{\SWtrGateLoneNewHarm}{14}
\newcommand{\SWtrGateLoneNet}{+2}
\newcommand{\SWtrGateLtwoNewRescue}{39}
\newcommand{\SWtrGateLtwoLostRescue}{30}
\newcommand{\SWtrGateLtwoRepaired}{11}
\newcommand{\SWtrGateLtwoNewHarm}{17}
\newcommand{\SWtrGateLtwoNet}{+3}
\newcommand{\SWtrGateLthreeNewRescue}{65}
\newcommand{\SWtrGateLthreeLostRescue}{56}
\newcommand{\SWtrGateLthreeRepaired}{32}
\newcommand{\SWtrGateLthreeNewHarm}{23}
\newcommand{\SWtrGateLthreeNet}{+18}
\newcommand{\SWtrGateMeanNewRescue}{41}
\newcommand{\SWtrGateMeanLostRescue}{67}
\newcommand{\SWtrGateMeanRepaired}{11}
\newcommand{\SWtrGateMeanNewHarm}{20}
\newcommand{\SWtrGateMeanNet}{$-$35}
\newcommand{\SWtrGateMeanNetLoss}{35}
\newcommand{\SWtrGateMeanNetRescueLoss}{26}
\newcommand{\SWtrGateFlipLow}{10.8}
\newcommand{\SWtrGateFlipHigh}{19.6}

\section{Correction Size at Inference for All Three Recipes}
\label{app:sw-corrsize}

All conditions in this section run on the 112-task checkpoints of the main text
(481.25k online actor updates); only the executed command changes at inference.
At each decision the base proposes a chunk $B$ and the editor outputs $A$.
The executed chunk is $B+\alpha(A-B)$, so $\alpha=1$ is the unmodified
editor. Every condition covers the 18 held-out tasks on the first pass,
with 50 initial states per task (900 pairs), and is paired with the base run
of its own batch. Outcome shares follow Appendix~\ref{app:sw-guide}. The
conditions are diagnostics of these editors: no editor is retrained
and no training setting changes.

\subsection{Does scaling the correction at inference help every recipe?}
\label{app:sw-corrsize-dose}

\begin{figure}[t]
\centering
\includegraphics[width=\linewidth]{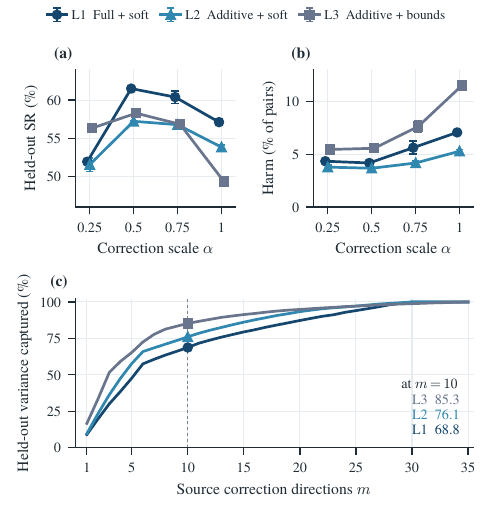}
\caption{\textbf{Correction scale and source directions for the three
112-task editors.} (a) Held-out SR and (b) harm share against the correction
scale $\alpha$ on the 18 held-out tasks; markers are the mean of the two
batches of Table~\ref{tab:sw-corrsize-dose} and bars span the two batches.
(c) Share of held-out correction variance captured by the first $m$
principal directions of source-task corrections
(Table~\ref{tab:sw-corrsize-subspace}).}
\label{fig:sw-corrsize-dose-subspace}
\end{figure}

\emph{Half scale raises held-out SR and lowers harm for all three recipes,
in both batches.}
Table~\ref{tab:libero-controls} summarizes scaling for L1, L2 and L3 as the
mean of two separately run batches, next to an earlier batch for L3 alone.
Table~\ref{tab:sw-corrsize-dose} and
Figure~\ref{fig:sw-corrsize-dose-subspace}a,b give the two batches
separately, with paired intervals.

\begin{table*}[t]
\centering\footnotesize
\caption{\textbf{Uniform scaling of the correction for the three 112-task editors.} 18 held-out
tasks, first pass, 900 pairs per cell. The executed chunk is $B+\alpha(A-B)$; $\alpha=1$ is the
unmodified editor. The two batches were run separately and each pairs every condition with its own
base run. Rescue and harm are shares of the 900 pairs; gain is SR minus base SR in points.
$\Delta$SR is the paired difference to $\alpha=1$ in the same batch; intervals are 95\% from 5,000
bootstrap draws that resample initial states within tasks.}
\label{tab:sw-corrsize-dose}
\setlength{\tabcolsep}{3.0pt}
\begin{tabularx}{\textwidth}{@{}>{\raggedright\arraybackslash}Xr@{\hspace{8pt}}rrrrl@{\hspace{10pt}}rrrrl@{}}
\toprule
 & & \multicolumn{5}{c}{First batch (base SR 23.78\%)} & \multicolumn{5}{c}{Second batch (base SR 24.00\%)} \\
\cmidrule(r{10pt}){3-7} \cmidrule{8-12}
Recipe & $\alpha$ & SR & Rescue & Harm & Gain & $\Delta$SR vs.\ $\alpha{=}1$ & SR & Rescue & Harm & Gain & $\Delta$SR vs.\ $\alpha{=}1$ \\
\midrule
\swatch{recfull}~L1 full action + soft penalty & 1.00 & 57.56 & 41.0 & 7.2 & +33.78 & -- & 56.67 & 39.6 & 6.9 & +32.67 & -- \\
 & 0.75 & 61.22 & 42.4 & 5.0 & +37.44 & +3.67 [+0.56, +6.78] & 59.56 & 41.8 & 6.2 & +35.56 & +2.89 [0.00, +5.78] \\
 & 0.50 & 61.89 & 42.2 & 4.1 & +38.11 & +4.33 [+1.11, +7.56] & 61.11 & 41.3 & 4.2 & +37.11 & +4.44 [+1.22, +7.78] \\
 & 0.25 & 51.78 & 32.3 & 4.3 & +28.00 & $-$5.78 [$-$9.11, $-$2.44] & 52.11 & 32.4 & 4.3 & +28.11 & $-$4.56 [$-$7.78, $-$1.33] \\
\addlinespace[2pt]
\swatch{recadd}~L2 additive + soft penalty & 1.00 & 54.11 & 35.8 & 5.4 & +30.33 & -- & 53.56 & 34.7 & 5.1 & +29.56 & -- \\
 & 0.75 & 56.56 & 37.1 & 4.3 & +32.78 & +2.44 [$-$0.44, +5.33] & 57.00 & 37.0 & 4.0 & +33.00 & +3.44 [+0.56, +6.33] \\
 & 0.50 & 57.22 & 37.0 & 3.6 & +33.44 & +3.11 [$-$0.22, +6.44] & 57.22 & 37.0 & 3.8 & +33.22 & +3.67 [+0.33, +6.89] \\
 & 0.25 & 50.67 & 30.9 & 4.0 & +26.89 & $-$3.44 [$-$6.89, +0.11] & 52.44 & 32.0 & 3.6 & +28.44 & $-$1.11 [$-$4.56, +2.33] \\
\addlinespace[2pt]
\swatch{recbnd}~L3 additive + output bounds & 1.00 & 49.22 & 36.6 & 11.1 & +25.44 & -- & 49.44 & 37.3 & 11.9 & +25.44 & -- \\
 & 0.75 & 56.89 & 40.2 & 7.1 & +33.11 & +7.67 [+4.78, +10.56] & 56.78 & 40.9 & 8.1 & +32.78 & +7.33 [+4.56, +10.22] \\
 & 0.50 & 58.22 & 40.1 & 5.7 & +34.44 & +9.00 [+5.78, +12.22] & 58.33 & 39.8 & 5.4 & +34.33 & +8.89 [+5.78, +12.00] \\
 & 0.25 & 56.89 & 38.3 & 5.2 & +33.11 & +7.67 [+4.44, +11.00] & 55.78 & 37.4 & 5.7 & +31.78 & +6.33 [+2.89, +9.78] \\
\bottomrule
\end{tabularx}
\end{table*}

Averaged over the two batches, half scale changes SR by
\SWcsHalfPooledLone{} points for L1 \SWcsHalfPooledCILone, by
\SWcsHalfPooledLtwo{} for L2 \SWcsHalfPooledCILtwo{} and by
\SWcsHalfPooledLthree{} for L3 \SWcsHalfPooledCILthree, relative to the
unmodified editor. Within single batches the interval excludes zero in five
of the six recipe--batch cells; the exception is L2 in the first batch, with
\SWcsHalfLtwoOneCI.

For L3, the earlier batch of Table~\ref{tab:libero-controls} reaches
48.22\% at full scale and 57.44\% at half scale. The two batches here give
\SWcsSRLthreeFullOne\% and \SWcsSRLthreeFullTwo\% at full scale. At half
scale they give \SWcsSRLthreeHalfOne\% and \SWcsSRLthreeHalfTwo\%. As shares of
pairs, that batch has L3 harm of 9.2\% at full scale and 4.9\% at half
scale (Table~\ref{tab:outcome-shares-rates}). In the two batches of
Table~\ref{tab:sw-corrsize-dose} the harm share is two to three points
higher at full scale and within one point at half scale; each batch has its
own base run, and repeated evaluations change several percent of pairs
(Appendix~\ref{app:sw-robustness-repeat}).

All four scales are reported for every recipe, and no scale is selected on
the held-out tasks. At half scale, L3 reaches \SWcsSRLthreeHalfOne\% and
\SWcsSRLthreeHalfTwo\% in the two batches and L2 \SWcsSRLtwoHalfOne\% in
both, so the bounded recipe is no longer below L2.

Harm falls and rescue rises under every recipe. From full to half scale the
harm share falls by \SWcsHarmDropLone{} points for L1, \SWcsHarmDropLtwo{}
for L2 and \SWcsHarmDropLthree{} for L3 (means of the two batches). Rescue
changes by \SWcsRescueChangeLone, \SWcsRescueChangeLtwo{} and
\SWcsRescueChangeLthree{} points.

Quarter scale removes part of the rescue of the soft-penalty recipes. L1 SR
changes by \SWcsQuarterLoneOne{} points in the first batch and by
\SWcsQuarterLoneTwo{} in the second, relative to full scale, while L3 at
quarter scale stays within \SWcsLthreeQuarterGap{} points of its half-scale
SR. L3 also has the largest corrections at full scale: its mean executed arm
correction norm is \SWcsArmNormLthree{} in command units, against
\SWcsArmNormLone{} for L1 and \SWcsArmNormLtwo{} for L2. Relative to full scale, every
smaller scale lowers harm under every recipe, and half scale does so
without losing rescue.

\newcommand{\SWagSRFull}{48.06}
\newcommand{\SWagSRBoth}{56.11}
\newcommand{\SWagSRArm}{58.33}
\newcommand{\SWagSRGrip}{37.78}
\newcommand{\SWagDArm}{+10.28}
\newcommand{\SWagDArmCI}{[+5.56, +15.28]}
\newcommand{\SWagDGrip}{$-$10.28}
\newcommand{\SWagDGripCI}{[$-$14.72, $-$5.83]}
\newcommand{\SWagInter}{+8.06}
\newcommand{\SWagInterCI}{[+1.94, +14.17]}
\newcommand{\SWagDBoth}{+8.06}
\newcommand{\SWagDBothCI}{[+3.06, +13.06]}
\newcommand{\SWagDArmAbs}{10.28}
\newcommand{\SWagDGripAbs}{10.28}
\newcommand{\SWagGripGivenArm}{$-$2.22}
\newcommand{\SWagGainBoth}{18.3}
\newcommand{\SWagLostBoth}{10.3}
\newcommand{\SWagGainArm}{18.6}
\newcommand{\SWagLostArm}{8.3}
\newcommand{\SWagGainGrip}{5.8}
\newcommand{\SWagLostGrip}{16.1}
\newcommand{\SWagTasksUpArm}{13}
\newcommand{\SWagTasksDownGrip}{11}
\newcommand{\SWagTabThreeOrig}{48.22}
\newcommand{\SWagTabThreeNoGrip}{39.89}

\subsection{Do the arm and the gripper correction respond alike to scaling?}
\label{app:sw-corrsize-armgrip}

\emph{No: for the bounded editor, halving the arm correction raises held-out
SR by \SWagDArmAbs{} points and halving the gripper correction lowers it by
\SWagDGripAbs{} points.}
The scale of Appendix~\ref{app:sw-corrsize-dose} multiplies all seven
coordinates of the correction. Here the six arm coordinates and the gripper
coordinate of the 112-task L3 editor are scaled separately
(Table~\ref{tab:sw-corrsize-armgrip}), on the 18 held-out tasks with the
first 20 initial states of each task in one pass. The batch is separate from
those of Table~\ref{tab:sw-corrsize-dose} and contains no base run, so the
table gives SR and paired differences between conditions in place of rescue
and harm.

\begin{table}[t]
\centering\footnotesize
\caption{\textbf{Arm and gripper correction scaled separately} for the 112-task bounded editor (L3, one
training run). 18 held-out tasks, the first 20 initial states of each task, one pass: 360 episodes per condition,
in one batch without a base run. Top: SR, and the shares of the 360 pairs that succeed only under the condition
or only at full scale. Bottom: paired differences in points; intervals resample initial states within tasks with
the same draws for all conditions (10,000 draws). The adjusted interval is the 98.3\% interval for the three
contrasts fixed in the evaluation plan; the interaction is both halved minus arm halved minus gripper halved
plus full.}
\label{tab:sw-corrsize-armgrip}
\setlength{\tabcolsep}{4.2pt}
\begin{tabular*}{\linewidth}{@{\extracolsep{\fill}}lrrrrr@{}}
\toprule
 & \multicolumn{2}{c}{Scale} & & \multicolumn{2}{c}{Succeeds only} \\
\cmidrule(lr){2-3} \cmidrule(l){5-6}
Condition & Arm & Gripper & SR & here & at full \\
\midrule
Full scale & 1 & 1 & 48.06 & -- & -- \\
Both halved & 0.5 & 0.5 & 56.11 & 18.3 & 10.3 \\
Arm halved & 0.5 & 1 & 58.33 & 18.6 & 8.3 \\
Gripper halved & 1 & 0.5 & 37.78 & 5.8 & 16.1 \\
\bottomrule
\end{tabular*}

\vspace{5pt}
\setlength{\tabcolsep}{4.2pt}
\begin{tabular*}{\linewidth}{@{\extracolsep{\fill}}lrr@{}}
\toprule
Contrast & Difference [95\% CI] & Adjusted CI \\
\midrule
Arm halved $-$ full & +10.28 [+5.56, +15.28] & [+4.44, +16.39] \\
Gripper halved $-$ full & $-$10.28 [$-$14.72, $-$5.83] & [$-$15.83, $-$5.00] \\
Interaction & +8.06 [+1.94, +14.17] & [+0.83, +15.28] \\
Both halved $-$ full & +8.06 [+3.06, +13.06] & -- \\
\bottomrule
\end{tabular*}
\end{table}

\paragraph{The gain of half scale comes from the arm coordinates.}
At full scale the editor reaches \SWagSRFull\%. Halving only the arm
correction gives \SWagSRArm\%, a difference of \SWagDArm{} points
\SWagDArmCI, and SR rises on \SWagTasksUpArm{} of the 18 tasks. Halving both
gives \SWagSRBoth\%, \SWagDBoth{} points above full scale \SWagDBothCI. On
the 900 pairs of Table~\ref{tab:sw-corrsize-dose}, uniform half scale
reaches \SWcsSRLthreeHalfOne\% and \SWcsSRLthreeHalfTwo\%.

\paragraph{Halving the gripper correction alone lowers SR.}
With the arm at full scale, halving the gripper correction gives
\SWagSRGrip\%, a difference of \SWagDGrip{} points \SWagDGripCI. In
\SWagLostGrip\% of the pairs the episode succeeds only at full scale, and in
\SWagGainGrip\% only with the halved gripper correction. Removing the
gripper correction entirely lowers SR from \SWagTabThreeOrig\% to
\SWagTabThreeNoGrip\% in Table~\ref{tab:libero-controls}. With the arm
correction halved, halving the gripper correction as well changes SR by
\SWagGripGivenArm{} points, and the interaction contrast is \SWagInter{}
points \SWagInterCI. All three contrasts of the evaluation plan exclude zero
at the adjusted level. For this editor and these initial states, a smaller
arm correction raises held-out SR and a smaller gripper correction lowers
it.

\subsection{Do source correction directions explain held-out corrections for every recipe?}
\label{app:sw-corrsize-subspace}

\emph{Ten source directions capture most held-out correction variance for
the bounded recipe and less for the soft-penalty recipes, which need
\SWcsSubNeedLtwo{} (L2) and \SWcsSubNeedLone{} (L1) directions to reach 85\%.}
Figure~\ref{fig:correction-support}d reports this statistic for L3.
Table~\ref{tab:sw-corrsize-subspace} and
Figure~\ref{fig:sw-corrsize-dose-subspace}c give it for all three recipes
(scaling, centering and formula of Appendix~\ref{app:correction-structure}),
on every held-out decision of the first batch at $\alpha=1$.

\begin{table}[t]
\centering\footnotesize
\caption{\textbf{Source correction directions on held-out corrections.} Share (\%) of held-out
correction variance captured by the first $m$ principal directions of each 112-task editor's
source-task corrections (scaling and centering of Appendix~\ref{app:correction-structure}; tasks and
episodes weighted equally). Source: the same statistic on source corrections. $m$ for 85\%: the smallest $m$ that
captures 85\% of held-out correction variance. Five-step mean: share
of squared held-out correction in the within-chunk mean.}
\label{tab:sw-corrsize-subspace}
\setlength{\tabcolsep}{4.2pt}
\begin{tabular}{@{}lrrrrrr@{}}
\toprule
 & \multicolumn{3}{c}{Held-out, first $m$} & Source & $m$ for & Five-step \\
\cmidrule(lr){2-4}
Recipe & $m{=}5$ & $m{=}10$ & $m{=}20$ & $m{=}10$ & 85\% & mean \\
\midrule
\swatch{recfull}~L1 & 47.7 & 68.8 & 87.3 & 72.3 & 19 & 56.4 \\
\swatch{recadd}~L2 & 57.7 & 76.1 & 93.4 & 77.0 & 15 & 65.1 \\
\swatch{recbnd}~L3 & 65.3 & 85.3 & 94.9 & 88.9 & 10 & 76.2 \\
\bottomrule
\end{tabular}
\end{table}

Ten directions capture \SWcsSubTenLthree\% for L3, \SWcsSubTenLtwo\% for L2
and \SWcsSubTenLone\% for L1. The mean over the five steps of a chunk contains \SWcsSubBlockLthree\%
of the squared held-out correction for L3, \SWcsSubBlockLtwo\% for L2 and
\SWcsSubBlockLone\% for L1. The statistic uses the per-coordinate scales of
Appendix~\ref{app:correction-structure}; in unscaled command units the gripper coordinates dominate the
variance and the L3 value at ten directions is \SWcsSubRawLthree\%.

Held-out corrections of the bounded recipe therefore lie closer to a fixed
set of source directions than those of the soft-penalty recipes. This agrees
with the source-mean control of Figure~\ref{fig:libero-controls}, where replacing each editor's corrections by its fixed source-mean
template changes SR by $-$3.22 points for L3, with an interval that includes
zero, and lowers it by 18.11 and 12.00 points for L1 and L2.

\subsection{Which actor-side signal separates harmed from kept episodes?}
\label{app:sw-corrsize-signals}

\emph{For the soft-penalty recipes the largest correction in an episode
separates harmed from kept episodes; nearest-source distance separates them
less well, and no signal ranks first for all three recipes.}
Held-out tasks have no critic, so only actor-side quantities are available
at inference. Table~\ref{tab:sw-corrsize-signals} gives, for the unmodified
editors of the first batch, the AUROC of five per-episode signals for
separating harmed from kept episodes among the \SWcsNBaseSucc{} pairs in
which the base succeeds, and rescued from still-failed episodes among the
other \SWcsNBaseFail.

\begin{table*}[t]
\centering\footnotesize
\caption{\textbf{Actor-side signals and paired outcomes at $\alpha=1$.} AUROC of the episode maximum
of each signal on the 18 held-out tasks (first batch, first pass). Left: harmed against kept episodes
among the 214 pairs in which the base succeeds. Right: rescued against still-failed episodes
among the 686 pairs in which the base fails. Correction signals are divided by their
source-task median; nearest-source distance compares the editor's pooled visual feature with its
source-task features; the disagreement signals compare the executed correction with corrections
predicted, but not executed, by two earlier snapshots of the same run and by the other two recipes.
Brackets: 95\% state-bootstrap intervals. Bold: highest value per recipe on the left.}
\label{tab:sw-corrsize-signals}
\setlength{\tabcolsep}{3.2pt}
\begin{tabularx}{\textwidth}{@{}>{\raggedright\arraybackslash}Xllllll@{}}
\toprule
 & \multicolumn{3}{c}{Harmed vs.\ kept (base succeeds)} & \multicolumn{3}{c}{Rescued vs.\ still failed (base fails)} \\
\cmidrule(lr){2-4} \cmidrule(l){5-7}
Signal (episode maximum) & L1 & L2 & L3 & L1 & L2 & L3 \\
\midrule
Arm correction norm / source median & \best{0.82}\,{\scriptsize[0.76, 0.88]} & 0.83\,{\scriptsize[0.77, 0.89]} & 0.62\,{\scriptsize[0.54, 0.69]} & 0.34\,{\scriptsize[0.31, 0.38]} & 0.32\,{\scriptsize[0.28, 0.35]} & 0.51\,{\scriptsize[0.47, 0.55]} \\
Gripper correction / source median & 0.80\,{\scriptsize[0.72, 0.87]} & \best{0.89}\,{\scriptsize[0.82, 0.94]} & 0.52\,{\scriptsize[0.44, 0.60]} & 0.38\,{\scriptsize[0.34, 0.42]} & 0.28\,{\scriptsize[0.24, 0.31]} & 0.50\,{\scriptsize[0.46, 0.54]} \\
Nearest-source distance & 0.64\,{\scriptsize[0.55, 0.72]} & 0.69\,{\scriptsize[0.60, 0.77]} & \best{0.75}\,{\scriptsize[0.68, 0.81]} & 0.51\,{\scriptsize[0.47, 0.55]} & 0.46\,{\scriptsize[0.42, 0.50]} & 0.29\,{\scriptsize[0.26, 0.32]} \\
Disagreement with earlier snapshots & 0.64\,{\scriptsize[0.56, 0.72]} & 0.65\,{\scriptsize[0.58, 0.72]} & 0.75\,{\scriptsize[0.69, 0.81]} & 0.35\,{\scriptsize[0.31, 0.38]} & 0.34\,{\scriptsize[0.30, 0.38]} & 0.25\,{\scriptsize[0.21, 0.28]} \\
Disagreement with the other recipes & 0.36\,{\scriptsize[0.27, 0.44]} & 0.45\,{\scriptsize[0.35, 0.55]} & 0.71\,{\scriptsize[0.64, 0.77]} & 0.55\,{\scriptsize[0.51, 0.59]} & 0.59\,{\scriptsize[0.55, 0.63]} & 0.31\,{\scriptsize[0.27, 0.34]} \\
\bottomrule
\end{tabularx}
\end{table*}

For L1 and L2 the episode maximum of the arm correction norm reaches an
AUROC of \SWcsAucArmLone{} and \SWcsAucArmLtwo. The gripper correction
reaches \SWcsAucGripLone{} and \SWcsAucGripLtwo. Nearest-source distance
gives \SWcsAucDistMaxLone{} and \SWcsAucDistMaxLtwo{} as an episode maximum.
As an episode mean it gives \SWcsAucDistMeanLone{} and
\SWcsAucDistMeanLtwo, and the arm correction norm has a higher AUROC than
nearest-source distance under all three aggregations that were computed
(mean, maximum, and mean over the first fifth of an episode's decisions).
Harmed episodes run longer than kept ones, which favors an episode
maximum; the first-fifth mean, which gives the same ordering, is less
affected.

For L3, whose output is bounded, the two magnitude signals give
\SWcsAucArmLthree{} and \SWcsAucGripLthree. Its highest values are
\SWcsAucDistLthree{} for nearest-source distance and \SWcsAucSnapLthree{}
for disagreement with earlier snapshots. Among base failures, the magnitude
signals of L1 and L2 lie between \SWcsAucRescueLow{} and
\SWcsAucRescueHigh: episodes with a large correction are rescued less often.
These are associations within one batch. For the soft-penalty recipes,
correction size has the highest AUROC of the five signals, which agrees
with the scaling result of Appendix~\ref{app:sw-corrsize-dose}.

\subsection{Do context-dependent gates exceed uniform scaling at the same mean scale?}
\label{app:sw-corrsize-gates}

\emph{No tested gate exceeds uniform scaling at its own mean scale: across
support, magnitude-cap and cross-recipe disagreement gates, no interval is
positively separated from matched uniform scaling, and uniform half scale
gives the higher SR.}
We evaluate three families of per-decision actor-side gates: a support gate
that lowers $\alpha$ as nearest-source distance rises, a magnitude cap that
limits arm and gripper corrections to a source percentile, and a cross-recipe
disagreement gate (applied to each single recipe and to the three-recipe mean
correction) that lowers $\alpha$ linearly as pairwise disagreement among the
three recipes rises between source quantiles. Settings are selected on
development tasks outside the 18 held-out tasks and evaluated once on the
112-task editors on the 18 held-out tasks (Table~\ref{tab:sw-corrsize-gates}).

\begin{table*}[t]
\centering\footnotesize
\caption{\textbf{Context-dependent gates on the 112-task editors.} 18 held-out tasks, first pass, 900 pairs
per row. Top block: support gate and magnitude cap from the second batch of Table~\ref{tab:sw-corrsize-dose}
(base SR 24.00\%); the support gate keeps $\alpha=1$ up to the first percentile of leave-one-task-out source
distances and lowers it linearly to the floor at the second, while the magnitude cap limits the arm norm and
gripper correction to the stated source percentile. Bottom block: cross-recipe disagreement gate (on each
recipe and on the three-recipe mean correction, separate 900-pair batch, base SR 25.22\%), which lowers
$\alpha$ linearly when pairwise disagreement across L1, L2 and L3 rises between source quantiles. Each setting
has the highest paired gain on development tasks outside the 18 held-out tasks. Mean $\alpha$: applied scale
averaged over decisions. Last column: difference to uniform scaling linearly interpolated at the same mean
$\alpha$. Intervals: 95\%, state bootstrap.}
\label{tab:sw-corrsize-gates}
\setlength{\tabcolsep}{3.6pt}
\begin{tabularx}{\textwidth}{@{}ll>{\raggedright\arraybackslash}Xrrrrll@{}}
\toprule
Recipe & Gate & Setting & Mean $\alpha$ & SR & Rescue & Harm & $\Delta$SR vs.\ ungated & $\Delta$SR vs.\ uniform, same mean $\alpha$ \\
\midrule
\swatch{recfull}~L1 & Support gate & $q_{75}$--$q_{99}$, floor 0.25 & 0.94 & 57.22 & 39.8 & 6.6 & +0.56 [$-$1.78, +2.89] & $-$0.17 [$-$2.29, +1.97] \\
 & Magnitude cap & source median & 0.87 & 60.33 & 41.2 & 4.9 & +3.67 [+0.89, +6.33] & +2.13 [$-$0.14, +4.48] \\
\addlinespace[2pt]
\swatch{recadd}~L2 & Support gate & $q_{75}$--$q_{99}$, floor 0.50 & 0.94 & 53.00 & 35.0 & 6.0 & $-$0.56 [$-$2.78, +1.67] & $-$1.35 [$-$3.46, +0.74] \\
 & Magnitude cap & source $q_{75}$ & 0.94 & 55.22 & 36.7 & 5.4 & +1.67 [$-$1.00, +4.56] & +0.85 [$-$1.61, +3.50] \\
\addlinespace[2pt]
\swatch{recbnd}~L3 & Support gate & $q_{75}$--$q_{99}$, floor 0.25 & 0.92 & 50.00 & 37.0 & 11.0 & +0.56 [$-$1.78, +3.00] & $-$1.77 [$-$4.02, +0.52] \\
 & Magnitude cap & source median & 0.92 & 51.00 & 36.8 & 9.8 & +1.56 [$-$1.44, +4.33] & $-$0.93 [$-$3.60, +1.63] \\
\midrule
\multicolumn{9}{@{}l}{\emph{Cross-recipe disagreement gate (separate 900-pair batch; base SR 25.22\%; ungated mean ensemble 60.22\%)}} \\
\swatch{recfull}~L1 & Disagreement & $q_{50}$--$q_{95}$, floor 0.50 & 0.90 & 59.00 & 39.6 & 5.8 & +0.22 [$-$2.11, +2.67] & $-$0.79 [$-$2.98, +1.40] \\
\swatch{recadd}~L2 & Disagreement & $q_{75}$--$q_{99}$, floor 0.50 & 0.97 & 54.11 & 35.3 & 6.4 & +0.33 [$-$1.78, +2.44] & $-$0.12 [$-$2.09, +1.93] \\
\swatch{recbnd}~L3 & Disagreement & $q_{50}$--$q_{95}$, floor 0.25 & 0.84 & 51.44 & 35.6 & 9.3 & +2.00 [$-$0.78, +4.89] & $-$2.63 [$-$5.00, $-$0.31] \\
Mean ens. & Disagreement & $q_{50}$--$q_{95}$, floor 0.00 & 0.76 & 56.33 & 36.4 & 5.3 & $-$3.89 [$-$6.33, $-$1.33] & $-$4.85 [$-$7.67, $-$1.97] \\
\bottomrule
\end{tabularx}
\end{table*}

For the support gate and magnitude cap (top block, base SR 24.00\%), the
selected settings leave most of the correction in place, with mean applied
scales between \SWcsGateAlphaLow{} and \SWcsGateAlphaHigh. Against the
unmodified editor, only the magnitude cap on L1 excludes zero
(\SWcsGateCapLone{} points \SWcsGateCapLoneCI). Against uniform scaling at
the same mean scale, the six differences range from \SWcsGateMatchedLow{} to
\SWcsGateMatchedHigh{} points and every interval includes zero. Uniform half
scale in the same batch gives \SWcsSRLoneHalfTwo\%, \SWcsSRLtwoHalfTwo\% and
\SWcsSRLthreeHalfTwo\% for L1, L2 and L3, above every gated condition. On the
development tasks, none of the \SWcsDevCandidates{} candidate settings exceeds
uniform scaling with an interval excluding zero, and \SWcsDevBelow{} fall
below it.

The cross-recipe disagreement gate (bottom block, separate 900-pair batch,
base SR 25.22\%) reaches the same conclusion. Ungated anchors score 58.78\%
(L1), 53.78\% (L2), 49.44\% (L3) and 60.22\% (three-recipe mean ensemble).
Gating L1, L2 and L3 leaves SR statistically indistinguishable from ungated
execution (+0.22, +0.33 and +2.00 points, all CIs spanning zero), and gating
the mean ensemble changes SR by $-$3.89 [$-$6.33, $-$1.33] points. Relative to
uniform scaling interpolated at the same mean $\alpha$, every disagreement
gate is negative: $-$0.79 (L1) and $-$0.12 (L2) include zero, while L3
($-$2.63 [$-$5.00, $-$0.31]) and the mean ensemble ($-$4.85 [$-$7.67,
$-$1.97]) fall below uniform scaling, with intervals that exclude zero.

The disagreement gates change which episodes succeed more than how many
(Table~\ref{tab:sw-corrsize-gate-transitions}). Between each gated editor
and the same editor without the gate, the outcome differs in
\SWtrGateFlipLow--\SWtrGateFlipHigh\% of the 900 scenes, a share that
includes the variation of a repeated evaluation
(Appendix~\ref{app:sw-robustness-repeat}). For L1, L2 and L3 the changes
nearly cancel: the gate repairs \SWtrGateLoneRepaired, \SWtrGateLtwoRepaired{}
and \SWtrGateLthreeRepaired{} harms and adds \SWtrGateLoneNewHarm,
\SWtrGateLtwoNewHarm{} and \SWtrGateLthreeNewHarm, while new and lost
rescues are also close in number. On the mean ensemble the gate loses
\SWtrGateMeanLostRescue{} rescues and adds \SWtrGateMeanNewRescue, so that
\SWtrGateMeanNetRescueLoss{} of its \SWtrGateMeanNetLoss{} net lost successes are lost rescues. Across all three gate
families, how much the correction is attenuated matters more than where these
context-dependent gates attenuate it.

\begin{table}[t]
\centering\footnotesize
\caption{\textbf{What the disagreement gates change.} Scenes whose outcome differs between each gated
editor of the bottom block of Table~\ref{tab:sw-corrsize-gates} and the same editor without the gate
(900 matched scenes). New and lost rescues are scenes in which the base fails; repaired and new harms are
scenes in which it succeeds. Changed: share of the 900 scenes; net: successes gained by the gate.}
\label{tab:sw-corrsize-gate-transitions}
\setlength{\tabcolsep}{3pt}
\begin{tabularx}{\linewidth}{@{}>{\raggedright\arraybackslash}Xrrrrrr@{}}
\toprule
 & \multicolumn{2}{c}{Rescues} & \multicolumn{2}{c}{Harms} & & \\
\cmidrule(lr){2-3} \cmidrule(lr){4-5}
Gated editor & New & Lost & Repaired & New & Changed (\%) & Net \\
\midrule
\swatch{recfull}~L1 & 49 & 49 & 16 & 14 & 14.2 & +2 \\
\swatch{recadd}~L2 & 39 & 30 & 11 & 17 & 10.8 & +3 \\
\swatch{recbnd}~L3 & 65 & 56 & 32 & 23 & 19.6 & +18 \\
Mean ens. & 41 & 67 & 11 & 20 & 15.4 & $-$35 \\
\bottomrule
\end{tabularx}
\end{table}

\section{Comparisons Extended to All Recipes}
\label{app:sw-completions}

Each subsection completes a comparison that the main text details for one
recipe or one cohort, using its checkpoints and saved snapshots of the same
runs. The added training seeds of the 112-task editors, the two-source soft-penalty
editors and the main-camera L1 and L3 editors are further training runs with
no change to the method or its hyperparameters; the adaptation runs of
Appendix~\ref{app:sw-adaptation-replay} add source replay to the batch.
Rescue and harm are shares of
all paired episodes (Appendix~\ref{app:sw-guide}).

\subsection{Does held-out transfer appear early for the soft-penalty recipes too?}
\label{app:sw-completions-evolution}

\begin{figure}[t]
    \centering
    \includegraphics[width=\linewidth]{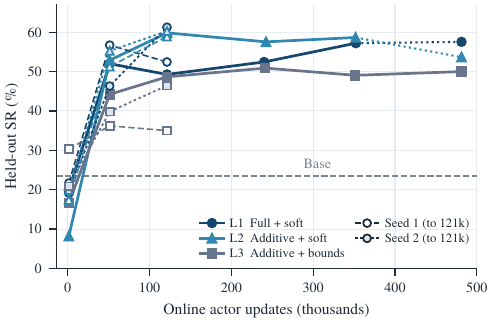}
    \caption{\textbf{Held-out SR across training for the three 112-task
    recipes} (18 held-out tasks, two passes). Filled markers are the
    runs of the main text: the L1 and L2 snapshots share one evaluation batch,
    dotted segments join them to their 481.25k endpoints from
    Table~\ref{tab:learner-classes}, which were evaluated separately, and the
    bounded recipe repeats Table~\ref{tab:libero-evolution}. Open and tinted
    markers trace two added training seeds (seeds~1 and~2), each run to
    121.25k updates. The dashed line marks the matched five-action base.}
    \label{fig:sw-completions-evolution}
\end{figure}

\emph{Yes: both soft-penalty editors reach at least 52\% held-out SR after a tenth of
training, and L2 is highest at 121.54k updates, before its endpoint.}
Figure~\ref{fig:libero-reference-training}a traces held-out SR across
training for all three recipes. For L1 and L2 it uses five saved snapshots
of each run, evaluated on the 18 held-out tasks under the two-pass protocol
and on 10 states of each source task (Table~\ref{tab:sw-completions-evolution},
Figure~\ref{fig:sw-completions-evolution}).

\begin{table}[t]
\centering\footnotesize
\caption{\textbf{Held-out and source SR (\%) of the two soft-penalty 112-task editors across training.} Held-out:
18 tasks, 1,800 pairs per snapshot, base SR 23.44\%; rescue and harm are shares of those pairs. Source:
10 states per task in one pass, base SR 21.07\%. Rows 121, 241 and 352 are the nearest saved snapshots
(L1: 121.25k, 240.00k, 352.20k;
L2: 121.54k, 242.27k, 351.81k).
$^\dagger$Endpoints of Table~\ref{tab:learner-classes}, from a separate batch (base SR 23.33\%;
source SR over 11,200 pairs). The 121k snapshots are also evaluated in the separate batch of
Table~\ref{tab:sw-robustness-seeds112}; the two evaluations of a snapshot differ by repeat-evaluation
variation (Appendix~\ref{app:sw-robustness-repeat}). Second and third training seeds: two further runs per recipe to 121.25k updates (base SR 23.44\% for seed~1 and 23.17\% for seed~2; source base 20.71\% for seed~2).}
\label{tab:sw-completions-evolution}
\setlength{\tabcolsep}{3.1pt}
\begin{tabular}{@{}lrrrrrrrr@{}}
\toprule
 & \multicolumn{4}{c}{L1 (full action + soft)} & \multicolumn{4}{c}{L2 (additive + soft)} \\
\cmidrule(lr){2-5} \cmidrule(l){6-9}
Updates (k) & SR & Resc. & Harm & Src. & SR & Resc. & Harm & Src. \\
\midrule
1.25 & 19.94 & 12.0 & 15.5 & 26.52 & 8.17 & 4.5 & 19.8 & 16.96 \\
51.25 & 52.00 & 32.8 & 4.2 & 63.12 & 52.89 & 37.4 & 8.0 & 70.00 \\
121 & 49.28 & 37.0 & 11.2 & 86.61 & 59.83 & 40.9 & 4.6 & 87.23 \\
241 & 52.44 & 37.1 & 8.1 & 92.95 & 57.56 & 40.7 & 6.6 & 93.21 \\
352 & 57.22 & 40.9 & 7.1 & 94.46 & 58.67 & 42.0 & 6.8 & 93.48 \\
\hlrow
481.25$^\dagger$ & 57.56 & 40.4 & 6.2 & 94.72 & 53.67 & 35.2 & 4.9 & 94.70 \\
\midrule
\multicolumn{9}{@{}l}{\emph{Second training seed}} \\
1.25 & 21.67 & 13.4 & 15.2 & -- & 17.39 & 8.3 & 14.3 & -- \\
51.25 & 56.72 & 38.8 & 5.5 & -- & 51.28 & 35.4 & 7.6 & -- \\
121 & 52.44 & 36.2 & 7.2 & 86.07 & 58.89 & 42.8 & 7.4 & 86.61 \\
\midrule
\multicolumn{9}{@{}l}{\emph{Third training seed}} \\
1.25 & 19.17 & 10.4 & 14.4 & -- & 19.89 & 9.1 & 12.3 & -- \\
51.25 & 46.33 & 32.1 & 8.9 & -- & 55.28 & 39.9 & 7.8 & -- \\
121 & 61.28 & 44.4 & 6.3 & 85.00 & 60.28 & 42.4 & 5.3 & 84.46 \\
\bottomrule
\end{tabular}
\end{table}

\paragraph{Transfer is present at 51.25k updates for all three recipes.}
At 1.25k updates both soft-penalty editors are below the base, L1 by 3.50
points [0.78, 6.22] and L2 by 15.28 points [12.83, 17.67], and harm exceeds
rescue. At 51.25k updates L1 reaches 52.00\% and L2 52.89\%, where the
bounded recipe is at 44.22\% (Table~\ref{tab:libero-evolution}). Both added
training seeds follow the same early-transfer pattern: at 51.25k updates, L1
reaches 56.72\% (seed~1) and 46.33\% (seed~2), L2 reaches 51.28\% and
55.28\%, and bounded L3 reaches 36.22\% and 39.83\%
(Appendix~\ref{app:sw-robustness}).

\paragraph{L1 keeps rising to its endpoint.}
L1's held-out SR at 352.20k updates is 7.94 points [5.17, 10.78] above that
of its 121.25k snapshot. The 352.20k snapshot differs from two repeat
evaluations of the 481.25k endpoint by $-$0.61 and +0.39 points, with
intervals that include zero.

\paragraph{L2 is highest before its endpoint.}
L2 reaches 59.83\% at 121.54k updates and does not rise afterwards: that
snapshot exceeds the 351.81k one by 1.17 points [$-$1.56, +3.94]. Its
endpoint is lower, at 53.67\% in the batch of
Table~\ref{tab:learner-classes}.

Paired on the same initial states with the two repeat evaluations of that
endpoint in Appendix~\ref{app:sw-robustness}, the 351.81k snapshot is 5.28
points [2.44, 8.06] and 4.39 points [1.61, 7.11] higher. This pairing
crosses evaluation batches, in which repeat evaluations of one checkpoint
differ by at most 1.00 point. L2's source SR is 93.48\% at 351.81k updates
and 94.70\% at the endpoint. The main text reports the last snapshot of each
run, so no snapshot is selected on held-out tasks.

At 121k updates L2 is above L1, and at the 481.25k endpoints of
Table~\ref{tab:learner-classes} L1 is above L2, so the order of the two
soft-penalty recipes depends on training length. Early transfer holds for
all three recipes, and for L2 held-out SR does not follow source SR late in
training.

\subsection{Does expert routing change held-out transfer?}
\label{app:sw-completions-moe}

\emph{The readout editors' held-out transfer is not distinguishable with and
without routing; the two routed encoder runs differ by 8.83 points, a
difference that one run per variant cannot attribute to routing
supervision.}
The bottom block of Table~\ref{tab:learner-classes} gives source SR for four
routed 16-source-task editors. We evaluate the same four checkpoints on the
18 held-out tasks under the two-pass protocol and re-evaluate them on the
source tasks, where SR is within 1.5 points of the values in that block
(Table~\ref{tab:sw-completions-moe}).

\begin{table*}[t]
\centering\footnotesize
\caption{\textbf{Held-out transfer of the routed 16-source-task editors} (additive + soft penalty; four experts, two
selected per decision; ``+ sup.'' adds task-cluster routing supervision). Each MoE variant is one training run,
evaluated on the 18 held-out tasks (1,800 pairs; base SR 24.22\%) and re-evaluated on the
16 source tasks. Rescue and harm are shares of the held-out pairs; intervals are 95\% initial-state-cluster
intervals. Cov.\,7 / Unc.\,11: SR on held-out tasks whose scene does or does not contain one of the 16 source tasks.
Entropy: mean routing entropy on held-out decisions in nats ($\ln 4 = 1.39$ when uniform). $^\dagger$MLP rows repeat
Tables~\ref{tab:learner-classes} and~\ref{tab:six-recipe-heldout-summary} (separate batch, own base).}
\label{tab:sw-completions-moe}
\setlength{\tabcolsep}{3pt}
\begin{tabular*}{\textwidth}{@{\extracolsep{\fill}}llrrrrrrrrrr@{}}
\toprule
 & & Source & \multicolumn{5}{c}{Held-out tasks (18)} & \multicolumn{2}{c}{Held-out SR by scene} & & + sup.\ minus MoE \\
\cmidrule(lr){4-8} \cmidrule(lr){9-10}
Representation & Actor & SR & Base & SR & Resc. & Harm & Gain [95\% CI] & Cov.\,7 & Unc.\,11 & Entropy & [95\% CI] \\
\midrule
R2 (readout) & MLP$^\dagger$ & 77.00 & 23.83 & 41.06 & 27.6 & 10.3 & +17.22 & 46.00 & 37.91 & -- & -- \\
 & MoE & 74.94 & 24.22 & 43.33 & 27.9 & 8.8 & +19.11 [+16.22, +22.00] & 61.00 & 32.09 & 1.35 & -- \\
 & MoE + sup. & 66.88 & 24.22 & 42.61 & 25.6 & 7.2 & +18.39 [+15.56, +21.22] & 60.71 & 31.09 & 1.37 & $-$0.72 [$-$3.56, +2.06] \\
\midrule
L2 (encoder) & MLP$^\dagger$ & 91.63 & 23.83 & 45.06 & 27.9 & 6.7 & +21.22 & 61.29 & 34.73 & -- & -- \\
 & MoE & 89.63 & 24.22 & 39.61 & 26.3 & 10.9 & +15.39 [+12.56, +18.17] & 49.57 & 33.27 & 1.33 & -- \\
 & MoE + sup. & 94.44 & 24.22 & 48.44 & 32.2 & 8.0 & +24.22 [+21.44, +27.06] & 59.86 & 41.18 & 1.37 & +8.83 [+5.94, +11.61] \\
\bottomrule
\end{tabular*}
\end{table*}

\paragraph{The readout editors transfer alike with and without routing.}
The readout MoE and its supervised variant gain 19.11 and 18.39 points over
the base; the supervised variant is 0.72 points lower [$-$3.56, +2.06]. The MLP editor
gains 17.22 points over its own base. The supervised MoE has the lowest
source SR of the three (66.88\%) and a held-out gain between the other two,
so source SR does not order these editors by transfer.

\paragraph{The two routed encoder runs differ, at the scale of seed spread.}
The encoder MoE reaches 39.61\% and its supervised variant 48.44\%, a
difference of +8.83 points [+5.94, +11.61] that is present in covered and
uncovered scenes. Each variant is one training run, and the interval covers
evaluation states only. Across three training seeds the L2 and L1 MLP
editors span 2.61 and 7.67 points of held-out SR
(Appendix~\ref{app:sw-robustness}), so a difference of this size between two
runs is not evidence that supervision produces it. Routing is also not
sharper under supervision: mean routing entropy on held-out tasks is 1.33 to
1.37 nats in all four variants, close to the uniform value. At 16 source
tasks, these single runs do not establish an effect of expert routing on
held-out transfer.

\subsection{Does the editor's instruction matter for any recipe?}
\label{app:sw-completions-instruction}

\emph{No: replacing the editor's instruction changes held-out SR by at most
1.67 points for any of the three recipes, and every interval includes zero.}
Appendix~\ref{app:libero-controls} replaces the instruction for the bounded recipe on six held-out tasks. Here each 112-task editor runs on all 18 held-out tasks with the
instruction of another task while the base keeps the target instruction. The
six earlier task pairs are kept, and each remaining task takes the
instruction of the nearest task in its suite
(Table~\ref{tab:sw-completions-instruction}).

\begin{table}[t]
\centering\footnotesize
\caption{\textbf{Replacing only the editor's instruction}, for the three 112-task editors on the 18 held-out tasks (first
pass, 900 pairs per condition; base SR 25.00\%). $\Delta$: replaced minus original SR in
points, with a 95\% initial-state-cluster interval, and on the 10 tasks in source-covered and the 8 in uncovered
scenes. Corr.\ change: change of the editor's correction on the same context, in percent of the correction norm
(mean and 90th percentile over decisions).}
\label{tab:sw-completions-instruction}
\setlength{\tabcolsep}{2.6pt}
\begin{tabularx}{\linewidth}{@{}Xrrrrrrr@{}}
\toprule
 & \multicolumn{2}{c}{SR} & & \multicolumn{2}{c}{$\Delta$ by scene} & \multicolumn{2}{c}{Corr.\ change} \\
\cmidrule(lr){2-3} \cmidrule(lr){5-6} \cmidrule(l){7-8}
 & Orig. & Repl. & $\Delta$ [95\% CI] & Cov. & Unc. & Mean & P90 \\
\midrule
L1 & 56.00 & 57.44 & +1.44 [$-$1.22, +4.22] & +1.20 & +1.75 & 0.05 & 0.14 \\
L2 & 55.56 & 53.89 & $-$1.67 [$-$4.44, +1.11] & $-$3.20 & +0.25 & 0.03 & 0.07 \\
L3 & 49.33 & 49.33 & 0.00 [$-$2.67, +2.44] & 0.00 & 0.00 & 0.03 & 0.07 \\
\bottomrule
\end{tabularx}
\end{table}

Replacing the instruction changes SR by +1.44 points for L1, $-$1.67 for L2
and 0.00 for L3, and no interval excludes zero in either scene cohort. The
recorded corrections show why. On the same context and base proposal, the
replaced instruction changes the editor's correction by 0.03\% to 0.05\% of
its norm on average and by at most 0.14\% at the 90th percentile. SR
differences of this size also arise between two evaluations of one policy
(Appendix~\ref{app:sw-robustness}). For all three recipes the correction is
insensitive to the editor's instruction, which extends that six-task result to every recipe and every held-out task.

\subsection{Does the source advantage of the encoder recipes survive with the main camera only?}
\label{app:sw-completions-cameras}

\emph{On source tasks every main-camera encoder still exceeds its readout
counterpart; on held-out tasks none of the three main-camera encoder recipes
exceeds its readout counterpart.}
The readout recipes see the main camera and the encoder recipes normally see
the main and wrist cameras. Table~\ref{tab:no-wrist-summary} reports L2
trained on the main camera only; we add the same 100k-update main-camera-only
training for both L1 and bounded L3 and place all three next to the readout
and dual-camera rows of Table~\ref{tab:learner-classes}
(Table~\ref{tab:sw-completions-cameras}).

\begin{table}[t]
\centering\footnotesize
\caption{\textbf{Camera-matched comparison of the readout (R) and encoder (L) recipes at 16 source tasks} (100k updates,
one training run per row; 1,600 source and 1,800 held-out pairs). Shaded rows train the encoder editor on the main
camera only; the others repeat Tables~\ref{tab:learner-classes} and~\ref{tab:six-recipe-heldout-cohorts}.
Source / held-out base SR is 22.06 / 23.83\% for the unshaded rows,
21.75 / 23.50\% for main-camera L1, 21.69 / 23.67\% for main-camera L2 and
23.75 / 24.39\% for main-camera L3. Rescue and harm are shares of the
held-out pairs; gains are in points. Cov.\,7 / Unc.\,11: held-out gain in scenes with and without one of the 16
source tasks (the 7 / 11 split of Table~\ref{tab:learner-classes}).}
\label{tab:sw-completions-cameras}
\setlength{\tabcolsep}{2.6pt}
\begin{tabularx}{\linewidth}{@{}>{\raggedright\arraybackslash}Xrrrrrrr@{}}
\toprule
 & Source & \multicolumn{4}{c}{Held-out tasks (18)} & \multicolumn{2}{c}{Gain by scene} \\
\cmidrule(lr){3-6} \cmidrule(l){7-8}
Recipe, cameras & SR & SR & Resc. & Harm & Gain & Cov.\,7 & Unc.\,11 \\
\midrule
R1, main & 61.75 & 40.28 & 26.4 & 10.0 & +16.44 & +24.57 & +11.27 \\
L1, main + wrist & 93.00 & 40.28 & 24.9 & 8.4 & +16.44 & +36.57 & +3.64 \\
\hlrow
L1, main & 87.31 & 34.17 & 22.7 & 12.0 & +10.67 & +31.86 & $-$2.82 \\
\midrule
R2, main & 77.00 & 41.06 & 27.6 & 10.3 & +17.22 & +16.57 & +17.64 \\
L2, main + wrist & 91.63 & 45.06 & 27.9 & 6.7 & +21.22 & +31.86 & +14.45 \\
\hlrow
L2, main & 79.94 & 40.56 & 27.0 & 10.1 & +16.89 & +26.43 & +10.82 \\
\midrule
R3, main & 56.88 & 39.22 & 24.1 & 8.7 & +15.39 & +27.57 & +7.64 \\
L3, main + wrist & 85.63 & 23.11 & 15.8 & 16.5 & $-$0.72 & +12.29 & $-$9.00 \\
\hlrow
L3, main & 65.69 & 27.17 & 17.7 & 14.9 & +2.78 & +19.43 & $-$7.82 \\
\bottomrule
\end{tabularx}
\end{table}

\paragraph{With one camera, all three encoders retain a source advantage.}
With the main camera only, L1 reaches 87.31\% on the 16 source tasks (25.56
points [22.75, 28.38] above R1 and 5.69 points [3.50, 7.88] below dual-camera
L1), L2 reaches 79.94\% (2.94 points [0.44, 5.38] above R2), and L3 reaches
65.69\% (+41.94 [+38.69, +45.19] over its 23.75\% batch base, against 56.88\%
for R3 and 85.63\% for dual-camera L3).

\paragraph{No main-camera encoder transfers better than its readout counterpart.}
On the 18 held-out tasks, main-camera L1 gains 10.67 points [8.00, 13.39]
(vs.\ +16.44 for R1 and dual-camera L1), with $-$2.82 [$-$5.82, +0.27] in
uncovered scenes. Main-camera L2 gains 16.89 points [14.00, 19.72] (close to
R2 at +17.22 and below dual-camera L2 at +21.22). Main-camera L3 reaches
27.17\% SR against its 24.39\% batch base (+2.78 [+0.06, +5.50]; rescue
17.7\%, harm 14.9\%; +19.43 in covered scenes and $-$7.82 in uncovered
scenes), slightly above dual-camera L3 ($-$0.72) yet 12.61 points below R3
(+15.39). Each main-camera row is one training run. Across all three editing
choices, matching the camera input preserves the encoder's source advantage
while leaving no held-out advantage over the fixed RL-token readout at 16
source tasks.

\subsection{Does narrow-source negative transfer hold for every recipe?}
\label{app:sw-completions-twosource}

\emph{Yes: trained on the same two source tasks, L1 and L2 also fall below
the frozen base on the 18 held-out tasks despite high source success.}
Section~\ref{sec:experiments:libero-reference} traces source breadth from 2 to
16 to 112 tasks under bounded L3. To test whether narrow-source negative
transfer is specific to output bounds, we train L1 and L2 on the same two
LIBERO-Long source tasks for 250k actor updates (seed~0) and evaluate them
under the two-pass protocol (Table~\ref{tab:sw-completions-twosource}).

\begin{table}[t]
\centering\footnotesize
\caption{\textbf{Source breadth across all three online-encoder recipes} (2, 16 and 112 source tasks; seed~0).
Two-source editors train on the same two LIBERO-Long source tasks for 250k actor updates (source base SR
18.00\%, 200 pairs; held-out base SR 23.00\% for L1, 23.22\% for L2 and 23.33\% for L3, 1,800
pairs). The 16- and 112-task rows repeat Table~\ref{tab:learner-classes} and
Tables~\ref{tab:six-recipe-heldout-summary} and~\ref{tab:libero-reference}. Rescue and harm are shares
of held-out pairs; intervals are 95\% initial-state-cluster bootstrap CIs on the fixed 18 held-out tasks.}
\label{tab:sw-completions-twosource}
\setlength{\tabcolsep}{2.4pt}
\begin{tabularx}{\linewidth}{@{}llrrrrr@{}}
\toprule
 & & Source & \multicolumn{4}{c}{Held-out tasks (18)} \\
\cmidrule(l){4-7}
Rec. & Src. & SR & SR & Resc. & Harm & Gain [95\% CI] \\
\midrule
L1 & 2 & 87.00 & 5.44 & 4.1 & 21.6 & $-$17.56 [$-$19.83, $-$15.22] \\
 & 16 & 93.00 & 40.28 & 24.9 & 8.4 & +16.44 [+13.6, +19.2] \\
 & 112 & 94.72 & 57.56 & 40.4 & 6.2 & +34.22 [+31.6, +36.8] \\
\midrule
L2 & 2 & 92.00 & 19.72 & 10.7 & 14.2 & $-$3.50 [$-$5.94, $-$1.06] \\
 & 16 & 91.63 & 45.06 & 27.9 & 6.7 & +21.22 [+18.6, +23.9] \\
 & 112 & 94.70 & 53.67 & 35.2 & 4.9 & +30.33 [+27.8, +32.8] \\
\midrule
L3 & 2 & 93.50 & 1.00 & 0.5 & 22.8 & $-$22.33 [$-$24.5, $-$20.2] \\
 & 16 & 85.63 & 23.11 & 15.8 & 16.5 & $-$0.72 [$-$3.3, +1.8] \\
 & 112 & 93.55 & 50.00 & 36.8 & 10.2 & +26.67 [+23.7, +29.5] \\
\bottomrule
\end{tabularx}
\end{table}

Both soft-penalty editors reach high success on the two source tasks (87.00\% for L1 and
92.00\% for L2, against 18.00\% base SR and 93.50\% for L3), yet both transfer
negatively to the 18 held-out tasks. Two-source L1 falls to 5.44\%
($-$17.56 [$-$19.83, $-$15.22] points below its 23.00\% base; rescue 4.1\%,
harm 21.6\%), and two-source L2 falls to 19.72\% ($-$3.50 [$-$5.94, $-$1.06]
below its 23.22\% base; rescue 10.7\%, harm 14.2\%). The additive soft-penalty
formulation (L2) loses fewer base successes than full-action L1 (harm 21.6\%) or
bounded L3 (held-out SR 1.00\%, harm 22.8\%), but widening source breadth from 2 to 16 to 112 tasks
raises rescue and lowers harm under all three recipes.

\subsection{Does pretraining speed adaptation under the soft-penalty recipes?}
\label{app:sw-adaptation}

\emph{Not consistently: pretrained runs are ahead at 1k updates on one of
the two targets, the two initializations are level from 5k updates on both,
and source-task success falls under both recipes.}

\begin{figure*}[t]
    \centering
    \includegraphics[width=\textwidth]{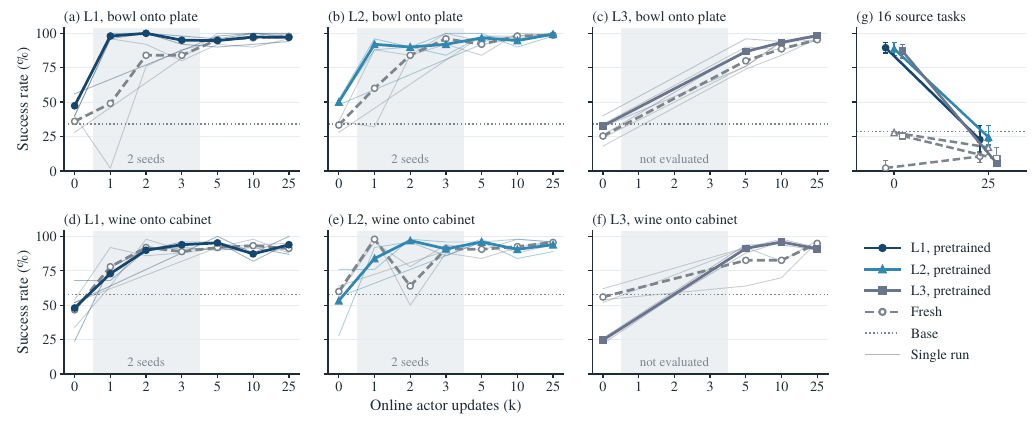}
    \caption{\textbf{Single-task adaptation by recipe and target over three
    training seeds.} (a)--(f) Target SR of L1, L2 and L3 on the two targets
    against online actor updates: thick lines are means over training seeds
    and thin lines are single runs. The shaded checkpoints were evaluated for
    two of the three seeds of L1 and L2 and not for L3. (g) SR on 16 source
    tasks before and after adaptation, as the mean and range over six runs
    per recipe and initialization. Dotted lines mark the base, measured in
    the batches of the two added L1 and L2 seeds.}
    \label{fig:sw-adaptation-curves}
\end{figure*}

Section~\ref{sec:experiments:adaptation} reports L1 and L2 adaptation as
means over three seeds per condition on two targets, placing the bowl beside
a plate onto that plate (bowl target) and placing wine onto a cabinet (wine
target), evaluated at 0, 5k, 10k and 25k online actor updates. This section
gives the runs behind those means. Seeds 1 and 2 of each of the eight
conditions start from the same 112-task editors or from fresh weights as
seed 0, use the same configuration without source replay, and are also
evaluated at 1k, 2k and 3k updates
(Figure~\ref{fig:sw-adaptation-curves},
Table~\ref{tab:sw-adaptation-seeds}). The figure also shows the three seeds
of the bounded recipe on the same targets
(Appendix~\ref{app:libero-adaptation}): its pretrained mean is above the
fresh mean at 5k and 10k updates on both targets, and below it at 25k
updates on the wine target.

\begin{table*}[t]
\centering\footnotesize
\caption{\textbf{Soft-penalty single-task adaptation over three training seeds.} The targets are placing the bowl beside a plate onto that plate and placing wine onto a cabinet. Each entry is the mean over training seeds; parentheses give the range over runs. Target SR uses 50 initial states (two passes at 25k); the 1k column uses the two seeds evaluated at that checkpoint. AUC is the trapezoid area under target SR at 0, 5k, 10k and 25k online actor updates divided by 25k (\%). Critic loss is the mean online critic loss ($\times 10^{-4}$). Source SR is the equal-task mean over 16 source tasks with ten states each, before and after adaptation; the matched base reaches 28.75\% on these episodes (31.25\% in one batch). Before update 0, fresh L1 performs 5,000 proposal-fitting updates and fresh L2 reproduces the base outcome on every episode of the two added seeds, which have a matched base. Section~\ref{sec:experiments:adaptation} reports the three-seed means; all runs use one configuration without source replay.}
\label{tab:sw-adaptation-seeds}
\setlength{\tabcolsep}{4.2pt}
\begin{tabularx}{\textwidth}{@{}>{\raggedright\arraybackslash}Xllrrrrrrr@{}}
\toprule
 & & & \multicolumn{3}{c}{Target SR at online actor updates} & & & \multicolumn{2}{c}{Source SR} \\
\cmidrule(lr){4-6} \cmidrule(l){9-10}
Target task & Recipe & Init. & 1k (2 seeds) & 5k & 25k & AUC & Critic loss & 0k & 25k \\
\midrule
Bowl onto plate & L1 & Pretrained & 98.00 (96--100) & 94.67 (90--98) & 97.00 (94--100) & 91.70 & 1.45 & 88.75 & 26.25 (17.50--33.13) \\
 &  & Fresh & 49.00 (2--96) & 95.33 (92--100) & 97.67 (96--99) & 90.63 & 2.65 & 4.17 & 9.17 (6.25--12.50) \\
\addlinespace[2pt]
 & L2 & Pretrained & 92.00 (88--96) & 96.67 (92--100) & 99.33 (98--100) & 92.00 & 1.37 & 89.17 & 22.71 (20.00--24.38) \\
 &  & Fresh & 60.00 (32--88) & 92.00 (84--98) & 98.67 (98--99) & 90.53 & 2.51 & 27.92 & 14.58 (11.88--17.50) \\
\midrule
Wine onto cabinet & L1 & Pretrained & 73.00 (68--78) & 95.33 (90--100) & 94.00 (87--100) & 87.00 & 1.90 & 90.21 & 19.17 (11.88--23.13) \\
 &  & Fresh & 78.00 (64--92) & 92.00 (92--92) & 91.33 (88--95) & 87.80 & 4.14 & 0.00 & 11.88 (10.00--13.75) \\
\addlinespace[2pt]
 & L2 & Pretrained & 84.00 (76--92) & 96.00 (96--96) & 94.00 (89--97) & 89.00 & 1.89 & 88.33 & 26.46 (18.75--33.13) \\
 &  & Fresh & 98.00 (96--100) & 90.67 (84--98) & 95.67 (94--97) & 89.90 & 3.59 & 27.71 & 19.58 (17.50--21.88) \\
\bottomrule
\end{tabularx}
\end{table*}

\paragraph{Pretrained runs lead at 1k updates on the bowl target only.}
On the bowl target, the four pretrained runs with a 1k checkpoint reach
88--100\% target SR. One fresh run of each recipe is still at 2\% (L1) and
32\% (L2) at that checkpoint. On the wine target the ranges overlap at 1k.
Pretrained runs reach 68--92\% there. Fresh runs reach 64--100\%. With two
seeds per condition at these checkpoints, the lead on the bowl target rests
on one fresh run per recipe.

\paragraph{From 5k updates the two initializations are level.}
At 5k updates, 22 of the 24 runs are at 90\% target SR or above; the two
exceptions are fresh L2 runs at 84\%. At 25k updates the pretrained runs reach 87--100\%. The
fresh runs reach 88--99\%.

\paragraph{Critic loss is lower under pretraining; the area under the SR curve is not higher.}
Mean online critic loss is lower under pretraining in all 12 run pairs
matched on recipe, target and seed, by 45.2--54.0\% in the three-seed means.
The three-seed mean AUC of pretrained runs differs from that of fresh runs by
$-$0.90 to +1.47 points across the four recipe--target pairs.

\paragraph{Source-task success falls during replay-free adaptation under both recipes.}
Before adaptation, every pretrained run is at 83.75\% SR or above on the 16
source tasks. After 25k updates on the target, the 12 pretrained runs reach
11.88--33.13\%. The three-seed means over both targets are 22.71\% for L1
and 24.58\% for L2.

\paragraph{Adapted editors end near or below the base on the source tasks.}
In the two added seeds, 7 of the 8 pretrained runs and all 8 fresh runs end
below the matched base of their own batch on the source tasks. On the
source tasks, both soft-penalty recipes end adaptation above the 5.42\%
source SR of the bounded recipe on the same targets (Section~\ref{sec:experiments:adaptation}).

\paragraph{Third target (Task~39, turn off stove, uncovered scene).}
To complete the three-target matrix shared with bounded L3
(Table~\ref{tab:libero-adaptation-full}), we run three seeds of pretrained
and fresh L1 and L2 on Task~39 under the same replay-free 25k-update protocol
with checkpoints at 0, 1k, 2k, 3k, 5k, 10k and 25k updates
(Table~\ref{tab:sw-adaptation-target39}). Pretraining lowers mean online
critic loss by 60.2\% for L1 (1.00 vs.\ 2.52, in units of $10^{-4}$) and 57.1\% for
L2 (1.21 vs.\ 2.81), with lower loss in all 6 seed-matched
pairs; combined with Tasks~98 and~112 and L3, pretraining lowers critic loss
in all 27 of 27 seed-matched pairs across the three targets and three
recipes. On target SR, pretrained L1 and L2 lead at 1k updates (90.67\% vs.\
64.00\% for L1; 80.67\% vs.\ 72.67\% for L2) and both initializations reach
91--99\% by 5k and 99--100\% by 25k. Over the seven checkpoints (AUC$_7$),
pretraining is slightly higher for both L1 (97.13 vs.\ 95.18) and L2 (95.96
vs.\ 93.61); over the four checkpoints shared with the main text (AUC$_4$),
pretrained L2 leads fresh L2 in all three seeds (91.40 vs.\ 88.27), while
fresh L1's 5k proposal-fitting warm start raises its 0k SR to 44.00\% (vs.\
15.33\% pretrained), making its 4-point AUC slightly higher (92.97 vs.\
91.27). Without source replay, source SR on the 16 tasks again falls
from 89.37\% to 20.83\% for L1 and from 87.50\% to 33.96\% for L2, both
above the 3.96\% of L3 on the same target.

\begin{table*}[t]
\centering\footnotesize
\caption{\textbf{Single-task adaptation on the third target (Task~39, turn off stove, uncovered scene; three training seeds).}
Matches the protocol of Table~\ref{tab:sw-adaptation-seeds} (base SR 5.0--6.0\% on Task~39, 24.4--24.8\% on
the 16 source tasks) with all three seeds evaluated at 0, 1k, 2k, 3k, 5k, 10k and 25k online actor updates for
L1 and L2. Parentheses give the range across the three seeds. AUC$_4$ uses the four checkpoints shared with
Section~\ref{sec:experiments:adaptation} (0, 5k, 10k, 25k); AUC$_7$ uses all seven checkpoints. Critic loss
is the mean online critic loss ($\times 10^{-4}$) over 999 logged records per run. $^\dagger$Bounded L3 rows
repeat Table~\ref{tab:libero-adaptation-full} (which omits 1k--3k evaluations).}
\label{tab:sw-adaptation-target39}
\setlength{\tabcolsep}{3.2pt}
\begin{tabularx}{\textwidth}{@{}llrrrrrrrrr@{}}
\toprule
 & & \multicolumn{4}{c}{Target SR at online actor updates} & & & & \multicolumn{2}{c}{Source SR} \\
\cmidrule(lr){3-6} \cmidrule(l){10-11}
Rec. & Init. & 0k & 1k & 5k & 25k & AUC$_4$ & AUC$_7$ & Critic loss & 0k & 25k \\
\midrule
L1 & Pretrained & 15.33 (12--22) & 90.67 (84--96) & 98.67 (96--100) & 100.00 (100--100) & 91.27 & 97.13 & 1.00 & 89.37 & 20.83 (18.13--23.13) \\
 & Fresh & 44.00 (30--52) & 64.00 (46--92) & 96.00 (90--100) & 99.67 (99--100) & 92.97 & 95.18 & 2.52 & 0.62 & 3.75 (0.63--10.00) \\
\addlinespace[2pt]
L2 & Pretrained & 23.33 (18--28) & 80.67 (72--94) & 96.67 (94--100) & 100.00 (100--100) & 91.40 & 95.96 & 1.21 & 87.50 & 33.96 (28.13--37.50) \\
 & Fresh & 6.00 (6--6) & 72.67 (50--94) & 90.67 (80--98) & 99.33 (99--100) & 88.27 & 93.61 & 2.81 & 24.58 & 16.46 (10.63--21.88) \\
\addlinespace[2pt]
\midrule
L3$^\dagger$ & Pretrained & 54.00 & -- & 88.67 & 99.67 & 91.97 & -- & 1.56 & 85.63 & 3.96 \\
 & Fresh & 6.00 & -- & 94.00 & 98.33 & 87.83 & -- & 3.33 & 24.38 & 1.67 \\
\bottomrule
\end{tabularx}
\end{table*}

\providecommand{\SWdrRetLtwo}{54.17}
\providecommand{\SWdrRetLtwoLo}{51.25}
\providecommand{\SWdrRetLtwoHi}{56.88}
\providecommand{\SWdrRetZeroLtwo}{78.75}
\providecommand{\SWdrNoRetLtwo}{24.58}
\providecommand{\SWdrNoRetLtwoLo}{18.75}
\providecommand{\SWdrNoRetLtwoHi}{33.13}
\providecommand{\SWdrNoRetZeroLtwo}{88.75}
\providecommand{\SWdrTgtLtwo}{98.00}
\providecommand{\SWdrNoTgtLtwo}{96.67}
\providecommand{\SWdrAucLtwo}{90.63}
\providecommand{\SWdrNoAucLtwo}{90.50}
\providecommand{\SWdrOneKLtwo}{82.33}
\providecommand{\SWdrOneKLtwoLo}{74}
\providecommand{\SWdrOneKLtwoHi}{88}
\providecommand{\SWdrAboveLtwo}{13}
\providecommand{\SWdrGpuhLoLtwo}{1.51}
\providecommand{\SWdrGpuhHiLtwo}{1.60}
\providecommand{\SWdrRetLthree}{20.63}
\providecommand{\SWdrRetLthreeLo}{13.13}
\providecommand{\SWdrRetLthreeHi}{26.88}
\providecommand{\SWdrRetZeroLthree}{74.69}
\providecommand{\SWdrNoRetLthree}{5.42}
\providecommand{\SWdrNoRetLthreeLo}{3.13}
\providecommand{\SWdrNoRetLthreeHi}{8.13}
\providecommand{\SWdrNoRetZeroLthree}{86.98}
\providecommand{\SWdrTgtLthree}{95.67}
\providecommand{\SWdrNoTgtLthree}{94.67}
\providecommand{\SWdrAucLthree}{86.10}
\providecommand{\SWdrNoAucLthree}{86.93}
\providecommand{\SWdrOneKLthree}{3.00}
\providecommand{\SWdrOneKLthreeLo}{0}
\providecommand{\SWdrOneKLthreeHi}{12}
\providecommand{\SWdrAboveLthree}{4}
\providecommand{\SWdrGpuhLoLthree}{1.43}
\providecommand{\SWdrGpuhHiLthree}{1.49}
\providecommand{\SWdrBase}{24.38}
\providecommand{\SWdrBaseLo}{22.50}
\providecommand{\SWdrBaseHi}{25.63}
\providecommand{\SWdrZeroLthree}{27.33}
\providecommand{\SWdrThreeKLthree}{78.33}
\providecommand{\SWdrFiveKLthree}{87.33}
\providecommand{\SWdrGapNo}{19.17}
\providecommand{\SWdrGapReplay}{33.54}
\providecommand{\SWdrPairLtwoBowl}{+31.88}
\providecommand{\SWdrPairLtwoBowlCI}{[25.63, 37.50]}
\providecommand{\SWdrPairLtwoWine}{+23.13}
\providecommand{\SWdrPairLtwoWineCI}{[15.63, 30.00]}
\providecommand{\SWdrPairLthreeLo}{+6.88}
\providecommand{\SWdrPairLthreeHi}{+23.13}
\providecommand{\SWdrPairTgtLtwoWine}{+9.00}
\providecommand{\SWdrPairTgtLtwoWineCI}{[3.00, 16.00]}
\providecommand{\SWdrPairTgtNeg}{$-$6.00}
\providecommand{\SWdrPairTgtNegCI}{[$-$11.00, $-$1.00]}
\providecommand{\SWdrGpuhLo}{1.43}
\providecommand{\SWdrGpuhHi}{1.60}
\providecommand{\SWdrRateLo}{5.04}
\providecommand{\SWdrRateHi}{5.31}

\subsection{Does source replay preserve source skills during single-task adaptation?}
\label{app:sw-adaptation-replay}

\emph{For the additive soft-penalty recipe, in part: source SR after
adaptation is \SWdrRetLtwo\% with replay against \SWdrNoRetLtwo\% without,
and target success is not lower. The bounded recipe reaches
\SWdrRetLthree\% and stays below the base.}

\begin{figure}[t]
    \centering
    \includegraphics[width=\linewidth]{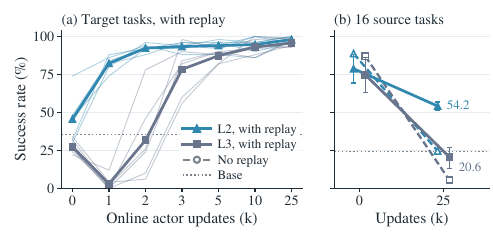}
    \caption{\textbf{Single-task adaptation with source replay.} (a) Target
    SR of pretrained L2 and L3 across online actor updates, as the mean over
    two targets and three seeds; thin lines are single runs. (b) SR on the 16
    source tasks before and after adaptation: mean and range over the six
    runs with replay, and the three-seed means without replay (dashed, open
    markers). Dotted lines mark the base.}
    \label{fig:sw-adaptation-replay}
\end{figure}

The adaptation runs of Section~\ref{sec:experiments:adaptation} and
Appendix~\ref{app:sw-adaptation} train on target data alone. Here every batch
of 256 rows is half target data and half replay of the 16 source tasks on
which retention is evaluated, one source task per batch in a reshuffled
cycle, with equal demonstration and online shares in each half. The source
replay is frozen: it is copied from the saved learner state of the 112-task
run (481.25k updates for L2, 465.11k for L3), and no source environment runs
during adaptation. The actor and encoder start from the 481.25k snapshots as
before, and the target critic and the 16 source critics start fresh. The
160k-sample critic and encoder warm-up uses the same mixture; collection, the
25k-update budget and the evaluation protocol follow
Appendix~\ref{app:sw-adaptation}. Pretrained L2 and L3 are run on the bowl
and wine targets with three seeds each
(Table~\ref{tab:sw-adaptation-replay},
Figure~\ref{fig:sw-adaptation-replay}); L1 was not run.

\begin{table*}[t]
\centering\footnotesize
\caption{\textbf{Single-task adaptation with source replay over three training seeds.} Pretrained L2 and L3 adapt
to the bowl and wine targets of Table~\ref{tab:sw-adaptation-seeds} with every batch half target data and half frozen
replay of the 16 source tasks. Each entry is the mean over training seeds; parentheses give the range over runs. Target SR
uses 50 initial states (two passes at 25k). AUC$_4$ uses the checkpoints at 0, 5k, 10k and 25k online actor updates
and AUC$_7$ all seven. Source SR is the equal-task mean over the 16 replayed source tasks with ten states each;
the base reaches 24.38\% on these episodes. The last three columns repeat the runs without replay
(Tables~\ref{tab:sw-adaptation-seeds} and~\ref{tab:libero-adaptation-full}).}
\label{tab:sw-adaptation-replay}
\setlength{\tabcolsep}{3.1pt}
\begin{tabularx}{\textwidth}{@{}l>{\raggedright\arraybackslash}Xrrrrrrrrrrrrrr@{}}
\toprule
 & & \multicolumn{7}{c}{Target SR at online actor updates} & & & \multicolumn{2}{c}{Source SR} & \multicolumn{3}{c}{Without replay} \\
\cmidrule(lr){3-9} \cmidrule(lr){12-13} \cmidrule(l){14-16}
Rec. & Target & 0k & 1k & 2k & 3k & 5k & 10k & 25k & AUC$_4$ & AUC$_7$ & 0k & 25k & 25k & AUC$_4$ & Source \\
\midrule
L2 & Bowl & 40.67 & 82.00 (80--84) & 92.00 & 95.33 & 96.00 & 97.33 & 99.67 & 92.10 & 95.77 & 77.92 & 52.92 (51.25--55.63) & 99.33 & 92.00 & 22.71 \\
 & Wine & 50.67 & 82.67 (74--88) & 92.67 & 91.33 & 92.00 & 92.00 & 96.33 & 89.17 & 92.09 & 79.58 & 55.42 (53.13--56.88) & 94.00 & 89.00 & 26.46 \\
\sumrow
 & Both & 45.67 & 82.33 (74--88) & 92.33 & 93.33 & 94.00 & 94.67 & 98.00 & 90.63 & 93.93 & 78.75 & 54.17 (51.25--56.88) & 96.67 & 90.50 & 24.58 \\
\midrule
L3 & Bowl & 28.00 & 2.00 (0--4) & 18.67 & 79.33 & 88.00 & 96.67 & 96.67 & 88.07 & 86.13 & 71.04 & 22.50 (18.13--26.88) & 98.33 & 87.43 & 5.83 \\
 & Wine & 26.67 & 4.00 (0--12) & 44.67 & 77.33 & 86.67 & 89.33 & 94.67 & 84.13 & 83.39 & 78.33 & 18.75 (13.13--26.25) & 91.00 & 86.43 & 5.00 \\
\sumrow
 & Both & 27.33 & 3.00 (0--12) & 31.67 & 78.33 & 87.33 & 93.00 & 95.67 & 86.10 & 84.76 & 74.69 & 20.63 (13.13--26.88) & 94.67 & 86.93 & 5.42 \\
\bottomrule
\end{tabularx}
\end{table*}

\paragraph{With replay, L2 keeps more than twice the source SR.}
After 25k updates L2 reaches \SWdrRetLtwo\% on the 16 source tasks
(\SWdrRetLtwoLo--\SWdrRetLtwoHi\% over the six runs), against
\SWdrNoRetLtwo\% (\SWdrNoRetLtwoLo--\SWdrNoRetLtwoHi\%) without replay and
\SWdrBase\% for the base on the same episodes; the two ranges do not overlap.
In the six-run mean, \SWdrAboveLtwo{} of the 16 source tasks are above the
base. Paired on initial states, seed~0 gains \SWdrPairLtwoBowl{}
\SWdrPairLtwoBowlCI{} points on the bowl target and \SWdrPairLtwoWine{}
\SWdrPairLtwoWineCI{} on the wine target
(Table~\ref{tab:sw-adaptation-replay-paired}).

\begin{table}[t]
\centering\footnotesize
\caption{\textbf{Seed-matched differences between adaptation with and without source replay} at 25k updates (with
minus without, in points), paired on initial states: 160 source-task and 100 target episodes per run. Intervals
resample initial states within tasks. The runs without replay have matched records for all three L3 seeds and for
seed 0 of L2.}
\label{tab:sw-adaptation-replay-paired}
\setlength{\tabcolsep}{3.2pt}
\begin{tabularx}{\linewidth}{@{}ll>{\centering\arraybackslash}Xrr@{}}
\toprule
Rec. & Target & Seed & Source SR [95\% CI] & Target SR [95\% CI] \\
\midrule
L2 & Bowl & 0 & +31.88 [25.63, 37.50] & 0.00 [0.00, 0.00] \\
 & Wine & 0 & +23.13 [15.63, 30.00] & +9.00 [3.00, 16.00] \\
\midrule
L3 & Bowl & 0 & +19.38 [14.38, 24.38] & 0.00 [$-$3.00, 3.00] \\
 & Bowl & 1 & +18.75 [13.75, 23.75] & +1.00 [$-$4.00, 7.00] \\
 & Bowl & 2 & +11.88 [6.88, 16.27] & $-$6.00 [$-$11.00, $-$1.00] \\
 & Wine & 0 & +23.13 [18.13, 28.13] & +3.00 [$-$3.00, 10.00] \\
 & Wine & 1 & +11.25 [6.88, 15.63] & +2.00 [$-$4.00, 8.00] \\
 & Wine & 2 & +6.88 [1.88, 11.25] & +6.00 [0.00, 12.00] \\
\bottomrule
\end{tabularx}
\end{table}

\paragraph{The bounded recipe gains from replay and stays below the base.}
L3 reaches \SWdrRetLthree\% (\SWdrRetLthreeLo--\SWdrRetLthreeHi\%), against
\SWdrNoRetLthree\% without replay. The paired difference is positive in all
six seed--target pairs, between \SWdrPairLthreeLo{} and \SWdrPairLthreeHi{}
points with every interval excluding zero. The mean is still below the base
and below L2 without replay, and \SWdrAboveLthree{} of the 16 source tasks
are above the base. L2 exceeds L3 by \SWdrGapNo{} points without replay and
by \SWdrGapReplay{} with it.

\paragraph{Target success is not lower with replay.}
At 25k updates target SR is \SWdrTgtLtwo\% for L2 and \SWdrTgtLthree\% for
L3, against \SWdrNoTgtLtwo\% and \SWdrNoTgtLthree\% without replay, and
AUC$_4$ is \SWdrAucLtwo{} and \SWdrAucLthree{} against \SWdrNoAucLtwo{} and
\SWdrNoAucLthree. Of the eight seed-matched pairs, one L3 run is lower with
replay (\SWdrPairTgtNeg{} \SWdrPairTgtNegCI{} points) and L2 on the wine
target is higher (\SWdrPairTgtLtwoWine{} \SWdrPairTgtLtwoWineCI); the other
six intervals include zero.

\paragraph{L3 loses the target early in adaptation; L2 does not.}
At 1k updates L3's target SR is \SWdrOneKLthree\%
(\SWdrOneKLthreeLo--\SWdrOneKLthreeHi\% over the six runs), below its
\SWdrZeroLthree\% at update~0; it returns to \SWdrThreeKLthree\% at 3k and
\SWdrFiveKLthree\% at 5k. L2 is at \SWdrOneKLtwo\%
(\SWdrOneKLtwoLo--\SWdrOneKLtwoHi\%) at 1k. The L3 runs without replay were
not evaluated between 0 and 5k updates, so the data do not attribute the dip
to the replay mixture.

\paragraph{Retention is partial, and its scope is the replayed tasks.}
At update~0, after the mixed warm-up, source SR is already
\SWdrRetZeroLtwo\% for L2 and \SWdrRetZeroLthree\% for L3, against
\SWdrNoRetZeroLtwo\% and \SWdrNoRetZeroLthree\% in the runs without replay,
and it falls further over the 25k updates. The replay covers the 16 evaluated
source tasks and not the other 96, and it holds recorded transitions only:
the result does not show what online source collection or the source critics
of the 112-task run would retain. Because 17 fresh critics are trained on the
mixture, critic losses are not comparable with those of
Appendix~\ref{app:sw-adaptation}. L2 and L3 are complete recipes that differ
in anchoring and in the associated loss scaling (Appendix~\ref{app:method}).
Each run takes \SWdrGpuhLo--\SWdrGpuhHi{} GPU-hours on one A100
(Appendix~\ref{app:sw-system-cost}). Within these runs, source replay raises
the retained source SR under both recipes and keeps it above the base only
under the additive soft penalty.

\section{ReSOP: System Design, Training Cost and Engineering Checks}
\label{app:sw-system}

Section~\ref{sec:method:resop} describes ReSOP in one paragraph. This
section lays out the components behind that paragraph, and lists the design
choices with the measurements that the main text and
Appendix~\ref{app:system-performance} report for them.
Appendix~\ref{app:sw-system-design} describes the system as it produced the
results of the main text. Appendix~\ref{app:sw-system-cost} gives the recorded cost
of the training runs, and Appendix~\ref{app:sw-system-engineering} covers
changes examined afterwards.

\subsection{How is the training system organized?}
\label{app:sw-system-design}

\emph{Collection, replay and learning are separated per task, and only a
small inference snapshot travels from the learner to the collectors.}

ReSOP is built on the RLinf codebase~\citep{yu2026rlinf} and adopts the
decoupled collection and task-balanced replay design of SOP~\citep{pan2026sop};
Figure~\ref{fig:method-overview} shows its data flow.
\emph{Collectors} run the simulators. Each collector holds resident
environments, queries the frozen base policy, applies the editor of its
current inference snapshot, and executes the resulting $K$-action chunk.
\emph{Replay} keeps one demonstration store and one first-in-first-out store
of online transitions for every task, so that data volume and sampling are
accounted per task. The \emph{learner} owns the shared actor with its encoder
or readout and, for every task, a critic ensemble with its target network and
optimizer state. It publishes \emph{inference snapshots} that contain the
actor and the inference-side representation only; collectors pull them and never
hold a critic.

\paragraph{The design choices address task imbalance, critic size and rendering cost.}
Table~\ref{tab:sw-system-design} lists the choices. Interaction differs by
7.07$\times$ between the tasks with the fewest and the most physical actions.
The replay schedule selects the tasks of every update and gives each the
same number of rows, and replay sampling differs by 1.06$\times$. Critics
are kept on the learner, and the snapshot that collectors pull contains none
of them. Rendering is costly and the policy is queried once per chunk, so
interior actions of a chunk step the simulator without images. Evaluation
pairs every editor episode with a base episode on the same task, initial
state and pass (Appendix~\ref{app:evaluation}). The 112-task bounded editor
was trained at the cost in the last row of the table.

\begin{table*}[!t]
\centering\footnotesize
\caption{\textbf{Design choices of ReSOP and the measurements reported for
them.} Measurements are those of Section~\ref{sec:method:resop} and of
Appendices~\ref{app:resfit-recipe} and~\ref{app:system-performance}.}
\label{tab:sw-system-design}
\setlength{\tabcolsep}{4pt}
\begin{tabularx}{\textwidth}{@{}>{\raggedright\arraybackslash}p{0.16\textwidth}>{\raggedright\arraybackslash}X>{\raggedright\arraybackslash}p{0.36\textwidth}@{}}
\toprule
Design choice & Mechanism & Measurement \\
\midrule
Task-balanced updates & After the critic warm-up, every update draws 256 rows from four tasks, 64 per task, half demonstrations and half online transitions. Task groups follow a reshuffled cycle in which every eligible task is selected equally often within a cycle. & Across 112 tasks, replay sampling differs by 1.06$\times$ between the most and the least sampled task, while interaction differs by 7.07$\times$. \\
\addlinespace[2pt]
Critics stay on the learner & Every task has its own critic heads, targets and optimizer state on the learner. Collectors pull the actor and the inference-side representation. & 1,120 critic heads for 112 tasks; 22.7\,MB inference snapshot (vs.\ 156.1\,GB full learner state). \\
\addlinespace[2pt]
Rendering at chunk boundaries & Cameras render at the last action of a $K$-action chunk and at episode ends. Physics, proprioception, reward and success checks run at every action. & On a $K{=}5$ two-task command replay: 79\% fewer render calls and 46.2 to 64.0 actions/s, with matching simulator states, rewards and boundary images. \\
\addlinespace[2pt]
Base-referenced actions & The executed chunk is the base proposal plus the scaled correction, with optional command bounds. The critic regression and the critic term of the actor average within a task and then across the tasks of a batch. & Eqs.~\ref{eq:soft-actor}--\ref{eq:resfit-actor}. \\
\addlinespace[2pt]
Paired evaluation & Base and editor run the same task, initial state and pass. Rescued and harmed episodes are counted per pair. & -- \\
\addlinespace[2pt]
Training cost & Four GPUs; one learner and three collectors in the throughput segment of Appendix~\ref{app:system-performance}. & 481,250 actor updates and 12.1M actions in 41.8 hours (167 GPU-hours) for the 112-task bounded editor. \\
\bottomrule
\end{tabularx}
\end{table*}

\paragraph{Rendering follows the chunk, and sampling follows the task.}
Within a chunk, an image is requested only for the last action or at the
horizon, an early episode end returns a fresh frame, and every chunk reports
its counts of rendered and skipped observations. This loop is behind the
reduction in render calls. With four tasks per update, tasks are drawn in
reshuffled cycles without repetition inside an update, and rows are split
equally over the selected tasks and over the two sources of each task. This
schedule is behind the sampling ratio. The action composition follows
Equation~\ref{eq:resfit-action}; the critic regression and the critic term of
the actor average equally over the tasks of a batch. The behavior-cloning
penalty of the full-action recipe is a plain mean, and the bounded recipe
has none.

\newcommand{\SWrcFullHours}{41.76}
\newcommand{\SWrcFullGpuh}{167.04}
\newcommand{\SWrcFullRate}{3.20}
\newcommand{\SWrcFullActions}{12.11}
\newcommand{\SWrcRollbackLow}{4.07}
\newcommand{\SWrcRollbackHigh}{4.10}
\newcommand{\SWrcFirstHours}{9.82}
\newcommand{\SWrcFirstGpuh}{39.29}
\newcommand{\SWrcFirstRate}{3.43}
\newcommand{\SWrcFirstActions}{3.58}
\newcommand{\SWrcSecondLoneHours}{11.84}
\newcommand{\SWrcSecondLoneGpuh}{47.36}
\newcommand{\SWrcSecondLoneRate}{2.84}
\newcommand{\SWrcSecondLoneActions}{3.61}
\newcommand{\SWrcSecondLoneBefore}{5.56}
\newcommand{\SWrcSecondLtwoHours}{11.76}
\newcommand{\SWrcSecondLtwoGpuh}{47.03}
\newcommand{\SWrcSecondLtwoRate}{2.86}
\newcommand{\SWrcSecondLtwoActions}{3.73}
\newcommand{\SWrcSecondLtwoBefore}{5.18}
\newcommand{\SWrcSecondLthreeHours}{9.88}
\newcommand{\SWrcSecondLthreeGpuh}{39.52}
\newcommand{\SWrcSecondLthreeRate}{3.41}
\newcommand{\SWrcSecondLthreeActions}{3.56}
\newcommand{\SWrcSixteenLoneOneHours}{5.47}
\newcommand{\SWrcSixteenLoneOneGpuh}{10.94}
\newcommand{\SWrcSixteenLoneOneRate}{5.08}
\newcommand{\SWrcSixteenLoneTwoHours}{4.82}
\newcommand{\SWrcSixteenLoneTwoGpuh}{9.63}
\newcommand{\SWrcSixteenLoneTwoRate}{5.77}
\newcommand{\SWrcSixteenLtwoOneHours}{5.12}
\newcommand{\SWrcSixteenLtwoOneGpuh}{10.24}
\newcommand{\SWrcSixteenLtwoOneRate}{5.43}
\newcommand{\SWrcSixteenLtwoTwoHours}{6.52}
\newcommand{\SWrcSixteenLtwoTwoGpuh}{13.04}
\newcommand{\SWrcSixteenLtwoTwoRate}{4.26}
\newcommand{\SWrcSixteenLthreeOneHours}{5.53}
\newcommand{\SWrcSixteenLthreeOneGpuh}{11.07}
\newcommand{\SWrcSixteenLthreeOneRate}{5.02}
\newcommand{\SWrcSixteenLthreeTwoHours}{6.50}
\newcommand{\SWrcSixteenLthreeTwoGpuh}{13.01}
\newcommand{\SWrcSixteenLthreeTwoRate}{4.27}
\newcommand{\SWrcSixteenGpuhMin}{9.63}
\newcommand{\SWrcSixteenGpuhMax}{13.04}
\newcommand{\SWrcCameraHours}{4.61}
\newcommand{\SWrcCameraGpuh}{9.22}
\newcommand{\SWrcCameraRate}{6.03}

\subsection{What do the training runs cost, and is the cost repeatable?}
\label{app:sw-system-cost}

\emph{A second run of the bounded 112-task configuration takes
\SWrcSecondLthreeGpuh{} GPU-hours to reach 121.25k updates, against
\SWrcFirstGpuh{} for the first run over the same range.}

Table~\ref{tab:sw-system-runcost} lists the recorded cost of the 112-task run
of Section~\ref{sec:method:resop} and of the training runs added in
Appendices~\ref{app:sw-robustness},~\ref{app:sw-completions},
and~\ref{app:sw-adaptation}. GPU-hours are allocated GPUs times process time.
The runs of the first two blocks use code from before the changes of
Appendix~\ref{app:sw-system-engineering}; the third-seed 112-task runs of the
third block use the admitted changes, as does the 16-task main-camera L3 run
described below.

\begin{table}[t]
\centering\footnotesize
\caption{\textbf{Recorded cost of the training runs.} GPU-hours are allocated GPUs times process time; updates are
online actor updates, and the last column divides them by the process time. 112-task rows count the whole training
process: the full run of the main text sums three process segments and includes work repeated after two restarts,
and the L1 and L2 rows include an initialization phase and one source-task evaluation before the online phase
(5.56 and 5.18 GPU-hours). 16-task rows count the three training stages of each run, without
initialization and intermediate evaluation; these runs use pairs of RTX~6000D or RTX~4090 GPUs, which change
between stages for some runs, and the main-camera run uses two A100 GPUs. Hardware therefore differs between the first two blocks. The third block lists the third-seed 112-task runs (four A100 GPUs) under the admitted exact-state path of Appendix~\ref{app:sw-system-engineering}, with the last column defined as above; L2 includes initialization and initial source evaluation (as in the top block), while L1 had an interrupted segment whose end time was not recorded and sums its two logged online segments only. The bottom block summarizes the 16 single-task adaptation runs of Appendix~\ref{app:sw-adaptation} on one GPU per run (1.21--1.39\,h on RTX~4090, 0.94--1.01\,h on RTX~6000D; training stages only) and the 12 source-replay runs of Appendix~\ref{app:sw-adaptation-replay} on one A100 each (1.43--1.60\,h including warm-up and checkpointing; rate over the online window).}
\label{tab:sw-system-runcost}
\setlength{\tabcolsep}{3.5pt}
\begin{tabular*}{\linewidth}{@{\extracolsep{\fill}}lrrrr@{}}
\toprule
Run & Updates & Hours & GPU-h & Upd./s \\
\midrule
\multicolumn{5}{@{}l}{\emph{112 source tasks, four A100 GPUs}} \\
L3, run of the main text & 481.25k & 41.76 & 167.04 & 3.20 \\
\quad its first 121.25k updates & 121.25k & 9.82 & 39.29 & 3.43 \\
L3, second seed & 121.25k & 9.88 & 39.52 & 3.41 \\
L1, second seed & 121.25k & 11.84 & 47.36 & 2.84 \\
L2, second seed & 121.25k & 11.76 & 47.03 & 2.86 \\
\midrule
\multicolumn{5}{@{}l}{\emph{16 source tasks, two GPUs, training stages only}} \\
L1, seed 1 & 100k & 5.47 & 10.94 & 5.08 \\
L1, seed 2 & 100k & 4.82 & 9.63 & 5.77 \\
L2, seed 1 & 100k & 5.12 & 10.24 & 5.43 \\
L2, seed 2 & 100k & 6.52 & 13.04 & 4.26 \\
L3, seed 1 & 100k & 5.53 & 11.07 & 5.02 \\
L3, seed 2 & 100k & 6.50 & 13.01 & 4.27 \\
L1, main camera only & 100k & 4.61 & 9.22 & 6.03 \\
\midrule
\multicolumn{5}{@{}l}{\emph{112 source tasks, third seed (admitted path, four A100 GPUs)}} \\
L1, third seed (online segs.) & 121.25k & 8.64 & 34.57 & 3.90 \\
L2, third seed & 121.25k & 9.29 & 37.18 & 3.62 \\
L3, third seed & 121.25k & 9.08 & 36.31 & 3.71 \\
\midrule
\multicolumn{5}{@{}l}{\emph{Single-task adaptation, one GPU (25k updates each)}} \\
L1/L2, RTX~4090 (14 runs, med.) & 25k & 1.31 & 1.31 & 5.28 \\
L1/L2, RTX~6000D (2 runs, mean) & 25k & 0.98 & 0.98 & 7.10 \\
L2/L3 + replay, A100 (12, med.) & 25k & 1.50 & 1.50 & 5.17 \\
\bottomrule
\end{tabular*}
\end{table}

\paragraph{A repeated run reproduces the cost per update.}
The run of the main text reaches 121.25k online actor updates after
\SWrcFirstHours{} process hours on four A100 GPUs, or \SWrcFirstGpuh{}
GPU-hours. The second seed of the same configuration takes
\SWrcSecondLthreeHours{} hours and \SWrcSecondLthreeGpuh{} GPU-hours to the
same point. It executes \SWrcSecondLthreeActions{}M environment actions,
against \SWrcFirstActions{}M in the first run. Over this range the two runs
use the same code revision and differ in training seed.

\paragraph{The full-run figure includes repeated work, and the added runs are shorter.}
The \SWrcFullGpuh{} GPU-hours of the full run include
\SWrcRollbackLow--\SWrcRollbackHigh{} process hours of work repeated after
two restarts (Appendix~\ref{app:libero-reference}). The second seeds of L1 and L2
at 112 tasks take \SWrcSecondLoneGpuh{} and \SWrcSecondLtwoGpuh{} GPU-hours
to 121.25k updates. The third-seed 112-task runs in the third block of
Table~\ref{tab:sw-system-runcost} ran on four A100-SXM4-80GB GPUs with all
admitted changes of Appendix~\ref{app:sw-system-engineering} enabled and all
rejected candidates disabled, taking 37.18 GPU-hours for L2 (3.62 updates/s over
total process time; 4.22 updates/s in the online window) and 36.31 GPU-hours for
L3 (3.71 updates/s over the native training phase). Third-seed L1 resumed from
57.8k updates after an interrupted segment whose end time was not recorded and
logged 34.57 GPU-hours across its two online segments (3.90 updates/s across
both segments, 4.29 in the resumed window; initialization and initial source
evaluation add 4.59 GPU-hours separately).

\paragraph{Matched online windows under the earlier and updated paths.}
Restricting to the identical logged online learner interval from 25 to 121,225
actor updates (121,200 updates on four A100 GPUs), the earlier-code L3 runs
(seeds~0 and~1) both average 3.66 actor updates/s (3.58M and 3.56M environment
actions; seed~1 takes 36.80 GPU-hours in the native online phase and 39.48
GPU-hours including startup and terminal checkpointing, excluding job setup
before the process starts), while the third-seed L3 run under the admitted path averages 4.00
actor updates/s (6.30M environment actions; 33.65 GPU-hours native online,
36.31 GPU-hours full native training phase). Because these runs also differ in
seed, physical GPU assignment, collector concurrency, snapshot cadence, and
total environment actions, the numbers record each deployment's observed cost
rather than an isolated causal speedup of individual code changes; no full
481.25k-update run has been timed under the updated path. The 16-task seed runs
take \SWrcSixteenGpuhMin--\SWrcSixteenGpuhMax{} GPU-hours for 100k updates on
two GPUs, and the main-camera L1 run takes \SWrcCameraGpuh{} GPU-hours.

\paragraph{Workstation and single-GPU RTX runs.}
Single-task adaptation does not require multi-GPU datacenter nodes: across all
16 completed 25k-update runs of Appendix~\ref{app:sw-adaptation} (L1 and L2,
Tasks~98 and~112, fresh and pretrained, seeds~1 and~2; bottom block of
Table~\ref{tab:sw-system-runcost}), each run finishes on one GPU in
1.21--1.39 GPU-hours on an RTX~4090 (14 runs, median 1.31~GPU-hours,
5.01--5.76 updates/s) and 0.94--1.01 GPU-hours on an RTX~6000D (2 runs,
mean 0.98~GPU-hours; 20.16~GPU-hours combined across all 16 runs). The 12
adaptation runs with source replay of Appendix~\ref{app:sw-adaptation-replay}
take \SWdrGpuhLo--\SWdrGpuhHi{} GPU-hours each on one A100, including warm-up
and checkpointing, at \SWdrRateLo--\SWdrRateHi{} updates/s in the online window. At 16
source tasks on two RTX~6000D GPUs (one learner, one collector), the
main-camera L3 run under the admitted path averages
6.06 actor updates/s (6,205 critic samples/s) over the matched native learner
window from 25 to 19,975 updates (19,950 updates), against 4.40 and 5.37
updates/s for the two earlier-code two-camera L3 seeds on pairs of RTX~6000D
GPUs; because the main-camera run omits the wrist image and uses a dedicated
two-GPU allocation, this comparison records deployment cost rather than an
isolated code speedup. Long-horizon 112-task continuation runs on four
RTX~6000D GPUs under the earlier native learner revision also sustain
3.74--3.85 actor updates/s (e.g., 148k completed L2 updates and 3.06M
environment actions in 10.98~hours at 3.74 updates/s). Finally, in paired 5k
16-task L2 verification runs on two RTX~4090 GPUs (50k--55k actor updates), the native
path finishes in 1,260.5\,s (0.70 GPU-hours), ahead of the rejected broadcast
layer-normalization (1,406.6\,s) and packed-critic (1,453.9\,s) candidates.

\subsection{Which later engineering changes were admitted, and on what evidence?}
\label{app:sw-system-engineering}

\emph{Changes that pass an equivalence check against the earlier path are
enabled for later runs; five candidates failed theirs and stay disabled.}

The changes in this subsection were examined after the experiments of the
main text. The results and the training cost in
Section~\ref{sec:method:resop} were measured before them; of the runs in
this paper, only the third 112-task seeds
(Appendix~\ref{app:sw-robustness-seeds112}) and the 16-task main-camera L3
run (Appendix~\ref{app:sw-completions-cameras}) were trained with the
admitted changes.
Table~\ref{tab:sw-system-engineering} lists the candidates.

\begin{table*}[t]
\centering\footnotesize
\caption{\textbf{Engineering changes examined after the experiments of the
main text.} Rows above the inner rule were admitted; rows below it were
rejected. Timing entries state their scope, and none is the rate of a whole
training run. Throughput differences are paired medians relative to the
earlier path.}
\label{tab:sw-system-engineering}
\setlength{\tabcolsep}{4pt}
\begin{tabularx}{\textwidth}{@{}>{\raggedright\arraybackslash}p{0.25\textwidth}>{\raggedright\arraybackslash}p{0.11\textwidth}>{\raggedright\arraybackslash}X@{}}
\toprule
Candidate & Check & Evidence and scope \\
\midrule
Per-update bookkeeping restricted to the tasks of the batch; one shared critic-training path & Exact state & Passes on trained 112-task L2 and L3 checkpoints. 16 fixed-batch windows on trained 16-task checkpoints, two RTX~4090: +4.82\% (L2), +4.53\% (L3). Synthetic critic panel, 32 windows, one RTX~6000D, 4 to 112 tasks: +5.00\% to +28.89\% (L2), +2.16\% to +28.13\% (L3). \\
Replay rows handed to the learner with task ranges & Exact state & 15 draws, including one after a restore, match in sampled rows and random-number state. \\
Sharded checkpoint format; conversion of existing checkpoints & Exact state; tolerance & Seven updates after a resume equal seven uninterrupted updates (two cases). Three full 16-task states: 2.40\,GB smaller each; save time $-9.93$, $-3.76$ and $+3.62$\,s. Two converted checkpoints: largest absolute difference $2.4\times10^{-7}$ on a fixed critic panel (absolute tolerance $10^{-6}$). \\
Shared rollout and intervention interface & Same inputs & 3,340 physical actions recorded on one task, none different. \\
\midrule
Broadcast layer normalization & Tolerance & Exact on 200 single-update comparisons; the 5k-update training curve falls outside the spread of earlier seeds. \\
Packed critic computation & Exact state & 14 (L2) and 25 (L3) mismatches after the updates. \\
Batched evaluation in several lanes & Paired outcomes & 166 of 800 serial--batched pairs differ; 0 of 400 between two serial runs (4 tasks, 50 states, base and L2). 220.81 against 158.63 episodes per GPU-hour. \\
Overlapped checkpoint readback & Timing & 28.30\% slower on network storage (paired median). \\
Asynchronous checkpoint I/O & Exact state & Large-state restore passes (18 shards, 14.24\,GiB, 2,385 tensor leaves, RNG state); four-arm ABBA run (48,080 critic updates) fails strict exact-state check at the 20-update warm-up boundary (cause unresolved); stays disabled. \\
\bottomrule
\end{tabularx}
\end{table*}

\paragraph{A change is admitted only at the level of equivalence it can meet.}
Three checks are used. \emph{Exact state}: the earlier path and the candidate
run the same updates from the same state, after which parameters, gradients,
optimizer state, clocks and random-number state must be identical, on CPU and
with deterministic GPU kernels. \emph{Tolerance}: outputs on a fixed panel of
inputs must agree within a numerical tolerance; a change to the update itself
also needs 200 independent single-update comparisons and a short training run
judged against the spread of earlier seeds. \emph{Paired outcomes}: under the
evaluation protocol, the candidate may disagree with the earlier path no more
often than two runs of the earlier path disagree with each other. A candidate
that fails its check stays disabled, and a resumed run keeps the settings
saved in its checkpoint.

\paragraph{The admitted changes pass their checks, and fixed-batch learner windows run a few percent faster.}
Restricting per-update bookkeeping to the tasks of the current batch, with
one critic-training path shared by the recipes, passes the exact-state check
on trained 112-task L2 and L3 checkpoints. On trained 16-task checkpoints the
paired median learner throughput is 4.82\% higher for L2 and 4.53\% higher
for L3, over 16 fixed-batch windows on two RTX~4090 GPUs. On a synthetic
critic panel the difference grows with the number of tasks and reaches
28.89\% and 28.13\% at 112 tasks. Both measurements time 500 critic and 125
actor updates on a fixed batch. Neither is the rate of a whole training run,
in which collection, replay and checkpointing also take time. Checkpoints of
three full 16-task training states are each 2.40\,GB smaller in the sharded
format and need less peak memory to write; saving is faster for two of them
and 3.62\,s slower for the third.

\paragraph{Five candidates are rejected, and exact closed-loop equality is not established.}
Broadcast layer normalization is exact on the 200 single-update comparisons,
but its 5k-update training curve falls outside the spread of earlier seeds.
Packed critic computation leaves 14 mismatches for L2 and 25 for L3 in the
exact-state check, and does not pass the tolerance check either. Batched
evaluation in several lanes completes more episodes per GPU-hour, but its
outcomes differ from serial evaluation on 166 of 800 pairs, while two serial
runs agree on all 400 pairs. Overlapped readback is 28.30\% slower on network
storage, so readback stays serial. Asynchronous checkpoint I/O passed
large-state restore checks (18 shards, 14.24\,GiB, 2,385 tensor leaves and
RNG state), but a four-arm ABBA check over 48,080 critic updates failed the
strict exact-state criterion at the 20-update warm-up boundary with the cause
unresolved, so asynchronous checkpointing stays disabled. Exact equality of
whole episodes is not established for any path: two serial runs of the
earlier path on the same task already diverge in their sensor outputs.
Appendix~\ref{app:sw-robustness-repeat} reports how often outcomes change
between repeated evaluations.

\section{Learner and Runtime Details}
\label{app:method}

This appendix specifies the configurations used in the reported LIBERO-130
and RoboTwin S4 runs (Table~\ref{tab:recipe-summary}). Shared notation describes their transition
semantics; architecture and optimization settings are specified per
configuration. The 16-task LIBERO six-recipe study crosses two representation
strategies with three editing choices using MLP actors (with top-two MoE
routing and offline cluster supervision evaluated separately in
Section~\ref{sec:experiments:design}), complementing the 112-task runs of
the three online-encoder recipes and the single-task adaptation runs. Online task-contrastive regularization
and route-entropy balancing belong to the RoboTwin S4 study
(Section~\ref{sec:experiments:robotwin}).

\paragraph{Actor-loss scaling in the six-recipe study.}
The scaling rule follows the presence of a behavior-cloning (BC) penalty,
not full-action versus residual parameterization. The four soft-penalty
recipes (R1, L1, R2, L2) use task-wise Q normalization alongside
the BC term toward the frozen base. The bounded-residual recipes (R3, L3)
have no BC term and optimize the unnormalized mean Q.
Normalization sets the value term's weight relative to BC; the soft-penalty and
bounded-residual recipes therefore differ in both their anchoring mechanism and
this associated loss scaling. The full-action objective and calibration
used in the S4 runs are specified below.

For the 16-task LIBERO six-recipe study, all four soft-penalty recipes use
$\beta_{\mathrm{BC}}=5$ ($\beta$ in Equation~\ref{eq:soft-actor}), with mean squared deviation over the complete
five-action proposal and the corresponding base reference. Their value term
uses a calibrated per-task anchor divided by the detached mean absolute
ensemble-mean Q, clamped below at $10^{-8}$. Thus the denominator is
$\max(\mathbb E_m[|\overline Q_m|],10^{-8})$, rather than an additive
epsilon. The actor objective averages the selected tasks' contributions.
These recipe-specific weights define $w_m$ in the value term of
Equation~\ref{eq:soft-actor}; the target of BC is the base proposal, not
the demonstration action.

\begin{table*}[t]
\centering
\caption{\textbf{Architecture and execution settings of the main and S4 runs.}
These settings identify the evaluated configurations, not fixed benchmark
architectures; the 16-task LIBERO routing study (Table~\ref{tab:six-actor-moe})
instead uses the four-expert, top-two MoE actor.}
\label{tab:recipe-summary}
\small
\begin{tabular}{lll}
\toprule
Setting & Main LIBERO-130 runs & RoboTwin S4 runs \\
\midrule
Benchmark & LIBERO-130 & RoboTwin S4 \\
Shared actor & MLP & MoE, four experts, top-two \\
Critic heads per task ($H_Q$) & 10 & 6 \\
Actions predicted per base query & 8 & 50 \\
Actions per editor decision ($K$) & 5 & 10 \\
\bottomrule
\end{tabular}
\end{table*}

\subsection{Chunk returns and bootstrap semantics}
\label{app:targets}

Let a learner transition span $L$ actually executed physical actions across
one or more editor decisions. With per-action reward $r_{t+j}$ and discount
$\gamma$, the stored return and bootstrap multiplier are
\begin{equation}
    R_t^{(L)}=\sum_{j=0}^{L-1}\gamma^j r_{t+j},
    \qquad
    D_t^{(L)}=(1-d_t)\gamma^L,
    \label{eq:chunk-return}
\end{equation}
where $d_t$ marks a true terminal reached within that span.
A time limit ends the collected episode but retains bootstrapping from its
valid final observation before reset. Returns do not cross episode or task
boundaries. A short executed prefix uses its actual length $L$, rather than
the nominal prediction horizon.

For task $m$, sample two target heads $\mathcal J$ from its ensemble,
using the random-subset minimum estimator of
REDQ~\citep{chen2021randomized}.
Both reported configurations use a target of the form
\begin{equation}
    y_t=R_t^{(L)}+
      D_t^{(L)}\min_{h\in\mathcal J}
          Q^-_{m,h}(x_{t+L},\widetilde A^-_{t+L}).
    \label{eq:chunk-target}
\end{equation}
The next command $\widetilde A^-$ differs between configurations. The S4 full-action learner uses
a sample from the current actor with fixed action noise; its joint warm-up
uses the next base reference. The LIBERO bounded-residual learner uses the target actor's residual,
smoothed with clipped noise and composed with the next base plan.
Neither target contains an entropy-temperature term.
Critics value the planned chunk, while return construction records how much
of that chunk was executed before termination.

\subsection{RL-token readout and full actions with a soft penalty}
\label{app:rlt-recipe}

The RoboTwin S4 runs use the representation and full-action learning
implementation adapted from RL Token~\citep{xu2026rltoken}.

\paragraph{Representation and action coordinates.}
A pretrained RL-token readout extracts a 2,048-dimensional token from
LingBot-VLA's final-layer visual--language token features. The base and readout
remain fixed during online RL. The actor additionally receives 14-dimensional
proprioception and the current ten-action, 14-dimensional reference chunk.
Its input width is therefore $2{,}048+14+140=2{,}202$.
The critic state instead concatenates the token, proprioception, the full
50-action base plan, and a five-way cursor encoding, giving
$2{,}048+14+700+5=2{,}767$ dimensions before its candidate action input.
The base plan is consumed through five ten-action decisions.

Action normalization uses fixed per-coordinate affine constants from the
base interface. If $\ell_a$ and $u_a$ denote those constants, the coordinate
map is
\begin{align}
    \mathcal N(a)&=2\frac{a-\ell_a}{u_a-\ell_a}-1,\nonumber\\
    \mathcal N^{-1}(b)&=\ell_a+\frac{b+1}{2}(u_a-\ell_a).
\end{align}
The statistical constants set the coordinate system; they are not physical
actuator limits. The inherited interface does not clip the affine coordinate
to its nominal interval. The actor predicts full commands in this coordinate,
without an additive residual scale.

\paragraph{Actor and critics.}
The S4 actor uses four experts with top-two routing.
Each expert has two 256-wide hidden layers with LayerNorm and ReLU.
Each model task has a six-head critic ensemble with two 256-wide hidden
layers with LayerNorm and ReLU. Actor training uses the mean across heads; target construction
uses the minimum of two sampled target heads. Task-contrastive regularization uses
a 128-dimensional embedding, temperature $0.1$, and loss weight $0.01$.
The task-contrastive configuration uses marginal-route entropy balancing;
the baseline uses Switch-style balancing. Their coefficients are reported in
Table~\ref{tab:rlt-s4-settings}.

\paragraph{Initialization and online updates.}
For task $m$, let $\overline Q_m$ denote the mean over critic heads,
$x$ the critic state, and $\widetilde A_\theta$ an actor sample with fixed
action noise. The online actor objective is
\begin{align}
    \mathcal L^{\mathrm{soft}}_{\pi,m}
      &= \beta_{\mathrm{BC}}\,
         \mathbb E_m\!\left[\operatorname{MSE}
               (\widetilde A_\theta,B)\right]\nonumber\\
      &\quad -w_m\,\mathbb E_m\!\left[
               \overline Q_m(x,\widetilde A_\theta)\right], \\
    w_m &= \frac{\kappa_m}
      {\operatorname{sg}\!\left(
          \mathbb E_m[|\overline Q_m(x,\widetilde A_\theta)|]
       \right)+\epsilon}.
    \label{eq:rlt-qnorm}
\end{align}
Here $\operatorname{sg}$ stops gradients, and $\operatorname{MSE}$ averages
over chunk entries. Task contributions are averaged with equal weight.
The recipe uses $\beta_{\mathrm{BC}}=5$ and $\epsilon=10^{-8}$.
When demonstration transitions enter replay, their recorded actions supervise
critic TD learning. Their states also enter actor optimization, whose
behavioral target is the frozen base proposal at those states.

The two S4 configurations have distinct initialization histories and budgets,
reported in Appendix~\ref{app:rlt-s4-training}.
Task-wise Q-normalization anchors $\kappa_m$ are calibrated as the mean absolute
ensemble-mean policy value at the end of joint training, then held fixed.
The joint actor objective uses a Q coefficient of $0.05$; online training
replaces that coefficient with Equation~\ref{eq:rlt-qnorm}.

The configuration comparison covers two million online critic updates;
the extended-training analysis follows the task-contrastive run to 2,823,723
updates. Both counts exclude initialization.
Actor and critic learning rates are $2.5\times10^{-5}$ and
$3\times10^{-5}$, respectively; the update ratio is one actor update for
four critic updates, with target interpolation coefficient $0.005$.
The baseline uses a global batch of 4,096 rows. The task-contrastive run
starts with 4,096 rows and increases to 16,384 after update 994,849;
tasks receive equal row counts within each batch.
The critic loss is the mean squared TD error across rows
and heads. Fixed action noise has standard deviation $0.002$.

\paragraph{Replay and return horizon.}
Online replay partitions transitions by the four S4 tasks,
with FIFO replacement and uniform within-task sampling.
Collector task allocation is proportional to the canonical task horizon,
while learner batches give tasks equal row counts.
The per-action discount is $0.92$. Tasks with canonical rounded horizon
at least 370 physical actions use two-decision returns ($2K=20$ physical
actions); the remaining tasks use one-decision returns ($K=10$ physical actions).
Sample presentations per new row are distinct from optimizer updates because
they also depend on global batch size.

\subsection{Online-learned encoder and bounded residual output}
\label{app:resfit-recipe}

The bounded-residual LIBERO configuration adapts the visual encoder and
residual learner from ResFiT~\citep{ankile2025residual} to shared chunk-level editing.

\paragraph{Frozen base and task partition.}
The base is the LIBERO-130 supervised OpenVLA-OFT
checkpoint released by RLinf~\citep{kim2025oft,rlinfLibero130Base},
revision \texttt{a4ea91b}, used without weight modification.
The model card reports 42.09\% success on LIBERO-130 under the publisher's
evaluation protocol. Base success rates in this study are measured under
the same protocol as the edited policy.
The 112-task editor uses 112 source tasks and 18 editor-held-out tasks.
Expert preparation uses 50 official demonstrations per source task,
re-rendered and queried at the five-action decision boundaries.
Editor-held-out tasks contribute neither prepared expert rows nor critic training.
Both base-only and editor evaluation consume five actions per new base query,
discarding the other three actions in its eight-action prediction.

\paragraph{Representation and networks.}
The separate MinVit visual encoder processes main and wrist images at
$84\times84$ resolution, using depths two and one, respectively,
embedding dimension 128 and four attention heads.
Training applies random-shift augmentation with padding four, as in
DrQ-v2~\citep{yarats2021drqv2}.
Proprioception contains eight values: end-effector position, axis-angle
orientation, and two finger positions.
A frozen sentence encoder supplies a 384-dimensional language feature,
projected through a trainable 256-wide layer to 64 dimensions.
In the bounded-residual configuration, the shared MLP actor uses two
1,024-wide hidden layers and LayerNorm;
its output is initialized to zero.
Each source task has ten critic heads, with two 1,024-wide hidden layers
and a 1,024-dimensional spatial readout.
The 112 task-specific ensembles contain 1,120 online critic heads and
1,120 target copies, in addition to online optimizer state. Their storage
grows linearly with the number of source tasks. A learner batch selects
four tasks, so critic forward and backward computation uses those tasks'
ensembles rather than all 112. Inference snapshots omit critics, target
networks, and optimizer state; editor-held-out evaluation uses the shared actor
and representation with the frozen base and creates no target-task critics.

\paragraph{Commands and noise.}
A residual chunk contains five seven-dimensional native commands:
three translation components, three rotation components, and a gripper
component. The scale vector is
\begin{equation}
    \Lambda=(0.1,0.1,0.1,0.1,0.1,0.1,2).
\end{equation}
The actor's unit-tanh mean is scaled by $\Lambda$, then composed with the
base and clipped to the native range $[-1,1]$ per coordinate.
Collection adds Gaussian noise of standard deviation $0.025$ after scaling.
Target-policy noise has the same standard deviation and is clipped at
$0.3$ before composition. Evaluation uses the actor mean without exploration.
The correction statistics in Appendix~\ref{app:correction-structure} divide
each executed correction (edited command minus base command) by its
corresponding entry in $\Lambda$: $0.1$ for the six pose coordinates and $2$
for the gripper.

\paragraph{Rewards, targets, and losses.}
The bounded-residual configuration uses delayed actor updates and target-policy
smoothing~\citep{fujimoto2018addressing}. Its actor maximizes the mean of the
active task's critic heads without an additional base-deviation penalty
(Equation~\ref{eq:resfit-actor}). Both expert and online replay contribute
to critic and encoder learning; states from both sources enter the actor's
value objective.

The environment supplies reward one upon success and zero otherwise.
Success is terminal. The maximum episode length is 512 physical actions;
time limits retain the final-observation bootstrap.
The per-action discount is $0.995$ and the backup spans three editor
chunks, normally 15 physical actions. Short final chunks use their actual
executed length in Equation~\ref{eq:chunk-return}.

With $H_Q=10$ heads and $N$ sampled rows, the configured critic loss is
\begin{equation}
    \mathcal L_Q =
       \frac{1}{N}\sum_{i=1}^{N}
       \left(\frac{1}{H_Q}\sum_{h=1}^{H_Q}
          |Q_{m_i,h}(x_i,A_i)-y_i|\right)^2.
    \label{eq:resfit-critic-loss}
\end{equation}
Thus the mean absolute error over heads is squared, rather than averaging
the squared errors of individual heads.
The encoder is updated through this critic loss.
The target uses the current encoder without gradients, a target actor,
and target critic heads.
Actor and critic learning rates are $10^{-6}$ and $10^{-4}$;
the actor updates once per four critic updates.
Target interpolation uses $0.005$, and gradient norms are clipped at one.

\paragraph{Data mixture and task participation.}
A global batch contains 256 rows from four tasks, with 64 rows per selected
task and equal expert/online shares after the expert-only critic warm-up.
Rows are sampled uniformly within each task and source.
An eligible online task must have data in replay; the runtime records
actual task participation during startup and subsequent updates.
Online FIFO capacity is 10,000 chunk rows per task.
Collection runs independently of a fixed update-to-data quota.
Global actor updates are the primary training-budget axis, accompanied by
critic updates and physical interaction counts.

\subsection{MoE architecture and auxiliary objectives}
\label{app:actor-routing}

Our mixture-of-experts (MoE) implementation adapts expert routing from
DyGRO-VLA~\citep{lin2026dygrovla}.
The router uses the current policy context to select the top-$k$ experts;
their outputs are combined using normalized routing weights.
Task-contrastive regularization is an optional auxiliary objective, separate
from the MoE architecture. We specify it and the routing-balancing objective
below.

Let $v_\theta(c)$ be a normalized embedding of the actor's input context
$c$, and $u_j$ a normalized learnable prototype for source task $j$.
With temperature $T>0$, the task-contrastive regularizer is
\begin{equation}
    \mathcal L_{\mathrm{task}}
      =-\mathbb E_{(c,m)}
       \log\frac{\exp(v_\theta(c)^\top u_m/T)}
       {\sum_j\exp(v_\theta(c)^\top u_j/T)}.
    \label{eq:task-aux}
\end{equation}
This prototype-based objective increases similarity to the labeled task's
prototype relative to the other task prototypes. Routing at execution is
computed from context features. Separately, with dense router probabilities $p_e(c)$,
the balancing term is the negative entropy of their batch average,
\begin{equation}
    \mathcal L_{\mathrm{bal}}
        = \sum_e \overline p_e\log\overline p_e,
    \qquad
    \overline p_e=\mathbb E_c[p_e(c)].
    \label{eq:router-entropy}
\end{equation}
This is computed before top-$k$ selection.
The combined actor objective adds
$\lambda_{\mathrm{task}}\mathcal L_{\mathrm{task}}
+\lambda_{\mathrm{bal}}\mathcal L_{\mathrm{bal}}$ to the configuration-specific
control loss. Setting $\lambda_{\mathrm{task}}=0$ removes task-contrastive
supervision while retaining the MoE architecture. The task-contrastive S4
run uses both auxiliary terms, with coefficients specified in
Appendix~\ref{app:rlt-recipe}; the main LIBERO transfer and adaptation runs
use a shared MLP.

\subsection{Runtime identities and counters}
\label{app:runtime}

Collector transitions retain task identity, policy version, reward, and
actual executed span. Online replay uses FIFO replacement within each task;
task-group size and the online/demonstration mixture follow the configuration.
Distributed execution preserves each logical optimizer update and its global
batch definition across ranks. The S4 runs bound learner progress relative
to accepted online rows; the bounded-residual LIBERO runs collect independently
of a fixed update-to-data quota.

ReSOP separates the identity of an algorithmic update from the hardware
used to execute it. A logical optimizer advance counts once, independent of
the number of devices or critic heads. A sample-presentation counter counts
the replay rows used by those updates. Accepted decision rows and physical
environment actions count data collection at different temporal resolutions.
Task-level counters make these quantities interpretable when learner
updates involve only a subset of tasks.
Physical actions per second measure environment throughput; decision rows
per second measure replay-data production at the specified chunk length.
Optimizer advances and sample presentations measure learner work separately
from collection.

An inference snapshot contains the actor and representation needed for
execution. A recoverable training checkpoint additionally binds optimizer
state, target networks, replay state, random-number state, and progress
counters. Restored collectors open new episodes and base-plan queues.
Pending returns are resolved without crossing the boundary to a newly
started simulator stream.

\section{Evaluation Protocols}
\label{app:experimental-details}

\subsection{Task exposure and transfer}
\label{app:task-exposure}

The reported full-action runs with an RL-token readout inherit a pretrained
readout and editor initialization, then learn jointly on the four S4 tasks. Their checkpoint
evaluations keep these tasks fixed and vary the scene initialization, while
Appendix~\ref{app:s4-followup} additionally tests zero-update transfer from
the 1.49M task-contrastive S4 checkpoint to 12 other RoboTwin tasks.

For LIBERO-130, the frozen OpenVLA-OFT base's supervised training covers 130
tasks; at execution, the frozen base takes a single main-view
$256\times256$ RGB image and the instruction, without proprioception, whereas
the L recipes receive main and wrist RGB images, proprioception, the
instruction, and base proposals, except in the main-camera ablation.
The editor split contains 112 source and 18 editor-held-out tasks.
It comprises 80/10 tasks from LIBERO-90 and 8/2 from each of the four
ten-task suites. Only the 112 source tasks enter editor data preparation,
warm-up, or online critic learning.

\subsection{Evaluation units and aggregation}
\label{app:evaluation}

For task $m$ with $n_m$ evaluation episodes and binary outcomes $s_{m,i}$,
the macro success rate over a task cohort $\mathcal T$ is
\begin{equation}
    \operatorname{SR}(\mathcal T)
      =\frac{1}{|\mathcal T|}
         \sum_{m\in\mathcal T}\frac{1}{n_m}
                   \sum_{i=1}^{n_m}s_{m,i}.
\end{equation}
Steps to success is the success-conditioned mean episode length. For episode
length $L_{m,i}$ in physical actions, the task-level value is
\begin{equation}
    \overline L_m^{+}
       = \frac{\sum_{i=1}^{n_m}s_{m,i}L_{m,i}}
                    {\sum_{i=1}^{n_m}s_{m,i}}.
\end{equation}
We average these task-level values with equal task weight. A task with no
successful episodes has an undefined value (N/A); when such tasks occur, we
report the number of defined tasks alongside the macro mean. This metric
counts environment actions, not policy decisions, learner updates, or
wall-clock time, and is interpreted jointly with SR.

The LIBERO protocol evaluates 50 native initial states in two passes,
giving 100 episodes per task and 50 distinct initial states.
Repeated passes are identified separately from distinct initial states.
Base/editor pairs use matching task, initial-state identity, action cadence,
and horizon. Each of the $N$ pairs in a cohort is rescued (base failure,
editor success), kept (both succeed), harmed (base success, editor failure),
or still failed. We report the four as shares of $N$, so that
$r+k+h+f=1$, editor SR is $r+k$, base SR is $k+h$, and the SR gain is $r-h$.
Rescue and harm percentages throughout are $r$ and $h$, shares of all pairs
in the stated cohort; tables give the counts beside them. Where two policies
other than the base are contrasted, the reference policy takes the base's role.

\begin{table*}[t]
\centering\footnotesize
\caption{\textbf{Paired-outcome shares and the equivalent conditional rates.}
With base SR $b=k+h$, the rate of rescue among base failures is $r/(1-b)$ and
the rate of harm among base successes is $h/b$. This paper reports shares; the rates are listed
for comparison with reports that condition on the base outcome. All values
are percentages.}
\label{tab:outcome-shares-rates}
\setlength{\tabcolsep}{7pt}
\begin{tabularx}{0.78\textwidth}{@{}>{\raggedright\arraybackslash}Xrrrrrr@{}}
\toprule
 & & & \multicolumn{2}{c}{Share of pairs} & \multicolumn{2}{c}{Conditional rate} \\
\cmidrule(lr){4-5} \cmidrule(l){6-7}
Editor / condition & Pairs & Base SR & Rescue & Harm & Rescue & Harm \\
\midrule
\multicolumn{7}{@{}l}{\emph{LIBERO editor-held-out tasks, two passes}} \\
L1, 112 sources & 1,800 & 23.33 & 40.4 & 6.2 & 52.8 & 26.7 \\
L2, 112 sources & 1,800 & 23.33 & 35.2 & 4.9 & 45.9 & 21.0 \\
L3, 112 sources & 1,800 & 23.33 & 36.8 & 10.2 & 48.0 & 43.6 \\
L3, 16 sources & 1,800 & 23.83 & 15.8 & 16.5 & 20.7 & 69.2 \\
L1, 2 sources & 1,800 & 23.00 & 4.1 & 21.6 & 5.3 & 94.0 \\
L2, 2 sources & 1,800 & 23.22 & 10.7 & 14.2 & 13.9 & 61.0 \\
L3, 2 sources & 1,800 & 23.33 & 0.5 & 22.8 & 0.7 & 97.9 \\
\midrule
\multicolumn{7}{@{}l}{\emph{112-task editors by scene coverage, two passes}} \\
L1, covered scenes & 1,000 & 22.40 & 42.4 & 8.6 & 54.6 & 38.4 \\
L1, uncovered scenes & 800 & 24.50 & 38.0 & 3.3 & 50.3 & 13.3 \\
L2, covered scenes & 1,000 & 22.40 & 42.0 & 6.1 & 54.1 & 27.2 \\
L2, uncovered scenes & 800 & 24.50 & 26.8 & 3.4 & 35.4 & 13.8 \\
L3, covered scenes & 1,000 & 22.40 & 44.5 & 9.9 & 57.3 & 44.2 \\
L3, uncovered scenes & 800 & 24.50 & 27.3 & 10.5 & 36.1 & 42.9 \\
\midrule
\multicolumn{7}{@{}l}{\emph{Six 16-task editors by scene coverage, two passes (lowest to highest)}} \\
Covered scenes & 700 & 29.43 & 28.0--43.6 & 7.0--15.7 & 39.7--61.7 & 23.8--53.4 \\
Uncovered scenes & 1,100 & 20.27 & 8.0--25.9 & 4.2--17.0 & 10.0--32.5 & 20.6--83.9 \\
\midrule
\multicolumn{7}{@{}l}{\emph{L3 correction scale, first pass}} \\
Original & 900 & 23.56 & 33.9 & 9.2 & 44.3 & 39.2 \\
Half scale & 900 & 23.56 & 38.8 & 4.9 & 50.7 & 20.8 \\
Quarter scale & 900 & 23.56 & 38.3 & 5.3 & 50.1 & 22.6 \\
No gripper correction & 900 & 23.56 & 27.8 & 11.4 & 36.3 & 48.6 \\
\midrule
\multicolumn{7}{@{}l}{\emph{RoboTwin}} \\
S4 editor, four source tasks & 800 & 74.63 & 15.0 & 5.5 & 59.1 & 7.4 \\
S4 editor, 12 other tasks & 600 & 71.00 & 4.2 & 50.0 & 14.4 & 70.4 \\
S4 editor, 12 other (matched, $\alpha=1.0$) & 600 & 68.67 & 6.2 & 48.3 & 19.7 & 70.4 \\
S4 editor, 12 other (matched, $\alpha=0.5$) & 600 & 68.67 & 5.3 & 40.5 & 17.0 & 59.0 \\
S4 MoE baseline, 12 other (matched) & 600 & 68.67 & 6.3 & 44.5 & 20.2 & 64.8 \\
\bottomrule
\end{tabularx}
\end{table*}

\subsection{RoboTwin task selection}
\label{app:robotwin-selection}

The RoboTwin reference cohort contains 43 tasks after seven temporary
exclusions. The exclusions address two
different concerns: invalid asset/grasp instances and success predicates that
can accept unintended behavior. They keep invalid instances out of
multi-task RL; the criterion is environment validity,
not the base policy's success rate. Table~\ref{tab:robotwin-exclusions}
lists the tasks and the specific reasons. The editor experiments reported
here use the four-task S4 subset, whose base and editor evaluations are
reported in Appendices~\ref{app:robotwin-base-reference} and~\ref{app:s4-followup}.

\begin{table*}[t]
\centering\small
\caption{\textbf{Seven tasks temporarily excluded from the RoboTwin reference
cohort.} Paired tasks share the same reported issue.}
\label{tab:robotwin-exclusions}
\setlength{\tabcolsep}{4pt}
\begin{tabular}{@{}p{0.32\linewidth}p{0.64\linewidth}@{}}
\toprule
Task(s) & Reason for temporary exclusion \\
\midrule
\texttt{pick\_diverse\_bottles} &
Pre-contact bottle toppling was observed in our runs, with its cause
unresolved. Separately, the annotated grasp of \texttt{001\_bottle} variant 19
exceeds the ALOHA gripper opening
(\href{https://github.com/RoboTwin-Platform/RoboTwin/issues/482}{report 482}). \\[3pt]
\texttt{put\_object\_cabinet} &
\texttt{077\_phone} variant 4 has grasp contacts below the object's bottom
and an approach direction through the table, causing expert planning failures
(\href{https://github.com/RoboTwin-Platform/RoboTwin/issues/486}{report 486}). \\[3pt]
\texttt{place\_a2b\_left},\newline
\texttt{place\_a2b\_right} &
\texttt{086\_woodenblock} can be selected as the manipulated object A despite
missing grasp contacts and exceeding the ALOHA gripper opening
(\href{https://github.com/RoboTwin-Platform/RoboTwin/issues/536}{report 536}). \\[3pt]
\texttt{stack\_blocks\_two},\newline
\texttt{stack\_blocks\_three} &
A transient stack configuration can satisfy the success predicate without
sustained stability. The public falling-stack demonstration used a
different runtime from our SAPIEN setup
(\href{https://github.com/RoboTwin-Platform/RoboTwin/issues/491}{report 491}). \\[3pt]
\texttt{put\_bottles\_dustbin} &
A reported rollout knocks a bottle off the table into the acceptance region,
bypassing the instructed pick-and-place process
(\href{https://arxiv.org/html/2609.37771v1\#S3.SS2.SSS2}{benchmark analysis, Sec.~3.2.2}). \\
\bottomrule
\end{tabular}
\end{table*}

The asset concerns principally involve \texttt{aloha-agilex}; their impact
depends on the environment and asset revisions, robot, and seed filtering.
The toppling observation is distinct from the bottle grasp annotation issue:
filtering variant 19 alone does not resolve its unknown cause.
The block-stack report motivates a stability check but is not a reproduction
of that failure in our SAPIEN setup. These exclusions reflect the
environment version used here.

\section{LIBERO 112-Task Editors: Transfer, Controls, and Adaptation}
\label{app:libero-112}

This appendix supports Sections~\ref{sec:experiments:libero-reference},
\ref{sec:experiments:transfer-structure}, and~\ref{sec:experiments:adaptation}. It covers the three
112-task editors: protocol and resources, per-task results, source breadth,
checkpoint evolution, correction structure, inference-time controls, task
gradients, and single-task adaptation.

\subsection{LIBERO bounded-residual evaluation}
\label{app:libero-reference}

The evaluated bounded-residual configuration uses a shared MLP actor, an online-learned
encoder, ten critic heads per source task, five-action correction chunks,
and three-chunk TD backups, with batch size 256.
Table~\ref{tab:libero-training-resources} reports its training resources and
budget at the evaluated checkpoint. Available compute limited this
configuration to one training run of 481,250 updates;
Appendix~\ref{app:sw-robustness-seeds112} adds two seeds trained to 121.25k updates. The training curve continues
to the last recorded update at 488,600.

\begin{table}[tb]
\centering\footnotesize
\caption{\textbf{Training resources and observed process accounting for all three 112-source-task runs} (L1, L2, L3) at 481,250 actor updates. Each run uses four A100 GPUs, one for the learner and three for collectors. For L1 and L2, times are lower bounds from the recorded online windows.}
\label{tab:libero-training-resources}
\setlength{\tabcolsep}{4pt}
\begin{tabularx}{\linewidth}{@{}>{\raggedright\arraybackslash}Xrrr@{}}
\toprule
Item & L1 & L2 & L3 \\
\midrule
Independent runs / seed & 1 / 0 & 1 / 0 & 1 / 0 \\
Online actor updates & 481,250 & 481,250 & 481,250 \\
Initial fitting updates & 5,000 & 0 & 0 \\
Critic samples & 501.76M & 501.76M & --- \\
Physical steps / actions & 13.38M & 13.00M & 12.11M \\
Online process time & 42.9\,h* & 42.6\,h* & 41.8\,h \\
Allocated GPU-hours & --- & --- & 167 \\
\bottomrule
\end{tabularx}
\par\smallskip\raggedright\footnotesize
*Recorded online process windows (154,489\,s for L1; 153,455\,s for L2), without the initialization windows of 0.87\,h and 0.83\,h and without segments whose end was not recorded.
\end{table}

Training-process time sums three process segments through the reported
checkpoint: 41.76 hours, including loading, warm-up, checkpoint writes,
and work repeated after two restarts. Four allocated GPUs give 167.04
GPU-hours. Calendar elapsed time is 59.79 hours, including 18.03 hours
between processes. The segments that lead to the reported checkpoint account for approximately
37.66--37.68 process hours. Offline preparation and between-process gaps
are excluded from training-process time; post-checkpoint shutdown is outside
the measurement.

\paragraph{Context distance.}
The context-coverage analysis in Figure~\ref{fig:correction-support}c
computes each editor-held-out context's Euclidean distance to its nearest source
context in the common 256-dimensional feature space. Mean NN averages
these distances within each task, then equally across tasks.
Appendix~\ref{app:libero-breadth} describes the reference encoder and the
sampled source and query contexts.

\paragraph{Evaluation population.}
Both the frozen base and the editor execute five actions per decision.
Each policy is evaluated on 50 native initial states in two passes for
each of 130 tasks, giving 26,000 episodes and 13,000 paired cases.
Pairs match task, initial-state identity, pass, and reset seed.
The two passes repeat the same state pool, rather than adding 50 distinct
states. The base has been fine-tuned on all 130 tasks; the 18 editor-held-out tasks
are held out from editor training.

\paragraph{Predefined named-scene cohort.}
The split's reporting group contains tasks 38, 39, 69--72, 126, and 128:
three from \texttt{KITCHEN\_SCENE8} and five from
\texttt{LIVING\_ROOM\_SCENE6}. It was fixed before training, and no source
task has either scene identifier. Across the 800 paired cases in this
cohort, the base succeeds 196 times and the editor 330 times, giving
24.50\% and 41.25\% success. The grouping uses named scene identifiers;
cross-suite geometric overlap has not been assessed. These eight tasks form
the uncovered-scene cohort of the tables below; the other ten
editor-held-out tasks lie in source-covered scenes.

Table~\ref{tab:libero-reference-suites} separates the results by suite.
Editor-held-out performance varies within and across suites: Object success rises
from 29.00\% to 95.00\%, whereas editor-held-out Goal success falls from 40.00\%
to 27.50\%. Each of these editor-held-out cohorts contains two tasks.
Across all 1,800 editor-held-out pairs, the editor succeeds on 663 base failures
and fails on 183 base successes. Their difference accounts for the net
increase of 480 successful episodes. Both policies succeed in 237 cases
and fail in 717. These counts distinguish recovered failures from lost
successes within the aggregate improvement.

\paragraph{Task-level paired uncertainty.}
For the fixed 481,250-update checkpoint, we compute each editor-held-out
task's editor-minus-base success-rate difference from its 100 paired
episodes (50 initial states, two passes). We draw 18 tasks with replacement
and equal probability per replicate, retaining all episode pairs and both
passes within each selected task, and average their differences.
The 2.5th and 97.5th percentiles of 100,000 replicates give a 95\% interval
of [\LiberoHeldoutCILow, \LiberoHeldoutCIHigh] percentage points around
the observed \LiberoHeldoutGain-point gain. The computation uses NumPy
PCG64 with seed 20261112 and linearly interpolated quantiles.
This treats the empirical task set as an exchangeable resampling population.
It describes task-composition uncertainty conditional on the trained
checkpoint and observed task results, not training-seed variability or
uncertainty from new evaluation episodes. Different configurations and
checkpoints do not provide independent training replications.

\begin{table*}[t]
\centering\small
\caption{\textbf{Suite-level success rates (\%) across all three 112-source-task editors} (L1, L2, L3) at 481,250 updates. LIBERO-90 has 80 source and 10 editor-held-out tasks; each other suite has eight and two, respectively.}
\label{tab:libero-reference-suites}
\setlength{\tabcolsep}{4pt}
\begin{tabular}{@{}lrrrrrrrr@{}}
\toprule
& \multicolumn{4}{c}{Source tasks (112)} & \multicolumn{4}{c}{Editor-held-out tasks (18)} \\
\cmidrule(lr){2-5}\cmidrule(lr){6-9}
Suite & Base & L1 & L2 & L3 & Base & L1 & L2 & L3 \\
\midrule
LIBERO-90 & 14.46 & \second{94.50} & \best{94.60} & 93.59 & 20.50 & \best{65.00} & 56.10 & \second{56.20} \\
Spatial & 40.38 & \second{97.38} & 94.50 & \best{97.75} & 35.50 & \best{50.00} & \second{48.00} & 38.50 \\
Object & 30.25 & \second{99.50} & \best{99.75} & 99.00 & 29.00 & \best{99.00} & \second{97.00} & 95.00 \\
Goal & 46.88 & \best{97.12} & \second{96.62} & 95.12 & 40.00 & 27.00 & \best{47.00} & \second{27.50} \\
Long & 5.12 & \second{87.12} & \best{88.88} & 82.00 & 3.00 & \best{17.00} & \second{10.50} & 8.00 \\
\bottomrule
\end{tabular}
\end{table*}

\begin{table}[t]
\centering\small
\caption{\textbf{Editor-held-out task results.} Steps to Success averages physical actions over each policy's successful episodes; N/A denotes no successes. Task labels are abbreviated.}
\label{tab:libero-reference-heldout}
\setlength{\tabcolsep}{3pt}
\begin{tabularx}{\linewidth}{@{}>{\raggedright\arraybackslash}Xrrrr@{}}
\toprule
& \multicolumn{2}{c}{SR (\%)} & \multicolumn{2}{c}{Steps} \\ \cmidrule(lr){2-3}\cmidrule(lr){4-5} Task & Base & Editor & Base & Editor \\
\midrule
Butter into drawer; close & 6 & 84 & 306.5 & 172.4 \\
Bowl onto plate & 2 & 76 & 320.5 & 95.8 \\
Stack front bowl & 7 & 6 & 162.1 & 179.2 \\
Right moka pot onto stove & 1 & 1 & 267.0 & 350.0 \\
Turn off stove & 5 & 37 & 195.4 & 152.9 \\
Pan under shelf & 0 & 82 & N/A & 119.5 \\
Pudding left of plate & 56 & 72 & 177.9 & 143.5 \\
Pudding right of plate & 76 & 56 & 181.4 & 77.5 \\
Red mug onto plate & 52 & 68 & 145.0 & 97.5 \\
White mug onto plate & 0 & 80 & N/A & 132.5 \\
Bowl from ramekin & 50 & 58 & 122.0 & 237.2 \\
Bowl beside plate & 21 & 19 & 145.2 & 207.8 \\
Salad dressing into basket & 24 & 96 & 179.9 & 80.7 \\
Butter into basket & 34 & 94 & 197.2 & 120.6 \\
Wine onto cabinet & 55 & 24 & 124.3 & 106.3 \\
Cream cheese into bowl & 25 & 31 & 127.0 & 110.0 \\
Mug and pudding placement & 4 & 16 & 265.5 & 211.1 \\
Both moka pots onto stove & 2 & 0 & 418.0 & N/A \\
\bottomrule
\end{tabularx}

\end{table}

Table~\ref{tab:libero-reference-heldout} reports all 18 editor-held-out tasks.
Steps to Success is undefined when a policy has no successful episodes.
On the source cohort, the base and editor have defined values for 96 and
112 tasks, respectively; on the editor-held-out cohort, the counts are 16 and 17.
An average over each policy's defined tasks would therefore compare different
task populations. We report conditional lengths per task alongside success
rates rather than interpret that difference as a common-population speedup.
Even within tasks where both policies have successes, these conditional
means can involve different initial states and passes.

Restricting the comparison to matched cases where both policies succeed
leaves 237 pairs across 13 editor-held-out tasks. The editor has a shorter
mean episode length on ten of these tasks; the equal-task mean falls from
183.84 to 138.93 physical actions. This comparison is conditional on joint
success, rather than a speedup over the full evaluation population.
For example, on \emph{pudding left of plate}, means over each policy's own
successes fall from 177.93 to 143.53 actions, whereas its 40 jointly
successful pairs increase from 144.95 to 175.70 actions.

\paragraph{Training curve.}
At each recorded actor-update position, we compute success over the latest
20 completed episodes of each task, then average the 112 task rates equally.
The macro curve starts at 40,000 updates, when every task has a full window.
No further smoothing is applied. The curve follows the run through its
resumptions and uses only the segments that lead to the reported checkpoint.
Training episodes include exploration, whereas the frozen-checkpoint
evaluation executes the deterministic editor. Overlapping rolling windows
and repeated evaluation passes are not independent training seeds.

\begin{figure}[t]
    \centering
    \includegraphics[width=0.90\linewidth]{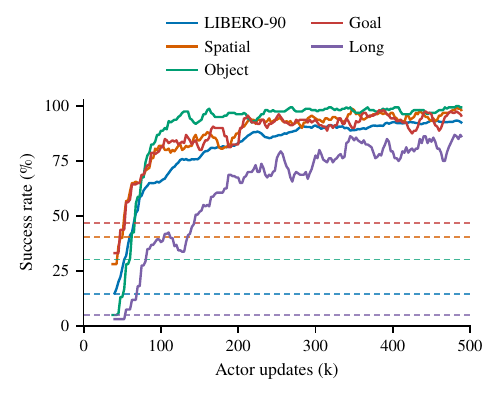}
    \caption{\textbf{LIBERO training across five suites.}
    Solid curves average each task's latest 20 training episodes with equal
    task weight. Same-color dashed lines show frozen-base evaluation on the
    corresponding source tasks: 80 for LIBERO-90 and eight per
    other suite. No additional smoothing is applied.}
    \label{fig:libero-training-suites}
\end{figure}

Figure~\ref{fig:libero-training-suites} separates this trajectory by suite.
LIBERO-90 contributes 80 source tasks, and each remaining suite contributes
eight. Within each suite, we average task rates equally and begin the curve
only when all its tasks have complete 20-episode windows.
The suite curves therefore begin at different update counts.
Each dashed reference uses frozen-base evaluation on the same suite's
source tasks under the five-action execution protocol.

\subsection{112-task soft-penalty recipes (L1 and L2) vs.\ bounded L3}
\label{app:libero-112-soft}

\begin{figure*}[t]
    \centering
    \includegraphics[width=0.92\textwidth]{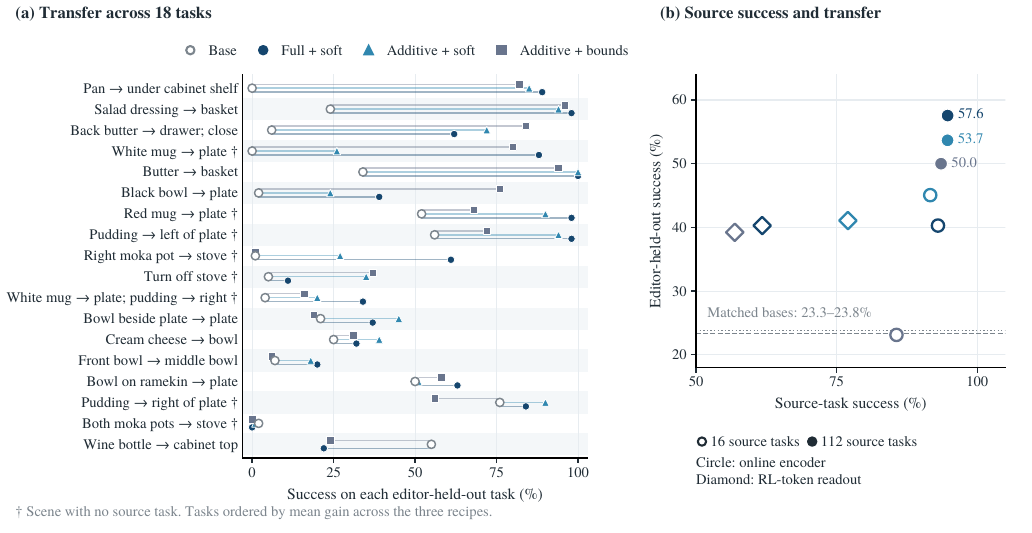}
    \caption{\textbf{Per-task transfer and source against held-out success.}
    (a)~Success on each of the 18 editor-held-out tasks for the base and the
    three 112-task editors (100 paired episodes per task).
    (b)~Source-task against editor-held-out success for the six 16-task
    editors and the three 112-task editors; dashed lines mark the matched
    bases. Each point is one training run.}
    \label{fig:libero112-task-transfer}
\end{figure*}

We train both soft-penalty online-encoder configurations on the exact same
112 LIBERO source tasks as bounded L3 using ReSOP through 481,250 online actor
updates (5,000 proposal-fitting updates for L1, no warm-up for L2; 22,715,427-byte
inference snapshots). Source evaluation covers all 112 tasks (50 states$\times$2
passes = 11,200 paired episodes); held-out evaluation covers all 18 editor-held-out
tasks (1,800 paired episodes, SHA-256 initial-state pairing verified identical
to Base and L3). Table~\ref{tab:libero112-terminal-summary} summarizes source,
held-out, paired recipe contrasts (5,000 intra-task initial-state-cluster bootstrap
draws), and first-pass controls against native eight-action base ($41.22\%$)
and L3 source-mean ($45.22\%$) across all 900 first-pass states; Table~\ref{tab:libero112-per-task}
reports all 18 per-task success rates, and
Figure~\ref{fig:libero112-task-transfer} plots them together with source
against held-out success for all nine editors.
Table~\ref{tab:libero112-p0-controls} gives the source-mean and frozen-frame
controls for L1 and L2. Each source-mean template is computed from that
recipe's own source-task corrections, so the comparison in
Section~\ref{sec:experiments:transfer-structure} is between each live editor
and its own context-free correction.

\begin{table*}[t]
\centering\footnotesize
\caption{\textbf{Comparison of the three 112-source-task configurations at 481,250
online updates} (11,200 source pairs; 1,800 held-out pairs from 50
states$\times$2 passes; 900 first-pass control pairs). Two-pass base SR is
23.33\% (All 18), 22.40\% (Cov.\ 10), and 24.50\% (Uncov.\ 8). Paired contrasts
use 5,000 state-cluster bootstrap draws; the L3 first-pass intervals are those
of Table~\ref{tab:libero-controls-summary}. Rescue (R) and harm (H) are shares of
the paired episodes in each cohort.}
\label{tab:libero112-terminal-summary}
\setlength{\tabcolsep}{2.6pt}
\begin{tabularx}{\textwidth}{@{}>{\raggedright\arraybackslash}Xrrrrrrrr@{}}
\toprule
Recipe & Src.\,112 & All\,18 & $\Delta$ Base [95\% CI] & Rescue & Harm & Cov.\,10 (R/H) & Unc.\,8 (R/H) & First pass vs.\ $K{=}8$/Mean \\
\midrule
L3 (Add.+bound) & 93.55 & 50.00 & +26.67 & 36.8 & 10.2 & 57.00 (44.5/9.9) & 41.25 (27.3/10.5) & +7.22 / +3.22 \\
L2 (Add.+soft) & 94.70 & \second{53.67} & \second{+30.33 [27.44, 33.33]} & 35.2 & \best{4.9} & \best{58.30} (42.0/\best{6.1}) & \second{47.88} (26.8/3.4) & \second{+12.33} / \second{+8.33} \\
\hlrow
L1 (Full+soft) & \best{94.72} & \best{57.56} & \best{+34.22 [31.39, 37.11]} & \best{40.4} & \second{6.2} & 56.20 (42.4/8.6) & \best{59.25} (\best{38.0}/\best{3.3}) & \best{+16.56} / \best{+12.56} \\
\bottomrule
\end{tabularx}
\par\smallskip
\raggedright
Cross-recipe paired two-pass contrasts (All 18 / Covered 10 / Uncovered 8):
L1$-$L3 = +7.56 [+4.72, +10.33] ($-$0.80 / +18.00);
L2$-$L3 = +3.67 [+0.78, +6.61] (+1.30 / +6.63);
L1$-$L2 = +3.89 [+1.11, +6.61].
First pass vs.\ $K=8$ Base (41.22\%): L1 57.78\% (+16.56 [13.11, 20.00]),
L2 53.56\% (+12.33 [8.89, 15.89]), L3 48.44\% (+7.22 [3.78, 10.67]);
vs.\ L3 source-mean (45.22\%): L1 +12.56 [9.33, 15.67], L2 +8.33 [4.78, 12.00], L3 +3.22 [$-$0.22, 6.67].
\end{table*}

\begin{table}[t]
\centering\footnotesize
\caption{\textbf{Per-task two-pass success rates (\%) across all 18 editor-held-out
LIBERO tasks} for Base, L1, L2, and L3 (100 paired episodes per task). Task~IDs
and scene-coverage cohorts match the predefined split.}
\label{tab:libero112-per-task}
\setlength{\tabcolsep}{2.2pt}
\begin{tabularx}{\linewidth}{@{}l>{\raggedright\arraybackslash}Xrrrr@{}}
\toprule
ID (Grp.) & Task & Base & L1 & L2 & L3 \\
\midrule
3 (Cov.) & Butter into drawer; close & 6 & 62 & 72 & \best{84} \\
9 (Cov.) & Bowl onto plate & 2 & 39 & 24 & \best{76} \\
16 (Cov.) & Stack front bowl & 7 & \best{20} & 18 & 6 \\
38 (Unc.) & Right moka pot onto stove & 1 & \best{61} & 27 & 1 \\
39 (Unc.) & Turn off stove & 5 & 11 & 35 & \best{37} \\
42 (Cov.) & Pan under shelf & 0 & \best{89} & 85 & 82 \\
69 (Unc.) & Pudding left of plate & 56 & \best{98} & 94 & 72 \\
70 (Unc.) & Pudding right of plate & 76 & 84 & \best{90} & 56 \\
71 (Unc.) & Red mug onto plate & 52 & \best{98} & 90 & 68 \\
72 (Unc.) & White mug onto plate & 0 & \best{88} & 26 & 80 \\
95 (Cov.) & Bowl from ramekin & 50 & \best{63} & 51 & 58 \\
98 (Cov.) & Bowl beside plate & 21 & 37 & \best{45} & 19 \\
102 (Cov.) & Salad dressing into basket & 24 & \best{98} & 94 & 96 \\
106 (Cov.) & Butter into basket & 34 & \best{100} & \best{100} & 94 \\
112 (Cov.) & Wine onto cabinet & \best{55} & 22 & \best{55} & 24 \\
116 (Cov.) & Cream cheese into bowl & 25 & 32 & \best{39} & 31 \\
126 (Unc.) & Mug and pudding placement & 4 & \best{34} & 20 & 16 \\
128 (Unc.) & Both moka pots onto stove & \best{2} & 0 & 1 & 0 \\
\midrule
Cov.\ (10) & Covered macro mean & 22.40 & 56.20 & \best{58.30} & 57.00 \\
Unc.\ (8) & Uncovered macro mean & 24.50 & \best{59.25} & 47.88 & 41.25 \\
\sumrow
All 18 & Held-out macro mean & 23.33 & \best{57.56} & \second{53.67} & 50.00 \\
\bottomrule
\end{tabularx}
\end{table}

\begin{table*}[t]
\centering\footnotesize
\caption{\textbf{Matched first-pass source-mean and frozen-frame controls for L1 and L2} (18 editor-held-out tasks, 900 paired states per condition, identical initial-state SHA-256 to Base). Rescue and harm are shares of the 900 pairs against Base (24.56\%, 221/900).}
\label{tab:libero112-p0-controls}
\setlength{\tabcolsep}{3pt}
\begin{tabularx}{\textwidth}{@{}>{\raggedright\arraybackslash}Xrrrrrr@{}}
\toprule
Condition (first pass) & All 18\,$\uparrow$ & Cov.\,10\,$\uparrow$ & Uncov.\,8\,$\uparrow$ & Rescue\,$\uparrow$ & Harm\,$\downarrow$ & $\Delta$ vs.\ own live [95\% CI] \\
\midrule
Base (5-action) & 24.56 & 24.60 & 24.50 & 0.0 & 0.0 & --- \\
\midrule
112-task L1 (live) & \best{57.78} & \best{56.20} & \best{59.75} & \best{40.3} & 7.1 & --- \\
L1 source-mean ($5\times 7$) & 39.67 & 43.60 & 34.75 & 21.0 & \best{5.9} & $-$18.11 [$-$21.22, $-$14.78] \\
L1 frozen editor frames & 45.33 & 48.20 & 41.75 & 29.3 & 8.6 & $-$12.44 [$-$15.67, $-$9.22] \\
\midrule
112-task L2 (live) & \best{53.56} & \best{58.40} & \best{47.50} & \best{34.2} & \best{5.2} & --- \\
L2 source-mean ($5\times 7$) & 41.56 & 44.20 & 38.25 & 23.6 & 6.6 & $-$12.00 [$-$15.67, $-$8.56] \\
L2 frozen editor frames & 37.44 & 45.60 & 27.25 & 25.1 & 12.2 & $-$16.11 [$-$19.34, $-$12.78] \\
\bottomrule
\end{tabularx}
\end{table*}

\paragraph{Structure of rescue and harm across the three recipes.}
The following analyses regroup the 1,800 paired held-out episodes of each
112-task recipe after evaluation; the groupings were not fixed in advance.
\emph{Base competence.} Table~\ref{tab:outcome-by-base} groups the 18 tasks
by base SR. On the nine tasks the base solves at most 7\% of the time, every
recipe gains 31--42 points with at most 2.0\% harm. On the five tasks it
solves at least half the time, harm rises to 11.2--26.8\%; L1 and L2 still
gain 15.2 and 18.2 points, whereas L3 loses 2.2.
\emph{Overlap.} At least one recipe rescues 1,000 of the 1,800 pairs (55.6\%)
and all three rescue 361 (20.1\%). At least one recipe harms 241 pairs
(13.4\%), but all three harm only 34 (1.9\%) and 133 are harmed by exactly
one recipe, so another recipe keeps most harmed successes.
\emph{Concentration.} The three tasks with the most harm hold 63.4\%, 53.4\%
and 53.6\% of the harm of L1, L2 and L3; wine onto the cabinet alone accounts
for 41 of 112, 18 of 88 and 38 of 183 harmed pairs.
\emph{Reproducibility of the base success.} The base's two passes disagree on
62 of the 900 initial states (6.9\%). Harm reaches 54.8\%, 35.5\% and 58.1\%
of the 62 base successes on those states under L1, L2 and L3, against
21.8\%, 18.4\% and 41.1\% of the 358 successes that the base reproduces in
both passes. As a scale for run-to-run variation, the base succeeds in the first pass
and fails in the second on 42 states (4.7\%) and the reverse on 20 (2.2\%); two
first-pass base evaluations on different machines differ on 113 of 900 pairs
(12.6\%). The passes use different reset seeds, so these figures are not a
matched null for the base--editor pairing.
\emph{Best scale per task.} For the L3 scaling control
(Table~\ref{tab:libero-l2-summary}), choosing in hindsight the best of the
original, half and quarter scales for each task gives 66.44\%, against 57.44\%
for half scale on every task. All eight uncovered-scene tasks prefer a reduced
scale, and five of the ten covered-scene tasks prefer the original one. This
is a hindsight bound, not a policy.

\begin{table*}[t]
\centering\small
\caption{\textbf{Rescue and harm by base competence (112-task recipes).}
The 18 editor-held-out tasks are grouped by two-pass base SR after evaluation;
rescue (R) and harm (H) are shares of each group's paired episodes, and SR is
the editor's success rate. All values are percentages.}
\label{tab:outcome-by-base}
\setlength{\tabcolsep}{4pt}
\begin{tabularx}{\textwidth}{@{}>{\raggedright\arraybackslash}Xrrrrrrrrrrr@{}}
\toprule
 & & & \multicolumn{3}{c}{L1} & \multicolumn{3}{c}{L2} & \multicolumn{3}{c}{L3} \\
\cmidrule(lr){4-6} \cmidrule(lr){7-9} \cmidrule(l){10-12}
Base SR group (tasks) & Pairs & Base & R & H & SR & R & H & SR & R & H & SR \\
\midrule
$\le 7\%$ (3, 9, 16, 38, 39, 42, 72, 126, 128) & 900 & 3.0 & 43.9 & 2.0 & 44.9 & 32.6 & 1.3 & 34.2 & 41.1 & 1.7 & 42.4 \\
21--34\% (98, 102, 106, 116) & 400 & 26.0 & 46.3 & 5.5 & 66.8 & 48.5 & 5.0 & 69.5 & 42.5 & 8.5 & 60.0 \\
$\ge 50\%$ (69, 70, 71, 95, 112) & 500 & 57.8 & 29.6 & 14.4 & 73.0 & 29.4 & 11.2 & 76.0 & 24.6 & 26.8 & 55.6 \\
\bottomrule
\end{tabularx}
\end{table*}

\paragraph{Replay in Figure~\ref{fig:correction-support}e.}
The replay case was fixed from the evaluation records before any frame was
viewed: on a predeclared task, the lowest-index initial state on which the
base fails and the L1 editor succeeds in both passes. On this task (placing
the frying pan under the cabinet shelf, Task~42), the two-pass evaluation
gives 0\% for the base and 89\% for L1. In the replay from initial state~0,
the base times out at 512 actions with a per-decision arm-command norm below
0.01 throughout; the editor succeeds at step 277 (254 in the original
evaluation pass: the outcomes agree, the lengths do not). The trace plots,
for each of the 56 editor decisions, the mean over valid chunk slots of the
$\ell_2$ norm of the six arm-command coordinates, for the base proposal and
for the executed command minus that proposal. The base proposal is below 0.01
at 28 of the 56 decisions, and the editor's change stays between 0.04 and
0.09. Rendered frames are $512\times512$ displays; policy inputs keep their
native resolution. This is one qualitative case, not an estimate of how often
base failures are stalls.

\subsection{Source breadth and editor-held-out transfer}
\label{app:libero-breadth}

The source-breadth comparison evaluates a two-task editor at 250,000 actor
updates and checkpoints of the 112-task editor. Both use the same frozen
base, the MLP actor and encoder architecture family, five-action chunks,
three-chunk backups, batch size 256, and training seed 0.
At 241,250 actor updates, the 112-task run has 1,000,000 total critic updates,
including warm-up, compared with 1,000,625 for the two-task run.
The 51,250-update checkpoint provides a second comparison at similar interaction
counts: 1,628,580 versus 1,529,962 physical actions, a 6.4\% difference.
All conditions use the same 18 editor-held-out tasks, 50 native initial states
and two passes per state, paired across policies (Table~\ref{tab:libero-reference}).

\begin{table}[t]
\centering\small
\caption{\textbf{Source-task learning and editor-held-out generalization on LIBERO-130.} All editor-held-out results use the same 18 tasks and 50 initial states in two passes per task. SR is reported in percent. Updates count actor updates; actions count physical training interactions; $\Delta$ is the gain over the base in percentage points.}
\label{tab:libero-reference}
\setlength{\tabcolsep}{3pt}
\begin{tabularx}{\linewidth}{@{}>{\raggedright\arraybackslash}Xrrr@{}}
\toprule
Task cohort & Base & Editor & $\Delta$ \\
\midrule
\multicolumn{4}{@{}l}{\textit{2 tasks; 250k updates; 1.530M actions}} \\
Source (2) & 18.00 & 93.50 & +75.50 \\
Editor-held-out (18) & 23.33 & 1.00 & $-$22.33 \\
\midrule \multicolumn{4}{@{}l}{\textit{112 tasks; 51.25k updates; 1.629M actions}} \\
Editor-held-out (18) & 23.33 & 44.22 & +20.89 \\
\midrule \multicolumn{4}{@{}l}{\textit{112 tasks; 241.25k updates; 6.629M actions}} \\
Editor-held-out (18) & 23.33 & \best{50.89} & \best{+27.56} \\
\midrule \multicolumn{4}{@{}l}{\textit{112 tasks; 481.25k updates; 12.111M actions}} \\
Source (112) & 19.09 & \best{93.55} & +74.46 \\
\hlrow
Editor-held-out (18) & 23.33 & \second{50.00} & \second{+26.67} \\
\bottomrule
\end{tabularx}

\end{table}

The two source tasks are the LIBERO-Long tasks of placing a black bowl in
the bottom drawer and closing it, and placing a book in the back compartment
of a caddy. The editor reaches 89\% and 98\%
success on them, respectively, versus 1\% and 35\% for the base.
Its high source-task success and low editor-held-out success show that learning
effective corrections on these source tasks does not ensure transfer.

The two-task and 112-task configurations use 625/35,000 warm-up critic
updates, 100/5,600 expert demonstrations, and two/four tasks per learner
batch, respectively. Each uses ten critic heads per source task and its
own recorded state normalization.

The base and final 112-task results reuse both passes of the evaluation
in Appendix~\ref{app:libero-reference}, rather than adding independent replications.
Figure~\ref{fig:correction-support}d uses the final editor's action
records. The visual coverage analysis in panel (c) instead uses one
common reference encoder from the 112-task policy at 241,250 updates.

\paragraph{Visual context coverage.}
The two-task and 112-task source pools contain 135 and 6,966 sampled
post-training contexts. Each source subset contains 112 observations:
56 per task in the two-task condition and one per task in the 112-task
condition. We repeat this subset sampling 32 times. All source observations
and the fixed 1,222 query contexts from the final policy's visits to
18 editor-held-out tasks are encoded in the same reference space.
For each camera, we mean-pool its spatial tokens, concatenate main and
wrist vectors, and L2-normalize the resulting 256-dimensional feature.
Nearest-neighbor distances are computed in this space and averaged with
equal target-task weights, giving 0.3264 for the two-task and 0.1593 for the
112-task source subsets.
The two-dimensional PCA is only a visualization: it is fitted on the
combined source subset for analysis seed zero and does not use query
observations for fitting. These subsets describe recorded post-training
visits, not the complete training replay, and the reference is a visual
proxy rather than the editor's full input context.

\subsection{Editor-held-out evolution within one training run}
\label{app:libero-evolution}

We evaluate six checkpoints of the same 112-task run
(Table~\ref{tab:libero-evolution}). Together with the frozen base and the
fixed two-task editor, the evaluation covers $8\times18\times50\times2=14{,}400$
episodes. Initial-state identities and hashes match across conditions;
evaluation seed is 0 and the reset seed equals the pass index (0 or 1).
The checkpoints use five-action chunks and three-chunk training backups.
Evaluation retains the 512-action horizon, ten settling no-ops,
normalization and post-policy clipping. Technical errors are retried as incomplete cases, not counted as
policy failures; duplicate provenance is retained separately.

For the checkpoint curves, we hold the 18 tasks fixed and draw 50 initial
states with replacement within each task for 10,000 bootstrap replicates.
Both passes of a state remain together, and all checkpoints and the base
share the same sampled indices. Pointwise intervals use the 2.5th and
97.5th percentiles. This measures evaluation-state uncertainty conditional
on the fixed tasks and trained models; the task-composition interval in
Appendix~\ref{app:libero-reference} uses a different resampling unit.

\begin{table}[t]
\centering\small
\caption{\textbf{Six checkpoints of one 112-task training run.} SR (\%) uses the same 18 editor-held-out tasks and 50 initial states in two passes. Brackets give pointwise 95\% initial-state-cluster bootstrap intervals. Actions are physical training interactions; 481.25k remains the fixed final checkpoint.}
\label{tab:libero-evolution}
\setlength{\tabcolsep}{3pt}
\begin{tabular}{@{}rrr@{}}
\toprule
Updates (k) & Actions (M) & SR [95\% CI] \\
\midrule
1.25 & 0.040 & 16.61 [14.78, 18.44] \\
51.25 & 1.629 & 44.22 [41.94, 46.50] \\
121.25 & 3.581 & 48.67 [46.33, 51.00] \\
241.25 & 6.629 & \best{50.89} [48.78, 53.11] \\
351.25 & 9.179 & 49.06 [46.83, 51.28] \\
\hlrow
481.25 & 12.111 & \second{50.00} [47.83, 52.17] \\
\bottomrule
\end{tabular}

\end{table}

For the six paired gains over the base, an approximate simultaneous band
uses the 95th percentile of the maximum absolute centered bootstrap
deviation across the six checkpoints. Among these evaluated checkpoints, 51.25k
is the first with a positive lower bound; this does not locate a continuous
onset time. The sampled maximum is 50.89\% at 241.25k, while the fixed final
481.25k model reaches 50.00\%. These correlated checkpoints do not supply
independent training seeds, and this evaluation does not select a new final model.

\begin{figure*}[t]
    \centering
    \includegraphics[width=0.82\textwidth]{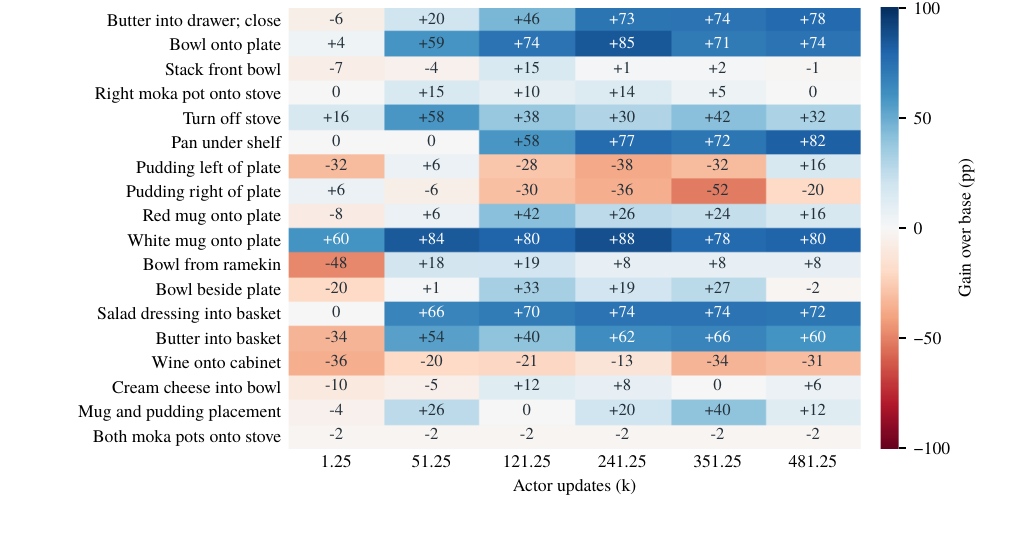}
    \caption{\textbf{Editor-held-out gains across six checkpoints.}
    Each cell is the task's editor-minus-base success-rate difference in
    percentage points, using 50 paired initial states in two passes.
    Rows retain the same 18 task identities; columns are ordered by
    actor updates. Blue indicates a gain and red a regression.}
    \label{fig:libero-heldout-evolution-tasks}
\end{figure*}

Figure~\ref{fig:libero-heldout-evolution-tasks} shows paired per-task gains
across training. The fixed final model still improves 12 tasks, reduces
success on five, and matches the base on one. The interval where aggregate
success stays near 50\% contains changes in individual task outcomes.

\subsection{Cross-task correction reconstruction}
\label{app:correction-structure}

Figure~\ref{fig:correction-support}d analyzes the effective correction, the
executed policy action minus the base action, at checkpoint 481,250.
Each of five consecutive seven-dimensional corrections is divided by the residual
scales (0.1 for pose coordinates and 2 for the gripper), then concatenated
into a 35-dimensional vector. The directions are fitted on
\SWcsSubNsrcLthree{} source-task chunks and applied to the
\SWcsSubNheldLthree{} chunks of every held-out decision in one first-pass
evaluation. Statistics weight tasks and, within a task, episodes equally.

Let $\mu$ and $U_r$ be the weighted mean and first $r$ principal directions
fitted on source-task corrections. For editor-held-out correction $e_i$ with weight $w_i$,
the captured fraction is
\begin{equation}
 C_r=\frac{\sum_i w_i\lVert U_r^\top(e_i-\mu)\rVert^2}
                {\sum_i w_i\lVert e_i-\mu\rVert^2}.
\end{equation}
At ten dimensions, this fraction is \SWcsSubSrcTenLthree\% for source-task
corrections and \SWcsSubTenLthree\% for editor-held-out corrections
(Table~\ref{tab:sw-corrsize-subspace} gives all three recipes). To separate
temporal structure, we also decompose each chunk into its five-step mean and
deviations from that mean; the within-chunk mean holds
\SWcsSubBlockLthree\% of the squared held-out correction. A separate
analysis on a smaller sample, up to five elapsed-progress bins per
trajectory (1,668 source-task and 270 held-out chunks), gives 89.2\% and
86.1\% at ten dimensions and 79.1\% and 73.1\% for the within-chunk mean.
These are scale-normalized command statistics, rather than physical energy.
The reconstruction describes overlap in output directions; whether a correction
is useful depends on its direction and magnitude in the task context.

\subsection{Inference-time and prefix controls on the 18 editor-held-out tasks}
\label{app:libero-controls}

\begin{table*}[t]
\centering\small
\caption{\textbf{Paired first-pass held-out control contrasts} across all 18 tasks and the predefined source-covered (10) and uncovered (8) cohorts (50 states per task). Rescues are pairs where $B$ fails and $A$ succeeds, harms
where $B$ succeeds and $A$ fails; percentages are shares of the cohort's pairs. 112-L3 denotes the 112-task L3 editor.}
\label{tab:libero-controls-summary}
\setlength{\tabcolsep}{4pt}
\begin{tabularx}{\textwidth}{@{}>{\raggedright\arraybackslash}Xrrrrr@{}}
\toprule
Contrast ($A - B$) & Cohort (eps) & SR $A$ / $B$ (\%) & $\Delta$ [95\% CI] (points) & Rescues (share) & Harms (share) \\
\midrule
Base $K=8$ $-$ Base $K=5$ & All 18 (900) & 41.22 / 24.56 & +16.67 [+13.78, +19.67] & 188 (20.9\%) & 38 (4.2\%) \\
Source-mean $-$ Base $K=5$ & All 18 (900) & 45.22 / 24.56 & +20.67 [+17.44, +23.89] & 248 (27.6\%) & 62 (6.9\%) \\
Source-mean $-$ Base $K=8$ & All 18 (900) & 45.22 / 41.22 & +4.00 [+0.56, +7.44] & 158 (17.6\%) & 122 (13.6\%) \\
112-L3 $-$ Base $K=5$ & All 18 (900) & 48.44 / 24.56 & +23.89 [+20.56, +27.22] & 315 (35.0\%) & 100 (11.1\%) \\
112-L3 $-$ Base $K=8$ & All 18 (900) & 48.44 / 41.22 & +7.22 [+3.78, +10.67] & 219 (24.3\%) & 154 (17.1\%) \\
Editor first-frame $-$ 112-L3 & All 18 (900) & 37.78 / 48.44 & $-$10.67 [$-$14.11, $-$7.33] & 113 (12.6\%) & 209 (23.2\%) \\
112-L3 $-$ Source-mean & All 18 (900) & 48.44 / 45.22 & +3.22 [$-$0.22, +6.67] & 186 (20.7\%) & 157 (17.4\%) \\
\midrule
\hlrow
112-L3 $-$ Source-mean & Covered 10 (500) & 55.20 / 49.20 & \best{+6.00 [+1.00, +10.80]} & 126 (25.2\%) & 96 (19.2\%) \\
112-L3 $-$ Base $K=8$ & Covered 10 (500) & 55.20 / 45.60 & +9.60 [+4.60, +14.40] & 145 (29.0\%) & 97 (19.4\%) \\
Editor first-frame $-$ 112-L3 & Covered 10 (500) & 43.00 / 55.20 & $-$12.20 [$-$16.60, $-$7.80] & 69 (13.8\%) & 130 (26.0\%) \\
112-L3 $-$ Source-mean & Uncovered 8 (400) & 40.00 / 40.25 & $-$0.25 [$-$5.00, +4.75] & 60 (15.0\%) & 61 (15.3\%) \\
112-L3 $-$ Base $K=8$ & Uncovered 8 (400) & 40.00 / 35.75 & +4.25 [$-$0.50, +9.00] & 74 (18.5\%) & 57 (14.3\%) \\
Editor first-frame $-$ 112-L3 & Uncovered 8 (400) & 31.25 / 40.00 & $-$8.75 [$-$13.75, $-$3.75] & 44 (11.0\%) & 79 (19.8\%) \\
\bottomrule
\end{tabularx}
\end{table*}

\begin{table}[t]
\centering\small
\caption{\textbf{Per-task first-pass success rates (\%) across all 18 editor-held-out tasks} (50 native initial states each). $K5$/$K8$: base with five/eight executed actions; Mean: source-mean correction; Frz.: frozen editor frames; L3: live 112-task L3 editor.}
\label{tab:libero-controls-tasks}
\setlength{\tabcolsep}{2.2pt}
\begin{tabularx}{\linewidth}{@{}l>{\raggedright\arraybackslash}Xrrrrr@{}}
\toprule
ID (Grp.) & Task & $K5$ & $K8$ & Mean & Frz. & L3 \\
\midrule
3 (Cov.) & Butter into drawer; close & 8 & 56 & 84 & 54 & 82 \\
9 (Cov.) & Bowl onto plate & 2 & 40 & 32 & 66 & 72 \\
16 (Cov.) & Stack front bowl & 8 & 18 & 46 & 0 & 4 \\
38 (Unc.) & Right moka pot onto stove & 0 & 8 & 32 & 0 & 0 \\
39 (Unc.) & Turn off stove & 6 & 12 & 14 & 2 & 28 \\
42 (Cov.) & Pan under shelf & 0 & 2 & 10 & 24 & 86 \\
69 (Unc.) & Pudding left of plate & 56 & 82 & 88 & 38 & 72 \\
70 (Unc.) & Pudding right of plate & 76 & 88 & 88 & 62 & 56 \\
71 (Unc.) & Red mug onto plate & 52 & 74 & 50 & 60 & 68 \\
72 (Unc.) & White mug onto plate & 0 & 10 & 44 & 58 & 80 \\
95 (Cov.) & Bowl from ramekin & 52 & 74 & 74 & 42 & 48 \\
98 (Cov.) & Bowl beside plate & 26 & 38 & 20 & 88 & 18 \\
102 (Cov.) & Salad dressing into basket & 24 & 40 & 56 & 34 & 96 \\
106 (Cov.) & Butter into basket & 34 & 70 & 90 & 100 & 94 \\
112 (Cov.) & Wine onto cabinet & 60 & 72 & 68 & 4 & 30 \\
116 (Cov.) & Cream cheese into bowl & 32 & 46 & 12 & 18 & 22 \\
126 (Unc.) & Mug and pudding placement & 4 & 6 & 0 & 30 & 16 \\
128 (Unc.) & Both moka pots onto stove & 2 & 6 & 6 & 0 & 0 \\
\bottomrule
\end{tabularx}
\end{table}

All five conditions in this control study share the exact first-pass schedule on
the 18 editor-held-out tasks (native initial states 0--49 per task, reset
seed~0, evaluation seed~0, 512-step horizon, 10 settling no-ops, 900 paired
episodes per arm; initial-state SHA-256 digests verified identical across
all arms). \emph{Base $K=5$} and \emph{112-L3 contextual} reuse the
first-pass subset of the 481,250-update evaluation without re-running. \emph{Base $K=8$} changes only the
executed chunk prefix of the frozen base policy from five to its full eight
predicted actions; note that the published 42.09\% model-card figure covers
all 130 tasks under the publisher's protocol and is not a direct 18-task
equivalent, nor does $K=8$ alone reproduce that full pipeline.
\emph{Source-mean ($5\times 7$)} replaces the editor network with the fixed
35-dimensional mean source correction from the 1,668 source-task chunks in
Appendix~\ref{app:correction-structure} (scaled once from actor units via
$[0.1{\times}6,\,2.0]$), added onto live $K=5$ base proposals under the same
post-composition clipping while the frozen base retains live vision.
\emph{Editor first-frame freeze} freezes only the editor's main and wrist
images at step~0 while leaving live base vision, proprioception, instruction,
and $B_t$ unchanged. Confidence intervals use 5,000 paired intra-task native
state resamples (seed~0); these measure state-sampling uncertainty conditional
on the fixed 18 tasks and single trained 112-task model rather than training-seed
variance.

Table~\ref{tab:libero-controls-summary} summarizes all predeclared and cohort
contrasts, and Table~\ref{tab:libero-controls-tasks} lists all 18 tasks. On
the 250 matched episodes where both \emph{Source-mean} and \emph{112-L3}
succeed, \emph{112-L3} completes trajectories 37.19 control steps faster
(126.75 vs.\ 163.94 steps). Per-task results differ although the aggregate
difference is small: for example, on covered Task~42 (86\% vs.\ 10\%) and
Task~102 (96\% vs.\ 56\%) the live editor is higher, whereas on Task~16
(4\% vs.\ 46\%) and Task~112 (30\% vs.\ 68\%) the constant mean is higher; under
first-frame freezing, Task~98 rises from 18\% to 88\%, so stale
editor frames do not lower every task.

\paragraph{Correction scaling, gripper, and instruction-swap controls.}
We next test whether the harm of the 112-task L3 editor on the first pass
stems from output amplitude, gripper corrections, or instruction sensitivity.
The scaling and gripper controls use 900 matched first-pass pairs across the
same 18 editor-held-out tasks per condition; the instruction swap uses 300
matched pairs across Tasks~69, 70, 71, 72, 112, and 116
(Tables~\ref{tab:libero-l2-summary}, \ref{tab:libero-l2-tasks}
and~\ref{tab:libero-l3-prompt-swap}). All
conditions keep the 481,250-update L3 weights frozen. The scaling controls
multiply the executed correction by 0.5 or 0.25, the gripper control zeroes its
seventh coordinate, and the instruction swap replaces only the editor's task
prompt while the base policy keeps its original instruction. These controls
ran on RTX~4090 GPUs, so they are compared against the base and the original
L3 editor re-evaluated in the same batch (base 23.56\%, 212/900; original L3
48.22\%, 434/900) rather than against the A100 evaluation (base 24.56\%,
221/900). Within this batch, the base and the original L3 ran
on different GPUs, and the GPU assignment per episode was not recorded for
the intervention runs; the L3 editor was trained with post-policy action
clipping and is executed without it. Confidence intervals use 2,000
task-stratified paired case bootstrap draws (seed~20261115).

\begin{table*}[t]
\centering\small
\caption{\textbf{Amplitude-scaling and gripper-ablation controls of the frozen 112-task
L3 editor} (first pass on RTX~4090, 18 held-out tasks $\times$ 50 states = 900
pairs per condition). The original L3 of the same batch is the comparator (48.22\%,
Cov.~10: 54.00\%, Uncov.~8: 41.00\%); Rescue and Harm are shares of the cohort's
pairs against the base of that batch (23.56\%, 212/900; Cov.~10: 22.60\%,
Uncov.~8: 24.75\%), with counts in parentheses.}
\label{tab:libero-l2-summary}
\setlength{\tabcolsep}{3.8pt}
\begin{tabularx}{\textwidth}{@{}>{\raggedright\arraybackslash}Xrrrrr@{}}
\toprule
Condition vs.\ original L3 & Cohort ($n$) & SR / Orig.\ L3\,(\%) & $\Delta$ SR [95\% CI]\,(points) & Rescue vs.\ Base\,(\%) & Harm vs.\ Base\,(\%) \\
\midrule
Original L3 & All 18 (900) & 48.22 / --- & --- & 33.89 (305) & 9.22 (83) \\
\hlrow
0.5$\times$ L3 scaling & All 18 (900) & \best{57.44} / 48.22 & \best{+9.22 [+6.00, +12.44]} & \best{38.78} (349) & \best{4.89} (44) \\
0.25$\times$ L3 scaling & All 18 (900) & \second{56.56} / 48.22 & \second{+8.33 [+4.89, +11.67]} & \second{38.33} (345) & \second{5.33} (48) \\
Zero-gripper L3 & All 18 (900) & 39.89 / 48.22 & $-$8.33 [$-$11.22, $-$5.33] & 27.78 (250) & 11.44 (103) \\
\midrule
Original L3 & Cov.\ 10 (500) & 54.00 / --- & --- & \best{41.60} (208) & 10.20 (51) \\
0.5$\times$ L3 scaling & Cov.\ 10 (500) & \best{57.00} / 54.00 & +3.00 [$-$1.80, +7.40] & 40.40 (202) & \best{6.00} (30) \\
0.25$\times$ L3 scaling & Cov.\ 10 (500) & \second{55.60} / 54.00 & +1.60 [$-$3.00, +6.60] & \second{40.80} (204) & \second{7.80} (39) \\
Zero-gripper L3 & Cov.\ 10 (500) & 40.60 / 54.00 & $-$13.40 [$-$17.41, $-$9.20] & 31.20 (156) & 13.20 (66) \\
\midrule
Original L3 & Uncov.\ 8 (400) & 41.00 / --- & --- & 24.25 (97) & 8.00 (32) \\
\hlrow
0.5$\times$ L3 scaling & Uncov.\ 8 (400) & \best{58.00} / 41.00 & \best{+17.00 [+12.25, +21.25]} & \best{36.75} (147) & \second{3.50} (14) \\
0.25$\times$ L3 scaling & Uncov.\ 8 (400) & \second{57.75} / 41.00 & \second{+16.75 [+11.75, +21.50]} & \second{35.25} (141) & \best{2.25} (9) \\
Zero-gripper L3 & Uncov.\ 8 (400) & 39.00 / 41.00 & $-$2.00 [$-$6.25, +2.25] & 23.50 (94) & 9.25 (37) \\
\bottomrule
\end{tabularx}
\end{table*}

\begin{table}[t]
\centering\footnotesize
\caption{\textbf{Per-task first-pass success rates (\%) in the scaling and
gripper-zeroing study} (50 paired native initial states per task; Base and L3
are the runs of the same RTX~4090 batch).}
\label{tab:libero-l2-tasks}
\setlength{\tabcolsep}{2.2pt}
\begin{tabularx}{\linewidth}{@{}l>{\raggedright\arraybackslash}Xrrrrr@{}}
\toprule
ID (Grp.) & Task & Base & L3 & 0.5$\times$ & 0.25$\times$ & NoGrip \\
\midrule
3 (Cov.) & Butter into drawer; close & 14 & 78 & 92 & \best{96} & 50 \\
9 (Cov.) & Bowl onto plate & 2 & \best{66} & 52 & 30 & 56 \\
16 (Cov.) & Stack front bowl & 4 & 6 & 6 & \best{16} & 2 \\
38 (Unc.) & Right moka pot onto stove & 0 & 0 & 32 & \best{42} & 0 \\
39 (Unc.) & Turn off stove & 4 & 14 & 40 & \best{48} & 26 \\
42 (Cov.) & Pan under shelf & 0 & \best{84} & 52 & 50 & 82 \\
69 (Unc.) & Pudding left of plate & 58 & 80 & 86 & \best{94} & 70 \\
70 (Unc.) & Pudding right of plate & 76 & 62 & 84 & \best{96} & 58 \\
71 (Unc.) & Red mug onto plate & 56 & 70 & \best{92} & 80 & 66 \\
72 (Unc.) & White mug onto plate & 2 & 78 & \best{88} & 80 & 70 \\
95 (Cov.) & Bowl from ramekin & 60 & 54 & 46 & 38 & \best{66} \\
98 (Cov.) & Bowl beside plate & 20 & 18 & 50 & \best{58} & 20 \\
102 (Cov.) & Salad dressing into basket & 26 & \best{98} & 76 & 80 & 32 \\
106 (Cov.) & Butter into basket & 36 & \best{92} & 84 & 62 & 76 \\
112 (Cov.) & Wine onto cabinet & 40 & 16 & 62 & \best{76} & 10 \\
116 (Cov.) & Cream cheese into bowl & 24 & 28 & \best{50} & \best{50} & 12 \\
126 (Unc.) & Mug and pudding placement & 0 & 24 & \best{40} & 16 & 22 \\
128 (Unc.) & Both moka pots onto stove & 2 & 0 & 2 & \best{6} & 0 \\
\midrule
\sumrow
All 18 & Macro mean (900) & 23.56 & 48.22 & \best{57.44} & \second{56.56} & 39.89 \\
\bottomrule
\end{tabularx}
\end{table}

Halving the executed L3 correction (half scale) raises 18-task success from
48.22\% to 57.44\% (+9.22 [+6.00, +12.44] points over the original L3 of the same batch),
lowering harm from 9.22\% to 4.89\% of pairs (83 to 44 of 900)
while raising rescue from 33.89\% to 38.78\% (305 to 349). A
stronger attenuation (quarter scale) performs comparably at 56.56\%
(+8.33 [+4.89, +11.67]; harm 5.33\%, 48 pairs; rescue
38.33\%, 345 pairs). Most gains concentrate on uncovered scenes,
where half scale rises from 41.00\% to 58.00\% (+17.00 [+12.25, +21.25])
and quarter scale lowers harm from 8.00\% to 2.25\% of pairs (32 to 9 of 400); on
covered scenes, gains are small and cross zero (+3.00 and +1.60), with Tasks~42
(from 84\% to 52\%) and 102 (from 98\% to 76\%) needing full-scale edits whereas
Task~112 recovers from 16\% to 76\% at quarter scale. Successful episodes
lengthen with attenuation (129.52\,$\pm$\,74.91 steps for the original L3,
$n=434$; 157.98\,$\pm$\,80.02 at half scale, $n=517$;
178.31\,$\pm$\,96.55 at quarter scale, $n=509$). Conversely, zeroing the
gripper channel lowers success to 39.89\% ($-$8.33 [$-$11.22, $-$5.33]; harm
11.44\%, 103 pairs), especially on covered scenes ($-$13.40
[$-$17.41, $-$9.20]; Task~102 from 98\% to 32\%). The learned gripper correction
therefore contributes to success on these tasks, and much of the L3 harm is
tied to correction amplitude.

\begin{table}[t]
\centering\footnotesize
\caption{\textbf{Editor prompt-swap control across six held-out tasks}
(300 paired first-pass episodes, 17,482 valid editor decisions). The original
L3 of the same batch is 55.67\% (167/300; base of that batch 42.67\%, A100 base 46.00\%); difference
+0.33 [$-$4.33, +5.33] percentage points.}
\label{tab:libero-l3-prompt-swap}
\setlength{\tabcolsep}{2.2pt}
\begin{tabularx}{\linewidth}{@{}l>{\raggedright\arraybackslash}Xrrrr@{}}
\toprule
ID (Grp.) & Task & Base & L3 & Swap & Rel.\ shift ($\times 10^{-5}$) \\
\midrule
69 (Unc.) & Pudding left of plate & 58 & 80 & 72 & 1.68 \\
70 (Unc.) & Pudding right of plate & 76 & 62 & 58 & 2.79 \\
71 (Unc.) & Red mug onto plate & 56 & 70 & 82 & 6.29 \\
72 (Unc.) & White mug onto plate & 2 & 78 & 80 & 5.48 \\
112 (Cov.) & Wine onto cabinet & 40 & 16 & 22 & 12.29 \\
116 (Cov.) & Cream cheese into bowl & 24 & 28 & 22 & 13.58 \\
\midrule
\sumrow
All 6 & Macro / total (300) & 42.67 & 55.67 & 56.00 & 8.29 \\
\bottomrule
\end{tabularx}
\end{table}

Replacing only the editor prompt (Table~\ref{tab:libero-l3-prompt-swap})
changes executed edits by only $8.29\times 10^{-5}$ in mean normalized
Frobenius norm across all 17,482 valid decisions (zero near-zero exclusions),
yielding 56.00\% (168/300; successful steps 113.38\,$\pm$\,59.41) vs.\ 55.67\%
(167/300; 114.27\,$\pm$\,55.65) for the original L3 of the same batch (+0.33 [$-$4.33, +5.33]
points; 29 rescues vs.\ 28 harms of 300 pairs, 9.7\% vs.\ 9.3\%). Given the small
magnitude shift alongside Task~112/116 ($1.23$--$1.36\times 10^{-4}$) exceeding
Task~69/70 ($1.68$--$2.79\times 10^{-5}$) and unrecorded per-episode GPU assignment
between the intervention and control runs, this control indicates
weak dependence on editor text rather than literal prompt invariance;
moreover, weighting all 300 episodes equally or averaging
task means equally yields the same mean relative shift of
$6.69\times 10^{-5}$ (vs.\ $8.29\times 10^{-5}$ pooled over decisions), so the
result does not depend on weighting by episode length.

\subsection{Task-gradient structure at a fixed checkpoint}
\label{app:libero-gradients}

Action-space regularity and shared-parameter learning address different
questions. Similar correction patterns can occur across tasks even when
those tasks request different parameter changes. We therefore examine the
local relation between task gradients in the shared actor and visual encoder.
This diagnostic uses the 465,112-update checkpoint, distinct from the
481,250-update policy evaluation, and takes no optimizer steps.

\paragraph{Sampling and gradient comparison.}
For each of the 112 source tasks, we sample two replay panels, each with
32 expert and 32 online transitions. We repeat this sampling four times.
For each parameter group, $g_m$ denotes the gradient of its task-local
training objective, averaged over the panel. We compare tasks using
\begin{equation}
 c_{mn}=\frac{g_m^\top g_n}{\lVert g_m\rVert\lVert g_n\rVert}.
\end{equation}
Positive and negative values indicate locally aligned and opposed gradients;
neither directly measures a change in success rate. Zero-gradient cosines
are undefined and excluded. The analysis covers all 6,216 unordered task
pairs. Editor-held-out tasks are excluded because they have no trained critic in
this checkpoint. Training and evaluation network modes provide a sensitivity
check on augmentation and dropout; they are not different policy evaluations.

\paragraph{Aggregate directions.}
In training mode, the actor's mean cross-task cosine is $-0.00019$, with
49.5\% of pairs negative. For the encoder, the corresponding values are
$0.00197$ and 48.8\%. A near-zero mean and roughly equal positive and
negative fractions do not establish strong destructive interference.
To interpret the aggregate magnitude, we compare the measured net gradient
with an orthogonal reference that preserves each task's gradient norm:
\begin{equation}
\label{eq:gradient-orthogonal-reference}
 R_{\perp}=\frac{\lVert\sum_m g_m\rVert}
                    {\sqrt{\sum_m\lVert g_m\rVert^2}}.
\end{equation}
Orthogonal task gradients give $R_{\perp}=1$. Ratios below or above one
indicate a negative or positive sum of cross-task inner products,
respectively. Table~\ref{tab:libero-gradient-norms} shows ratios close to
one for L3 and extends the same 112-task probe to the 481.25k-update L1 and
L2 checkpoints, decomposing the actor gradient into its total objective, raw
$Q$ term and soft-penalty term. Across all three recipes, total actor
$R_\perp$ lies between 0.98 and 1.07 in both network modes, with mean
cross-task cosines near zero ($-0.0002$ to $+0.0042$) and 48.4--49.5\%
negative task pairs. Separating the L1 and L2 actor terms shows positive
cross-task alignment, again over both modes, in the $Q$ term ($R_\perp = 1.25$--$1.29$,
46.0--47.6\% negative) and the soft-penalty term ($R_\perp = 2.35$--$2.57$,
31.8--39.8\% negative), while their within-task opposing pull leaves the
combined actor gradient near the orthogonal reference. Encoder $R_\perp$
stays above one in all three recipes (1.05--1.50). Unlike the raw
approximately 0.10--0.11 ratio obtained by dividing by the sum of individual
norms (which also occurs for orthogonal gradients), $R_\perp$ shows little
net destructive cancellation across the 112 tasks under any recipe.

\begin{table}[t]
\centering\footnotesize
\caption{\textbf{112-task gradient probe across all three recipes.} Net gradient norm relative to the
magnitude-matched orthogonal reference $R_\perp$ (Eq.~\ref{eq:gradient-orthogonal-reference}; $R_\perp=1$ for
orthogonal gradients), mean cross-task cosine and negative-pair share (\%) across all 6,216 unordered task pairs
(four resamples $\times$ two panels, equal expert/online mixtures). L3 cosines and negative shares use training
mode (Appendix~\ref{app:libero-gradients}); L1 and L2 use evaluation mode (training mode is within 0.003 in
mean cosine and 0.9 points in negative share).}
\label{tab:libero-gradient-norms}
\setlength{\tabcolsep}{2.4pt}
\begin{tabularx}{\linewidth}{@{}l>{\raggedright\arraybackslash}Xrrrr@{}}
\toprule
 & & \multicolumn{2}{c}{$R_\perp$} & Mean & Neg. \\
\cmidrule(lr){3-4}
Recipe & Group / term & Train & Eval & cos. & (\%) \\
\midrule
L3 (465k) & Actor (total) & 0.9828 & 1.0300 & $-$0.0002 & 49.5 \\
 & Encoder & 1.0558 & 1.1088 & +0.0020 & 48.8 \\
\midrule
L1 (481k) & Actor (total) & 1.0579 & 1.0450 & +0.0036 & 49.0 \\
 & Actor ($Q$ term) & 1.2497 & 1.2648 & +0.0105 & 47.2 \\
 & Actor (soft penalty) & 2.3733 & 2.3488 & +0.0478 & 39.8 \\
 & Encoder & 1.5040 & 1.3370 & +0.0065 & 47.6 \\
\midrule
L2 (481k) & Actor (total) & 1.0140 & 1.0714 & +0.0042 & 48.4 \\
 & Actor ($Q$ term) & 1.2678 & 1.2930 & +0.0105 & 46.0 \\
 & Actor (soft penalty) & 2.5725 & 2.5427 & +0.0585 & 32.1 \\
 & Encoder & 1.0529 & 1.0477 & +0.0013 & 49.6 \\
\bottomrule
\end{tabularx}
\end{table}

\paragraph{Repeatable task-pair relationships.}
We next ask whether the same task pairs retain the sign of their cosine
when the replay samples change. For each repeat, we compare task $m$ in
one panel with task $n$ in the other and average the two panel assignments.
A pair is persistent if all four cosines are defined, nonzero, and share
the same sign. For the actor in training mode, 745 pairs are persistently
negative and 745 are persistently positive.
Independently permuting task identities across repeats preserves each
repeat's gradient matrix but breaks task-pair correspondence. Across 512
such permutations, the persistent-negative count averages 382.1 and reaches
at most 433; the positive count averages 394.3 and reaches at most 449.
Thus, task identity carries information about local gradient relationships
beyond the overall fraction of negative cosines.

These repeated associations describe one checkpoint and its replay
distribution. Resampling is not replication across training seeds, and the
permutation comparison is descriptive rather than an inferential significance
test. The probe does not reproduce optimizer updates or the evolving
critic--actor training sequence. It therefore does not establish that these
relationships cause the observed success gains or regressions, or that a
particular architecture or task-supervision objective resolves them.

\subsection{Single-task adaptation from the 112-task editors}
\label{app:libero-adaptation}

\begin{table*}[t]
\centering\small
\caption{\textbf{Complete single-task adaptation metrics for the bounded recipe} across checkpoints, budgets, diagnostics, and source-task success on 16 source tasks.
Success rates (\%) and steps-to-success average across 3 training seeds.
90\% and 95\% crossings record first observed actor updates across
$\{0,5\text{k},10\text{k},25\text{k}\}$ (without interpolation; NA indicates
Task~112 pretrained seed~0 peaked at 94.00\% and did not reach 95\%).}
\label{tab:libero-adaptation-full}
\setlength{\tabcolsep}{3.5pt}
\begin{tabularx}{\textwidth}{@{}>{\raggedright\arraybackslash}Xrrrrrrrrrr@{}}
\toprule
Task / Init & 0k & 5k & 10k & 25k & \shortstack{Steps@\\10k/25k} & \shortstack{AUC\\0--25k} & \shortstack{First $\ge$90\%\\(k upd)} & \shortstack{First $\ge$95\%\\(k upd)} & \shortstack{Mean critic\\loss ($\times 10^{-4}$)} & \shortstack{Source SR\\0k $\to$ 25k} \\
\midrule
\hlrow
Task 39 / Pretrained & \best{54.00} & 88.67 & \best{97.33} & \best{99.67} & 99.4 / 89.3 & \best{91.97} & 6.7 (3/3) & \best{6.7 (3/3)} & \best{1.56} & 85.63 $\to$ 3.96 \\
Task 39 / Fresh & 6.00 & \best{94.00} & \best{97.33} & 98.33 & 95.4 / 87.3 & 87.83 & \best{5.0 (3/3)} & 11.7 (3/3) & 3.33 & 24.38 $\to$ 1.67 \\
\addlinespace[2pt]
\hlrow
Task 98 / Pretrained & \best{32.67} & \best{86.67} & \best{93.33} & \best{98.33} & 82.9 / 79.8 & \best{87.43} & \best{8.3 (3/3)} & \best{18.3 (3/3)} & \best{2.48} & 86.67 $\to$ 5.83 \\
Task 98 / Fresh & 25.33 & 80.00 & 88.67 & 95.33 & 91.8 / 83.4 & 82.60 & 13.3 (3/3) & 25.0 (3/3) & 5.13 & 25.21 $\to$ 5.21 \\
\addlinespace[2pt]
\hlrow
Task 112 / Pretrained & 24.67 & \best{91.33} & \best{96.00} & 91.00 & 62.6 / 58.9 & \best{86.43} & \best{6.7 (3/3)} & NA (2/3) & \best{3.07} & 87.29 $\to$ 5.00 \\
Task 112 / Fresh & \best{56.00} & 82.67 & 82.67 & \best{95.00} & 74.5 / 68.2 & 83.70 & 11.7 (3/3) & \best{20.0 (3/3)} & 5.82 & 25.63 $\to$ 12.29 \\
\bottomrule
\end{tabularx}
\end{table*}

We evaluate whether the 112-task L3 inference snapshot (481,250 actor
updates, 22,664,547 bytes carrying only the shared encoder, MLP actor, and
normalization statistics) accelerates downstream single-task RL relative to
fresh (actor, encoder) weights. The three adaptation targets were
predeclared from the 112-task evaluation to cover positive (Task~39,
uncovered scene \texttt{KITCHEN\_SCENE8}, from 5.00\% for the base to
37.00\% for the editor), neutral (Task~98, covered, from 21.00\% to 19.00\%), and negative
(Task~112, covered, from 55.00\% to 24.00\%) zero-update transfer. Each
condition trains three adaptation seeds for 25k online actor updates with
160k critic/encoder warm-up samples before actor update 0, a freshly
initialized 10-head task critic, shared source normalization, L3 action
bounds, and zero source-task replay (10,296 evaluation episodes across
18 runs). Evaluations use 50 native initial states at 0k, 5k, and 10k and two passes (100 episodes) at
25k, plus 10-state probes per task on 16 source tasks at 0k and 25k
(Figure~\ref{fig:sw-adaptation-curves}c,f for the two targets shared
with the soft-penalty recipes; complete breakdown for all three targets in
Table~\ref{tab:libero-adaptation-full}). The cross-recipe comparison in
Section~\ref{sec:experiments:adaptation} averages the bounded recipe over the
two targets it shares with the soft-penalty runs (Tasks~98 and 112):
94.67\% vs.\ 85.67\% at 10k and 94.67\% vs.\ 95.17\% at 25k for pretrained
vs.\ fresh initialization, with 5.42\% source SR after pretrained adaptation.

Across all nine task--seed runs, pretrained initialization requires fewer
physical control steps to reach exploratory thresholds: 31.7k vs.\ 42.6k
steps for $\ge 80\%$ (6.11k vs.\ 8.33k updates, all 9/9 runs), 31.7k vs.\
49.8k for $\ge 85\%$ (6.11k vs.\ 10.00k updates, 9/9), and 36.3k vs.\ 49.8k
for $\ge 90\%$ (7.22k vs.\ 10.00k updates, 9/9). On Task~112 at 10k
updates, pretrained wins 24 vs.\ 4 of 150 matched evaluation states (122 ties),
consistent with its shorter successful trajectories (62.6 vs.\ 74.5 steps at 10k;
58.9 vs.\ 68.2 at 25k) and 47.3\% lower online critic loss despite starting
at negative zero-update transfer. Because single-task updates omit source
replay, both initializations lose source-task success between 0k and 25k.

\paragraph{Soft-penalty adaptation on the common targets (seed 0).}
Seed 0 of the L1 and L2 runs in Section~\ref{sec:experiments:adaptation}
covers Tasks 98 and 112 with pretrained and fresh initialization (eight
runs, 4,608 evaluation episodes). Target
evaluations use 50 initial states at 0k, 5k and 10k online actor updates and
two passes over those states at 25k. Source retention uses ten states on
each of 16 source tasks at 0k and 25k. Fresh L1 performs 5,000 proposal-fitting
updates before the online axis; the other three recipe--initialization
conditions do not. Both initializations reach at least 90\% by the first
5k checkpoint in each of these eight runs. That evaluation spacing does not
resolve earlier threshold crossings; Appendix~\ref{app:sw-adaptation} reports
seeds 1 and 2 of each condition with checkpoints at 1k, 2k and 3k.
AUC is the trapezoid area under target SR at 0k, 5k, 10k and 25k online actor
updates divided by 25k. For pretrained vs.\ fresh initialization in seed 0 it is
94.30 vs.\ 92.20 (Task~98) and 90.40 vs.\ 86.10 (Task~112) for L1, and
91.60 vs.\ 90.20 and 85.10 vs.\ 89.40 for L2. The three-seed L3 means on
the same targets are 87.43 vs.\ 82.60 and 86.43 vs.\ 83.70; per-seed fresh
L3 AUC on Task~112 spans 75.00--89.60, so single-seed differences of a few
points are not resolved. Mean online critic loss is lower under pretraining in
all six recipe--target pairs: by 36.1\% and 58.0\% for L1 and 54.5\% and
42.4\% for L2 in seed 0, and by 51.7\% and 47.3\% for L3;
Figure~\ref{fig:libero-adaptation} averages the two targets and the three
seeds of each recipe.
Table~\ref{tab:soft-adaptation-complete} reports the eight seed-0 runs,
including lower pretrained outcomes on wine placement, and
Table~\ref{tab:sw-adaptation-seeds} gives the three-seed means for L1 and L2.

\begin{table*}[t]
\centering\small
\caption{\textbf{Complete soft-penalty adaptation results (success percentages).}
Task 98 places the bowl beside a plate onto that plate; Task 112 places wine
onto the cabinet. P/F denote pretrained/fresh initialization. Source SR is
the equal-task mean on 16 source tasks, not a fraction of the initial SR.
Each row is one seed-0 training run.}
\label{tab:soft-adaptation-complete}
\setlength{\tabcolsep}{6pt}
\begin{tabular}{@{}lllrrrrrr@{}}
\toprule
Task & Recipe & Init. & 0k & 5k & 10k & 25k & Source 0k & Source 25k \\
\midrule
98 & L1 & P & 56 & 98 & 100 & 97 & 85.63 & 28.13 \\
98 & L1 & F & 28 & 100 & 100 & 98 & 7.50 & 8.75 \\
112 & L1 & P & 52 & 100 & 88 & 100 & 89.38 & 11.88 \\
112 & L1 & F & 52 & 92 & 88 & 91 & 0.00 & 11.88 \\
\midrule
98 & L2 & P & 48 & 92 & 96 & 100 & 88.13 & 23.75 \\
98 & L2 & F & 28 & 98 & 96 & 98 & 26.25 & 17.50 \\
112 & L2 & P & 56 & 96 & 84 & 89 & 91.88 & 33.13 \\
112 & L2 & F & 64 & 98 & 88 & 94 & 25.63 & 21.88 \\
\bottomrule
\end{tabular}
\end{table*}

\paragraph{Frozen-encoder mechanism arms (seed 0).}
To test whether representation transfer explains lower critic loss and
whether representation drift explains source-task forgetting, we run two
additional seed-0 mechanism arms on Tasks~98 and 112 using the same
112-task L3 checkpoint at 481,250 updates, 160k critic warm-up, fresh 10-head task
critics, L3 bounds, and 25k actor updates without source replay:
\emph{Pretrained actor + frozen encoder} loads both modules and freezes
the complete visual encoder (dual-camera backbones and
language projection; parameters verified unchanged before warm-up and at
5k, 10k, and 25k), whereas \emph{Fresh actor + frozen encoder} loads only
that frozen encoder with a freshly initialized actor. Table~\ref{tab:libero-adaptation-frozen}
compares these seed-0 runs with seed~0 of the trainable
reference arms (without treating single-seed outcomes as multi-seed repeats).
Freezing the shared encoder with a fresh actor lowers mean online critic loss
over 25k updates by 52.0\% on Task~98 (from 0.000471 to 0.000226) and 47.1\%
on Task~112 (from 0.000503 to 0.000266), nearly matching full pretraining
while avoiding Task~112's 0k actor drop (58.00\% vs.\ 18.00\%); yet even
with an unchanged encoder, source SR falls from 92.50\% to 7.50\% (Task~98)
and 6.88\% (Task~112), so actor updates alone suffice to overwrite
source-task behavior.

\begin{table*}[t]
\centering\small
\caption{\textbf{Matched seed-0 encoder-freezing adaptation ablations on Tasks~98 and 112} vs.\ seed~0 of the trainable reference arms (25k online actor updates, 50 states at 0k/5k/10k, two passes at 25k). Mean critic loss averages all 997 logged online records across 25--24,975 updates.}
\label{tab:libero-adaptation-frozen}
\setlength{\tabcolsep}{4pt}
\begin{tabularx}{\textwidth}{@{}>{\raggedright\arraybackslash}lXrrrrrrr@{}}
\toprule
Task & Arm (seed 0) & 0k & 5k & 10k & 25k & \shortstack{AUC\\0--25k} & \shortstack{Mean critic\\loss ($\times 10^{-4}$)} & \shortstack{Source SR\\0k $\to$ 25k} \\
\midrule
\multirow{4}{*}{Task 98}
 & Fresh (trainable enc.) & 34.00 & 90.00 & 88.00 & 95.00 & 85.10 & 4.71 & 25.00 $\to$ 4.38 \\
 & Fresh actor + frozen enc. & 26.00 & 78.00 & 86.00 & 97.00 & 81.70 & 2.26 & 26.25 $\to$ 15.00 \\
 & Pretrained actor + frozen enc. & 38.00 & 86.00 & \best{98.00} & 93.00 & 88.10 & \best{1.78} & 92.50 $\to$ 7.50 \\
 & Pretrained (trainable enc.) & \best{40.00} & \best{96.00} & 94.00 & \best{99.00} & \best{90.50} & 2.02 & 88.12 $\to$ 3.12 \\
\midrule
\multirow{4}{*}{Task 112}
 & Fresh (trainable enc.) & 52.00 & 92.00 & \best{96.00} & 92.00 & \best{89.60} & 5.03 & 25.62 $\to$ 9.38 \\
 & Fresh actor + frozen enc. & \best{58.00} & 86.00 & 92.00 & 92.00 & 87.40 & \best{2.66} & 25.00 $\to$ 5.62 \\
 & Pretrained actor + frozen enc. & 18.00 & 90.00 & 92.00 & \best{94.00} & 84.80 & 2.67 & 92.50 $\to$ 6.88 \\
 & Pretrained (trainable enc.) & 26.00 & \best{94.00} & 94.00 & 90.00 & 86.00 & 2.87 & 81.88 $\to$ 3.12 \\
\bottomrule
\end{tabularx}
\end{table*}

\section{LIBERO Recipe Study on 16 Source Tasks}
\label{app:recipe-study}

This appendix supports Section~\ref{sec:experiments:design}: the six recipes
trained on 16 source tasks, their editor-held-out transfer, learner
throughput, and gradient diagnostics.

\subsection{Fixed-budget six-recipe study}
\label{app:six-recipes}

\paragraph{Cohort and uncertainty.}
The 16 LIBERO source-task IDs are 7, 21, 34, 64, 65, 66, 74, 75,
90, 94, 108, 109, 111, 115, 120, and 123.
Each of the six 100k-online-actor-update checkpoints and the frozen base
is evaluated on 50 native initial states in two passes per task.
Initial-state identities match across all seven policies; the two passes
are repeated evaluations, not additional distinct states.
We use 10,000 bootstrap draws with seed 0, independently resampling the
50 states within each fixed task while grouping both passes and pairing
policies. Percentile intervals therefore describe evaluation-state
uncertainty on this cohort, not variation across trained seeds or task sets.
Pairwise contrasts are exploratory and not multiplicity-adjusted.

\begin{table}[tb]
\centering
\caption{\textbf{Six-recipe evaluation and training accounting.} Success and its
95\% initial-state-cluster interval are percentages. All editors have
100k online actor and 400k online critic updates. Preparation lists additional
actor updates; actions count physical training interactions.}
\label{tab:six-recipe-detail}
\small\setlength{\tabcolsep}{3pt}
\begin{tabular}{@{}lrrr@{}}\toprule
Recipe & SR [95\% CI]\,$\uparrow$ & Prep. & Actions \\\midrule
Base & 22.06 [19.94, 24.25] & --- & --- \\
R1 & 61.75 [59.44, 64.06] & 5k & 1,575,672 \\
R2 & 77.00 [75.00, 78.94] & 0 & 679,362 \\
R3 & 56.88 [54.56, 59.19] & 0 & 608,254 \\
L1 & 93.00 [91.44, 94.44] & 5k & 610,293 \\
L2 & 91.63 [90.13, 93.06] & 0 & 614,413 \\
L3 & 85.63 [83.63, 87.50] & 0 & 1,326,131 \\\bottomrule
\end{tabular}
\end{table}

\paragraph{Budget interpretation.}
The shared endpoint fixes online actor updates, not total training cost
(Table~\ref{tab:six-recipe-detail}). R1 and L1 each execute 5,000 proposal-fitting
updates before online learning. Physical interaction counts also differ.
Training episode logs lack a reliable actor-update alignment across
restarts, so we do not reconstruct update-aligned success curves or infer
end-to-end training time from timestamp envelopes. These results compare
the trained recipes at the stated endpoint; they do not establish
time-to-threshold or asymptotic performance rankings.

\paragraph{Feature probes.}
A common panel contains 512 expert observations from 16 tasks, 16 expert
episodes per task, and two observations per episode. Four episode-grouped
folds evaluate linear prediction of expert-minus-base action corrections.
The controls include task identity, proprioception, and base proposals;
task identity is a diagnostic input, not an added actor input.
Mean squared error is 1.6343 for zero correction and 1.0378 for controls
alone. Adding the base's prefix features, RL-token features, or L1 encoder
features yields 1.0400, 1.2911, and 1.3107, respectively.
Thus these probes do not explain L1's closed-loop advantage, nor establish
that a frozen representation lacks task information. Linear readout on
expert states need not capture the nonlinear information used during online
control. The folds hold out probe-fitting episodes, not representation-training
exposure; the representations also use their native camera inputs.

\paragraph{Task-gradient diagnostics.}
For 16 tasks there are 120 unordered task pairs. At the measured checkpoints,
the numbers of negative actor-gradient inner products for the native total
objectives are 1, 48, 54, 59, 54, and 50 for R1--R3 and L1--L3.
For the raw-Q components they are 51, 49, 54, 55, 52, and 50.
The shared anchoring term can therefore align gradients without implying
better success. These diagnostics hold representations fixed and measure
gradients before clipping and AdamW, not realized parameter updates.
Neither conflict counts nor the offline critic response to interpolated
expert corrections establish the cause of the performance differences.

\paragraph{Panel magnitudes, wrist-camera ablation, and MoE routing.}
Table~\ref{tab:six-recipe-panel} reports scaled correction magnitudes on
the 512 expert-observation panel alongside paired rescue and harm outcomes
from the 1,600-episode evaluation (353 base successes, 1,247 base
failures). Table~\ref{tab:wrist-freeze-tasks} gives the per-task outcomes
of freezing the wrist camera to frame~0 for L2 on 25 states per task
(100 paired episodes total; $-52$ [$-61, -43$] points) alongside R2 on
the same 25 states (first pass). Table~\ref{tab:six-actor-moe} summarizes the
16-task MoE and cluster-supervision evaluations over 1,600 paired episodes
per policy.

\begin{table}[tb]
\centering\footnotesize
\caption{\textbf{Panel correction statistics} (512 expert observations) and
paired rescue/harm counts (16 tasks; percentages are shares of the 1,600
pairs) for all six recipes.
Near-bound reports the fraction of decisions with at least one coordinate
exceeding $0.9\Lambda_j$.}
\label{tab:six-recipe-panel}
\setlength{\tabcolsep}{2pt}
\begin{tabularx}{\linewidth}{@{}Xrrrrr@{}}
\toprule
Recipe & \shortstack{Arm\\$|A-B|/\Lambda$} & \shortstack{Grip.\\$|A-B|/\Lambda$} & \shortstack{Near\\bound} & \shortstack{Rescue\,$\uparrow$\\/1,600} & \shortstack{Harm\,$\downarrow$\\/1,600} \\
\midrule
R1 & 0.206 & 0.002 & 14.8\% & 679 (42.4\%) & 44 (2.8\%) \\
R2 & 0.292 & 0.004 & 30.3\% & 896 (56.0\%) & 17 (1.1\%) \\
R3 & 0.657 & 0.451 & 99.8\% & 635 (39.7\%) & 78 (4.9\%) \\
\midrule
L1 & 0.293 & 0.001 & 34.6\% & \best{1,151 (71.9\%)} & \second{16 (1.0\%)} \\
\hlrow
L2 & 0.268 & 0.004 & 26.8\% & \second{1,122 (70.1\%)} & \best{9 (0.6\%)} \\
L3 & 0.631 & 0.213 & 98.8\% & 1,041 (65.1\%) & 24 (1.5\%) \\
\bottomrule
\end{tabularx}
\end{table}

\begin{table}[tb]
\centering\footnotesize
\caption{\textbf{Per-task wrist-camera freeze intervention on L2} (4 tasks $\times$
25 states, 100 episodes total). R2 values show matched first-pass states.}
\label{tab:wrist-freeze-tasks}
\setlength{\tabcolsep}{2pt}
\begin{tabularx}{\linewidth}{@{}>{\raggedright\arraybackslash}Xrrr@{}}
\toprule
Task (ID) & \shortstack{L2\\control} & \shortstack{L2 frozen\\wrist} & \shortstack{R2\\(matched)} \\
\midrule
Stack bowls in tray (64) & 24/25 & 1/25 & 13/25 \\
Book in right caddy (75) & 22/25 & 1/25 & 14/25 \\
Bowl from drawer to plate (94) & 23/25 & 17/25 & 16/25 \\
Two items in basket (120) & 21/25 & 19/25 & 0/25 \\
\midrule
\sumrow
Total (100 episodes) & \best{90/100} & 38/100 & 43/100 \\
\bottomrule
\end{tabularx}
\end{table}

\begin{table}[tb]
\centering\footnotesize
\caption{\textbf{Expert routing and cluster supervision on 16 LIBERO source tasks}
(100k online updates, 1,600 paired episodes). CS denotes cluster supervision.}
\label{tab:six-actor-moe}
\setlength{\tabcolsep}{2pt}
\begin{tabularx}{\linewidth}{@{}>{\raggedright\arraybackslash}Xrr@{}}
\toprule
Actor configuration & SR [95\% CI]\,$\uparrow$ & \shortstack{Paired $\Delta\,\uparrow$ vs.\\MLP [95\% CI]} \\
\midrule
R2-MLP & 77.00 [75.00, 78.94] & --- \\
R2-MoE & 75.75 [73.69, 77.81] & $-$1.25 [$-$3.94, +1.44] \\
R2-MoE+CS & 68.31 [66.31, 70.38] & $-$8.69 [$-$11.31, $-$6.06] \\
\midrule
L2-MLP & \second{91.63 [90.13, 93.06]} & --- \\
L2-MoE & 90.25 [88.56, 91.88] & $-$1.38 [$-$3.50, +0.69] \\
\hlrow
L2-MoE+CS & \best{93.75 [92.38, 95.06]} & \best{+2.13 [+0.19, +4.13]} \\
\bottomrule
\end{tabularx}
\end{table}

\paragraph{Main-camera L2 training ablation.}
To test whether L2's source advantage over R2 comes purely from wrist camera
access rather than online visual-encoder training, we train an independent
seed-0 L2 editor for 100k online actor updates (522,112 online environment
steps; 16,472.7\,s of online training on two A100-SXM4-80GB
GPUs; 15.0\,MB inference snapshot) using only the main camera
input (81 vs.\ 162 visual tokens). Removing the wrist stream reduces both
input information and encoder capacity, so this tests the full
main-camera-only training architecture rather than a fixed-parameter
pure-information intervention. On five-state development probes across the
16 source tasks, SR progresses across 22.50\% (0k), 52.50\% (20k), 72.50\%
(50k), and 77.50\% (100k; base 22.50--23.75\%). Under full 50-state
two-pass evaluation (6,800 paired episodes across all 34 tasks;
Table~\ref{tab:no-wrist-summary}), main-camera L2 reaches 79.94\%
(1,279/1,600; 95\% CI [78.25, 81.63]) on source tasks (+58.25 [+55.50,
+60.94] over paired 21.69\% base; 963 rescues, 31 harms) and 40.56\%
(730/1,800; [38.33, 42.72]) on editor-held-out tasks (+16.89 [+14.00,
+19.72] over paired 23.67\% base; 486 rescues, 182 harms; 47.90\% and
31.38\% on the ten covered and eight uncovered tasks of the 112-task scene
split, where Table~\ref{tab:sw-completions-cameras} uses the 7/11 split of
the 16 source tasks). Against exact state-matched
dual-view L2 on the 16 source tasks (91.63\%, 1,466/1,600), main-camera
L2 changes by $-$11.69 [$-$13.94, $-$9.63] points (84 rescues, 271 harms);
Task~64 (from 94\% to 0\%) and Task~75 (from 83\% to 29\%) account for 148 of the
187 net lost successes (79.1\%; Table~\ref{tab:no-wrist-tasks}). Conversely,
with identical single-view main input, main-camera L2 still solves
Task~120 at 84.0\% (vs.\ 2.0\% for R2 and 81.0\% for dual-view L2) and
Task~94 at 92.0\% (vs.\ 67.0\% for R2), separating online visual-encoder
representation from wrist-camera visibility. The dual-view run has no
per-episode held-out records in this evaluation batch, so held-out
comparisons pair against the base only.

\begin{table}[tb]
\centering\footnotesize
\caption{\textbf{Main-camera L2 training ablation} (100k updates,
seed~0; 50 states$\times$2 passes, 2,000 paired intra-task state-cluster
bootstrap draws). The source comparison pairs against 16-task dual-view L2 and
the base; no per-episode dual-view held-out records are available. Covered
and uncovered follow the 112-task scene split (10/8 tasks).}
\label{tab:no-wrist-summary}
\setlength{\tabcolsep}{2.5pt}
\begin{tabularx}{\linewidth}{@{}>{\raggedright\arraybackslash}Xrrr@{}}
\toprule
Cohort / Metric & Base & Main-camera L2 & Dual L2 \\
\midrule
Source 16 SR [95\% CI] & 21.69 & 79.94 [78.25, 81.63] & \best{91.63} \\
Source $\Delta$ vs.\ Base / Dual & --- & +58.25 / $-$11.69 & --- \\
Held-out 18 SR [95\% CI] & 23.67 & 40.56 [38.33, 42.72] & --- \\
Held-out Covered 10 / Uncov.\ 8 & 23.30 / 24.13 & 47.90 / 31.38 & --- \\
Held-out rescue / harm (shares of pairs) & --- & 27.0\% / 10.1\% & --- \\
\bottomrule
\end{tabularx}
\end{table}

\begin{table}[tb]
\centering\footnotesize
\caption{\textbf{Per-task 16-source-task success rates (\%) across R2,
main-camera L2, and dual-view L2} (100 episodes/task). The four tasks of the
wrist-freeze intervention (Table~\ref{tab:wrist-freeze-tasks}) are highlighted.}
\label{tab:no-wrist-tasks}
\setlength{\tabcolsep}{2.2pt}
\begin{tabularx}{\linewidth}{@{}l>{\raggedright\arraybackslash}Xrrrr@{}}
\toprule
ID & Task & Base & R2 & Main L2 & Dual L2 \\
\midrule
7 & Top drawer open & 10 & 84 & \best{92} & 83 \\
21 & Stove on; pan & 1 & 71 & 82 & \best{93} \\
34 & Mug in front & 34 & 86 & 95 & \best{96} \\
\hlrow
64 & Stack bowls tray & 0 & 52 & 0 & \best{94} \\
65 & Red mug left plate & 30 & \best{100} & 88 & 98 \\
66 & Red mug right plate & 20 & 86 & 90 & \best{99} \\
74 & Book left caddy & 34 & 98 & 99 & \best{100} \\
\hlrow
75 & Book right caddy & 0 & 58 & 29 & \best{83} \\
90 & Bowl middle plate & 74 & \best{96} & 76 & \best{96} \\
\hlrow
94 & Bowl drawer plate & 9 & 67 & 92 & \best{94} \\
108 & Pudding basket & 4 & \best{100} & \best{100} & \best{100} \\
109 & Juice basket & 60 & 90 & 94 & \best{100} \\
111 & Bowl on stove & 50 & \best{100} & \best{100} & \best{100} \\
115 & Plate front stove & 20 & 89 & 93 & \best{94} \\
\hlrow
120 & Two items basket & 0 & 2 & \best{84} & 81 \\
123 & Bowl bottom drawer & 1 & 53 & \best{65} & 55 \\
\midrule
\sumrow
Avg & Source 16 mean & 21.69 & 77.00 & 79.94 & \best{91.63} \\
\bottomrule
\end{tabularx}
\end{table}

\subsection{Six-recipe editor-held-out transfer and scene cohorts}
\label{app:six-recipe-heldout}

\paragraph{Evaluation hardware and paired base re-evaluation.}
The 18 editor-held-out tasks (50 native initial states in two passes per task,
1,800 paired episodes per policy, zero editor updates) for the six
100k-update 16-source-task editors were evaluated on a single RTX 5090 GPU,
whereas source-task evaluations ran on A100 GPUs; no cross-machine
equivalence check was run. To ensure exact episode-level
pairing within this evaluation batch, the frozen base policy was
re-evaluated under the identical protocol, scoring 429/1,800 successes
(23.83\%, 1,371 failures). Across the 18 editor-held-out tasks, the two base
evaluations differ by 2.6 percentage points on average (and 10 points on
task~112). All six-recipe editor-held-out comparisons pair strictly against
this matched 23.83\% base run rather than subtracting across evaluation
batches.
Confidence intervals in Table~\ref{tab:six-recipe-heldout-summary} are paired
95\% initial-state-cluster bootstrap intervals conditional on the fixed
18-task cohort and are not multiplicity-adjusted. Across the five recipes
that improve over the base, per-task success spreads (max minus min) have a
median of 40 percentage points and reach at least 30 points on 11 of 18
tasks. On the editor-held-out set, paired initial-state-cluster differences
between L and R recipes are 0.0 [$-$2.8, 2.9] points (L1 minus R1), +4.0
[1.1, 6.9] (L2 minus R2), and $-$16.1 [$-$18.8, $-$13.4] (L3 minus R3).

\begin{table}[tb]
\centering\footnotesize
\caption{\textbf{Six-recipe editor-held-out transfer summary} (18 tasks, 1,800 paired
episodes; 100k online actor updates on 16 source tasks). Paired $\Delta$ and
95\% initial-state-cluster CIs compare against the matched frozen-base
re-evaluation (23.83\%, 429/1,800).}
\label{tab:six-recipe-heldout-summary}
\setlength{\tabcolsep}{2pt}
\begin{tabularx}{\linewidth}{@{}Xrrrr@{}}
\toprule
Recipe & \shortstack{Held-out\\SR\,$\uparrow$} & \shortstack{Paired $\Delta$ vs.\\Base [95\% CI]\,$\uparrow$} & \shortstack{Rescue\,$\uparrow$\\/1,800} & \shortstack{Harm\,$\downarrow$\\/1,800} \\
\midrule
R1 & 40.28 & +16.4 [13.4, 19.4] & 476 (26.4\%) & 180 (10.0\%) \\
R2 & \second{41.06} & \second{+17.2 [14.1, 20.2]} & \second{496 (27.6\%)} & 186 (10.3\%) \\
R3 & 39.22 & +15.4 [12.7, 18.2] & 434 (24.1\%) & 157 (8.7\%) \\
\midrule
L1 & 40.28 & +16.4 [13.6, 19.2] & 448 (24.9\%) & \second{152 (8.4\%)} \\
\hlrow
L2 & \best{45.06} & \best{+21.2 [18.6, 23.9]} & \best{503 (27.9\%)} & \best{121 (6.7\%)} \\
L3 & 23.11 & $-$0.7 [$-$3.3, 1.8] & 284 (15.8\%) & 297 (16.5\%) \\
\bottomrule
\end{tabularx}
\end{table}

\begin{table*}[tb]
\centering\small
\caption{\textbf{Scene-coverage decomposition for the six 16-source-task editors on
the 18 editor-held-out tasks.} Covered scenes contain 7 tasks (494 base
failures, 206 base successes; base SR 29.4\%); uncovered scenes contain 11 tasks
(877 base failures, 223 base successes; base SR 20.3\%).}
\label{tab:six-recipe-heldout-cohorts}
\setlength{\tabcolsep}{4pt}
\begin{tabularx}{\textwidth}{@{}>{\raggedright\arraybackslash}Xrrrr@{}}
\toprule
 & \multicolumn{2}{c}{Source-covered scenes (7 tasks; 700 ep.)} & \multicolumn{2}{c}{Uncovered scenes (11 tasks; 1,100 ep.)} \\
\cmidrule(lr){2-3} \cmidrule(l){4-5}
Recipe & Rescue\,$\uparrow$ (/700) & Harm\,$\downarrow$ (/700) & Rescue\,$\uparrow$ (/1,100) & Harm\,$\downarrow$ (/1,100) \\
\midrule
R1 & 36.7\% (257) & 12.1\% (85) & \second{19.9\% (219)} & 8.6\% (95) \\
R2 & 30.1\% (211) & 13.6\% (95) & \best{25.9\% (285)} & 8.3\% (91) \\
R3 & 38.4\% (269) & 10.9\% (76) & 15.0\% (165) & \second{7.4\% (81)} \\
\midrule
L1 & \best{43.6\% (305)} & \best{7.0\% (49)} & 13.0\% (143) & 9.4\% (103) \\
\hlrow
L2 & \second{42.6\% (298)} & \second{10.7\% (75)} & 18.6\% (205) & \best{4.2\% (46)} \\
L3 & 28.0\% (196) & 15.7\% (110) & 8.0\% (88) & 17.0\% (187) \\
\bottomrule
\end{tabularx}
\end{table*}

\paragraph{Sixteen-task scene-coverage definition and breakdown.}
For the 16-source-task split, \emph{source-covered scenes} comprise the
seven editor-held-out tasks sharing a named LIBERO-90 scene or single-scene
10-task suite with the 16 source tasks (tasks 9, 95, 98, 102, 106, 112, and
116; base 29.4\%, 206 successes and 494 failures), while \emph{uncovered
scenes} comprise the remaining 11 tasks (3, 16, 38, 39, 42, 69, 70, 71, 72,
126, and 128; base 20.3\%, 223 successes and 877 failures). Because these
observational cohorts contain different tasks and different base rates,
Table~\ref{tab:six-recipe-heldout-cohorts} reports rescue and harm as shares
of each cohort's paired episodes, with counts. Across all six recipes, rescue is higher on
source-covered scenes, whereas harm has no consistent direction across the
two cohorts. Under the 112-task split (10 covered tasks, base 21.8\% in this
re-run vs.\ 8 uncovered, base 26.4\%), rescue remains higher for all six
recipes on covered tasks (20.5--35.7\% vs.\ 9.9--24.4\%).
Table~\ref{tab:six-recipe-heldout-tasks} gives per-task success across all
18 editor-held-out tasks.

\begin{table*}[tb]
\centering\footnotesize
\caption{\textbf{Per-task editor-held-out success (\%) across the six
16-source-task recipes} (100 episodes per task, paired against the RTX 5090
Base re-run at 23.83\%). Tasks marked $\dagger$ lie in 16-source-covered
scenes; tasks marked $\star$ lie in the eight 112-task uncovered scene
cohort. The final column reproduces the 112-task L3 run (paired with the
23.33\% A100 base run) for side-by-side reference.}
\label{tab:six-recipe-heldout-tasks}
\setlength{\tabcolsep}{4pt}
\begin{tabularx}{\textwidth}{@{}>{\raggedright\arraybackslash}Xrrrrrrrr@{}}
\toprule
Task (ID) & Base & R1\,$\uparrow$ & R2\,$\uparrow$ & R3\,$\uparrow$ & L1\,$\uparrow$ & L2\,$\uparrow$ & L3\,$\uparrow$ & 112-L3\,$\uparrow$ \\
\midrule
Butter into drawer; close (3) & 7 & 32 & \best{81} & 57 & 43 & 7 & 10 & 84 \\
Bowl onto plate (9)$^\dagger$ & 3 & \best{90} & 55 & 51 & 33 & 66 & 6 & 76 \\
Stack front bowl (16) & 5 & \best{64} & 5 & 2 & 9 & 13 & 0 & 6 \\
Right moka pot onto stove (38)$^\star$ & 0 & \best{10} & 8 & 6 & 7 & 6 & 0 & 1 \\
Turn off stove (39)$^\star$ & 4 & 26 & 41 & 16 & 37 & \best{78} & 32 & 37 \\
Pan under shelf (42) & 0 & 9 & \best{10} & 2 & 0 & 0 & 0 & 82 \\
Pudding left of plate (69)$^\star$ & 60 & 40 & 26 & 10 & 24 & \best{72} & 0 & 72 \\
Pudding right of plate (70)$^\star$ & 80 & 62 & 64 & \best{90} & 72 & 86 & 18 & 56 \\
Red mug onto plate (71)$^\star$ & 60 & 76 & \best{92} & 74 & 68 & 86 & 24 & 68 \\
White mug onto plate (72)$^\star$ & 2 & 28 & \best{58} & 26 & 2 & 30 & 34 & 80 \\
Bowl from ramekin (95)$^\dagger$ & 52 & 61 & 39 & 33 & \best{67} & 40 & 24 & 58 \\
Bowl beside plate (98)$^\dagger$ & 25 & 36 & 9 & \best{45} & 38 & 29 & 1 & 19 \\
Salad dressing into basket (102)$^\dagger$ & 22 & 86 & 48 & 74 & 88 & 96 & \best{98} & 96 \\
Butter into basket (106)$^\dagger$ & 36 & 30 & 80 & \best{100} & \best{100} & \best{100} & \best{100} & 94 \\
Wine onto cabinet (112)$^\dagger$ & 45 & 52 & 57 & 48 & \best{92} & 73 & 25 & 24 \\
Cream cheese into bowl (116)$^\dagger$ & 23 & 23 & 34 & \best{48} & 44 & 25 & 38 & 31 \\
Mug and pudding placement (126)$^\star$ & 4 & 0 & \best{28} & 24 & 0 & 4 & 6 & 16 \\
Both moka pots onto stove (128)$^\star$ & 1 & 0 & \best{4} & 0 & 1 & 0 & 0 & 0 \\
\midrule
\sumrow
Mean (all 18) & 23.83 & 40.28 & \second{41.06} & 39.22 & 40.28 & \best{45.06} & 23.11 & 50.00 \\
\bottomrule
\end{tabularx}
\end{table*}

\subsection{Learner throughput and preparation}
\label{app:learner-throughput}

We time the native R2 and L2 learners restored from their 100k-online-update
checkpoints (Table~\ref{tab:learner-throughput}). Both use a global batch of 256, four tasks with 64 samples
each, ten critic heads per task, and four critic updates per actor update.
Each timed block contains 500 critic updates and 125 actor updates, preceded
by five untimed updates. Four rounds exchange the two methods between two
A100-SXM4-80GB GPUs in an ABBA assignment, so each method runs on both
devices. The reported rates use synchronized wall time; the repeated timing
blocks hold trained initialization fixed. Computation uses FP32 and
cycles through four fixed expert batches covering all 16 tasks.

The GPU-resident condition places inputs on the device before timing;
the transfer-inclusive condition moves each batch from CPU memory within
the timed interval. Both include forward and backward computation,
optimizer steps, and target updates, but exclude collection, base inference,
new feature extraction, checkpointing, and parameter publication.
R2 uses cached 2,048-dimensional RL-token features and base-proposal critic
conditioning; L2 uses its native two-view RGB inputs and updates the encoder
through the critic. These are complete learner-path comparisons, including
their different inputs and network shapes. With GPU-resident inputs the
R2 learner completes 6.223 and the L2 learner 5.960 actor updates/s.

Online pipeline throughput includes waiting for collection and therefore
need not follow the isolated learner ordering. In the final continuations,
R2 and L2 complete 4.837 and 5.014 actor updates/s, respectively. In the
shorter 95k--99,975-update windows, the rates are 4.976 and 4.902.
The continuations overlap in time, but the shorter windows do not;
GPU utilization during training was not recorded. These observations
give no consistent end-to-end throughput ordering.

Preparation is accounted for separately. R2's recorded readout and prefix
preparation takes 1,175.61\,s, including 24.15\,s of base inference;
feature-cache construction time is unrecorded. Both recipes perform 5k
critic warm-up updates without actor updates. Their initialization stages
take 327.08\,s for R2 and 310.24\,s for L2, including work beyond those
warm-up updates. Neither the isolated timing nor these partial preparation
records measures total time to target success.

\begin{table}[tb]
\centering\small
\caption{\textbf{Isolated learner throughput in actor updates/s}: medians and
ranges across four timing blocks on matched expert samples.}
\label{tab:learner-throughput}
\begin{tabularx}{\linewidth}{@{}>{\raggedright\arraybackslash}Xrr@{}}
\toprule
Input timing & R2 & L2 \\
\midrule
GPU-resident & 6.223 & 5.960 \\
Range & [6.062, 6.445] & [5.867, 5.970] \\
Including transfer & 6.009 & 5.506 \\
Range & [5.708, 6.276] & [5.488, 5.555] \\
\bottomrule
\end{tabularx}
\end{table}

\subsection{State-dependent task-gradient relationships}
\label{app:gradient-phases}

At the fixed R2 and L2 checkpoints, we partition 50 expert episodes per
source task into four disjoint subsets by episode ordinal modulo four.
The subsets contain 13, 13, 12, and 12 episodes per task. Each episode
contributes its initial observation and the decision observation at or
immediately before its temporal midpoint. We compute gradients on the
initial observations, midpoint observations, and their union separately.
R2 and L2 use matched observations from the prepared expert data. The
four subsets measure sensitivity to episode sampling at fixed weights.

For each task, we separate the raw value term, the base-proposal penalty,
and the complete actor objective with its native scaling. Gradients are
taken with respect to all actor parameters, with the representation held
fixed, before clipping or optimizer updates. The aggregate ratio
$R_{\perp}$ defined in Equation~\ref{eq:gradient-orthogonal-reference}
compares the net gradient with a magnitude-matched orthogonal reference.
We compute it within each model's parameter space and then summarize
across episode subsets (Table~\ref{tab:gradient-phases}).

The representation ordering changes with the objective and state phase.
For the complete objective at episode starts, L2 has more aggregate
cancellation than R2; at midpoints their mean ratios are nearly equal.
Within R2, placing a red mug on the left versus right plate has positive
Q-gradient cosine at episode starts in all four subsets (mean $+0.278$),
but negative cosine at midpoints (mean $-0.333$). A task pair's local
gradient relation can therefore reverse along its trajectories.

We also form four task groups by average-linkage clustering with distance
$\sqrt{2(1-C_{ij})}$, where $C$ averages the two models' Q-gradient cosine
matrices. These partitions vary across episode subsets. Mean adjusted Rand indices
over the six subset pairs are 0.061 for both phases, 0.511 for initial
observations, and 0.305 for midpoints. This sensitivity limits a fixed
task-group interpretation. The measurements describe local gradients,
not the change in another task's loss after an optimizer step or the
benefit of an expert-routing policy.

\begin{table}[tb]
\centering\small
\caption{\textbf{Aggregate actor-gradient ratio $R_{\perp}$}: mean [minimum, maximum]
across four expert-episode subsets. ``Both'' combines initial and midpoint
observations. Values above one indicate net positive cross-task inner products.}
\label{tab:gradient-phases}
\setlength{\tabcolsep}{3pt}
\begin{tabularx}{\linewidth}{@{}>{\raggedright\arraybackslash}Xrr@{}}
\toprule
Objective, phase & R2 & L2 \\
\midrule
Q, both & 1.320 [1.193, 1.472] & 1.110 [1.051, 1.216] \\
Q, initial & 1.317 [1.169, 1.538] & 1.009 [0.943, 1.107] \\
Q, midpoint & 1.068 [1.027, 1.091] & 1.084 [0.976, 1.234] \\
\midrule
Penalty, both & 1.584 [1.472, 1.764] & 1.120 [1.102, 1.135] \\
Penalty, initial & 1.573 [1.506, 1.681] & 1.268 [1.244, 1.300] \\
Penalty, midpoint & 1.069 [1.029, 1.130] & 0.979 [0.945, 1.047] \\
\midrule
Total, both & 1.205 [1.055, 1.447] & 1.129 [1.070, 1.280] \\
Total, initial & 1.038 [0.932, 1.205] & 0.859 [0.778, 1.000] \\
Total, midpoint & 1.148 [1.018, 1.239] & 1.147 [1.035, 1.323] \\
\bottomrule
\end{tabularx}
\end{table}

\subsection{RL-token editing on two source tasks}
\label{app:rlt-two-source}

A separate LIBERO study uses a fixed RL-token readout, full-action
prediction with a soft penalty, and a four-expert, top-two MoE actor
with task-label supervision. It trains jointly on closing the top drawer
and opening the bottom drawer, with 5k preparation actor updates followed
by 45k online actor updates. Evaluation pairs the frozen base and editor
on 50 initial states in two passes per task. Table~\ref{tab:rlt-two-source}
reports both recorded discount settings. The higher-discount configuration
reaches 94.50\% mean success, including 90\% on bottom-drawer opening
where its base succeeds in 4\% of cases. This demonstrates effective
correction with the fixed readout on these tasks, without identifying
which information in the representation enables it. The cohort and MoE
architecture differ from the six-recipe MLP study and the two-task L3
transfer comparison.

\begin{table}[tb]
\centering\small
\caption{\textbf{Two-source-task RL-token study.} Success rates (\%) use 100 paired
episodes per task. Discount factors apply per physical action.}
\label{tab:rlt-two-source}
\begin{tabularx}{\linewidth}{@{}>{\raggedright\arraybackslash}Xrrr@{}}
\toprule
Task & Base & $\gamma=0.92$ & $\gamma=0.995$ \\
\midrule
Close top drawer & 80.00 & 91.00 & \best{99.00} \\
Open bottom drawer & 4.00 & 4.00 & \best{90.00} \\
\midrule
\sumrow
Mean & 42.00 & 47.50 & \best{94.50} \\
\bottomrule
\end{tabularx}
\end{table}

\section{RoboTwin S4}
\label{app:robotwin}

This appendix supports Section~\ref{sec:experiments:robotwin}, which asks
whether the narrow-source pattern of LIBERO (gains on the source tasks
without transfer) recurs with a different base and simulator. It covers the
training configurations, matched evaluations, routing interventions,
checkpoint selection, and the 200-scene and 12-task evaluations.

\subsection{Full-action editing on four tasks}
\label{app:rlt-s4-training}

\paragraph{Tasks and configurations.}
The S4 cohort comprises \texttt{beat\_block\_hammer},
\texttt{hanging\_mug}, \texttt{move\_playingcard\_away}, and
\texttt{turn\_switch}. Simulation throughput limited source training to
these four tasks; this is the narrow end of the source-breadth
range of Section~\ref{sec:experiments:libero-reference}. Both runs use MoE
actors. The task-contrastive
configuration uses Equation~\ref{eq:task-aux} and marginal-route entropy
balancing; the baseline omits the task-contrastive term and uses
Switch-style balancing. Table~\ref{tab:rlt-s4-settings} records their training
settings. The runs also differ in initialization history and training seed.
The comparison therefore characterizes two training configurations rather
than isolating a single auxiliary loss.

\begin{table}[!htbp]
\centering\small
\caption{\textbf{Training settings of the S4 comparison.} TC denotes the
task-contrastive configuration. Online updates exclude initialization.}
\label{tab:rlt-s4-settings}
\setlength{\tabcolsep}{3pt}
\begin{tabularx}{\linewidth}{@{}>{\raggedright\arraybackslash}Xrr@{}}
\toprule
Setting & Baseline & MoE+TC \\
\midrule
Task-contrastive weight & 0 & 0.01 \\
Route-balancing weight & 0.001 & 0.02452 \\
Initialization critic updates & 300k & 150k \\
Initial online batch & 4,096 & 4,096 \\
Batch after update 994,849 & 4,096 & 16,384 \\
Online critic budget & 2M & 2M \\
Training seed & 20261024 & 20261008 \\
\bottomrule
\end{tabularx}
\end{table}

\begin{figure}[t]
    \centering
    \includegraphics[width=\linewidth]{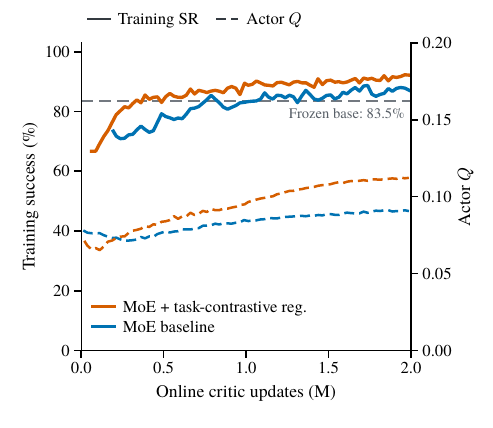}
    \caption{\textbf{Full-action editing with task-contrastive regularization.}
    Solid lines: four-task mean of rolling success over each task's latest
    100 training episodes (left axis). Colored dashed lines: raw actor Q (right axis).
    The gray horizontal line marks the earlier frozen-base evaluation
    (83.50\%, 50-action execution interface).
    Both metrics use 25k-update bins.}
    \label{fig:rlt-s4-training}
\end{figure}

Over 1.5--2.0M online critic updates, the task-contrastive configuration
reaches \SFourRegularizedSR\% mean training success, versus
\SFourBaselineSR\% for the baseline, a difference of \SFourGain{}
percentage points (Figure~\ref{fig:rlt-s4-training}).
This is a comparison of training configurations with the differences
listed above, not an isolated auxiliary-loss ablation.
Frozen-policy comparisons at two fixed analysis checkpoints are reported
below; the training-success difference is not their evaluation difference,
as frozen evaluations differ by $-0.25$ [$-5.00, +4.26$] points at 1.00M
updates and $+3.50$ [$-0.75, +7.75$] points at 1.49M updates.

\paragraph{Training success and aggregation.}
At each recorded snapshot, a task's rolling success rate uses its latest
100 completed training episodes with known outcomes. A four-task mean is
defined only when all four windows are complete. These are online training
statistics, separate from frozen-checkpoint evaluation.
We average snapshots within fixed, non-overlapping bins of 25,000 online
critic updates over 0--2M. Each plotted abscissa is the mean recorded update
within its bin; empty bins remain gaps. No further smoothing is applied.
The late-training summary averages the 20 bin means spanning 1.5--2.0M
with equal bin weight. Within-bin averaging weights recorded snapshots
equally. Update counters are attached to log records by nearest-time
matching within \mbox{2\,s}. Each curve represents one training run;
overlapping windows are not independent replicates.

Raw actor Q is extracted separately from each run's critic telemetry.
The four-task value is the arithmetic mean of finite task values at the same
snapshot; missing task values leave the macro value undefined.
We retain the latest timestamp at each update and bin Q using the same
25k-update boundaries as success, with its own logging cadence.
All Q axes span 0--0.20. Each critic is learned on its run's evolving replay
distribution, so raw Q describes value-learning dynamics rather than a
shared calibrated scale for ranking the two policies.

\begin{figure}[t]
    \centering
    \includegraphics[width=\linewidth]{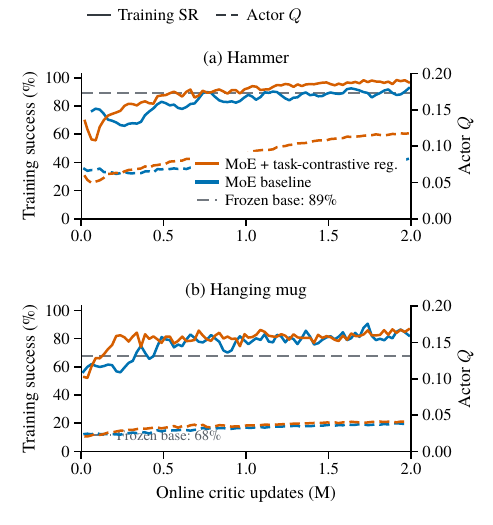}
    \caption{\textbf{S4 configuration comparison on Hammer and Hanging Mug.}
    Solid lines show rolling training success (left axis) and colored dashed
    lines show raw actor Q (right axis). Like Figure~\ref{fig:rlt-s4-training},
    both metrics use 25k-online-critic-update bins. Both configurations use
    MoE actors; Table~\ref{tab:rlt-s4-settings} specifies their settings.
    Gray horizontal lines mark each task's frozen-base evaluation success.}
    \label{fig:rlt-s4-tasks}
\end{figure}

\begin{figure}[t]
    \centering
    \includegraphics[width=\linewidth]{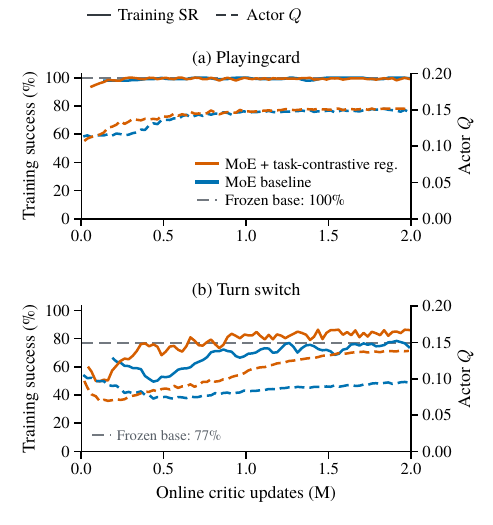}
    \caption{\textbf{S4 configuration comparison on Playingcard and Turn Switch.}
    Colors identify configurations; solid lines show training success and
    colored dashed lines show raw actor Q; gray horizontal lines mark
    frozen-base evaluation success. Axis limits and aggregation match
    Figure~\ref{fig:rlt-s4-tasks}.}
    \label{fig:rlt-s4-tasks-card-switch}
\end{figure}

\paragraph{Task-level differences.}
Over the same late-training interval, the task-contrastive configuration
exceeds the baseline by \SFourSwitchGain{} percentage points on
\texttt{turn\_switch} and \SFourHammerGain{} points on
\texttt{beat\_block\_hammer}. The differences on \texttt{hanging\_mug} and
\texttt{move\_playingcard\_away} are each less than one point in magnitude
(Figures~\ref{fig:rlt-s4-tasks} and~\ref{fig:rlt-s4-tasks-card-switch}).

\subsection{Matched frozen-policy evaluation and routing interventions}
\label{app:s4-matched-mechanisms}

\paragraph{Fixed checkpoints and matched base.}
The analysis fixes 1.00M and 1.49M online critic updates and compares
both configurations on the same four tasks and 100 scenarios per task
(Table~\ref{tab:s4-matched-checkpoints}). We reuse the earlier
task-contrastive evaluations and newly evaluate the MoE baseline and
frozen base with matching horizons, prompts, and per-scenario VLA sampling.
The frozen base succeeds in 328 of 400 episodes (82.00\%).
At 1.49M, the task-contrastive configuration rescues 49 of the 400 paired
episodes (12.3\%) and harms 25 (6.3\%); the MoE baseline rescues 43 (10.8\%)
and harms 33 (8.3\%).
The task-contrastive minus baseline differences are $-0.25$
[$-5.00$, $+4.26$] percentage points at 1.00M and $+3.50$
[$-0.75$, $+7.75$] at 1.49M. These paired percentile intervals use
2,000 within-task scenario-bootstrap draws, preserving the pairing across
policies and equally weighting tasks.

\begin{table}[t]
\centering\small
\caption{\textbf{Matched S4 evaluation}, with 100 scenarios per task. TC denotes
task-contrastive supervision; H, M, P, and S denote Hammer, Hanging Mug,
Playingcard, and Turn Switch. Entries are success percentages.}
\label{tab:s4-matched-checkpoints}
\setlength{\tabcolsep}{3pt}
\begin{tabular}{@{}lrrrrr@{}}
\toprule
Policy (updates) & H & M & P & S & Mean\,$\uparrow$ \\
\midrule
Frozen base & 91 & 61 & \best{100} & 76 & 82.00 \\
\midrule
MoE (1.00M) & 86 & \best{87} & 99 & 74 & \second{86.50} \\
MoE+TC (1.00M) & 86 & 81 & \best{100} & 78 & 86.25 \\
MoE (1.49M) & 84 & 81 & \best{100} & 73 & 84.50 \\
\hlrow
MoE+TC (1.49M) & \best{92} & 78 & 98 & \best{84} & \best{88.00} \\
\bottomrule
\end{tabular}
\end{table}

The two checkpoints were fixed for this analysis; these evaluation
scenarios had previously participated in checkpoint screening.
Repeating the task-contrastive 1.49M evaluation on the first 50 scenarios
per task changes 16 of 200 binary outcomes. These repeated executions
are not additional independent scenarios. Figure~\ref{fig:s4-extended-training}b,c
uses 10,000 paired within-task draws for marginal pointwise intervals;
the six checkpoints in panel (a) retain their original
100,000-draw intervals. These quantify scene uncertainty for fixed policies.

\paragraph{Fixed routing mixtures.}
For each 1.49M model, we average its valid top-two dispatch weights over
decisions in 50 recorded episodes per task. The task-mean intervention
uses this fixed mixture; the swapped intervention exchanges
Hammer with Switch and Mug with Playingcard; the uniform intervention
assigns weight $1/4$ to each expert. Mixtures combine expert outputs
before the actor's original $\tanh$ action transform. Task means are
generally dense, so fixed mixtures change expert participation as well
as state dependence.
Each intervention uses 30 scenarios per task, paired with newly recorded
unmodified controls: 102/120 successes for the MoE baseline and 106/120
for the task-contrastive configuration. Their task-mean intervention
effects are $-32.50$ [$-42.50$, $-22.50$] and $-74.17$
[$-82.50$, $-65.00$] percentage points, respectively, using 2,000 paired
bootstrap draws. Swapped and uniform mixtures also lower success
(Figure~\ref{fig:s4-extended-training}c and Table~\ref{tab:s4-routing-interventions}).
The new task-contrastive control is distinct from the 400-episode result in panel (b).

\begin{table}[tb]
\centering\footnotesize
\caption{\textbf{RoboTwin S4 routing interventions at 1.49M critic updates}
(4 tasks $\times$ 30 scenarios = 120 paired episodes per model). Entries
report success percentages.}
\label{tab:s4-routing-interventions}
\setlength{\tabcolsep}{2.5pt}
\begin{tabularx}{\linewidth}{@{}Xrrrr@{}}
\toprule
Model & \shortstack{Unmodified\\top-two} & \shortstack{Task\\mean} & \shortstack{Swapped\\mean} & Uniform \\
\midrule
MoE baseline & \best{85.00} & 52.50 & 36.67 & 30.83 \\
\hlrow
MoE+TC & \best{88.33} & 14.17 & 30.00 & 27.50 \\
\bottomrule
\end{tabularx}
\end{table}

\subsection{Extended training and checkpoint selection}
\label{app:s4-checkpoint-selection}

Figure~\ref{fig:s4-extended-training} follows the same task-contrastive S4
run through 2,823,723 online critic updates. The training seed and batch
schedule follow Table~\ref{tab:rlt-s4-settings}; the 2M budget there defines
the earlier configuration comparison. Both training success and raw actor Q
are averaged in non-overlapping 25k-update bins, retaining the final partial
bin. Q is measured on sampled training states, whose distribution changes
with training, rather than being a calibrated success probability.
Training uses exploration noise, whereas checkpoint evaluation executes
the deterministic actor.

\begin{figure}[tb]
    \centering
    \includegraphics[width=\linewidth]{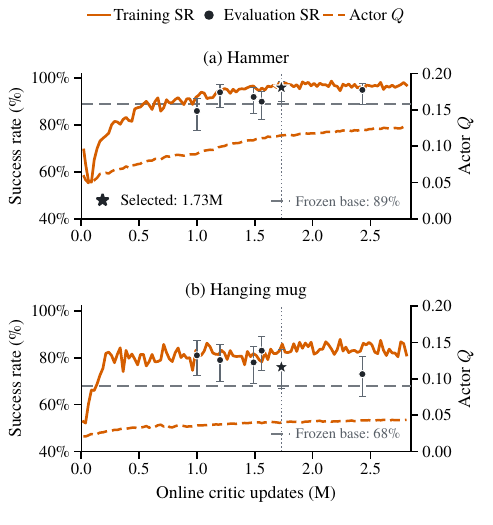}
    \caption{\textbf{Extended training on Hammer and Hanging Mug.}
    Training SR and actor Q follow the same 25k-update aggregation as
    Figure~\ref{fig:s4-extended-training}. Each evaluation point uses
    100 evaluation scenarios, with pointwise 95\% Wilson intervals.
    Stars and vertical lines mark the shared 1.73M checkpoint selected
    by four-task success. Solid lines show training SR; orange dashed lines show
    Q on the right axis. Gray horizontal lines mark frozen-base evaluation success.}
    \label{fig:s4-extended-hammer-mug}
\end{figure}

\begin{figure}[tb]
    \centering
    \includegraphics[width=\linewidth]{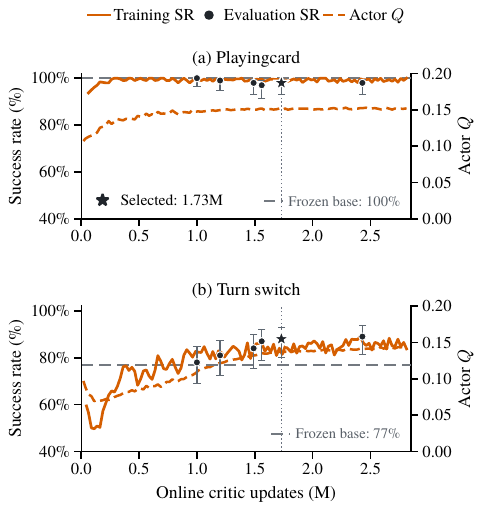}
    \caption{\textbf{Extended training on Playingcard and Turn Switch.}
    Evaluation uses 100 scenarios per checkpoint, with pointwise
    95\% Wilson intervals. Stars mark the same 1.73M checkpoint as in
    Figure~\ref{fig:s4-extended-hammer-mug}; gray horizontal lines mark
    frozen-base evaluation success. All panels share axis limits
    and the solid-SR / dashed-Q encoding.
    The six checkpoints retain distinct evaluation points even where
    success rates coincide.}
    \label{fig:s4-extended-card-switch}
\end{figure}

Figures~\ref{fig:s4-extended-hammer-mug} and
\ref{fig:s4-extended-card-switch} separate these trajectories by task.
The later 2.43M checkpoint loses three successful Mug episodes and one
Hammer episode relative to 1.73M, gains one Switch episode, and leaves
Playingcard success unchanged (Table~\ref{tab:s4-e100-finalists}).
Thus, the lower aggregate evaluation score reflects task-dependent changes
rather than a uniform decline across the cohort. Every panel marks the
same checkpoint selected by four-task success, not a task-specific optimum.

\paragraph{Staged search.}
We retrospectively screen saved checkpoints at 10k-update spacing in two
windows: 1.50--2.80M and an extension to 1.00--1.50M.
The windows contain 131 and 51 candidates, respectively, with the shared
1.50M checkpoint giving 181 distinct candidates from one training run.
First-stage screening evaluates 20 Hanging Mug scenarios. We retain the top eight checkpoints
by success count, including all ties at the eighth-place boundary, yielding
17 and 15 candidates in the two windows.
Second-stage screening evaluates 50 scenarios each for Hanging Mug and Turn Switch, then retains
three candidates per window by equal-task SR. Ties use lower equal-task
steps to success and then earlier update count.
Four-task checkpoint evaluation assesses these six finalists with 100 scenarios
per task, using the same ranking rules to select the final checkpoint.
Table~\ref{tab:s4-e100-finalists} reports every finalist.

\paragraph{Evaluation scenarios and uncertainty.}
All checkpoints use a fixed scenario plan, with seeds that do not appear in the
online rollout logs. These evaluations compare checkpoints within
the same four tasks; they do not test transfer to additional tasks.
Separation from offline demonstration seeds is not fully established.
The two screening stages reuse prefixes of the four-task evaluation
scenarios; the four-task scores also rank the finalists, so the reported scores are
selected-checkpoint evaluations. The anchor-task screening trades full-cohort
coverage for evaluation cost: 7,980 distinct checkpoint/task/scenario cases
replace 72,400 required to evaluate all 181 candidates on 100 scenarios
for each of the four tasks.
The six checkpoint intervals use 100,000 task-stratified empirical bootstrap
replicates and the 2.5th and 97.5th percentiles. They describe finite-scenario
uncertainty conditional on each checkpoint, not variation across training
seeds or a correction for checkpoint selection.
The latest complete four-task evaluation is at 2.43M; later screening
measurements cover only the first-stage Hanging Mug scenarios.

\begin{table}[t]
\centering\small
\caption{\textbf{Four-task evaluation of all six S4 finalists.} Each task has 100 scenarios;
task columns report success percentages. H: Hammer; M: Hanging Mug;
P: Playingcard; S: Turn Switch. Bold identifies the selected checkpoint.}
\label{tab:s4-e100-finalists}
\setlength{\tabcolsep}{4pt}
\begin{tabular}{@{}rrrrrr@{}}
\toprule
Updates (M) & H & M & P & S & Mean (\%)\,$\uparrow$ \\
\midrule
1.00 & 86 & 81 & \best{100} & 78 & 86.25 \\
1.20 & 94 & 79 & 99 & 81 & 88.25 \\
1.49 & 92 & 78 & 98 & 84 & 88.00 \\
1.56 & 90 & \best{83} & 97 & 87 & \second{89.25} \\
\hlrow
\textbf{1.73} & \best{96} & 76 & 98 & 88 & \textbf{89.50} \\
2.43 & 95 & 73 & 98 & \best{89} & 88.75 \\
\bottomrule
\end{tabular}
\end{table}

\subsection{Earlier RoboTwin S4 evaluation protocol}
\label{app:robotwin-base-reference}

Table~\ref{tab:robotwin-s4-results} reports the frozen LingBot-VLA base and
the shared full-action editor selected at 1.73M online critic updates.
Both evaluations contain 100 episodes per task. Success rates average
83.50\% for the base and 89.50\% for the editor. Steps to Success averages
physical actions over each policy's successful episodes within
each task, followed by an equal-task mean.
These evaluations use the earlier protocol; the base's
horizon and VLA sampling differ from the matched comparison in
Table~\ref{tab:s4-matched-checkpoints}. The older base result is retained
as the gray reference in the training plots of this appendix
(Figures~\ref{fig:rlt-s4-training}--\ref{fig:rlt-s4-tasks-card-switch}).

\begin{table}[tb]
\centering\small
\caption{\textbf{RoboTwin S4 Clean evaluation of the frozen base and the jointly trained
full-action editor} at the selected 1.73M-update checkpoint. Each policy is evaluated
on 100 episodes per task. Steps denotes Steps to Success: mean physical
environment actions over successful episodes. The last row averages tasks equally.}
\label{tab:robotwin-s4-results}
\setlength{\tabcolsep}{4pt}
\begin{tabular}{@{}lrrrr@{}}
\toprule
& \multicolumn{2}{c}{Frozen base} & \multicolumn{2}{c}{Full-action editor} \\
\cmidrule(lr){2-3}\cmidrule(l){4-5}
Task & SR (\%) & Steps & SR (\%) & Steps \\
\midrule
Hammer & 89.00 & 112.53 & \best{96.00} & 108.03 \\
Hanging Mug & 68.00 & 304.75 & \best{76.00} & 303.91 \\
Playingcard & \best{100.00} & 107.36 & 98.00 & 103.65 \\
Turn Switch & 77.00 & 65.12 & \best{88.00} & 69.61 \\
\midrule
\sumrow
Mean & 83.50 & 147.44 & \best{89.50} & 146.30 \\
\bottomrule
\end{tabular}
\end{table}

The recorded base interface executes its full 50-action plan with ten
denoising steps. Initialization failures can be replaced by official same-task
reserve seeds before a rollout starts; failures after rollout starts count
as failures. The editor executes ten-action chunks under the evaluation
settings described in Appendix~\ref{app:s4-checkpoint-selection}.

\subsection{200-scene evaluation, matched supervision ablation, phase routing, and transfer to 12 other tasks}
\label{app:s4-followup}

We extend the RoboTwin S4 evaluation along four axes. Each policy is one
training run; bootstrap intervals use 2,000 stratified scene resamples with
seed~20261115 and measure finite-scene uncertainty rather than training-seed
variance.

\begin{table*}[t]
\centering\small
\caption{\textbf{Evaluation of the matched base and five S4 editor checkpoints on
200 scenes per task} (4 tasks $\times$ 200 scenes = 800 episodes per policy;
4,800 in total). H: Hammer; M: Hanging Mug; P: Playingcard; S: Turn Switch.
Rescue and harm are shares of the 800 pairs against the matched base (counts
in parentheses).}
\label{tab:s4-ra-200}
\setlength{\tabcolsep}{4pt}
\begin{tabularx}{\textwidth}{@{}>{\raggedright\arraybackslash}Xrrrrrrrr@{}}
\toprule
Policy (updates) & H & M & P & S & Macro SR\,$\uparrow$ & $\Delta$ vs.\ Base [95\% CI] & Rescue\,$\uparrow$ & Harm\,$\downarrow$ \\
\midrule
Matched base & 80.5 & 57.5 & \best{100.0} & 60.5 & 74.63 & --- & --- & --- \\
\midrule
MoE baseline (1.00M) & 72.0 & 76.0 & 99.0 & 58.0 & 76.25 & +1.63 [$-$1.25, +4.63] & 10.3\% (82) & 8.6\% (69) \\
MoE baseline (1.49M) & 77.0 & \best{79.0} & \second{99.5} & 57.5 & 78.25 & +3.63 [+0.63, +6.50] & 11.4\% (91) & 7.8\% (62) \\
MoE+TC (1.00M) & 89.0 & 73.5 & 99.0 & 71.5 & \second{83.25} & \second{+8.63 [+5.63, +11.75]} & \best{15.1\% (121)} & 6.5\% (52) \\
\hlrow
MoE+TC (1.49M) & \best{91.5} & 72.0 & 98.5 & \best{74.5} & \best{84.13} & \best{+9.50 [+6.62, +12.50]} & \second{15.0\% (120)} & \best{5.5\% (44)} \\
MoE+TC (1.73M) & \second{89.5} & 73.5 & 98.0 & 71.0 & 83.00 & +8.38 [+5.25, +11.38] & 14.3\% (114) & \second{5.9\% (47)} \\
\bottomrule
\end{tabularx}
\end{table*}

\begin{table}[t]
\centering\footnotesize
\caption{\textbf{Matched task-contrastive supervision ablation on the first 100
scenes per task.} Both arms share one 150k joint warm-up and differ only in the
task-contrastive weight (0.01 vs.\ 0); 400 scenes per endpoint and arm, 1,600
episodes in total. SUP/UNSUP: with/without task-contrastive supervision.}
\label{tab:s4-rb-matched}
\setlength{\tabcolsep}{2.4pt}
\begin{tabularx}{\linewidth}{@{}>{\raggedright\arraybackslash}Xrrrrrr@{}}
\toprule
Endpoint & Arm & H & M & P & S & SR\,$\uparrow$ / $\Delta$ [95\% CI] \\
\midrule
125k & SUP & 80 & 65 & 94 & 45 & 71.00 / +4.50 \\
125k & UNSUP & 62 & 65 & 97 & 42 & 66.50 [$-$0.26, +9.75] \\
\midrule
250k & SUP & 81 & 70 & 93 & 48 & \best{73.00} / +2.00 \\
250k & UNSUP & 66 & 69 & 98 & 51 & 71.00 [$-$3.00, +7.25] \\
\bottomrule
\end{tabularx}
\end{table}

\paragraph{200-scene evaluation.}
Evaluating the six frozen S4 policies across 200 scenes per task
(4,800 episodes; Table~\ref{tab:s4-ra-200}) yields 74.63\% (597/800)
for the matched base, 78.25\% for the 1.49M MoE baseline (+3.63 [+0.63,
+6.50]), and 84.13\% for 1.49M MoE+TC (+9.50 [+6.62, +12.50]; 120
rescues and 44 harms of 800 pairs). Its gain is largest on the two tasks
the base solves least often, +14.5 points on Hanging Mug (base 57.5\%) and
+14.0 on Turn Switch (60.5\%), against +11.0 on Hammer (80.5\%) and $-$1.5
on Playingcard, which the base always solves. MoE+TC leads the MoE baseline by +7.00
[+4.00, +10.00] at 1.00M and +5.88 [+2.88, +8.75] at 1.49M, with the largest differences on
Hammer (+14.5 points at 1.49M) and Turn Switch (+17.0 points). As noted in
Table~\ref{tab:rlt-s4-settings}, these two runs also differ in warm-up,
route balancing, batch schedule, and seed, and the disjointness of evaluation
scenes from training demonstrations is not established. Additional paired
200-scene comparisons across the MoE+TC training trajectory show $+0.88$
[$-$2.00, $+$3.50] percentage points from 1.00M to 1.49M, $-$1.13 [$-$3.88,
$+$1.50] from 1.49M to 1.73M, and $-$0.25 [$-$3.13, $+$2.63] from 1.00M to
1.73M, indicating a late-training plateau rather than continued net
improvement beyond 1.00M updates.

\paragraph{Matched task-contrastive supervision ablation.}
To isolate task-contrastive regularization, two arms start from the same
150k-update warm-up checkpoint and differ only in
the task-contrastive weight (0.01 vs.\ 0 under identical DyGRO entropy
balancing, batch schedule, and seed; Table~\ref{tab:s4-rb-matched}). Across
400 matched scenes at each fixed horizon, SUP minus UNSUP differs by +4.50
[$-$0.26, +9.75] points at 125k online updates (284 vs.\ 266 successes) and
+2.00 [$-$3.00, +7.25] at 250k (292 vs.\ 284). On the next 100 scenes per
task, the same 250k pair differs by +5.50 [+0.50, +10.50] points (299/400 vs.\
277/400; Hammer 70 vs.\ 58, Hanging Mug 72 vs.\ 66, Playingcard 99 vs.\ 99,
Turn Switch 58 vs.\ 54; 63 SUP-only and 41 UNSUP-only successes). These are
two evaluation cohorts for one trained pair, not two training runs. Two repeat
evaluations of the first cohort at the 250k endpoint yield paired
SUP minus UNSUP differences of +4.50 [$-$0.25, +9.25] (74.5\% vs.\ 70.0\%)
and +4.50 [$-$0.25, +9.00] (75.5\% vs.\ 71.0\%), with both intervals still
crossing zero; a repeat evaluation changes the outcome of 20/400 episodes (5.0\%, 12 on
\texttt{hanging\_mug}) for SUP and 22/400 (5.5\%, 12 on \texttt{hanging\_mug})
for UNSUP, the level of simulator variation. Per-task effects diverge:
\texttt{beat\_block\_hammer} gains +18.0 [+7.0, +29.0] at 125k and +15.0
[+4.0, +27.0] at 250k, whereas \texttt{move\_playingcard\_away} (93\% vs.\
98\%, $-$5.0 [$-$10.0, 0.0]) and \texttt{turn\_switch} (48\% vs.\ 51\%) do
not improve. One failed 34-episode batch of the UNSUP 125k evaluation was excluded and
rerun before its replacement outcomes were inspected. On the first cohort
every overall 95\% interval includes zero; on the second cohort the interval
excludes zero.

\paragraph{Phase-coupled routing diagnostics.}
Across 400 offline traces (50 per task for each of the MoE baseline and MoE+TC
at 1.49M, checked against 200 base traces), we define physical
phases via measured gripper closing ($e_1$: openness $<0.8$) and
task-native contact/lift events ($e_3$: hammer raised $\ge 0.05$\,m, mug
within 0.10\,m of functional point, switch contact; Playingcard lacks an
$e_3$ getter and is evaluated only on secondary $e_1$ splits). Table~\ref{tab:s4-rd-phase}
shows that mutual information between unordered top-two expert selection and
task phase exceeds all 1,000 within-episode circular-shift nulls in every
feasible complete-phase cohort (6/6, unadjusted $p=0.001$; all six remain
significant after Holm adjustment at $p_{\text{Holm6}}=0.006$ and across
12 tests at $p_{\text{Holm12}}=0.012$) and all secondary $e_1$ cohorts (8/8,
$p=0.001$). For instance, on \texttt{beat\_block\_hammer} under the MoE baseline,
Expert~1 carries 0.535 mean weight before $e_1$ and drops to 0.044 after
$e_3$, while Expert~2 rises from 0.029 to 0.597; under
MoE+TC, Expert~1 drops from 0.704 before $e_1$ to 0.000
between $e_1$ and $e_3$. Top-two switches can only occur on the ten-action decision
grid; Table~\ref{tab:s4-rd-grid} therefore compares observed switch distances
against decision-grid nulls with Holm-6 adjustment. Switches lie closer to
phase events than the null for both models on Hammer
($p_{\text{Holm6}}=0.006$) and for the MoE baseline on Hanging Mug
($p_{\text{Holm6}}=0.006$). Routing mass
measures expert utilization rather than causal expert skill.

\begin{table}[t]
\centering\footnotesize
\caption{\textbf{Offline phase-routing mutual information and event-switch proximity}
(50 traces per task and model; 1,000 within-episode nulls). Playingcard
lacks an $e_3$ getter and uses the secondary $e_1$-only cohort.}
\label{tab:s4-rd-phase}
\setlength{\tabcolsep}{2.2pt}
\begin{tabularx}{\linewidth}{@{}>{\raggedright\arraybackslash}lXrrrr@{}}
\toprule
Model & Task (cohort) & MI (nats) & Null mean & \shortstack{Switch\\steps} & \shortstack{Null\\steps} \\
\midrule
MoE & Hammer ($e_1,e_3$) & 0.5430 & 0.0238 & 15.77 & 25.75 \\
MoE & Hanging Mug ($e_1,e_3$) & 0.2541 & 0.0077 & 39.37 & 47.32 \\
MoE & Turn Switch ($e_1,e_3$) & 0.2526 & 0.0286 & 17.99 & 19.27 \\
MoE & Playingcard ($e_1$) & 0.3070 & 0.0087 & --- & --- \\
\midrule
MoE+TC & Hammer ($e_1,e_3$) & 0.3969 & 0.0197 & 14.19 & 24.18 \\
MoE+TC & Hanging Mug ($e_1,e_3$) & 0.1684 & 0.0071 & 48.95 & 47.01 \\
MoE+TC & Turn Switch ($e_1,e_3$) & 0.1861 & 0.0185 & 20.85 & 23.31 \\
MoE+TC & Playingcard ($e_1$) & 0.3129 & 0.0120 & --- & --- \\
\bottomrule
\end{tabularx}
\end{table}

\begin{table}[t]
\centering\footnotesize
\caption{\textbf{Ten-action decision-grid switch-proximity nulls and Holm-6
adjusted $p$-values for the six complete-phase cohorts.} All
six corresponding phase-MI tests survive at $p_{\text{Holm6}}=0.006$ and
$p_{\text{Holm12}}=0.012$.}
\label{tab:s4-rd-grid}
\setlength{\tabcolsep}{2.2pt}
\begin{tabularx}{\linewidth}{@{}>{\raggedright\arraybackslash}lXrrrr@{}}
\toprule
Model & Task & Obs.\,(steps) & Grid null & $p_{\text{grid}}$ & $p_{\text{Holm6}}$ \\
\midrule
MoE & Hammer & 15.77 & 23.82 & 0.0010 & 0.0060 \\
MoE+TC & Hammer & 14.19 & 22.28 & 0.0010 & 0.0060 \\
MoE & Hanging Mug & 39.37 & 47.02 & 0.0010 & 0.0060 \\
MoE+TC & Hanging Mug & 48.95 & 46.73 & 0.9890 & 0.9890 \\
MoE & Turn Switch & 17.99 & 19.37 & 0.1938 & 0.3876 \\
MoE+TC & Turn Switch & 20.85 & 23.39 & 0.0529 & 0.1588 \\
\bottomrule
\end{tabularx}
\end{table}

\paragraph{Zero-update transfer to 12 other RoboTwin tasks.}
Evaluating the four-task MoE+TC checkpoint at 1.49M without updates against
the matched base on 12 other RoboTwin tasks
(50 scenes/task = 600 paired outcomes/policy; Table~\ref{tab:s4-re-transfer})
lowers macro SR from 71.00\% (426/600) to 25.17\% (151/600), a paired
macro difference of $-$45.83 [$-$49.83, $-$42.00] percentage points. The
editor rescues 25 of the 600 paired episodes (4.2\%) while harming 300
(50.0\%): four multi-step tasks fall from 54--98\% to 0\%,
whereas compact button/stapler tasks retain most performance
(\texttt{press\_stapler} from 90\% to 86\%, \texttt{click\_alarmclock}
from 64\% to 62\%, \texttt{click\_bell} from 66\% to 60\%). Harm is 6--24\% of
the pairs on these three press and click tasks and 34--98\% on the nine
others. Successful-episode
lengths average 230.26\,$\pm$\,149.13 steps ($n=426$) for Base vs.\
140.29\,$\pm$\,141.50 ($n=151$) for the editor, reflecting different
surviving task subsets rather than faster completion. This four-task
full-action result parallels the two-task LIBERO editors
(Appendix~\ref{app:sw-completions-twosource}); the RoboTwin editors here have
four source tasks.

\begin{table}[t]
\centering\footnotesize
\caption{\textbf{Zero-update transfer from the four-task S4 editor} (MoE+TC, 1.49M)
to 12 other RoboTwin tasks (50 matched scenes per task, 600 pairs). Rescue
and harm are counts of each task's 50 pairs.}
\label{tab:s4-re-transfer}
\setlength{\tabcolsep}{2.2pt}
\begin{tabularx}{\linewidth}{@{}>{\raggedright\arraybackslash}Xrrrr@{}}
\toprule
Task & Base\,(\%) & Ed.\,(\%) & Rescue & Harm \\
\midrule
\texttt{blocks\_ranking\_rgb} & 84.0 & 32.0 & 2 & 28 \\
\texttt{blocks\_ranking\_size} & 62.0 & 18.0 & 0 & 22 \\
\texttt{click\_alarmclock} & 64.0 & 62.0 & 5 & 6 \\
\texttt{click\_bell} & 66.0 & 60.0 & 9 & 12 \\
\texttt{move\_can\_pot} & 76.0 & 10.0 & 1 & 34 \\
\texttt{move\_stapler\_pad} & 70.0 & 20.0 & 6 & 31 \\
\texttt{open\_microwave} & 54.0 & 0.0 & 0 & 27 \\
\texttt{place\_can\_basket} & 68.0 & 0.0 & 0 & 34 \\
\texttt{place\_dual\_shoes} & 98.0 & 0.0 & 0 & 49 \\
\texttt{press\_stapler} & 90.0 & 86.0 & 1 & 3 \\
\texttt{stack\_bowls\_three} & 74.0 & 0.0 & 0 & 37 \\
\texttt{stamp\_seal} & 46.0 & 14.0 & 1 & 17 \\
\midrule
\sumrow
Mean / Total (12) & 71.00 & 25.17 & 25 & 300 \\
\bottomrule
\end{tabularx}
\end{table}

\begin{table*}[t]
\centering\footnotesize
\caption{\textbf{Amplitude scaling and unsupervised MoE on 12 other RoboTwin tasks (600 matched scenes).}
All three editor rows share the same 600 matched initial states (12 tasks $\times$ 50 scenes; fresh matched base
SR 68.67\%, 412/600) and evaluate fixed 1.49M-update S4 checkpoints without updates.
Halving the MoE+TC correction ($\alpha=0.5$) raises SR by +7.00 points over full scale ($\alpha=1.0$) and lowers
harm by 7.8 percentage points (47 fewer harmed scenes), while transfer remains negative. The unsupervised MoE
baseline (1.49M) also transfers negatively; MoE+TC minus MoE is
$-$4.00 [$-$7.67, $-$0.33] points,
reflecting the full training-configuration difference of Table~\ref{tab:rlt-s4-settings} rather than an
isolated supervision ablation. R, K, H and F are shares of the 600 paired scenes (counts in parentheses).}
\label{tab:s4-b4-g6-overall}
\setlength{\tabcolsep}{3.5pt}
\begin{tabularx}{\textwidth}{@{}>{\raggedright\arraybackslash}Xrrrrrr@{}}
\toprule
Editor (1.49M) & SR\,(\%) & $\Delta$ vs.\ Base [95\% CI] & Rescue (R) & Kept (K) & Harm (H) & Still failed (F) \\
\midrule
MoE+TC, full ($\alpha=1.0$) & 26.50 & $-$42.17 [$-$46.17, $-$38.00] & 6.2\% (37) & 20.3\% (122) & 48.3\% (290) & 25.2\% (151) \\
\hlrow
MoE+TC, half ($\alpha=0.5$) & 33.50 & $-$35.17 [$-$39.17, $-$31.33] & 5.3\% (32) & 28.2\% (169) & 40.5\% (243) & 26.0\% (156) \\
MoE baseline, full ($\alpha=1.0$) & 30.50 & $-$38.17 [$-$42.50, $-$33.83] & 6.3\% (38) & 24.2\% (145) & 44.5\% (267) & 25.0\% (150) \\
\bottomrule
\end{tabularx}
\end{table*}

\begin{table}[t]
\centering\footnotesize
\caption{\textbf{Per-task success rates (\%) on the 12 other RoboTwin tasks in the matched 600-scene cohort}
(50 scenes/task; base, MoE+TC at full and half amplitude, and unsupervised MoE baseline at 1.49M updates).}
\label{tab:s4-b4-g6-tasks}
\setlength{\tabcolsep}{2.4pt}
\begin{tabularx}{\linewidth}{@{}>{\raggedright\arraybackslash}Xrrrr@{}}
\toprule
Task & Base & \shortstack{MoE+TC\\$\alpha=1.0$} & \shortstack{MoE+TC\\$\alpha=0.5$} & \shortstack{MoE\\$\alpha=1.0$} \\
\midrule
\texttt{blocks\_ranking\_rgb} & 88.0 & 34.0 & 76.0 & 48.0 \\
\texttt{blocks\_ranking\_size} & 46.0 & 24.0 & 42.0 & 18.0 \\
\texttt{click\_alarmclock} & 58.0 & 60.0 & 64.0 & 72.0 \\
\texttt{click\_bell} & 50.0 & 60.0 & 34.0 & 44.0 \\
\texttt{move\_can\_pot} & 90.0 & 4.0 & 10.0 & 30.0 \\
\texttt{move\_stapler\_pad} & 66.0 & 22.0 & 32.0 & 12.0 \\
\texttt{open\_microwave} & 60.0 & 0.0 & 10.0 & 4.0 \\
\texttt{place\_can\_basket} & 62.0 & 4.0 & 12.0 & 8.0 \\
\texttt{place\_dual\_shoes} & 76.0 & 0.0 & 0.0 & 0.0 \\
\texttt{press\_stapler} & 92.0 & 92.0 & 88.0 & 92.0 \\
\texttt{stack\_bowls\_three} & 86.0 & 2.0 & 8.0 & 26.0 \\
\texttt{stamp\_seal} & 50.0 & 16.0 & 26.0 & 12.0 \\
\midrule
\sumrow
Mean (12 tasks) & 68.67 & 26.50 & 33.50 & 30.50 \\
\bottomrule
\end{tabularx}
\end{table}

\begin{table}[t]
\centering\footnotesize
\caption{\textbf{Outcomes of the two S4 editors in the same scenes.} The 600 matched scenes of
Table~\ref{tab:s4-b4-g6-overall} (12 other tasks, full correction), split by the base outcome and by which
editor changes it. The base succeeds in 412 scenes and fails in 188. Counts, with shares of
the 600 scenes.}
\label{tab:s4-g6-transitions}
\setlength{\tabcolsep}{3pt}
\begin{tabularx}{\linewidth}{@{}>{\raggedright\arraybackslash}Xrrrr@{}}
\toprule
 & Both & \shortstack[r]{MoE+TC\\only} & \shortstack[r]{MoE\\only} & Neither \\
\midrule
Harmed by & 233 (38.8\%) & 57 (9.5\%) & 34 (5.7\%) & 88 (14.7\%) \\
Rescued by & 16 (2.7\%) & 21 (3.5\%) & 22 (3.7\%) & 129 (21.5\%) \\
\bottomrule
\end{tabularx}
\end{table}

\paragraph{Amplitude scaling and unsupervised MoE on the same 12 other tasks.}
To test whether shrinking the correction reduces harm on RoboTwin and whether
negative narrow-source transfer also occurs without task-contrastive routing
supervision, we run a matched 600-scene follow-up on the same 12 tasks
(50 scenes per task; fresh matched base SR 68.67\%, 412/600) comparing the
1.49M MoE+TC editor at full ($\alpha=1.0$) and half ($\alpha=0.5$) correction
amplitude alongside the 1.49M unsupervised MoE baseline ($\alpha=1.0$;
Tables~\ref{tab:s4-b4-g6-overall} and~\ref{tab:s4-b4-g6-tasks}). Halving the
MoE+TC correction raises SR by +7.00 points (26.50\% to 33.50\%) and reduces
harm from 48.3\% to 40.5\% (290 to 243 harmed scenes, 47 fewer), mirroring
the LIBERO scaling direction while transfer stays negative ($-$35.17
[$-$39.17, $-$31.33] points vs.\ base). The unsupervised MoE baseline likewise
transfers negatively, falling to 30.50\% ($-$38.17 [$-$42.50, $-$33.83];
rescue 6.3\%, harm 44.5\%). The paired difference MoE+TC minus MoE baseline
is $-$4.00 [$-$7.67, $-$0.33] points, reflecting the full training-configuration
differences of Table~\ref{tab:rlt-s4-settings} (route balancing, warm-up,
batch schedule and seed) rather than an isolated supervision ablation.

\paragraph{The two editors harm largely the same scenes.}
Table~\ref{tab:s4-g6-transitions} splits the same \SWtrGScenes{} scenes by
what each full-scale editor does to the base outcome. In \SWtrGBothHarm{}
scenes (\SWtrGBothHarmShare\%) the base succeeds and both editors fail,
which is \SWtrGOverlapTC\% of the \SWtrGHarmTC{} scenes harmed by MoE+TC
and \SWtrGOverlapMoE\% of the \SWtrGHarmMoE{} harmed by the MoE baseline.
Harm outside the source tasks therefore falls largely on the same scenes under
either routing configuration. The net difference of \SWtrGNet{} scenes between
MoE+TC and the MoE baseline arises on base successes (\SWtrGNetSucc{} scenes, against
\SWtrGNetFail{} on base failures): the two configurations differ in harm,
and rescue the same number of base failures to within one scene.

\section{System Measurements and the Coffee Study}
\label{app:system}

This appendix reports the ReSOP measurements cited in
Section~\ref{sec:method:resop} and the single-task Coffee study in which the
chunk length was set.

\subsection{System throughput and task allocation}
\label{app:system-performance}

Table~\ref{tab:resop-performance} summarizes measurements of the three
ReSOP system mechanisms detailed below.

\begin{table}[tb]
\centering\small
\caption{\textbf{ReSOP mechanisms and measurements.}
Max/min ratios compare per-task cumulative totals. Environment stepping
is measured by replaying recorded commands.}
\label{tab:resop-performance}
\setlength{\tabcolsep}{3pt}
\begin{tabularx}{\linewidth}{@{}>{\raggedright\arraybackslash}p{0.26\linewidth}>{\raggedright\arraybackslash}X>{\raggedright\arraybackslash}p{0.22\linewidth}@{}}
\toprule
Mechanism & Evidence & Scope \\
\midrule
Task-indexed replay & Interaction: $7.07\times$; critic samples: $1.06\times$ & 112-task training logs \\
Inference snapshots & 22.7\,MB & Published snapshot \\
Chunk-aware stepping & 79\% fewer render calls; 46.2 to 64.0 actions/s & Two-task replay \\
\bottomrule
\end{tabularx}
\end{table}

\paragraph{Sustained 112-task operation.}
The online pipeline measurement spans actor updates 308,775--483,775
within one resumed training segment. Its 35 consecutive windows each
contain 5k actor updates and together span 48,615.54\,s. One learner and
three collectors occupy four A100-SXM4-80GB GPUs; each collector runs
two environments. The interval contains 175k actor updates and 700k
global critic updates, giving 3.60 and 14.40 updates/s, respectively.
A global critic update consumes 256 replay samples across its selected
tasks. The resulting 3,686 replay samples/s counts repeated sample use,
not newly collected transitions. This interval measures continuous online
operation rather than the duration of the entire training run.

\paragraph{Task allocation.}
The cumulative records cover 66,254 completed episodes across 112 source
tasks. Episode counts range from 583 to 601 per task, with a coefficient
of variation of 1.00\%. Their cumulative lengths range from 38,221 to
270,227 physical actions per task (7.07-fold), totaling 12,272,524 actions.
Critic replay sampling instead ranges from 4,383,936 to 4,646,272 samples
per task, with a coefficient of variation of 1.63\% and a maximum/minimum
ratio of 1.060. These sampling records total 500,318,464 sample uses and
cover a different cumulative interval from the throughput measurement.
The allocation statistic measures critic replay sample use.
Among the completed episodes, all 11,654 failures reach the 512-action
limit. The smallest interaction total belongs to closing the top drawer:
586 episodes and 38,221 actions, including 12 failures. The largest belongs
to stacking bowls in the tray: 590 episodes and 270,227 actions, including
477 failures. Successful episode lengths also differ across tasks.

\paragraph{Inference snapshot size.}
The published snapshot at 481,250 actor updates occupies 22,664,547 bytes
(22.7 decimal MB) and contains the actor and inference-side representation.
Task-specific critics, target networks, and optimizer state remain at the
learner. The retained full resume state occupies 156.1 GB at 465,112 updates
and also includes replay and runtime state.

\paragraph{Rendering benchmark.}
We replay recorded commands from the 481,250-update editor on tasks 9 and
102, with one initial state per task and five-action chunks. Each task has
one untimed validation replay and eight timed paired repetitions in
alternating order. The command traces contain 113 and 86 physical actions,
respectively, yielding 1,592 timed actions per rendering mode.
Full rendering takes 34.4443 seconds; boundary rendering takes 24.8758 seconds,
corresponding to 46.22 and 64.00 actions per second of environment stepping.
Camera render calls fall from 3,184 to 656, a reduction of 79.4\%.
All replays match in simulator state, numeric observations, rewards,
termination flags, and both camera images at chunk boundaries.

Timing covers native chunk stepping in one serial environment on an
A100 GPU while other jobs run on the same machine. It excludes policy
inference, transport, learning, reset and settling, validation hashing,
and video writing. The adapter uses the rendering path enabled in the
112-task LIBERO training run: camera updates occur at chunk boundaries
and episode end, with physical dynamics and control checks at every action.

\paragraph{Collector-count comparison.}
We compare one and four collectors in short Coffee runs, with the learner
and collectors sharing one A100-SXM4-80GB GPU. Within each state--action
coordinate setting, the pair uses the same physical GPU and code revision.
Each run completes 400 actor updates at batch size 256 and four critic
updates per actor update. Table~\ref{tab:coffee-collectors} reports online
pipeline throughput, including waits for collection.

\begin{table}[tb]
\centering\small
\caption{\textbf{Coffee collector-count measurements.} Each row is one short
400-actor-update run. Environment rates use the logged physical-action counter;
they are not independently deduplicated interaction counts.}
\label{tab:coffee-collectors}
\setlength{\tabcolsep}{4pt}
\begin{tabularx}{\linewidth}{@{}>{\raggedright\arraybackslash}Xrrr@{}}
\toprule
Coordinates & Collectors & \shortstack{Actor\\updates/s} & \shortstack{Logged physical\\actions/s} \\
\midrule
EE & 1 & 12.04 & 24.30 \\
EE & 4 & 10.12 & 87.67 \\
Joint position & 1 & 11.58 & 24.62 \\
Joint position & 4 & 9.17 & 92.17 \\
\bottomrule
\end{tabularx}
\end{table}

Increasing the collector count raises logged physical-action throughput
to 3.61 times its one-collector rate in EE coordinates and 3.74 times
in joint-position coordinates,
while actor-update throughput falls in both cases. Under this shared-GPU
placement, higher collection throughput coexists with slower learner
progress in wall time. The comparison changes collector count, not the
rendering policy, and measures no time-to-success advantage.

\newcommand{\swVideoClips}{389}
\newcommand{\swVideoSame}{314}
\newcommand{\swVideoSameLen}{178}
\newcommand{\swVideoFiles}{10}

\begin{table*}[t]
\centering\footnotesize
\caption{\textbf{The replay clips}, numbered in the order of the project page and of
the ancillary files. Every panel is a replay whose outcome equals the formal
outcome of the same policy, task and initial state. L1, L2 and L3 are the
112-task editors unless a row says otherwise; scene coverage refers to their
source tasks.}
\label{tab:sw-videos-files}
\setlength{\tabcolsep}{4pt}
\begin{tabularx}{\textwidth}{@{}r>{\raggedright\arraybackslash}p{0.34\textwidth}>{\raggedright\arraybackslash}p{0.25\textwidth}>{\raggedright\arraybackslash}X@{}}
\toprule
Clip & Task & Panels, left to right & What the clip shows \\
\midrule
1 & Put the chocolate pudding to the left of the plate (uncovered scene) & Base: failure; L1: success; L2: success; L3: success & All three editors rescue a base failure \\
2 & Pick up the salad dressing and place it in the basket (source-covered scene) & Base: failure; L1: success; L2: success; L3: success & All three editors rescue a base failure \\
3 & Put the right moka pot on the stove (uncovered scene) & Base: failure; L1: success; L2: failure; L3: failure & Only the full-action editor rescues the base failure \\
4 & Put the chocolate pudding to the right of the plate (uncovered scene) & Base: success; L1: success; L2: success; L3: failure & The bounded editor loses a base success that both soft-penalty editors keep \\
5 & Put the wine bottle on top of the cabinet (source-covered scene) & Base: success; L3 full: failure; L3 half: success & Halving the bounded correction restores the base success \\
6 & Put the red mug on the plate (uncovered scene) & Base: success; L3 full: failure; L3 half: success & Halving the bounded correction restores the base success \\
7 & Pick up the salad dressing and place it in the basket (source-covered scene) & Base: failure; L3 live: success; L3 frozen: failure & The rescue disappears when the editor's camera frames are frozen at the first decision \\
8 & Put the red mug on the plate & Base: success; L3, 2 source tasks: failure & An editor trained on two source tasks loses a base success \\
9 & Beat block hammer (one of the editor's four source tasks) & Base: failure; Editor: success & The shared editor rescues a base failure on a source task \\
10 & Move can pot (not among the editor's four source tasks) & Base: success; Editor: failure & Without any update, the four-task editor loses a base success on another task \\
\bottomrule
\end{tabularx}

\end{table*}

\paragraph{Cross-runtime parity and memory on an RTX~4090.}
On 200 paired native episodes replayed on one RTX~4090 across Tasks~7, 64,
94, and 120 (100 Base, 100 16-task L2; 25 states/task), outcomes agree on
185/200 pairs (92.5\%: Base 4\% vs.\ 4\% in the A100 evaluation; L2 86\% vs.\
89\%, $-$3.0 percentage points), within the parity criterion fixed beforehand.
The replay runs at 269.4 episodes/hour including startup, with 14.525\,GiB peak Torch-allocated
memory (15,595,670,016 bytes; excludes external EGL/driver allocations).

\subsection{Coffee configurations and learning budgets}
\label{app:coffee-update-budget}

\begin{figure}[t]
    \centering
    \includegraphics[width=\linewidth]{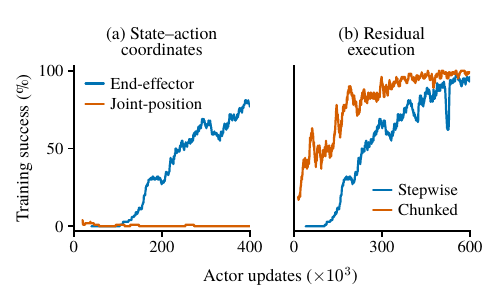}
    \caption{\textbf{Residual-learning configurations on Coffee.}
    (a) End-effector versus joint-position state--action coordinates.
    (b) Stepwise versus chunk-level editing in end-effector coordinates.
    Curves show success over the latest 100 training episodes.
    The blue curve is the same run in both
    panels; the horizontal axis counts actor optimizer updates.}
    \label{fig:coffee-design}
\end{figure}

We use DexMimicGen's Coffee task~\citep{jiang2025dexmimicgen}
(\texttt{TwoArmCoffee}), in which a fixed-base GR1 humanoid places a coffee
pod into a coffee maker and closes its lid through coordinated bimanual
manipulation. Our residual-learning implementation builds on
ResFiT~\citep{ankile2025residual}.
Each Coffee configuration is trained with seed 0. Training success is
measured over the latest 100 completed episodes, without additional
smoothing. The same end-effector stepwise run appears in both panels of
Figure~\ref{fig:coffee-design}. The coordinate comparison changes
proprioceptive state and action coordinates together: end-effector (EE) states
and corrections are compared with joint-position states and corrections.

The EE configuration learns successful corrections, whereas the
joint-position configuration remains near zero training success over the
displayed update budget.

The temporal comparison uses end-effector coordinates for both
configurations. Stepwise editing executes one action per decision and
uses a five-action TD backup. Chunk-level editing executes up to
five actions per decision and uses a three-chunk backup, spanning 15
physical actions when all three chunks are complete. Both use a batch
size of 256 during the first 400k actor updates.

Actor-update and interaction budgets describe different aspects of
learning efficiency. Near 400k actor updates, chunk-level and stepwise
editing reach 97\% and 77\% rolling training success, respectively;
at 600k actor updates these become 99\% and 96\%.
At approximately 400k physical actions, the corresponding
rates are 73\% and 77\%; at 600k actions, they are 85\% and 91\%.
Values at each budget use the most recent logged observation at or before
that budget, without interpolation. The chunk-level configuration thus
reaches high training success with fewer actor updates, while stepwise
editing attains higher success at these two matched interaction budgets.
The five-action chunk length used by all LIBERO recipes was first
established in this comparison.

\section{Video Replays}
\label{app:sw-videos}

The project page shows \swVideoFiles{} side-by-side replay clips
(Table~\ref{tab:sw-videos-files}); they are also ancillary files of this
paper. Each clip shows several policies on one task from one initial state,
in the main camera for LIBERO and the head camera for RoboTwin, at the
recorded 20 frames per second. The header above each panel names the policy and states the
outcome of that replay, and a panel that ends early holds its last frame.

\paragraph{The clips are replays, separate from the formal evaluations.}
Formal evaluations run without video. The clips come from separate runs of
the same checkpoints on the same tasks and initial states with rendering
enabled. Of the \swVideoClips{} recorded replays, \swVideoSame{} end in the
same success or failure as the formal episode and \swVideoSameLen{} also
have the same length (Table~\ref{tab:sw-videos-counts}). The states of the
group that compares full and reduced correction were selected because the
bounded editor lost a base success there in the formal evaluation. Every quantitative result in this paper comes from the formal evaluations, and none is computed from
a replay.

\paragraph{Only replays that reproduce the formal outcome are shown.}
A group of clips is eligible only if, for every panel, the replay ends in the
same outcome as the formal episode of the same policy, task and initial
state. From the eligible groups we chose cases that show each kind of
outcome: rescue by all three 112-task editors, rescue by the full-action
editor alone, harm by the bounded editor, the effect of halving its
correction, the frozen-vision control, an editor trained on two source tasks,
and the RoboTwin editor on a source task and on another task. The clips
illustrate these outcomes and are not a random sample.

\ifdefined\AppendixInOnePDF
\par\vspace{1.4\baselineskip}
\noindent\begin{minipage}{\columnwidth}
\centering\footnotesize
\captionsetup{type=table,hypcap=false}%
\captionof{table}{\textbf{Agreement between recorded replays and formal evaluations}, over all
recorded clips. Outcome: replays with the same success or failure as the
formal episode of the same policy, task and initial state. Length: those that
also have the same number of steps.}%
\label{tab:sw-videos-counts}%
\setlength{\tabcolsep}{5pt}%
\begin{tabularx}{\linewidth}{@{}>{\raggedright\arraybackslash}Xrrr@{}}
\toprule
Replay group & Clips & Outcome & Length \\
\midrule
Four policies on the 18 held-out tasks & 196 & 165 & 89 \\
Bounded editor at full and reduced correction & 125 & 88 & 50 \\
Live and frozen editor vision & 8 & 8 & 4 \\
Editor trained on two source tasks & 12 & 10 & 6 \\
RoboTwin base and shared editor & 48 & 43 & 29 \\
\midrule
All groups & 389 & 314 & 178 \\
\bottomrule
\end{tabularx}
\captionsetup{hypcap=true}%
\end{minipage}
\else
\begin{table}[t]
\centering\footnotesize
\caption{\textbf{Agreement between recorded replays and formal evaluations}, over all
recorded clips. Outcome: replays with the same success or failure as the
formal episode of the same policy, task and initial state. Length: those that
also have the same number of steps.}
\label{tab:sw-videos-counts}
\setlength{\tabcolsep}{5pt}
\begin{tabularx}{\linewidth}{@{}>{\raggedright\arraybackslash}Xrrr@{}}
\toprule
Replay group & Clips & Outcome & Length \\
\midrule
Four policies on the 18 held-out tasks & 196 & 165 & 89 \\
Bounded editor at full and reduced correction & 125 & 88 & 50 \\
Live and frozen editor vision & 8 & 8 & 4 \\
Editor trained on two source tasks & 12 & 10 & 6 \\
RoboTwin base and shared editor & 48 & 43 & 29 \\
\midrule
All groups & 389 & 314 & 178 \\
\bottomrule
\end{tabularx}

\end{table}
\fi

\end{document}